\RequirePackage{fix-cm}

\documentclass[twocolumn]{svjour3}          

\pdfoutput=1 

\smartqed  

\usepackage{graphicx}

\usepackage{latexsym}
\usepackage{amsmath}
\usepackage[pagebackref,colorlinks]{hyperref}
\usepackage[numbers]{natbib}
\usepackage{paralist}
\usepackage{caption}
\usepackage{subcaption}
\usepackage{booktabs}
\usepackage{afterpage}
\usepackage{rotating}
\usepackage{multirow}
\usepackage{amssymb}

\makeatletter
\newcommand\footnoteref[1]{\protected@xdef\@thefnmark{\ref{#1}}\@footnotemark}
\makeatother

\begin{document}

\title{A Unified Framework for Compositional Fitting of \mbox{Active Appearance Models}}


\author{Joan Alabort-i-Medina
        \and
        Stefanos Zafeiriou}


\institute{J. Alabort-i-Medina \at
           Department of Computing, Imperial College London, \\
           180 Queen's Gate, London SW7 2AZ, UK \\
           \email{ja310@imperial.ac.uk}
           \and
           Stefanos Zafeiriou \at
           \email{s.zafeiriou@imperial.ac.uk}}

\date{Received: date / Accepted: date}

\maketitle

\begin{abstract}+
Active Appearance Models (AAMs) are one of the most popular and well-established techniques for modeling deformable objects in computer vision. In this paper, we study the problem of fitting AAMs using Compositional Gradient Descent (CGD) algorithms. We present a unified and complete view of these algorithms and classify them with respect to three main characteristics:
\begin{inparaenum}[\itshape i\upshape)]
	\item \emph{cost function};
	\item type of \emph{composition}; and
	\item \emph{optimization method}.
\end{inparaenum}
Furthermore, we extend the previous view by:
\begin{inparaenum}[\itshape a\upshape)]
	\item proposing a novel \emph{Bayesian cost function} that can be interpreted as a general probabilistic formulation of the well-known project-out loss;
	\item introducing two new types of composition, \emph{asymmetric} and \emph{bidirectional}, that combine the gradients of both image and appearance model to derive better convergent and more robust CGD algorithms; and
	\item providing new valuable insights into existent CGD algorithms by reinterpreting them as direct applications of the \emph{Schur complement} and the \emph{Wiberg method}.
\end{inparaenum}
Finally, in order to encourage open research and facilitate future comparisons with our work, we make the implementation of the algorithms studied in this paper publicly available as part of the Menpo Project\footnote{\label{menpo_url}\url{http://www.menpo.org}}.

\end{abstract}

\section{Introduction}
\label{sec:intro}

Active Appearance Models (AAMs) \cite{Cootes2001, Matthews2004} are one of the most popular and well-established techniques for modeling and segmenting deformable objects in computer vision. AAMs are generative parametric models of shape and appearance that can be \emph{fitted} to images to recover the set of model parameters that best describe a particular instance of the object being modeled.

Fitting AAMs is a non-linear optimization problem that requires the  minimization (maximization) of a global error (similarity) measure between the input image and the appearance model. Several approaches \cite{Cootes2001, Hou2001, Matthews2004, Batur2005, Gross2005, Donner2006, Papandreou2008, Liu2009, Saragih2009, Amberg2009, Tresadern2010, Martins2010, Sauer2011, Tzimiropoulos2013, Kossaifi2014, Antonakos2014} have been proposed to define and solve the previous optimization problem. Broadly speaking, they can be divided into two different groups: 
\begin{itemize}
\item \emph{Regression} based \cite{Cootes2001, Hou2001, Batur2005, Donner2006, Saragih2009, Tresadern2010, Sauer2011}
\item \emph{Optimization} based \cite{Matthews2004, Gross2005, Papandreou2008, Amberg2009, Martins2010, Tzimiropoulos2013, Kossaifi2014}
\end{itemize}

Regression based techniques attempt to solve the problem by learning a direct function mapping between the error measure and the optimal values of the parameters. Most notable approaches include variations on the original \cite{Cootes2001} fixed linear regression approach of \cite{Hou2001, Donner2006}, the adaptive linear regression approach of \cite{Batur2005}, and the works of \cite{Saragih2009} and \cite{Tresadern2010} which considerably improved upon previous techniques by using boosted regression. Also, Cootes and Taylor \cite{Cootes2001b} and Tresadern et al. \cite{Tresadern2010} showed that the use of non-linear gradient-based and Haar-like appearance representations, respectively, lead to better fitting accuracy in regression based AAMs. 

Optimization based methods for fitting AAMs were proposed by Matthews and Baker in \cite{Matthews2004}. These techniques are known as Compositional Gradient Decent (CGD) algorithms and are based on direct analytical optimization of the error measure. Popular CGD algorithms include the very efficient project-out Inverse Compositional (PIC) algorithm \cite{Matthews2004}, the accurate but costly Simultaneous Inverse Compositional (SIC) algorithm \cite{Gross2005}, and the more efficient versions of SIC presented in \cite{Papandreou2008} and \cite{Tzimiropoulos2013}. Lucey et al. \cite{Lucey2013} extended these algorithms to the Fourier domain to efficiently enable convolution with Gabor filters, increasing their robustness; and the authors of \cite{Antonakos2014} showed that optimization based AAMs using non-linear feature based (e.g. SIFT\cite{Lowe1999} and HOG \cite{Dalal2005}) appearance models were competitive with modern state-of-the-art techniques in non-rigid face alignment \cite{Xiong2013, Asthana2013} in terms of fitting accuracy.

AAMs have often been criticized for several reasons: 
\begin{inparaenum}[\itshape i\upshape)] 
\item the limited representational power of their linear appearance model; 
\item the difficulty of optimizing shape and appearance parameters simultaneously; and
\item the complexity involved in handling occlusions. 
\end{inparaenum}
However, recent works in this area \cite{Papandreou2008, Saragih2009,Tresadern2010, Lucey2013, Tzimiropoulos2013, Antonakos2014} suggest that these limitations might have been over-stressed in the literature and that AAMs can produce highly accurate results if appropriate training data \cite{Tzimiropoulos2013}, appearance representations \cite{Tresadern2010, Lucey2013, Antonakos2014} and fitting strategies \cite{Papandreou2008, Saragih2009, Tresadern2010, Tzimiropoulos2013} are employed.

In this paper, we study the problem of fitting AAMs using CGD algorithms thoroughly. Summarizing, our main contributions are:
\begin{itemize}
	\item To present a unified and complete overview of the most relevant and recently published CGD algorithms for fitting AAMs \cite{Matthews2004, Gross2005, Papandreou2008, Amberg2009, Martins2010, Tzimiropoulos2012, Tzimiropoulos2013, Kossaifi2014}. To this end, we classify CGD algorithms with respect to three main characteristics: 
	\begin{inparaenum}[\itshape i\upshape)] 
		\item the \emph{cost function} defining the fitting problem; 
		\item the type of \emph{composition} used; and 
		\item the \emph{optimization method} employed to solve the non-linear optimization problem. 
	\end{inparaenum}

	\item To review the probabilistic interpretation of AAMs and propose a novel \emph{Bayesian formulation}\footnote{A preliminary version of this work \cite{Alabort2014} was presented at CVPR 2014.} of the fitting problem. We assume a probabilistic model for appearance generation with both Gaussian noise and a Gaussian prior over a latent appearance space. Marginalizing out the latent appearance space, we derive a novel cost function that only depends on shape parameters and that can be interpreted as a valid and more general probabilistic formulation of the well-known project-out cost function \cite{Matthews2004}. Our Bayesian formulation is motivated by seminal works on probabilistic component analysis and object tracking \cite{Moghaddam1997, Roweis1998, Tipping1999}.

	\item To propose the use of two novel types of composition for AAMs:
	\begin{inparaenum}[\itshape i\upshape)] 
		\item \emph{asymmetric}; and 
		\item \emph{bidirectional}. 
	\end{inparaenum} These types of composition have been widely used in the related field of parametric image alignment \cite{Malis2004, Megret2008, Autheserre2009, Megret2010} and use the gradients of both image and appearance model to derive better convergent and more robust CGD algorithms.

	\item To provide valuable insights into existent strategies used to derive fast and exact simultaneous algorithms for fitting AAMs by reinterpreting them as direct applications of the \emph{Schur complement} \cite{Boyd2004} and the \emph{Wiberg method} \cite{Okatani2006, Strelow2012}.
\end{itemize}

The remainder of the paper is structured as follows. Section \ref{sec:aam} introduces AAMs and reviews their probabilistic interpretation. Section \ref{sec:fitting} constitutes the main section of the paper and contains the discussion and derivations related to the cost functions \ref{sec:cost_function}; composition types \ref{sec:composition}; and optimization methods \ref{sec:optimization}. Implementation details and experimental results are reported in Section \ref{sec:experiment}. Finally, conclusions are drawn in Section \ref{sec:conclusion}.

\section{Active Appearance Models}
\label{sec:aam}

\begin{figure*}
	\centering
	\includegraphics[width=0.16\textwidth]{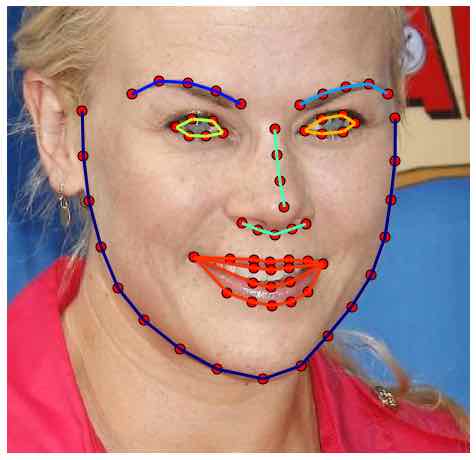}
	\includegraphics[width=0.16\textwidth]{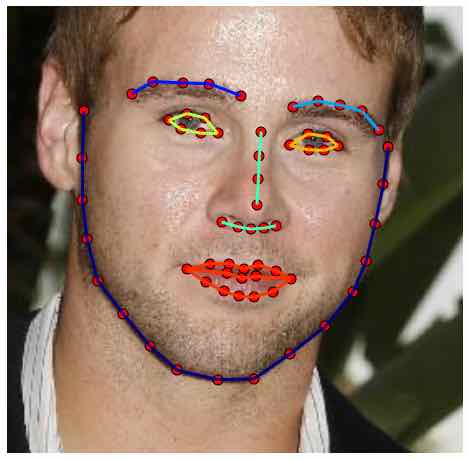}
	\includegraphics[width=0.16\textwidth]{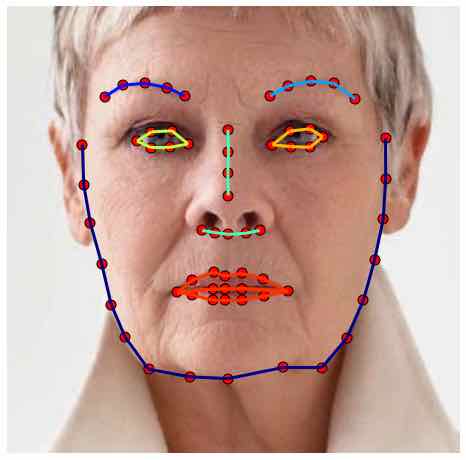}
	\includegraphics[width=0.16\textwidth]{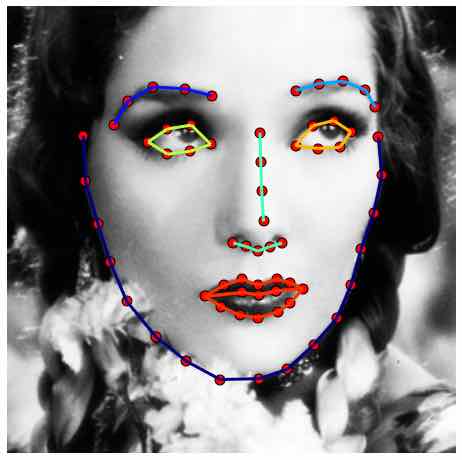}
	\includegraphics[width=0.16\textwidth]{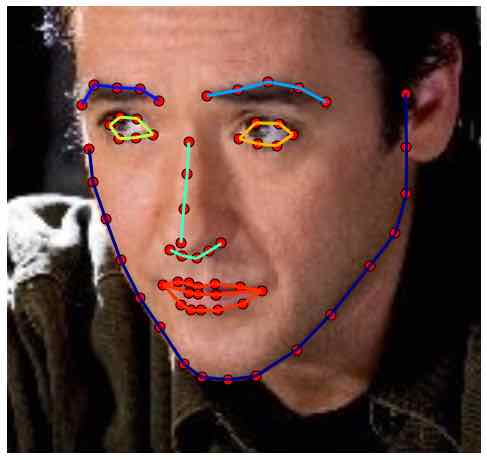}
	\includegraphics[width=0.16\textwidth]{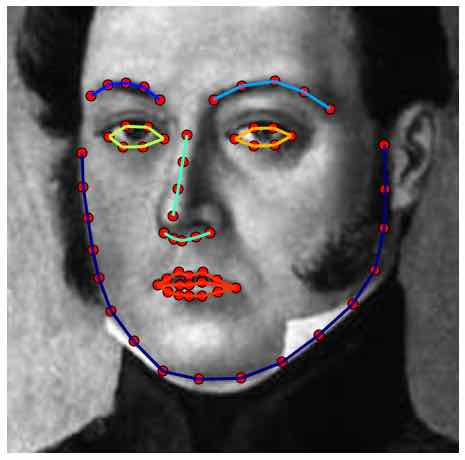}
	\caption{Exemplar images from the Labeled Faces in-the-Wild (LFPW) dataset \cite{Belhumeur2011} for which a consistent set of sparse landmarks representing the shape of the object being model (human face) has been manually defined \cite{Sagonas2013,Sagonas2013b}.}
	\label{fig:lfpw_images}
\end{figure*}

AAMs \cite{Cootes2001,Matthews2004} are generative parametric models that explain visual variations, in terms of shape and appearance, within a particular object class. AAMs are built from a collection of images for which the spatial position of a sparse set of $v$ landmark points $\mathbf{x}_i = (x_i, y_i)^T \in \mathbb{R}^2$ representing the shape $\mathbf{s} = (x_1, y_1, \dots, x_v, y_v)^T \in \mathbb{R}^{2v \times 1}$ of the object being modeled have been manually defined a priori. 

AAMs are themselves composed of three different models:
\begin{inparaenum}[(i)]
	\item shape model; 
	\item appearance model; and
	\item motion model. 
\end{inparaenum} 

The shape model, which is also referred to as Point Distribution Model (PDM), is obtained by typically applying Principal Component Analysis (PCA) to the set of object's shapes. The resulting shape model is mathematically expressed as:
\begin{equation}
	\begin{aligned}
		\mathbf{s} & = \mathbf{\bar{s}} + \sum_{i=1}^n p_i \mathbf{s}_i 
        \\
        & = \mathbf{\bar{s}} + \mathbf{S} \mathbf{p}
	\end{aligned}
\end{equation}
where $\mathbf{\bar{s}} \in \mathbb{R}^{2v \times 1}$ is the mean shape, and $\mathbf{S} \in \mathbb{R}^{2v \times  n}$ and $\mathbf{p} \in \mathbb{R}^{n \times 1}$ denote the shape bases and shape parameters, respectively. In order to allow a particular shape instance $\mathbf{s}$ to be arbitrarily positioned in space, the previous model can be augmented with a global similarity transform. Note that this normally requires the initial shapes to be normalized with respect to the same type of transform (typically using Procrustes Analysis (PA)) before PCA is applied. This results in the following expression for each landmark point of the shape model:
\begin{equation}
	\begin{aligned}
		\mathbf{x}_i & = s \mathbf{R} \left( \mathbf{\bar{x}}_i + \mathbf{X}_i \mathbf{p} \right) + \mathbf{t}
	\end{aligned}
\end{equation}
where $s$, $\mathbf{R} \in \mathbb{R}^{2 \times 2}$ and $\mathbf{t} \in \mathbb{R}^2$  denote the scale, rotation and translation applied by the global similarity transform, respectively. Using the orthonormalization procedure described in \cite{Matthews2004} the final expression for the shape model can be compactly written as the linear combination of a set of bases:
\begin{equation}
	\begin{aligned}
		\mathbf{s} & = \mathbf{\bar{s}} + \sum_{i=1}^4 p^*_i \mathbf{s}^*_i + \sum_{i=1}^n p_i \mathbf{s}_i 
        \\
        & = \mathbf{\bar{s}} + \mathbf{S} \mathbf{p}
	\end{aligned}
    \label{eq:shape_model}
\end{equation}
where $\mathbf{S} = (\mathbf{s}^*_1, \dots, \mathbf{s}^*_4, \mathbf{s}_1, \cdots, \mathbf{s}_n) \in \mathbb{R}^{2v \times (n+4)}$ and $\mathbf{p} = (p^*_1, \dots, p^*_4, p_1, \dots, p_n)^T \in \mathbb{R}^{(n+4) \times 1}$ are redefined as the concatenation of the similarity bases $\mathbf{s}^*_i$ and similarity parameters $p^*_i$ with the original $\mathbf{S}$ and $\mathbf{p}$, respectively.

The appearance model is obtained by warping the original images onto a common reference frame (typically defined in terms of the mean shape $\mathbf{\bar{s}}$) and applying PCA to the obtained warped images. Mathematically, the appearance model is defined by the following expression:
\begin{equation}
	\begin{aligned}
		A(\mathbf{x}) & = \bar{A}(\mathbf{x}) + \sum_{i=1}^m c_i A_i(\mathbf{x})
	\end{aligned}
    \label{eq:app_model}
\end{equation}
where $\mathbf{x} \in \Omega$ denote all pixel positions on the reference frame, and $\bar{A}(\mathbf{x})$, $A_i(\mathbf{x})$ and $c_i$ denote the mean texture, the appearance bases and appearance parameters, respectively. Denoting $\mathbf{a} = \text{vec}(A(\mathbf{x}))$ as the vectorized version of the previous appearance instance, Equation \ref{eq:app_model} can be concisely written in vector form as:
\begin{equation}
	\begin{aligned}
		\mathbf{a} & = \mathbf{\bar{a}} + \mathbf{A} \mathbf{c}
	\end{aligned}
    \label{eq:app_model_vec}
\end{equation}
where $\mathbf{a} \in \mathbb{R}^{F \times 1}$ is the mean appearance, and $\mathbf{A} \in \mathbb{R}^{F \times  m}$ and $\mathbf{c} \in \mathbb{R}^{m \times 1}$ denote the appearance bases and appearance parameters, respectively.

The role of the motion model, denoted by $\mathcal{W}(\mathbf{x}; \mathbf{p})$, is to extrapolate the position of all pixel positions $\mathbf{x} \in \Omega$ from the reference frame to a particular shape instance $\mathbf{s}$ (and vice-versa) based on their relative position with respect to the sparse set of landmarks defining the shape model (for which direct correspondences are always known). Classic motion models for AAMs are PieceWise Affine (PWA) \cite{Cootes2004,Matthews2004} and Thin Plate Splines (TPS) \cite{Cootes2004,Papandreou2008} warps. 

Given an image $I$ containing the object of interest, its manually annotated ground truth shape $\mathbf{s}$, and a particular motion model $\mathcal{W}(\mathbf{x}, \mathbf{p})$; the two main assumptions behind AAMs are:
\begin{enumerate}
	\item The ground truth shape of the object can be well approximated by the shape model
	\begin{eqnarray}
		\begin{aligned}
			\mathbf{s} & \approx \mathbf{\bar{s}} + \mathbf{S} \mathbf{p}
		\end{aligned}
	    \label{eq:aam_1}
	\end{eqnarray}

	\item The object's appearance can be well approximated by the appearance model after the image is warped, using the motion model and the previous shape approximation, onto the reference frame:
	\begin{eqnarray}
		\begin{aligned}
			\mathbf{i}[\mathbf{p}] & \approx \mathbf{\bar{a}} + \mathbf{A} \mathbf{c} 
		\end{aligned}
	    \label{eq:aam_2}
	\end{eqnarray}
	where $\mathbf{i}[\mathbf{p}] = \mathrm{vec}(I(\mathcal{W}(\mathbf{x}; \mathbf{p})))$ denotes the vectorized version of the warped image. Note that, the warp $\mathcal{W}(\mathbf{x}; \mathbf{p})$ which explicitly depends on the shape parameters $\mathbf{p}$, relates the shape and appearance models and is a central part of the AAMs formulation.
\end{enumerate}

Because of the explicit use of the motion model, the two previous assumptions provide a concise definition of AAMs. At this point, it is worth mentioning that the vector notation of Equations \ref{eq:aam_1} and \ref{eq:aam_2} will be, in general, the preferred notation in this paper.


\subsection{Probabilistic Formulation}
\label{sec:probabilistic_aam}

A probabilistic interpretation of AAMs can be obtained by rewriting Equations \ref{eq:aam_1} and \ref{eq:aam_2} assuming probabilistic models for shape and appearance generation. In this paper, motivated by seminal works on Probabilistic Component Analysis (PPCA) and object tracking \cite{Tipping1999, Roweis1998, Moghaddam1997}, we will assume probabilistic models for shape and appearance generation with both Gaussian noise and Gaussian priors over the latent shape and appearance spaces\footnote{This formulation is generic and one could assume other probabilistic generative models \cite{vanderMaaten2010, Bach2005, Prince2012, Nicolau2014} to define novel probabilistic versions of AAMs.}:
\begin{equation}
	\begin{aligned}
		\mathbf{s} & = \bar{\mathbf{s}} + \mathbf{S} \mathbf{p} + \boldsymbol{\varepsilon}
		\\
		\mathbf{p} & \sim \mathcal{N} \left( \mathbf{0}, \mathbf{\Lambda} \right) 
		\\
		\boldsymbol{\varepsilon} & \sim \mathcal{N} \left( \mathbf{0}, \varsigma ^2 \mathbf{I} \right) 
	\end{aligned}
	\label{eq:paam_1}
\end{equation}
\begin{equation}
	\begin{aligned}
		\mathbf{i}[\mathbf{p}] & = \bar{\mathbf{a}} + \mathbf{A} \mathbf{c} + \boldsymbol{\epsilon}
		\\
		\mathbf{c} & \sim \mathcal{N} \left( \mathbf{0}, \mathbf{\Sigma} \right) 
		\\
		\boldsymbol{\epsilon} & \sim \mathcal{N} \left( \mathbf{0}, \sigma^2 \mathbf{I} \right) 
	\end{aligned}
	\label{eq:paam_2}
\end{equation}
where the diagonal matrices $\mathbf{\Lambda} = \textrm{diag}(\lambda_{\mathbf{s}_1}, \cdots, \lambda_{\mathbf{s}_m})$ and $\mathbf{\Sigma} = \textrm{diag}(\lambda_{\mathbf{a}_1}, \cdots, \lambda_{\mathbf{a}_m})$ contain the eigenvalues associated to shape and appearance eigenvectors respectively and where $\varsigma ^2$ and $\sigma^2$ denote the estimated shape and image noise\footnote{\label{foot:noise}Theoretically, the optimal value for $\varsigma ^2$ and $\sigma^2$ is the average value of the eigenvalues associated to the discarded shape and appearance eigenvectors respectively i.e. \mbox{$\varsigma ^{2} = \frac{1}{N-n}\sum_{i=n}^N \lambda_{\mathbf{s}_i}$} and \mbox{$\sigma^{2} = \frac{1}{M-m}\sum_{i=m}^M \lambda_{\mathbf{a}_i}$} \cite{Moghaddam1997}.} respectively.

This probabilistic formulation will be used to derive Maximum-Likelihood (ML), Maximum A Posteriori (MAP) and Bayesian cost functions for fitting AAMs in Sections \ref{sec:rssd} and \ref{sec:rpo}.


\section{Fitting Active Appearance Models}
\label{sec:fitting}


Several techniques have been proposed to fit AAMs to images \cite{Cootes2001, Hou2001, Matthews2004, Batur2005, Gross2005, Donner2006, Papandreou2008, Liu2009, Saragih2009, Amberg2009, Tresadern2010, Martins2010, Sauer2011, Tzimiropoulos2013, Kossaifi2014, Antonakos2014}. In this paper, we will center the discussion around Compositional Gradient Descent (CGD) algorithms \cite{Matthews2004, Gross2005, Papandreou2008, Amberg2009, Martins2010, Tzimiropoulos2013, Kossaifi2014} for fitting AAMs. Consequently, we will not review regression based approaches. For more details on these type of methods the interested reader is referred to the existent literature \cite{Cootes2001, Hou2001, Batur2005, Donner2006, Liu2009, Saragih2009, Tresadern2010, Sauer2011}.

The following subsections present a unified and complete view of CGD algorithms by classifying them with respect to their three main characteristics: 
\begin{inparaenum}[\itshape a\upshape)]
\item \emph{cost function} (Section \ref{sec:cost_function}); 
\item type of \emph{composition} (Section \ref{sec:composition}); and 
\item \emph{optimization method} (Section \ref{sec:optimization}).
\end{inparaenum}

\subsection{Cost Function}
\label{sec:cost_function}

AAM fitting is typically formulated as the (regularized) search over the shape and appearance parameters that minimize a global error measure between the vectorized warped image and the appearance model:
\begin{equation}
    \begin{aligned}
        \mathbf{p}^*, \mathbf{c}^* & = \underset{\mathbf{p}, \mathbf{c}} {\mathrm{arg\, min\;}} \mathcal{R} (\mathbf{p}, \mathbf{c}) + \mathcal{D} (\mathbf{i}[\mathbf{p}], \mathbf{c}) 
        \end{aligned}
    \label{eq:aam_fitting}
\end{equation}
where $\mathcal{D}$ is a data term that quantifies the global error measure between the vectorized warped image and the appearance model and $\mathcal{R}$ is an \emph{optional} regularization term that penalizes complex shape and appearance deformations.

\subsubsection{Sum of Squared Differences}
\label{sec:rssd}

Arguably, the most natural choice for the previous data term is the \emph{Sum of Squared Differences} (SSD) between the vectorized warped image and the linear appearance model\footnote{This choice of $\mathcal{D}$ is naturally given by second main assumption behind AAMs, Equation \ref{eq:aam_2} and by the linear generative model of appearance defined by Equation \ref{eq:paam_2}.}. Consequently, the \emph{classic} AAM fitting problem is defined by the following non-linear optimization problem\footnote{The residual $\mathbf{r}$ in Equation \ref{eq:ssd} is linear with respect to the appearance parameters $\mathbf{c}$ and non-linear with respect to the shape parameters $\mathbf{p}$ through the warp $\mathcal{W}(\mathbf{x}; \mathbf{p})$}:
\begin{equation}
    \begin{aligned}
        \mathbf{p}^*, \mathbf{c}^* & = \underset{\mathbf{p}, \mathbf{c}} {\mathrm{arg\, min\;}} \frac{1}{2} \mathbf{r}^T\mathbf{r}
        \\
        & = \underset{\mathbf{p}, \mathbf{c}} {\mathrm{arg\, min\;}} 
        \underbrace{\frac{1}{2} \left\| \mathbf{i}[\mathbf{p}] - \left( \bar{\mathbf{a}} + \mathbf{A} \mathbf{c} \right) \right\|^2}_{\mathcal{D} (\mathbf{i}[\mathbf{p}], \mathbf{c})} 
    \end{aligned}
    \label{eq:ssd}
\end{equation}

On the other hand, considering regularization, the most natural choice for $\mathcal{R}$ is the sum of ${\ell_2}^2$-norms over the shape and appearance parameters. In this case, the \emph{regularized} AAM fitting problem is defined as follows:
\begin{equation}
    \begin{aligned}
        \mathbf{p}^*, \mathbf{c}^* & = \underset{\mathbf{p}, \mathbf{c}} {\mathrm{arg\, min\;}} \frac{1}{2}||\mathbf{p}||^2 + \frac{1}{2}||\mathbf{c}||^2 + \frac{1}{2} \mathbf{r}^T\mathbf{r}
        \\
        & =\underset{\mathbf{p}, \mathbf{c}} {\mathrm{arg\, min\;}} \underbrace{\frac{1}{2}||\mathbf{p}||^2 + \frac{1}{2}||\mathbf{c}||^2}_{\mathcal{R} (\mathbf{p}, \mathbf{c})} +
        \\
        & \qquad \qquad \quad \underbrace{\frac{1}{2}|| \mathbf{i}[\mathbf{p}] - (\mathbf{\bar{a}} + \mathbf{A} \mathbf{c}) ||^2}_{\mathcal{D} (\mathbf{i}[\mathbf{p}], \mathbf{c})}
    \end{aligned}
    \label{eq:rssd}
\end{equation}

\subsubsection*{Probabilistic Formulation}
\label{sec:rssd_pi}

A probabilistic formulation of the previous cost function can be naturally derived using the probabilistic generative models introduced in Section \ref{sec:probabilistic_aam}. Denoting the models' parameters as \mbox{$\Theta = \{\mathbf{\bar{s}}, \mathbf{S}, \mathbf{\Lambda}, \mathbf{\bar{a}}, \mathbf{A}, \mathbf{\Sigma}, \sigma^2\}$} a ML formulation can be derived as follows:
\begin{equation}
    \begin{aligned}
        \mathbf{p}^*, \mathbf{c}^* & = \underset{\mathbf{p}, \mathbf{c}}{\mathrm{arg\,max\;}} p(\mathbf{i}[\mathbf{p}] |  \mathbf{p}, \mathbf{c}, \Theta) 
        \\
        & = \underset{\mathbf{p}, \mathbf{c}}{\mathrm{arg\,max\;}} \ln p(\mathbf{i}[\mathbf{p}] | \mathbf{p}, \mathbf{c}, \Theta)
        \\
        & = \underset{\mathbf{p}, \mathbf{c}}{\mathrm{arg\,min\;}} 
        \underbrace{ \frac{1}{2\sigma^2} || \mathbf{i}[\mathbf{p}] - (\mathbf{\bar{a}} + \mathbf{A} \mathbf{c}) ||^2}_{\mathcal{D}(\mathbf{i}[\mathbf{p}], \mathbf{c})} 
    \end{aligned}
    \label{eq:prob_ssd}
\end{equation}
and a MAP formulation can be similarly derived by taking into account the prior distributions over the shape and appearance parameters:
\begin{equation}
    \begin{aligned}
        \mathbf{p}^*, \mathbf{c}^* & = \underset{\mathbf{p}, \mathbf{c}}{\mathrm{arg\,max\;}} p(\mathbf{p}, \mathbf{c}, \mathbf{i}[\mathbf{p}] | \Theta) 
        \\
        & = \underset{\mathbf{p}, \mathbf{c}}{\mathrm{arg\,max\;}}  p(\mathbf{p} | \mathbf{\Lambda})  p(\mathbf{c} | \mathbf{\Sigma}) p(\mathbf{i}[\mathbf{p}] |
        \mathbf{p}, \mathbf{c}, \Theta)  
        \\
        & = \underset{\mathbf{p}, \mathbf{c}}{\mathrm{arg\,max\;}}  \ln p(\mathbf{p} | \mathbf{\Lambda}) + \ln p(\mathbf{c} | \mathbf{\Sigma}) +
        \\
        & \qquad \qquad \quad \ln p(\mathbf{i}[\mathbf{p}] | \mathbf{p}, \mathbf{c}, \Theta)
        \\
        & = \underset{\mathbf{p}, \mathbf{c}}{\mathrm{arg\,min\;}}  \underbrace{\frac{1}{2} ||\mathbf{p}||^2_{\mathbf{\Lambda}^{-1}} + \frac{1}{2}||\mathbf{c}||^2_{\mathbf{\Sigma}^{-1}}}_{\mathcal{R}(\mathbf{p}, \mathbf{c})} +
        \\
        & \qquad \qquad \quad \underbrace{ \frac{1}{2\sigma^2} || \mathbf{i}[\mathbf{p}] - (\mathbf{\bar{a}} + \mathbf{A} \mathbf{c}) ||^2}_{\mathcal{D}(\mathbf{i}[\mathbf{p}], \mathbf{c})} 
    \end{aligned}
    \label{eq:prob_rssd}
\end{equation}
where we have assumed the shape and appearance parameters to be independent\footnote{This is a common assumption in CGD algorithms \cite{Matthews2004}, however, in reality, some degree of dependence between these parameters is to be expected \cite{Cootes2001}.}.

The previous ML and MAP formulations are weighted version of the optimization problem defined by Equation \ref{eq:ssd} and \ref{eq:rssd}. In both cases, the maximization of the conditional probability of the vectorized warped image given the shape, appearance and model parameters leads to the minimization of the data term $\mathcal{D}$ and, in the MAP case, the maximization of the prior probability over the shape and appearance parameters leads to the minimization of the regularization term $\mathcal{R}$.

\subsubsection{Project-Out}
\label{sec:rpo}

Matthews and Baker showed in \cite{Matthews2004} that one could express the SSD between the vectorized warped image and the linear PCA-based\footnote{The use of PCA ensures the orthonormality of the appearance bases and, consequently, $\mathbf{A}^T\mathbf{A} = \mathbf{I}$ (where $\mathbf{I} $ denotes the identity matrix). Similarly, the use of PCA also ensures orthogonality between the appearance mean and the appearance bases and, hence, $\mathbf{A}^T\bar{\mathbf{a}} =  \mathbf{0}$.} appearance model as the sum of two different terms:
\begin{equation}
    \begin{aligned}
        \frac{1}{2}\mathbf{r}^T \mathbf{r} & = \frac{1}{2}\mathbf{r}^T (\mathbf{A}\mathbf{A}^T + \mathbf{I} - \mathbf{A}\mathbf{A}^T) \mathbf{r}
        \\
        & = \frac{1}{2}\mathbf{r}^T (\mathbf{A}\mathbf{A}^T) \mathbf{r} + \frac{1}{2}\mathbf{r}^T (\mathbf{I} - \mathbf{A}\mathbf{A}^T) \mathbf{r}
        \\
        & = \frac{1}{2}\left\| \mathbf{i}[\mathbf{p}] - \left( \bar{\mathbf{a}} + \mathbf{A} \mathbf{c} \right) \right\|_{\mathbf{A}\mathbf{A}^T}^2 \, + 
        \\
        & \quad \, \frac{1}{2}\left\| \mathbf{i}[\mathbf{p}] - \left( \bar{\mathbf{a}} + \mathbf{A} \mathbf{c} \right) \right\|_{\mathbf{I} - \mathbf{A}\mathbf{A}^T}^2 
        \\
        & = f_1(\mathbf{p}, \mathbf{c}) + f_2(\mathbf{p}, \mathbf{c})
    \label{eq:ssd_terms}
    \end{aligned}
\end{equation}
The first term defines the distance \emph{within} the appearance subspace and it is always $0$ regardless of the value of the shape parameters $\mathbf{p}$:
\begin{equation}
    \begin{aligned}
        f_1(\mathbf{p}, \mathbf{c}) & = \frac{1}{2}\left\| \mathbf{i}[\mathbf{p}] - \left( \bar{\mathbf{a}} + \mathbf{A} \mathbf{c} \right) \right\|_{\mathbf{A}\mathbf{A}^T}^2
        \\
        & = \frac{1}{2} \left( \underbrace{\mathbf{i}[\mathbf{p}]^T \mathbf{A}}_{\mathbf{c}^T} \underbrace{\mathbf{A}^T \mathbf{i}[\mathbf{p}]}_{\mathbf{c}} - \underbrace{2\overbrace{\mathbf{i}[\mathbf{p}]^T \mathbf{A}}^{\mathbf{c}^T} \overbrace{\mathbf{A}^T \bar{\mathbf{a}}}^{\mathbf{0}}}_{0} \, - \right.
        \\
        & \quad \,\, 2\underbrace{\mathbf{i}[\mathbf{p}]^T \mathbf{A}}_{\mathbf{c}^T} \underbrace{\overbrace{\mathbf{A}^T \mathbf{A}}^{\mathbf{I}} \mathbf{c}}_{\mathbf{c}} + \underbrace{\overbrace{\bar{\mathbf{a}}^T \mathbf{A}}^{\mathbf{0}^T} \overbrace{\mathbf{A}^T \bar{\mathbf{a}}}^{\mathbf{0}}}_{0} \, +
        \\
        & \quad \,\, \left. \underbrace{\mathbf{c}^T \overbrace{\mathbf{A}^T \mathbf{A}}^{\mathbf{I}}}_{\mathbf{c}^T} \underbrace{\overbrace{\mathbf{A}^T \mathbf{A}}^{\mathbf{I}} \mathbf{c}}_{\mathbf{c}} \right)
        \\
        & = \frac{1}{2}(\mathbf{c}^T\mathbf{c} - 2\mathbf{c}^T\mathbf{c} + \mathbf{c}^T\mathbf{c})
        \\
        & = 0
    \label{eq:ssd_term1}
    \end{aligned}
\end{equation}
The second term measures the distance \emph{to} the appearance subspace i.e. the distance within its orthogonal complement. After some algebraic manipulation, one can show that this term reduces to a function that only depends on the shape parameters $\mathbf{p}$:
\begin{equation}
    \begin{aligned}
        f_2(\mathbf{p}, \mathbf{c}) & = \frac{1}{2}\left\| \mathbf{i}[\mathbf{p}] - \left( \bar{\mathbf{a}} + \mathbf{A} \mathbf{c} \right) \right\|_{\bar{\mathbf{A}}}^2
        \\
        & = \frac{1}{2} \left( \mathbf{i}[\mathbf{p}]^T \bar{\mathbf{A}} \mathbf{i}[\mathbf{p}] - 2\mathbf{i}[\mathbf{p}]^T \bar{\mathbf{A}} \bar{\mathbf{a}} \, - \right.
        \\
        & \quad \,\, \left. \underbrace{2\mathbf{i}[\mathbf{p}]^T \bar{\mathbf{A}} \mathbf{A}\mathbf{c}}_{0} + \bar{\mathbf{a}}^T \bar{\mathbf{A}} \bar{\mathbf{a}} + \underbrace{\mathbf{c}^T \mathbf{A}^T \bar{\mathbf{A}} \mathbf{A}\mathbf{c}}_{0} \right)
        \\
        & = \frac{1}{2} (\mathbf{i}[\mathbf{p}]^T \bar{\mathbf{A}} \mathbf{i}[\mathbf{p}] - 2\mathbf{i}[\mathbf{p}]^T \bar{\mathbf{A}} \bar{\mathbf{a}} + \bar{\mathbf{a}}^T \bar{\mathbf{A}} \bar{\mathbf{a}} )
        \\
        & = \frac{1}{2} \left\| \mathbf{i}[\mathbf{p}] - \bar{\mathbf{a}} \right\|_{\bar{\mathbf{A}}}^2
    \label{eq:ssd_term2}
    \end{aligned}
\end{equation}
where, for convenience, we have defined the orthogonal complement to the appearance subspace as $\bar{\mathbf{A}}= \mathbf{I} -\mathbf{A}\mathbf{A}^T$. Note that, as mentioned above, the previous term does not depend on the appearance parameters $\mathbf{c}$:
\begin{equation}
    \begin{aligned}
        f_2(\mathbf{p}, \mathbf{c}) & = \hat{f}_2(\mathbf{p}) = \frac{1}{2}\left\| \mathbf{i}[\mathbf{p}] - \bar{\mathbf{a}} \right\|_{\bar{\mathbf{A}}}^2
    \label{eq:po}
    \end{aligned}
\end{equation}

Therefore, using the previous \emph{project-out trick}, the minimization problems defined by Equations \ref{eq:ssd} and \ref{eq:rssd} reduce to:
\begin{equation}
    \begin{aligned}
        \mathbf{p}^* & = \underset{\mathbf{p}} {\mathrm{arg\, min\;}} \underbrace{\frac{1}{2}|| \mathbf{i}[\mathbf{p}] - \mathbf{\bar{a}} ||^2_{\bar{\mathbf{A}}}}_{\mathcal{D} (\mathbf{i}[\mathbf{p}])}
    \label{eq:po}
    \end{aligned}
\end{equation}
and
\begin{equation}
    \begin{aligned}
        \mathbf{p}^* & = \underset{\mathbf{p}} {\mathrm{arg\, min\;}} \underbrace{\frac{1}{2}||\mathbf{p}||^2}_{\mathcal{R} (\mathbf{p})} + \underbrace{\frac{1}{2}|| \mathbf{i}[\mathbf{p}] - \mathbf{\bar{a}} ||^2_{\bar{\mathbf{A}}}}_{\mathcal{D} (\mathbf{i}[\mathbf{p}])}
    \label{eq:rpo}
    \end{aligned}
\end{equation}
respectively.

\subsubsection*{Probabilistic Formulation}
\label{sec:po_pi}
Assuming the probabilistic models defined in Section \ref{sec:probabilistic_aam}, a \emph{Bayesian} formulation of the previous project-out data term can be naturally derived by marginalizing over the appearance parameters to obtain the following marginalized density:
\begin{equation}
    \begin{aligned}
        p(\mathbf{i}[\mathbf{p}] | \mathbf{p}, \Theta) & = \int_c p(\mathbf{i}[\mathbf{p}] | \mathbf{p}, \mathbf{c}, \Theta) p(\mathbf{c}|\mathbf{\Sigma}) d\mathbf{c} 
        \\
        & = \mathcal{N}(\bar{\mathbf{a}}, \mathbf{A}\mathbf{\Sigma}\mathbf{A}^T + \sigma^2 \mathbf{I})
    \end{aligned}
    \label{eq:marginal}
\end{equation}
and applying the Woodbury formula\footnote{Using the Woodbury formula:
\begin{equation*}
    \begin{aligned}
    \small
    	(\mathbf{A} \mathbf{\Sigma} \mathbf{A}^T + \sigma^2 \mathbf{I})^{-1} & = \frac{1}{\sigma^2} \mathbf{I} - \frac{1}{\sigma^4} \mathbf{A} \underbrace{(\mathbf{\Sigma}^{-1} + \frac{1}{\sigma^2}\mathbf{I})^{-1}}_{\textrm{reapply Woodbury}} \mathbf{A}^T
    	\\
    	& = \frac{1}{\sigma^2} \mathbf{I} - \frac{1}{\sigma^4} \mathbf{A} (\sigma^2\mathbf{I} - \sigma^4(\mathbf{\Sigma + \sigma^2\mathbf{I}})^{-1}) \mathbf{A}^T
    	\\
    	& = \frac{1}{\sigma^2} \mathbf{I} - \frac{1}{\sigma^4} \mathbf{A} (\sigma^2\mathbf{I} - \sigma^4\mathbf{D}^{-1}) \mathbf{A}^T
    	\\
    	& = \mathbf{A}\mathbf{D}^{-1}\mathbf{A}^T + \frac{1}{\sigma^2} (\mathbf{I} - \mathbf{A}\mathbf{A}^T)
    \end{aligned}
    \label{eq:woodbury}
\end{equation*}} \cite{Woodbury1950} to decompose the natural logarithm of the previous density into the sum of two different terms:
\begin{equation}
    \begin{aligned}
        \ln p(\mathbf{i}[\mathbf{p}] | \mathbf{p}, \Theta) & = \frac{1}{2}|| \mathbf{i}[\mathbf{p}] - \mathbf{\bar{a}} ||^2_{(\mathbf{A}\mathbf{\Sigma}\mathbf{A}^T + \sigma^2 \mathbf{I})^{-1}}
        \\
        & = \frac{1}{2}|| \mathbf{i}[\mathbf{p}] - \mathbf{\bar{a}} ||^2_{\mathbf{A}\mathbf{D}^{-1}\mathbf{A}^T} + 
        \\
        & \quad \, \frac{1}{2\sigma^2}|| \mathbf{i}[\mathbf{p}] - \mathbf{\bar{a}} ||^2_{\bar{\mathbf{A}}}
    \end{aligned}
    \label{eq:prob_po}
\end{equation}
where $\mathbf{D} = \textrm{diag}(\lambda_{\mathbf{a}_1} + \sigma^2, \cdots, \lambda_{\mathbf{a}_m} + \sigma^2)$.

\begin{figure}
    \centering
    \includegraphics[width=0.50\textwidth]{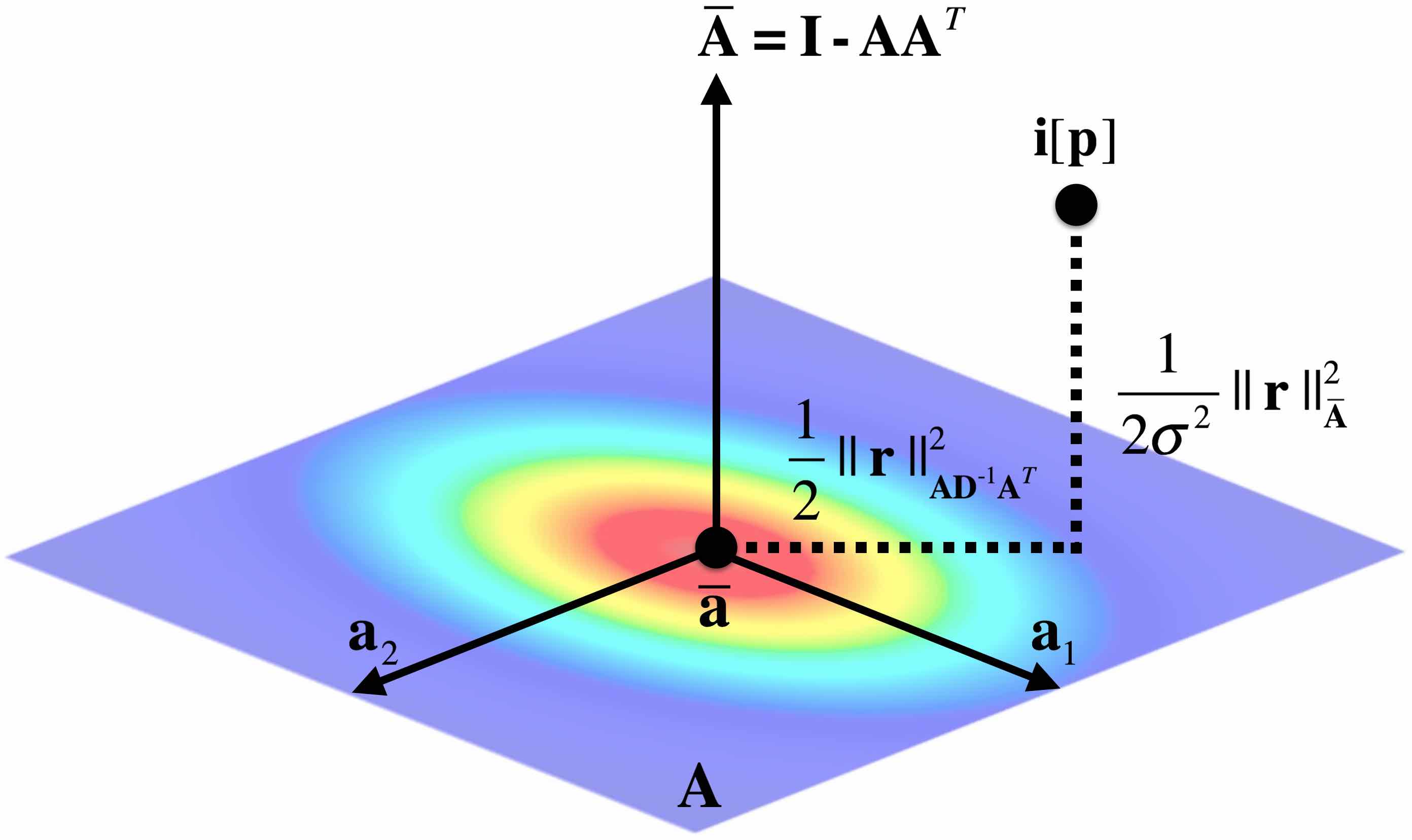}
    \caption{The Bayesian project-out formulation fits AAMs by minimizing two different distances: $i)$ the Mahalanobis distance \emph{within} the linear appearance subspace; and $ii)$ the Euclidean distance \emph{to} the linear appearance subspace (i.e. the Euclidean distance within its orthogonal complement) weighted by the inverse of the estimated image noise.}
    \label{fig:bayes}
\end{figure}

As depicted by Figure \ref{fig:bayes}, the previous two terms define respectively: 
\begin{inparaenum}[\itshape i\upshape)]
    \item the Mahalanobis distance \emph{within} the linear appearance subspace; and 
    \item the Euclidean distance \emph{to} the linear appearance subspace (i.e. the Euclidean distance within its orthogonal complement) weighted by the inverse of the estimated image noise.
\end{inparaenum}
Note that when the variance $\mathbf{\Sigma}$ of the prior distribution over the latent appearance space increases (and especially as $\mathbf{\Sigma} \rightarrow \infty$) $\mathbf{c}$ becomes uniformly distributed and the contribution of the first term $\frac{1}{2}|| \mathbf{i}[\mathbf{p}] - \mathbf{\bar{a}} ||^2_{\mathbf{A}\mathbf{D}^{-1}\mathbf{A}^T}$ vanishes; in this case, we obtain a weighted version of the project-out data term defined by Equation \ref{eq:po}. Hence, given our Bayesian formulation, the project-out loss arises naturally by assuming a uniform prior over the latent appearance space.

The probabilistic formulations of the minimization problems defined by Equations \ref{eq:po} and \ref{eq:rpo} can be derived, from the previous Bayesian Project-Out (BPO) cost function, as
\begin{equation}
    \begin{aligned}
        \mathbf{p}^* & = \underset{\mathbf{p}}{\mathrm{arg\,max\;}} \ln p(\mathbf{i}[\mathbf{p}] | \mathbf{p}, \Theta)
        \\
        & = \underset{\mathbf{p}}{\mathrm{arg\,min\;}} \underbrace{\frac{1}{2\sigma^2}|| \mathbf{i}[\mathbf{p}] - \mathbf{\bar{a}} ||^2_{\mathbf{Q}}}_{\mathcal{D}(\mathbf{i}[\mathbf{p}])}
    \end{aligned}
    \label{eq:prob_rpo}
\end{equation}
and 
\begin{equation}
    \begin{aligned}
        \mathbf{p}^* & = \underset{\mathbf{p}}{\mathrm{arg\,max\;}} p(\mathbf{p}, \mathbf{i}[\mathbf{p}] | \Theta) 
        \\
        & = \underset{\mathbf{p}}{\mathrm{arg\,max\;}}  p(\mathbf{p} | \mathbf{\Lambda})  p(\mathbf{i}[\mathbf{p}] |
        \mathbf{p}, \Theta)  
        \\
        & = \underset{\mathbf{p}}{\mathrm{arg\,max\;}}  \ln p(\mathbf{p} | \mathbf{\Lambda}) + \ln p(\mathbf{i}[\mathbf{p}] | \mathbf{p}, \Theta)
        \\
        & = \underset{\mathbf{p}}{\mathrm{arg\,min\;}}  \underbrace{\frac{1}{2} ||\mathbf{p}||^2_{\mathbf{\Lambda}^{-1}}}_{\mathcal{R}(\mathbf{p})} + \underbrace{\frac{1}{2\sigma^2}|| \mathbf{i}[\mathbf{p}] - \mathbf{\bar{a}} ||^2_{\mathbf{Q}}}_{\mathcal{D}(\mathbf{i}[\mathbf{p}])}
    \end{aligned}
    \label{eq:prob_rpo}
\end{equation}
respectively. Where we have defined the BPO operator as $\mathbf{Q} = \mathbf{I} - \mathbf{A}(\mathbf{I} - \sigma^2\mathbf{D}^{-1})\mathbf{A}^T$.

\subsection{Type of Composition}
\label{sec:composition}

Assuming, for the time being, that the true appearance parameters $\mathbf{c}^*$ are known, the problem defined by Equation \ref{eq:ssd} reduces to a non-rigid image alignment problem \cite{Baker2004, Munoz2014} between the particular instance of the object present in the image and its optimal appearance reconstruction by the appearance model:
\begin{equation}
    \begin{aligned}
        \mathbf{p}^* & = \underset{\mathbf{p}}{\mathrm{arg\,min\;}} 
        \frac{1}{2}\left\| \mathbf{i}[\mathbf{p}] - \mathbf{a} \right\|^2 
    \label{eq:ssd_shape}
    \end{aligned}
\end{equation}
where $\mathbf{a} = \bar{\mathbf{a}} + \mathbf{A} \mathbf{c}^*$ is obtained by directly evaluating Equation \ref{eq:app_model} given the true appearance parameters $\mathbf{c}^*$.

CGD algorithms iteratively solve the previous non-linear optimization problem with respect to the shape parameters $\mathbf{p}$ by:
\begin{enumerate}
    \item Introducing an incremental warp $\mathcal{W}(\mathbf{x}; \Delta\mathbf{p})$ according to the particular composition scheme being used.
    \label{it:step_1}
    
    \item Linearizing the previous incremental warp around the identity warp $\mathcal{W}(\mathbf{x}; \Delta\mathbf{p}) = \mathcal{W}(\mathbf{x}; \mathbf{0}) = \mathbf{x}$.
    \label{it:step_2}
    
    \item Solving for the parameters $\Delta\mathbf{p}$ of the incremental warp.
    \label{it:step_3}

    \item Updating the current warp estimate by using an appropriate compositional update rule.
    \label{it:step_4}

    \item Going back to Step \ref{it:step_1} until a particular convergence criterion is met.
    \label{it:step_5}
\end{enumerate}



Existent CGD algorithms for fitting AAMs have introduced the incremental warp either on the image or the model sides in what are known as \emph{forward} and \emph{inverse} compositional frameworks \cite{Matthews2004, Gross2005, Papandreou2008, Amberg2009, Martins2010, Tzimiropoulos2013} respectively. Inspired by related works in field of image alignment \cite{Malis2004, Megret2008, Autheserre2009, Megret2010}, we notice that novel CGD algorithms can be derived by introducing incremental warps on both image and model sides simultaneously. Depending on the exact relationship between these incremental warps we define two novel types of composition: \emph{asymmetric} and \emph{bidirectional}.

The following subsections explain how to introduce the incremental warp into the cost function and how to update the current warp estimate for the four types of composition considered in this paper: 
\begin{inparaenum}[\itshape i\upshape)]
    \item forward; 
    \item inverse;
    \item asymmetric; and
    \item bidirectional.
\end{inparaenum} For convenience, in these subsections we will use the simplified cost function defined by Equation \ref{eq:ssd_shape}. Furthermore, to maintain consistency with the vector notation used through out the paper, we will abuse the notation and write the operations of warp composition\footnote{\label{foot:warp}Further details regarding composition, $\mathbf{p} \circ \Delta \mathbf{p}$, and inversion, $\Delta \mathbf{p}^{-1}$, of typical AAMs' motion models such as PWA and TPS warps can be found in \cite{Matthews2004, Papandreou2008}.} $\mathcal{W}(\mathbf{x}; \mathbf{p}) \circ \mathcal{W}(\mathbf{x}; \Delta\mathbf{p})$ and inversion\footnoteref{foot:warp} $\mathcal{W}(\mathbf{x}; \mathbf{p})^{-1}$ as simply $\mathbf{p} \circ \Delta \mathbf{p}$ and $\mathbf{p}^{-1}$ respectively.

\subsubsection{Forward}
\label{sec:forward}

In the forward compositional framework the incremental warp $\Delta \mathbf{p}$ is introduced on the image side at each iteration by composing it with the current warp estimate $\mathbf{p}$:
\begin{equation}
    \begin{aligned}
        \Delta \mathbf{p}^* & = \underset{\Delta \mathbf{p}} {\mathrm{arg\, min\;}} \frac{1}{2}|| \mathbf{i}[\mathbf{p} \circ \Delta \mathbf{p}] - \mathbf{a} ||^2
    \label{eq:ssd_fc}
    \end{aligned}
\end{equation}

Once the optimal values for the parameters of the incremental warp are obtained, the current warp estimate is updated according to the following compositional update rule:
\begin{equation}
 	\begin{aligned}
    	\mathbf{p} \leftarrow \mathbf{p} \circ \Delta \mathbf{p}
    \label{eq:fc_update}
    \end{aligned}
\end{equation}

\subsubsection{Inverse}
\label{sec:inverse}

On the other hand, the inverse compositional framework inverts the roles of the image and the model by introducing the incremental warp on the model side:
\begin{equation}
    \begin{aligned}
        \Delta \mathbf{p}^* & = \underset{\Delta \mathbf{p}} {\mathrm{arg\, min\;}} \frac{1}{2}|| \mathbf{i}[\mathbf{p}] - \mathbf{a} [\Delta \mathbf{p}] ||^2
    \label{eq:ssd_ic}
    \end{aligned}
\end{equation}
Note that, in this case, the model is the one we seek to deform using the incremental warp.

Because the incremental warp is introduced on the model side, the solution $\Delta \mathbf{p}$ needs to be inverted before it is composed with the current warp estimate:
\begin{equation}
 	\begin{aligned}
    	\mathbf{p} \leftarrow \mathbf{p} \circ \Delta {\mathbf{p}}^{-1} 
    \label{eq:ic_update}
    \end{aligned}
\end{equation}

\subsubsection{Asymmetric}
\label{sec:asymmetric}

Asymmetric composition introduces two related incremental warps onto the cost function; one on the image side (forward) and the other on the model side (inverse): 
\begin{equation}
    \begin{aligned}
        \Delta \mathbf{p}^* & = \underset{\Delta \mathbf{p}} {\mathrm{arg\, min\;}} \frac{1}{2}|| \mathbf{i}[\mathbf{p} \circ \alpha \Delta \mathbf{p}] - \mathbf{a} [\beta \Delta \mathbf{p}^{-1}] ||^2
    \label{eq:ssd_ac}
    \end{aligned}
\end{equation}
Note that the previous two incremental warps are defined to be each others inverse. Consequently, using the first order approximation to warp inversion for typical AAMs warps $\Delta\mathbf{p}^{-1} = -\Delta\mathbf{p}$ defined in \cite{Matthews2004}, we can rewrite the previous asymmetric cost function as:
\begin{equation}
    \begin{aligned}
        \Delta \mathbf{p}^* & = \underset{\Delta \mathbf{p}} {\mathrm{arg\, min\;}} \frac{1}{2}|| \mathbf{i}[\mathbf{p} \circ \alpha \Delta \mathbf{p}] - \mathbf{a} [-\beta \Delta \mathbf{p} ||^2
    \label{eq:ssd_ac2}
    \end{aligned}
\end{equation}
Although this cost function will need to be linearized around both incremental warps, the parameters $\Delta \mathbf{p}$ controlling both warps are the same. Also, note that the parameters $\alpha \in [0, 1]$ and $\beta=(1-\alpha)$ control the relative contribution of both incremental warps in the computation of the optimal value for $\Delta \mathbf{p}$. 

In this case, the update rule for the current warp estimate is obtained by combining the previous forward and inverse compositional update rules into a single compositional update rule:
\begin{equation}
 	\begin{aligned}
    	\mathbf{p} & \leftarrow \mathbf{p} \circ \alpha \Delta \mathbf{p} \circ \beta \Delta \mathbf{p}
    \label{eq:ac_update}
    \end{aligned}
\end{equation}

Note that, the special case in which $\alpha = \beta = 0.5$ is also referred to as \emph{symmetric} composition \cite{Megret2008, Autheserre2009, Megret2010} and that the previous forward and inverse compositions can also be obtained from asymmetric composition by setting $\alpha = 1$ , $\beta = 0$ and $\alpha = 0$ , $\beta = 1$ respectively.

\subsubsection{Bidirectional}
\label{sec:bidirectional}

Similar to the previous asymmetric composition, bidirectional composition also introduces incremental warps on both image and model sides. However, in this case, the two incremental warps are assumed to be independent from each other:
\begin{equation}
    \begin{aligned}
        \Delta \mathbf{p}^*, \Delta \mathbf{q}^*  & = \underset{\Delta \mathbf{p}, \Delta \mathbf{q}} {\mathrm{arg\, min\;}} \frac{1}{2}|| \mathbf{i}[\mathbf{p}\circ \Delta \mathbf{p}] - \mathbf{a} [\Delta \mathbf{q}] ||^2
    \label{eq:ssd_bc}
    \end{aligned}
\end{equation}
Consequently, in Step \ref{it:step_4}, the cost function needs to be linearized around both incremental warps and solved with respect to the parameters controlling both warps, $\Delta \mathbf{p}$ and $\Delta \mathbf{q}$. 

Once the optimal value for both sets of parameters is recovered, the current estimate of the warp is updated using:
\begin{equation}
 	\begin{aligned}
    	\mathbf{p} \leftarrow \mathbf{p} \circ \Delta \mathbf{p} \circ \Delta {\mathbf{q}}^{-1} 
    \label{eq:bc_update}
    \end{aligned}
\end{equation}

\subsection{Optimization Method}
\label{sec:optimization}

Step \ref{it:step_2} and \ref{it:step_3} in CGD algorithms, i.e. linearizing the cost and solving for the incremental warp respectively, depend on the specific optimization method used by the algorithm. In this paper, we distinguish between three main optimization methods\footnote{Amberg et al. proposed the use of the \emph{Steepest Descent} method \cite{Boyd2004} in \cite{Amberg2009}. However, their approach requires a special formulation of the motion model and it performs poorly using the standard independent AAM formulation \cite{Matthews2004} used in this work.}:
\begin{inparaenum}[\itshape i\upshape)]
    \item \emph{Gauss-Newton} \cite{Boyd2004, Matthews2004, Gross2005, Martins2010, Papandreou2008, Tzimiropoulos2013};
    \item \emph{Newton} \cite{Boyd2004, Kossaifi2014}; and
    \item \emph{Wiberg} \cite{Okatani2006, Strelow2012, Papandreou2008, Tzimiropoulos2013}.
\end{inparaenum}

These methods can be used to iteratively solve the non-linear optimization problems defined by Equations \ref{eq:prob_rssd} and \ref{eq:prob_po}. The main differences between them are:
\begin{enumerate}
    \item The term being linearized. Gauss-Newton and Wiberg linearize the residual $\mathbf{r}$ while Newton linearizes the whole data term $\mathcal{D}$.

    \item The way in which each method solves for the incremental parameters $\Delta \mathbf{c}$, $\Delta \mathbf{p}$ and $\Delta \mathbf{q}$. Gauss-Newton and Newton can either solve for them \emph{simultaneously} or in an \emph{alternated} fashion while Wiberg defines its own procedure to solve for different sets of parameters\footnote{Wiberg reduces to Gauss-Newton when only a single set of parameters needs to be inferred.}.
\end{enumerate}

The following subsections thoroughly explain how the previous optimization methods are used in CGD algorithms. In order to simplify their comprehension full derivations will be given for all methods using the SSD data term (Equation \ref{eq:ssd}) with both asymmetric (Section \ref{sec:asymmetric}) and bidirectional (Section \ref{sec:bidirectional}) compositions\footnote{These represent the most general cases because the derivations for forward, inverse and symmetric compositions can be directly obtained from the asymmetric one and they require solving for both shape and appearance parameters.} while only direct solutions will be given for the Project-Out data term (Equation \ref{eq:po}).

\subsubsection{Gauss-Newton}
\label{sec:gauss_newton}

When \emph{asymmetric} composition is used, the optimization problem defined by the SSD data term is:
\begin{equation}
    \begin{aligned}
        \Delta \mathbf{c}^*, \Delta \mathbf{p}^* & = \underset{\Delta \mathbf{c}, \Delta \mathbf{p}}{\mathrm{arg\,min\;}} \frac{1}{2} \mathbf{r}_a^T\mathbf{r}_a
    \label{eq:asymmetric_ssd}
    \end{aligned}
\end{equation}
with the asymmetric residual $\mathbf{r}_a$ defined as:
\begin{equation}
    \begin{aligned}
		\mathbf{r}_a & = \mathbf{i}[\mathbf{p} \circ \alpha \Delta \mathbf{p}] - (\mathbf{a} + \mathbf{A}(\mathbf{c} + \Delta\mathbf{c})) [\beta \Delta \mathbf{p}^{-1}]
    \label{eq:asymmetric_residual}
    \end{aligned}
\end{equation}
and where we have introduced the incremental appearance parameters $\Delta\mathbf{c}$\footnote{The value of the current estimate of appearance parameters is updated at each iteration using the following additive update rule: $\mathbf{c} \leftarrow \mathbf{c} + \Delta \mathbf{c}$}. 
The Gauss-Newton method solves the previous optimization problem by performing a \emph{first} order Taylor expansion of the residual:
\begin{equation}
    \begin{aligned}
		\mathbf{r}_a(\Delta \boldsymbol{\ell}) & \approx \hat{\mathbf{r}}_a(\Delta \boldsymbol{\ell})
		\\
		& \approx \mathbf{r}_a + \frac{\partial \mathbf{r}_a}{\partial \Delta \boldsymbol{\ell}} \Delta \boldsymbol{\ell}
    \label{eq:asymmetric_residual_taylor}
    \end{aligned}
\end{equation}
and solving the following approximation of the original problem:
\begin{equation}
    \begin{aligned}
        \Delta \boldsymbol{\ell}^* & = \underset{\Delta \boldsymbol{\ell}}{\mathrm{arg\,min\;}} \frac{1}{2} \hat{\mathbf{r}}_a^T\hat{\mathbf{r}}_a
    \label{eq:asymmetric_ssd_taylor}
    \end{aligned}
\end{equation}
where, in order to unclutter the notation, we have defined $\Delta \boldsymbol{\ell} = (\Delta \mathbf{c}^T, \Delta \mathbf{p}^T)^T$ and the partial derivative of the residual with respect to the previous parameters, i.e. the \emph{Jacobian} of the residual, is defined as:
\begin{equation}
    \begin{aligned}
		\frac{\partial \mathbf{r}_a}{\partial \Delta \boldsymbol{\ell}}& = \left( \frac{\partial \mathbf{r}_a}{\partial \Delta \mathbf{c}}, \frac{\partial \mathbf{r}_a}{\partial \Delta \mathbf{p}} \right)
		\\
		& = \left( -\mathbf{A}, \nabla \mathbf{t} \frac{\partial \mathcal{W}}{\partial \Delta \mathbf{p}} \right)
		\\
		& = \left( -\mathbf{A}, \mathbf{J}_\mathbf{t} \right)
    \label{eq:asymmetric_jacobian}
    \end{aligned}
\end{equation}
where $\nabla\mathbf{t} = \left( \alpha \nabla \mathbf{i}[\mathbf{p}]  + \beta \nabla (\mathbf{a} + \mathbf{A}\mathbf{c}) \right)$.

When \emph{bidirectional} composition is used, the optimization problem is defined as:
\begin{equation}
    \begin{aligned}
        \Delta \mathbf{c}^*, \Delta \mathbf{p}^*, \Delta \mathbf{q}^*& = \underset{\Delta \mathbf{c}, \Delta \mathbf{p}, \Delta \mathbf{q}}{\mathrm{arg\,min\;}} \frac{1}{2} \mathbf{r}_b^T\mathbf{r}_b
    \label{eq:bidirectional_ssd}
    \end{aligned}
\end{equation}
where the bidirectional residual $\mathbf{r}_b$ reduces to:
\begin{equation}
    \begin{aligned}
		\mathbf{r}_b & = \mathbf{i}[\mathbf{p} \circ \Delta \mathbf{p}] - (\mathbf{a} + \mathbf{A}(\mathbf{c} + \Delta\mathbf{c})) [\Delta \mathbf{q}]
    \label{eq:bidirectional_residual}
    \end{aligned}
\end{equation}
The Gauss-Newton method proceeds in exactly the same manner as before, i.e. performing a first order Taylor expansion:
\begin{equation}
    \begin{aligned}
		\mathbf{r}_b(\Delta \boldsymbol{\ell}) & \approx \hat{\mathbf{r}}_b(\Delta \boldsymbol{\ell})
		\\
		& \approx \mathbf{r}_b + \frac{\partial \mathbf{r}_b}{\partial \Delta \boldsymbol{\ell}} \Delta \boldsymbol{\ell}
    \label{eq:idirectional_residual_taylor}
    \end{aligned}
\end{equation}
and solving the approximated problem:
\begin{equation}
    \begin{aligned}
        \Delta \boldsymbol{\ell}^* & = \underset{\Delta \boldsymbol{\ell}}{\mathrm{arg\,min\;}} \frac{1}{2} \hat{\mathbf{r}}_b^T\hat{\mathbf{r}}_b
    \label{eq:bidirectional_ssd_taylor}
    \end{aligned}
\end{equation}
where, in this case, $\Delta \boldsymbol{\ell} = (\Delta \mathbf{c}^T, \Delta \mathbf{p}^T, \Delta \mathbf{q}^T)^T$ and the Jacobian of the residual is defined as:
\begin{equation}
    \begin{aligned}
		\frac{\partial \mathbf{r}_b}{\partial \Delta \boldsymbol{\ell}}& = \left( \frac{\partial \mathbf{r}_b}{\partial \Delta \mathbf{c}}, \frac{\partial \mathbf{r}_b}{\partial \Delta \mathbf{p}}, \frac{\partial \mathbf{r}_b}{\partial \Delta \mathbf{q}} \right)
		\\
		& = \left( -\mathbf{A}, \mathbf{J}_{\mathbf{i}}, -\mathbf{J}_{\mathbf{a}} \right)
    \label{eq:bidirectional_jacobian}
    \end{aligned}
\end{equation}
where $\mathbf{J}_\mathbf{i} = \nabla \mathbf{i}[\mathbf{p}] \frac{\partial \mathcal{W}}{\partial \Delta \mathbf{p}}$ and $\mathbf{J}_\mathbf{a} = \nabla (\mathbf{a} + \mathbf{A}\mathbf{c}) \frac{\partial \mathcal{W}}{\partial \Delta \mathbf{q}}.$

\subsubsection*{Simultaneous}
\label{sec:gauss_newton_simultaneous}

The optimization problem defined by Equations \ref{eq:asymmetric_ssd_taylor}  and \ref{eq:bidirectional_ssd_taylor} can be solved with respect to all parameters simultaneously by simply equating their derivative to $0$:
\begin{equation}
    \begin{aligned}
		0 & = \frac{\partial\frac{1}{2}\hat{\mathbf{r}}^T \hat{\mathbf{r}}}{\partial \Delta \boldsymbol{\ell}}
		\\
		& = \frac{\partial\frac{1}{2}(\mathbf{r} + \frac{\partial \mathbf{r}}{\partial \Delta \boldsymbol{\ell}} \Delta \boldsymbol{\ell})^T(\mathbf{r} + \frac{\partial \mathbf{r}}{\partial \Delta \boldsymbol{\ell}} \Delta \boldsymbol{\ell})}{\partial \Delta \boldsymbol{\ell}}
		\\
		& = \left( \mathbf{r} + \frac{\partial \mathbf{r}}{\partial \Delta \boldsymbol{\ell}} \Delta \boldsymbol{\ell} \right) \frac{\partial \mathbf{r}}{\partial \Delta \boldsymbol{\ell}}^T
    \label{eq:ssd_bc}
    \end{aligned}
\end{equation}
The solution is given by:
\begin{equation}
    \begin{aligned}
		\Delta \boldsymbol{\ell}^* & =  -\left( \frac{\partial \mathbf{r}}{\partial \Delta \boldsymbol{\ell}}^T \frac{\partial \mathbf{r}}{\partial \Delta \boldsymbol{\ell}} \right)^{-1} \frac{\partial \mathbf{r}}{\partial \Delta \boldsymbol{\ell}}^T \mathbf{r}
    \label{eq:sim_solution}
    \end{aligned}
\end{equation}
where $\left( \frac{\partial \mathbf{r}}{\partial \Delta \boldsymbol{\ell}}^T \frac{\partial \mathbf{r}}{\partial \Delta \boldsymbol{\ell}} \right)$ is known as the Gauss-Newton approximation to the \emph{Hessian} matrix.

Directly inverting $\left( \frac{\partial \mathbf{r}}{\partial \Delta \boldsymbol{\ell}}^T \frac{\partial \mathbf{r}}{\partial \Delta \boldsymbol{\ell}} \right)$ has complexity\footnote{\label{foot:complexity}$m$ and $n$ denote the number of shape and appearance parameters respectively while $F$ denotes the number of pixels on the reference frame.} $O((n + m)^3)$ for asymmetric composition and $O((2n + m)^3)$ for bidirectional composition. However, one can take advantage of the problem structure and derive an algorithm with smaller complexity by using the \emph{Schur complement}\footnote{
Applying the Schur complement to the following system of equations:
\begin{equation*}
    \begin{aligned}
        \mathbf{A} \mathbf{x} + \mathbf{B} \mathbf{y} = \mathbf{a}
        \\
        \mathbf{C} \mathbf{x} + \mathbf{D} \mathbf{y} = \mathbf{b}
    \label{eq:schur_system}
    \end{aligned}
\end{equation*}
the solution for $\mathbf{x}$ is given by:
\begin{equation*}
    \begin{aligned}
        (\mathbf{A} - \mathbf{B}\mathbf{D}^{-1}\mathbf{C}) \mathbf{x} = \mathbf{a} - \mathbf{B}\mathbf{D}^{-1}\mathbf{b}
    \label{eq:schur_system}
    \end{aligned}
\end{equation*}
and the solution for $\mathbf{y}$ is obtained by substituting the value of $\mathbf{x}$ into the original system.}
\cite{Boyd2004}.

For \emph{asymmetric} composition we have:
\begin{equation}
    \begin{aligned}
    	-\left( \frac{\partial \mathbf{r}_a}{\partial \Delta \boldsymbol{\ell}}^T \frac{\partial \mathbf{r}_a}{\partial \Delta \boldsymbol{\ell}} \right) \Delta \boldsymbol{\ell} & = \frac{\partial \mathbf{r}_a}{\partial \Delta \boldsymbol{\ell}}^T \mathbf{r}
    	\\
        \begin{pmatrix}
            -\underbrace{\mathbf{A}^T \mathbf{A}}_{\mathbf{I}} & \mathbf{A}^T \mathbf{J}_{\mathbf{t}}
            \\
            \mathbf{J}_\mathbf{t}^T \mathbf{A} & -\mathbf{J}_{\mathbf{t}}^T \mathbf{J}_{\mathbf{t}}
        \end{pmatrix}
        \begin{pmatrix}
            \Delta\mathbf{c}
            \\
            \Delta\mathbf{p}
        \end{pmatrix}
        & =
        \begin{pmatrix}
            -\mathbf{A}^T
            \\
            \mathbf{J}_{\mathbf{t}}^T
        \end{pmatrix} \mathbf{r}_a
    \label{eq:asymmetric_structure}
    \end{aligned}
\end{equation}
Applying the Schur complement, the solution for $\Delta\mathbf{p}$ is given by:
\begin{equation}
    \begin{aligned}
        -(\mathbf{J}_{\mathbf{t}}^T\mathbf{J}_{\mathbf{t}} + \mathbf{J}_{\mathbf{t}}^T\mathbf{A}\mathbf{A}^T\mathbf{J}_{\mathbf{t}}^T) \Delta \mathbf{p} & = \mathbf{J}_{\mathbf{t}}^T\mathbf{r} - \mathbf{J}_{\mathbf{t}}^T\mathbf{A} \mathbf{A}^T \mathbf{r}_a
        \\
        -\mathbf{J}_{\mathbf{t}}^T(\mathbf{I} - \mathbf{A} \mathbf{A}^T)\mathbf{J}_{\mathbf{t}} \Delta \mathbf{p} & = \mathbf{J}_{\mathbf{t}}^T(\mathbf{I} - \mathbf{A} \mathbf{A}^T)\mathbf{r}_a
        \\
        -\mathbf{J}_{\mathbf{t}}^T\bar{\mathbf{A}}\mathbf{J}_{\mathbf{t}} \Delta \mathbf{p} & = \mathbf{J}_{\mathbf{t}}^T\bar{\mathbf{A}}\mathbf{r}_a
        \\
        \Delta \mathbf{p} & = -\left( \mathbf{J}_{\mathbf{t}}^T\bar{\mathbf{A}}\mathbf{J}_{\mathbf{t}} \right)^{-1} \mathbf{J}_{\mathbf{t}}^T\bar{\mathbf{A}}\mathbf{r}_a
    \label{eq:asymmetric_schur_solution1}
    \end{aligned}
\end{equation}
and plugging the solution for $\Delta\mathbf{p}$ into equation \ref{eq:asymmetric_structure} the optimal value for $\Delta\mathbf{c}$ is obtained by:
\begin{equation}
    \begin{aligned}
        -\Delta \mathbf{c} + \mathbf{A}^T\mathbf{J}_{\mathbf{t}}\Delta\mathbf{p}^* & = -\mathbf{A}^T \mathbf{r}_a
        \\
        \Delta \mathbf{c} & = \mathbf{A}^T \left( \mathbf{r}_a + \mathbf{J}_{\mathbf{t}} \Delta\mathbf{p}^* \right)
    \label{eq:asymmetric_schur_solution2}
    \end{aligned}
\end{equation}
Using the above procedure the complexity\footnoteref{foot:complexity} of solving each Gauss-Newton step is reduced to:
\begin{equation}
    \begin{aligned}
        O(
        \underbrace{nmF}_{\mathbf{J}_{\mathbf{t}}^T\bar{\mathbf{A}}}
        +
        \underbrace{n^2F + n^3}_{\left( \mathbf{J}_{\mathbf{t}}^T\bar{\mathbf{A}}\mathbf{J}_{\mathbf{t}} \right)^{-1}}
        )
    \label{eq:complexity_schur_asymmetric}
    \end{aligned}
\end{equation}

Using \emph{bidirectional} composition, we can apply the Schur complement either one or two times in order to take advantage of the $3\times3$ block structure of the matrix $\left( \frac{\partial \mathbf{r}_b}{\partial \Delta \boldsymbol{\ell}}^T \frac{\partial \mathbf{r}_b}{\partial \Delta \boldsymbol{\ell}} \right)$:
\begin{equation}
    \begin{aligned}
    	-\left( \frac{\partial \mathbf{r}_b}{\partial \Delta \boldsymbol{\ell}}^T \frac{\partial \mathbf{r}_b}{\partial \Delta \boldsymbol{\ell}} \right) \Delta \boldsymbol{\ell} & = \frac{\partial \mathbf{r}_b}{\partial \Delta \boldsymbol{\ell}}^T \mathbf{r}_b
    	\\
        \left(\begin{array}{c|cc}
            -\underbrace{\mathbf{A}^T \mathbf{A}}_{\mathbf{I}} & \mathbf{A}^T \mathbf{J}_{\mathbf{i}} & -\mathbf{A}^T \mathbf{J}_{\mathbf{a}}
            \\ \hline
            \mathbf{J}_{\mathbf{i}}^T \mathbf{A} & -\mathbf{J}_{\mathbf{i}}^T \mathbf{J}_{\mathbf{i}} & \mathbf{J}_{\mathbf{i}}^T \mathbf{J}_{\mathbf{a}}
            \\
            -\mathbf{J}_{\mathbf{a}}^T \mathbf{A} & \mathbf{J}_{\mathbf{a}}^T \mathbf{J}_{\mathbf{i}} & -\mathbf{J}_{\mathbf{a}}^T \mathbf{J}_{\mathbf{a}}
        \end{array} \right)
        \begin{pmatrix}
            \Delta\mathbf{c}
            \\ \hline
            \Delta\mathbf{p}
            \\
            \Delta\mathbf{q}
        \end{pmatrix}
        & =
        \begin{pmatrix}
            -\mathbf{A}^T
            \\ \hline
            \mathbf{J}_{\mathbf{i}}^T
            \\
            -\mathbf{J}_{\mathbf{a}}^T
        \end{pmatrix} \mathbf{r}_b
    \label{eq:bidirectional_structure}
    \end{aligned}
\end{equation}
Applying the Schur complement once, the combined solution for $(\Delta\mathbf{p}^T, \Delta\mathbf{q}^T)^T$ is given by:
\begin{equation}
    \begin{aligned}
        \begin{pmatrix}
            -\mathbf{J}_{\mathbf{i}}^T \bar{\mathbf{A}}\mathbf{J}_{\mathbf{i}} & \mathbf{J}_{\mathbf{i}}^T \bar{\mathbf{A}}\mathbf{J}_{\mathbf{a}}
            \\
            \mathbf{J}_{\mathbf{a}}^T \bar{\mathbf{A}}\mathbf{J}_{\mathbf{i}} & -\mathbf{J}_{\mathbf{a}}^T \bar{\mathbf{A}}\mathbf{J}_{\mathbf{a}}
            \\
        \end{pmatrix}
        \begin{pmatrix}
            \Delta\mathbf{p}
            \\
            \Delta\mathbf{q}
        \end{pmatrix} & =
        \begin{pmatrix}
            \mathbf{J}_{\mathbf{i}}^T\bar{\mathbf{A}}
            \\
            -\mathbf{J}_{\mathbf{a}}^T\bar{\mathbf{A}}
        \end{pmatrix} \mathbf{r}_b
       \\
       \begin{pmatrix}
            \Delta\mathbf{p}
            \\
            \Delta\mathbf{q}
        \end{pmatrix} & =
        \begin{pmatrix}
            -\mathbf{J}_{\mathbf{i}}^T \bar{\mathbf{A}}\mathbf{J}_{\mathbf{i}} & \mathbf{J}_{\mathbf{i}}^T \bar{\mathbf{A}}\mathbf{J}_{\mathbf{a}}
            \\
            \mathbf{J}_{\mathbf{a}}^T \bar{\mathbf{A}}\mathbf{J}_{\mathbf{i}} & -\mathbf{J}_{\mathbf{a}}^T \bar{\mathbf{A}}\mathbf{J}_{\mathbf{a}}
            \\
        \end{pmatrix}^{-1}
        \\
        & \quad \,
        \begin{pmatrix}
            \mathbf{J}_{\mathbf{i}}^T\bar{\mathbf{A}}
            \\
            -\mathbf{J}_{\mathbf{a}}^T\bar{\mathbf{A}}
        \end{pmatrix} \mathbf{r}_b
    \label{eq:bidirectional_schur_solution1}
    \end{aligned}
\end{equation}
Note that the complexity of inverting this new approximation to the Hessian matrix is $O((2n)^3)$\footnote{This is an important reduction in complexity because usually $m >> n$ in CGD algorithms.}. Similar to before, plugging the solutions for $\Delta\mathbf{p}$ and $\Delta\mathbf{q}$ into Equation \ref{eq:bidirectional_structure} we can infer the optimal value for $\Delta\mathbf{c}$ using:
\begin{equation}
    \begin{aligned}
        \Delta\mathbf{c} & = \mathbf{A}^T \left( \mathbf{r}_b - \mathbf{J}_{\mathbf{i}} \Delta\mathbf{p} + \mathbf{J}_{\mathbf{a}} \Delta\mathbf{q} \right)
    \label{eq:bidirectional_schur_solution2}
    \end{aligned}
\end{equation}
The total complexity per iteration of the previous approach is:
\begin{equation}
    \begin{aligned}
        O(
        \underbrace{2nmF}_{
        \begin{pmatrix}
            \mathbf{J}_{\mathbf{i}}^T\bar{\mathbf{A}}
            \\
            -\mathbf{J}_{\mathbf{a}}^T\bar{\mathbf{A}}
        \end{pmatrix}}
	    +
        \underbrace{(2n)^2F + (2n)^3}_{
        \begin{pmatrix}
            -\mathbf{J}_{\mathbf{i}}^T \bar{\mathbf{A}}\mathbf{J}_{\mathbf{i}} & \mathbf{J}_{\mathbf{i}}^T \bar{\mathbf{A}}\mathbf{J}_{\mathbf{a}}
            \\
            \mathbf{J}_{\mathbf{a}}^T \bar{\mathbf{A}}\mathbf{J}_{\mathbf{i}} & -\mathbf{J}_{\mathbf{a}}^T \bar{\mathbf{A}}\mathbf{J}_{\mathbf{a}}
        \end{pmatrix}^{-1}}
	    )
    \label{eq:complexity_schur_bidirectional1}
    \end{aligned}
\end{equation}

The Schur complement can be re-applied to Equation \ref{eq:bidirectional_schur_solution1} to derive a solution for $\Delta\mathbf{q}$ that only requires inverting a Hessian approximation matrix of size $n \times n$:
\begin{equation}
    \begin{aligned}
        \left( \mathbf{J}_{\mathbf{a}}^T\mathbf{P}\mathbf{J}_{\mathbf{a}} \right) \Delta \mathbf{q} & = \mathbf{J}_{\mathbf{a}}^T\mathbf{P}\mathbf{r}_b
        \\
        \Delta \mathbf{q} & = \left( \mathbf{J}_{\mathbf{a}}^T\mathbf{P}\mathbf{J}_{\mathbf{a}} \right)^{-1} \mathbf{J}_{\mathbf{a}}^T\mathbf{P}\mathbf{r}_b
    \label{eq:bidirectional_schur_solution3}
    \end{aligned}
\end{equation}
where we have defined the projection matrix $\mathbf{P}$ as:
\begin{equation}
    \begin{aligned}
		\mathbf{P} &= \bar{\mathbf{A}} - \bar{\mathbf{A}}\mathbf{J}_{\mathbf{i}} \left( \mathbf{J}_{\mathbf{i}}^T\bar{\mathbf{A}}\mathbf{J}_{\mathbf{i}} \right)^{-1} \mathbf{J}_{\mathbf{i}}^T\bar{\mathbf{A}}
		\label{eq:bidirectional_schur_projection}
    \end{aligned}
\end{equation}
and the solutions for $\Delta\mathbf{p}$ and $\Delta\mathbf{c}$ can be obtained by plugging the solutions for $\Delta\mathbf{q}$ into Equation \ref{eq:bidirectional_schur_solution1} and the solutions for $\Delta\mathbf{q}$ and $\Delta\mathbf{p}$ into Equation \ref{eq:bidirectional_structure} respectively:
\begin{equation}
    \begin{aligned}
        \Delta \mathbf{p} & = -\left( \mathbf{J}_{\mathbf{i}}^T\bar{\mathbf{A}}\mathbf{J}_{\mathbf{i}} \right)^{-1} \mathbf{J}_{\mathbf{i}}^T\bar{\mathbf{A}} \left(\mathbf{r}_b - \mathbf{J}_{\mathbf{a}}\Delta \mathbf{q} \right)
        \\
        \Delta\mathbf{c} & = \mathbf{A}^T \left( \mathbf{r}_b + \mathbf{J}_{\mathbf{i}} \Delta\mathbf{p} - \mathbf{J}_{\mathbf{a}} \Delta\mathbf{q} \right)
    \label{eq:bidirectional_schur_solution4}
    \end{aligned}
\end{equation}
The total complexity per iteration of the previous approach reduces to:
\begin{equation}
    \begin{aligned}
        O(
        \underbrace{2nmF}_{
            \mathbf{J}_{\mathbf{a}}^T\mathbf{P}
            \, \& \,
            \mathbf{J}_{\mathbf{i}}^T\bar{\mathbf{A}}}
        +
        \underbrace{2n^2F + 2n^3}_{
            \left( \mathbf{J}_{\mathbf{a}}^T\mathbf{P}\mathbf{J}_{\mathbf{a}} \right)^{-1}
            \, \& \,
            \left( \mathbf{J}_{\mathbf{i}}^T \bar{\mathbf{A}} \mathbf{J}_{\mathbf{i}} \right)^{-1}}
        )
    \label{eq:complexity_schur_bidirectional2}
    \end{aligned}
\end{equation}
Note that because of their reduced complexity, the solutions defined by Equations \ref{eq:bidirectional_schur_solution3} and \ref{eq:bidirectional_schur_solution4} are preferred over the ones defined by Equations \ref{eq:bidirectional_schur_solution1} and \ref{eq:bidirectional_schur_solution2}.

Finally, the solutions using the Project-Out cost function are:
\begin{itemize}
	\item For \emph{asymmetric} composition:
	\begin{equation}
	    \begin{aligned}
	        \Delta \mathbf{p} & = -\left( \mathbf{J}_{\mathbf{t}}^T\bar{\mathbf{A}}\mathbf{J}_{\mathbf{t}} \right)^{-1} \mathbf{J}_{\mathbf{t}}^T\bar{\mathbf{A}}\mathbf{r}
	    \label{eq:asymmetric_schur_po_solution}
	    \end{aligned}
	\end{equation}
	with complexity\footnote{\label{foot:ic}In practice, the solutions for the Project-Out cost function can be computed slightly faster than those for the SSD because they do not need to explicitly solve for $\Delta\mathbf{c}$. This is specially important in the \emph{inverse} compositional case because expressions of the form $(\mathbf{J}^T\mathbf{U}\mathbf{J})^{-1}\mathbf{J}^T\mathbf{U}$ can be completely precomputed and the computational cost per iteration reduces to $O(nF)$.} given by Equation \ref{eq:complexity_schur_asymmetric}.

	\item For \emph{bidirectional} composition:
	\begin{equation}
	    \begin{aligned}
	        \Delta \mathbf{q} & = \left( \mathbf{J}_{\bar{\mathbf{a}}}^T\mathbf{P}\mathbf{J}_{\bar{\mathbf{a}}} \right)^{-1} \mathbf{J}_{\bar{\mathbf{a}}}^T\mathbf{P}\mathbf{r}
	        \\
	        \Delta \mathbf{p} & = -\left( \mathbf{J}_{\mathbf{i}}^T\bar{\mathbf{A}}\mathbf{J}_{\mathbf{i}} \right)^{-1} \mathbf{J}_\mathbf{i}^T \bar{\mathbf{A}} \left( \mathbf{r} - \mathbf{J}_{\mathbf{a}} \Delta\mathbf{q} \right)
	    \label{eq:bidirectional_schur_po_solution}
	    \end{aligned}
	\end{equation}
	with complexity\footnoteref{foot:ic} given by Equation \ref{eq:complexity_schur_bidirectional2}.
\end{itemize}
where, in both cases, $\mathbf{r} = \mathbf{i}[\mathbf{p}] - \bar{\mathbf{a}}$.

\subsubsection*{Alternated}
\label{sec:gauss_newton_alternated}

Another way of solving optimization problems with two or more sets of variables is to use alternated optimization \cite{DelaTorre2012}.
Hence, instead of solving the previous problem simultaneously with respect to all parameters, we can update one set of parameters at a time while keeping the other sets fixed.

More specifically, using \emph{asymmetric} composition we can alternate between updating $\Delta\mathbf{c}$ given the previous $\Delta\mathbf{p}$ and then update $\Delta\mathbf{p}$ given the updated $\Delta\mathbf{c}$ in an alternate manner. Taking advantage of the structure of the problem defined by Equation \ref{eq:asymmetric_structure}, we can obtain the following system of equations:
\begin{equation}
    \begin{aligned}
        -\Delta\mathbf{c} + \mathbf{A}^T \mathbf{J}_{\mathbf{t}} \Delta\mathbf{p} & = -\mathbf{A}^T \mathbf{r}_a
        \\
        \mathbf{J}_{\mathbf{t}}^T \mathbf{A} \Delta\mathbf{c} - \mathbf{J}_{\mathbf{t}}^T \mathbf{J}_{\mathbf{t}} \Delta\mathbf{p} & = \mathbf{J}_\mathbf{t}^T \mathbf{r}_a
    \label{eq:asymmetric_alt_system}
    \end{aligned}
\end{equation}
which we can rewrite as:
\begin{equation}
    \begin{aligned}
        \Delta\mathbf{c} & = \mathbf{A}^T \left( \mathbf{r}_a + \mathbf{J}_{\mathbf{t}} \Delta\mathbf{p} \right)
        \\
        \Delta\mathbf{p}& = -\left(\mathbf{J}_{\mathbf{t}}^T\mathbf{J}_{\mathbf{t}}\right)^{-1} \mathbf{J}_{\mathbf{t}}^T \left( \mathbf{r}_a - \mathbf{A} \Delta\mathbf{c} \right)
        \label{eq:asymmetric_alt_solution}
    \end{aligned}
\end{equation}
in order to obtain the analytical expression for the previous alternated update rules. The complexity at each iteration is dominated by:
\begin{equation}
    \begin{aligned}
        O(\underbrace{n^2F + n^3}_{(\mathbf{J}_{\mathbf{t}}^T\mathbf{J}_{\mathbf{t}})^{-1}})
    \label{eq:complexity_alternated_asymmetric}
    \end{aligned}
\end{equation}

In the case of \emph{bidirectional} composition we can proceed in two different ways:
\begin{inparaenum}[\itshape a\upshape)]
\item update $\Delta\mathbf{c}$ given the previous $\Delta\mathbf{p}$ and $\Delta\mathbf{q}$ and then update $(\Delta\mathbf{p}^T, \Delta\mathbf{q}^T)^T$ from the updated $\Delta\mathbf{c}$, or
\item update $\Delta\mathbf{c}$ given the previous $\Delta\mathbf{p}$ and $\Delta\mathbf{q}$, then  $\Delta\mathbf{p}$ given the updated $\Delta\mathbf{c}$ and the previous $\Delta\mathbf{q}$ and, finally, $\Delta\mathbf{q}$ given the updated $\Delta\mathbf{c}$ and $\Delta\mathbf{p}$.
\end{inparaenum}

From Equation \ref{eq:bidirectional_structure}, we can derive the following system of equations:
\begin{equation}
    \begin{aligned}
        -\Delta\mathbf{c} + \mathbf{A}^T \mathbf{J}_\mathbf{i} \Delta\mathbf{p} - \mathbf{A}^T \mathbf{J}_\mathbf{a} \Delta\mathbf{q} & = -\mathbf{A}^T \mathbf{r}_b
        \\
        \mathbf{J}_{\mathbf{i}}^T \mathbf{A} \Delta\mathbf{c} - \mathbf{J}_{\mathbf{i}}^T \mathbf{J}_\mathbf{i} \Delta\mathbf{p} + \mathbf{J}_{\mathbf{i}}^T \mathbf{J}_\mathbf{a} \Delta\mathbf{q} & = \mathbf{J}_{\mathbf{i}}^T \mathbf{r}_b
        \\
        -\mathbf{J}_{\mathbf{a}}^T \mathbf{A} \Delta\mathbf{c} + \mathbf{J}_{\mathbf{a}}^T \mathbf{J}_\mathbf{i} \Delta\mathbf{p} - \mathbf{J}_{\mathbf{a}}^T \mathbf{J}_\mathbf{a} \Delta\mathbf{q} & = -\mathbf{J}_{\mathbf{a}}^T \mathbf{r}_b
    \label{eq:bidirectional_alt_system}
    \end{aligned}
\end{equation}
from which we can define the alternated update rules for the first of the previous two options:
\begin{equation}
    \begin{aligned}
        \Delta\mathbf{c}& = \mathbf{A}^T \left( \mathbf{r}_b + \mathbf{J}_{\mathbf{i}} \Delta\mathbf{p} - \mathbf{J}_{\mathbf{a}} \Delta\mathbf{q} \right)
        \\
        \begin{pmatrix}
            \Delta\mathbf{p}
            \\
            \Delta\mathbf{q}
        \end{pmatrix} & =
        \begin{pmatrix}
            -\mathbf{J}_{\mathbf{i}}^T \mathbf{J}_{\mathbf{i}} & \mathbf{J}_{\mathbf{i}}^T \mathbf{J}_{\mathbf{a}}
            \\
            \mathbf{J}_{\mathbf{a}}^T \mathbf{J}_{\mathbf{i}} & -\mathbf{J}_{\mathbf{a}}^T \mathbf{J}_{\mathbf{a}}
            \\
        \end{pmatrix}^{-1}
        \begin{pmatrix}
            \mathbf{J}_{\mathbf{i}}^T
            \\
            -\mathbf{J}_{\mathbf{a}}^T
        \end{pmatrix} \left(\mathbf{r}_b - \mathbf{A} \Delta\mathbf{c}\right)
        \label{eq:bidirectional_alt_solution1}
    \end{aligned}
\end{equation}
with complexity:
\begin{equation}
    \begin{aligned}
        O(\underbrace{(2n)^2F + (2n)^3}_{
        \begin{pmatrix}
            -\mathbf{J}_{\mathbf{i}}^T \mathbf{J}_{\mathbf{i}} & \mathbf{J}_{\mathbf{i}}^T \mathbf{J}_{\mathbf{a}}
            \\
            \mathbf{J}_{\mathbf{a}}^T \mathbf{J}_{\mathbf{i}} & -\mathbf{J}_{\mathbf{a}}^T \mathbf{J}_{\mathbf{a}}
            \\
        \end{pmatrix}^{-1}
        })
    \label{eq:complexity_alternated_bidirectional1}
    \end{aligned}
\end{equation}
The rules for the second options are:
\begin{equation}
    \begin{aligned}
        \Delta\mathbf{c} & = \mathbf{A}^T \left( \mathbf{r}_b + \mathbf{J}_{\mathbf{i}} \Delta\mathbf{p} - \mathbf{J}_{\mathbf{a}} \Delta\mathbf{q} \right)
        \\
        \Delta\mathbf{p} & = -(\mathbf{J}_{\mathbf{i}}^T\mathbf{J}_{\mathbf{i}})^{-1} \mathbf{J}_{\mathbf{i}}^T \left( \mathbf{r}_b - \mathbf{A} \Delta\mathbf{c} - \mathbf{J}_{\mathbf{a}} \Delta\mathbf{q} \right)
        \\
        \Delta\mathbf{q} & = (\mathbf{J}_{\mathbf{a}}^T\mathbf{J}_{\mathbf{a}})^{-1} \mathbf{J}_{\mathbf{a}}^T \left( \mathbf{r}_b - \mathbf{A} \Delta\mathbf{c} + \mathbf{J}_{\mathbf{i}} \Delta\mathbf{p} \right)
        \label{eq:bidirectional_alt_solution2}
    \end{aligned}
\end{equation}
and their complexity is dominated by:
\begin{equation}
    \begin{aligned}
        O(\underbrace{2n^2F + 2n^3}_{(\mathbf{J}_{\mathbf{i}}^T\mathbf{J}_{\mathbf{i}})^{-1} \, \& \, (\mathbf{J}_{\mathbf{a}}^T\mathbf{J}_{\mathbf{a}})^{-1}})
    \label{eq:complexity_alternated_bidirectional2}
    \end{aligned}
\end{equation}

On the other hand, the alternated update rules using the Project-Out cost function are:
\begin{itemize}
	\item For \emph{asymmetric} composition:
	There is no proper alternated rule because the Project-Out cost function only depends on one set of parameters, $\Delta \mathbf{p}$.
	\item For \emph{bidirectional} composition:
	\begin{equation}
	    \begin{aligned}
	        \Delta \mathbf{q} & = \left( \mathbf{J}_{\bar{\mathbf{a}}}^T\bar{\mathbf{A}}\mathbf{J}_{\bar{\mathbf{a}}} \right)^{-1} \mathbf{J}_{\bar{\mathbf{a}}}^T\bar{\mathbf{A}} \left( \mathbf{r} + \mathbf{J}_{\mathbf{i}} \Delta\mathbf{p} \right)
	        \\
	        \Delta \mathbf{p} & = -\left( \mathbf{J}_{\mathbf{i}}^T\bar{\mathbf{A}}\mathbf{J}_{\mathbf{i}} \right)^{-1} \mathbf{J}_\mathbf{i}^T \bar{\mathbf{A}} \left( \mathbf{r} - \mathbf{J}_{\mathbf{a}} \Delta\mathbf{q} \right)
	    \label{eq:bidirectional_alternated_po_solution}
	    \end{aligned}
	\end{equation}
	with equivalent complexity to the one given by Equation \ref{eq:complexity_schur_asymmetric} because, in this case, the term $\left( \mathbf{J}_{\bar{\mathbf{a}}}^T \bar{\mathbf{A}} \mathbf{J}_{\bar{\mathbf{a}}} \right)^{-1} \mathbf{J}_{\bar{\mathbf{a}}}^T \bar{\mathbf{A}}$ can be completely precomputed.
\end{itemize}

Note that all previous alternated update rules, Equations \ref{eq:asymmetric_alt_solution}, \ref{eq:bidirectional_alt_solution1}, \ref{eq:bidirectional_alt_solution2} and \ref{eq:bidirectional_alternated_po_solution}, are similar but slightly different from their simultaneous counterparts, Equations \ref{eq:asymmetric_schur_solution1} and \ref{eq:asymmetric_schur_solution2}, \ref{eq:bidirectional_schur_solution1} and \ref{eq:bidirectional_schur_solution2}, \ref{eq:bidirectional_schur_solution3} and \ref{eq:bidirectional_schur_solution4}, and \ref{eq:bidirectional_schur_po_solution}.

\subsubsection{Newton}
\label{sec:newton}

The Newton method performs a \emph{second} order Taylor expansion of the entire data term $\mathcal{D}$:
\begin{equation}
    \begin{aligned}
		\mathcal{D}(\Delta \boldsymbol{\ell}) & \approx \hat{\mathcal{D}}(\Delta \boldsymbol{\ell})
		\\
		& \approx \mathcal{D} + \frac{\partial \mathcal{D}}{\partial \Delta \boldsymbol{\ell}} \Delta \boldsymbol{\ell} + \frac{1}{2}\Delta \boldsymbol{\ell}^T \frac{\partial^2 \mathcal{D}}{\partial^2 \Delta \boldsymbol{\ell}} \Delta \boldsymbol{\ell}
    \label{eq:newton_taylor}
    \end{aligned}
\end{equation}
and solves the approximate problem:
\begin{equation}
    \begin{aligned}
        \Delta \boldsymbol{\ell}^* & = \underset{\Delta \boldsymbol{\ell}}{\mathrm{arg\,min\;}} \hat{\mathcal{D}}
    \label{eq:newton_optimization_problem}
    \end{aligned}
\end{equation}

Assuming \emph{asymmetric} composition, the previous data term is defined as:
\begin{equation}
    \begin{aligned}
		\mathcal{D}_a(\Delta \boldsymbol{\ell}) & = \frac{1}{2}\mathbf{r}_a^T \mathbf{r}_a
    \label{eq:asymmetric_data}
    \end{aligned}
\end{equation}
and the matrix containing the first order partial derivatives with respect to the parameters, i.e. the data term's \emph{Jacobian}, can be written as:
\begin{equation}
    \begin{aligned}
		\frac{\partial \mathcal{D}_a}{\partial \Delta \boldsymbol{\ell}} & = \left( \frac{\partial \mathcal{D}_a}{\partial \Delta \mathbf{c}}, \frac{\partial \mathcal{D}_a}{\partial \Delta \mathbf{p}} \right)
		\\
		& = \left( -\mathbf{A}^T \mathbf{r}_a, \mathbf{J}_{\mathbf{t}}^T \mathbf{r}_a \right)
    \label{eq:asymmetric_newton_jacobian}
    \end{aligned}
\end{equation}
On the other hand, the matrix $\frac{\partial^2 \mathcal{D}_a}{\partial^2 \Delta \boldsymbol{\ell}}$ of the second order partial derivatives, i.e. the \emph{Hessian} of the data term, takes the following form:
\begin{equation}
    \begin{aligned}
		\frac{\partial^2 \mathcal{D}_a}{\partial^2 \Delta \boldsymbol{\ell}} & =
		\begin{pmatrix}
			\frac{\partial^2 \mathcal{D}_a}{\partial^2 \Delta \mathbf{c}} & \frac{\partial^2 \mathcal{D}_a}{\partial \Delta \mathbf{c} \partial \Delta \mathbf{p}}
			\\
			\frac{\partial^2 \mathcal{D}_a}{\partial \Delta \mathbf{p} \partial \Delta \mathbf{c}} & \frac{\partial^2 \mathcal{D}_a}{\partial^2 \Delta \mathbf{p}}
		\end{pmatrix}
        \\
        & =
        \begin{pmatrix}
            \frac{\partial^2 \mathcal{D}_a}{\partial^2 \Delta \mathbf{c}} & \frac{\partial^2 \mathcal{D}_a}{\partial \Delta \mathbf{c} \partial \Delta \mathbf{p}}
            \\
            \left(\frac{\partial^2 \mathcal{D}_a}{\partial \Delta \mathbf{c} \partial \Delta \mathbf{p}}\right)^T & \frac{\partial^2 \mathcal{D}_a}{\partial^2 \Delta \mathbf{p}}
        \end{pmatrix}
    \label{eq:asymmetric_newton_hessian}
    \end{aligned}
\end{equation}
Note that the Hessian matrix is, by definition, symmetric. The definition of its individual terms is provided in Appendix \ref{sec:app11}.

A similar derivation can be obtained for \emph{bidirectional} composition where, as expected, the data term is defined as:
\begin{equation}
    \begin{aligned}
		\mathcal{D}_b(\Delta \boldsymbol{\ell}) & = \frac{1}{2}\mathbf{r}_b^T \mathbf{r}_b
    \label{eq:bidirectional_data}
    \end{aligned}
\end{equation}
In this case, the Jacobian matrix becomes:
\begin{equation}
    \begin{aligned}
		\frac{\partial \mathcal{D}_b}{\partial \Delta \boldsymbol{\ell}} & = \left( \frac{\partial \mathcal{D}_b}{\partial \Delta \mathbf{c}}, \frac{\partial \mathcal{D}_b}{\partial \Delta \mathbf{p}}, \frac{\partial \mathcal{D}_b}{\partial \Delta \mathbf{q}} \right)
		\\
		& = \left( -\mathbf{A}^T \mathbf{r}_a, \mathbf{J}_{\mathbf{i}}^T \mathbf{r}_a, -\mathbf{J}_{\mathbf{a}}^T \mathbf{r}_a \right)
    \label{eq:bidirectional_newton_jacobian}
    \end{aligned}
\end{equation}
and the Hessian matrix takes the following form:
\begin{equation}
    \begin{aligned}
		\frac{\partial^2 \mathcal{D}_b}{\partial^2 \Delta \boldsymbol{\ell}} & =
		\begin{pmatrix}
			\frac{\partial^2 \mathcal{D}_b}{\partial^2 \Delta \mathbf{c}} & \frac{\partial^2 \mathcal{D}_b}{\partial \Delta \mathbf{c} \partial \Delta \mathbf{p}} & \frac{\partial^2 \mathcal{D}_b}{\partial \Delta \mathbf{c} \partial \Delta \mathbf{q}}
			\\
			\frac{\partial^2 \mathcal{D}_b}{\partial \Delta \mathbf{p} \partial \Delta \mathbf{c}} & \frac{\partial^2 \mathcal{D}_b}{\partial^2 \Delta \mathbf{p}} & \frac{\partial^2 \mathcal{D}_b}{\partial \Delta \mathbf{p} \partial \Delta \mathbf{q}}
			\\
			\frac{\partial^2 \mathcal{D}_b}{\partial \Delta \mathbf{q} \partial \Delta \mathbf{c}} & \frac{\partial^2 \mathcal{D}_b}{\partial \Delta \mathbf{q} \partial \Delta \mathbf{p}}
			& \frac{\partial^2 \mathcal{D}_b}{\partial^2 \Delta \mathbf{q}}    
		\end{pmatrix}
        \\
        & =
        \begin{pmatrix}
            \frac{\partial^2 \mathcal{D}_b}{\partial^2 \Delta \mathbf{c}} & \frac{\partial^2 \mathcal{D}_b}{\partial \Delta \mathbf{c} \partial \Delta \mathbf{p}} & \frac{\partial^2 \mathcal{D}_b}{\partial \Delta \mathbf{c} \partial \Delta \mathbf{q}}
            \\
            \left(\frac{\partial^2 \mathcal{D}_b}{\partial \Delta \mathbf{c} \partial \Delta \mathbf{p}}\right)^T & \frac{\partial^2 \mathcal{D}_b}{\partial^2 \Delta \mathbf{p}} & \frac{\partial^2 \mathcal{D}_b}{\partial \Delta \mathbf{p} \partial \Delta \mathbf{q}}
            \\
            \left(\frac{\partial^2 \mathcal{D}_b}{\partial \Delta \mathbf{c} \partial \Delta \mathbf{q}}\right)^T & \left(\frac{\partial^2 \mathcal{D}_b}{\partial \Delta \mathbf{p} \partial \Delta \mathbf{q}}\right)^T
            & \frac{\partial^2 \mathcal{D}_b}{\partial^2 \Delta \mathbf{q}}
        \end{pmatrix}
    \label{eq:bidirectional_newton_hessian}
    \end{aligned}
\end{equation}
Notice that the previous matrix is again symmetric. The definition of its individual terms is provided in Appendix \ref{sec:app12}.

\subsubsection*{Simultaneous}
\label{sec:newton_simultaneous}

Using the Newton method we can solve for all parameters simultaneously by equating the partial derivative of Equation \ref{eq:newton_optimization_problem} to $0$:
\begin{equation}
    \begin{aligned}
    	0 & = \frac{\partial\hat{\mathcal{D}}}{\partial \Delta \boldsymbol{\ell}}
    	\\
    	& = \frac{\partial \left(\mathcal{D} + \frac{\partial \mathcal{D}}{\partial \Delta \boldsymbol{\ell}} \Delta \boldsymbol{\ell} + \frac{1}{2} \Delta \boldsymbol{\ell}^T \frac{\partial^2 \mathcal{D}}{\partial^2 \Delta \boldsymbol{\ell}} \Delta \boldsymbol{\ell} \right)}{\partial \Delta \boldsymbol{\ell}}
    	\\
		& = \frac{\partial \mathcal{D}}{\partial \Delta \boldsymbol{\ell}} + \frac{\partial^2 \mathcal{D}}{\partial^2 \Delta \boldsymbol{\ell}} \Delta \boldsymbol{\ell}
    \label{eq:ssd_bc}
    \end{aligned}
\end{equation}
with the solution given by:
\begin{equation}
    \begin{aligned}
    	\Delta \boldsymbol{\ell}^* & = -\frac{\partial^2 \mathcal{D}}{\partial^2 \Delta \boldsymbol{\ell}}^{-1} \frac{\partial \mathcal{D}}{\partial \Delta \boldsymbol{\ell}}
    \label{eq:ssd_bc}
    \end{aligned}
\end{equation}

Note that, similar to the Gauss-Newton method, the complexity of inverting the Hessian matrix $\frac{\partial^2 \mathcal{D}}{\partial^2 \Delta \boldsymbol{\ell}}$ is $O((m+2n)^3)$ for asymmetric composition and $O((2n + m)^3)$ for bidirectional composition. As shown by Kossaifi et al. \cite{Kossaifi2014}\footnote{In \cite{Kossaifi2014}, Kossaifi et al. applied the Schur complement to the Newton method using \emph{only} inverse composition while we apply it here using the more general \emph{asymmetric} (which includes \emph{forward}, \emph{inverse} and \emph{symmetric}) and \emph{bidirectional} compositions.}, we can take advantage of the structure of the Hessian in Equations \ref{eq:asymmetric_newton_hessian} and \ref{eq:bidirectional_newton_hessian} and apply the Schur complement to obtain more efficient solutions.

The solutions for $\Delta\mathbf{p}$ and $\Delta\mathbf{c}$ using \emph{asymmetric} composition are given by the following expressions:
\begin{equation}
    \begin{aligned}
        \Delta \mathbf{p} & = \left(\frac{\partial^2 \mathcal{D}_a}{\partial^2 \Delta \mathbf{p}} - \frac{\partial^2 \mathcal{D}_a}{\partial \Delta \mathbf{p} \Delta \mathbf{c}} \frac{\partial^2 \mathcal{D}_a}{\partial \Delta \mathbf{c} \Delta \mathbf{p}} \right)^{-1}
        \\
        & \quad \, \left( \frac{\partial \mathcal{D}_a}{\partial \Delta \mathbf{p}} - \frac{\partial^2 \mathcal{D}_a}{\partial \Delta \mathbf{p} \Delta \mathbf{c}} \frac{\partial \mathcal{D}_a}{\partial \Delta \mathbf{c}} \right)
    	\\
        \Delta \mathbf{c} & = \frac{\partial \mathcal{D}_a}{\partial \Delta \mathbf{c}} - \frac{\partial^2 \mathcal{D}_a}{\partial \Delta \mathbf{c} \partial \Delta \mathbf{p}} \Delta \mathbf{p}^*
    \label{eq:asymmetric_newton_schur_solutions}
    \end{aligned}
\end{equation}
with complexity:
\begin{equation}
    \begin{aligned}
        O(
        \underbrace{nmF}_{\frac{\partial^2 \mathcal{D}_a}{\partial \Delta \mathbf{p} \partial \Delta \mathbf{c}}}
        +
        \underbrace{n^2m}_{\frac{\partial^2 \mathcal{D}_a}{\partial \Delta \mathbf{p} \partial \Delta \mathbf{c}} \frac{\partial^2 \mathcal{D}_a}{\partial \Delta \mathbf{c} \partial \Delta \mathbf{p}}}
        +
        \underbrace{2n^2F}_{\frac{\partial^2 \mathcal{D}_a}{\partial^2 \Delta \mathbf{p}}}
        +
        \underbrace{n^3}_{\mathbf{H}^{-1}}
        )
    \label{eq:complexity_sim_asymmetric_newton}
    \end{aligned}
\end{equation}
where we have defined $\mathbf{H} =\left(\frac{\partial^2 \mathcal{D}_a}{\partial^2 \Delta \mathbf{p}} - \frac{\partial^2 \mathcal{D}_a}{\partial \Delta \mathbf{p} \Delta \mathbf{c}} \frac{\partial^2 \mathcal{D}_a}{\partial \Delta \mathbf{c} \Delta \mathbf{p}} \right)^{-1}$ in order to unclutter the notation.

On the other hand, the solutions for \emph{bidirectional} composition are given either by:
\begin{equation}
    \begin{aligned}
        \begin{pmatrix}
            \Delta\mathbf{p}
            \\
            \Delta\mathbf{q}
        \end{pmatrix} & =
        \begin{pmatrix}
            \mathbf{V} & \mathbf{W}^T
            \\
            \mathbf{W} & \mathbf{U}
            \\
        \end{pmatrix}^{-1}
        \begin{pmatrix}
            \mathbf{v}
            \\
            \mathbf{u}
        \end{pmatrix}
        \\
        \Delta \mathbf{c} & = \frac{\partial \mathcal{D}_b}{\partial \Delta \mathbf{c}} - \frac{\partial^2 \mathcal{D}_b}{\partial \Delta \mathbf{c} \partial \Delta \mathbf{p}} \Delta \mathbf{p} - \frac{\partial^2 \mathcal{D}_b}{\partial \Delta \mathbf{c} \partial \Delta \mathbf{q}} \Delta \mathbf{q}
    \label{eq:bidirectional_newton_schur_solutions1}
    \end{aligned}
\end{equation}
or
\begin{equation}
    \begin{aligned}
        \Delta \mathbf{p} & = \left( \mathbf{U} - \mathbf{W} \mathbf{V}^{-1} \mathbf{W}^T \right)^{-1} \left(\mathbf{u} -  \mathbf{W} \mathbf{V}^{-1}\mathbf{v} \right)
    	\\
        \Delta \mathbf{p} & = \mathbf{V}^{-1} \left( \mathbf{v} - \mathbf{W}^T \Delta\mathbf{q}\right)
    	\\
        \Delta \mathbf{c} & = \frac{\partial \mathcal{D}_b}{\partial \Delta \mathbf{c}} - \frac{\partial^2 \mathcal{D}_b}{\partial \Delta \mathbf{c} \partial \Delta \mathbf{p}} \Delta \mathbf{p} - \frac{\partial^2 \mathcal{D}_b}{\partial \Delta \mathbf{c} \partial \Delta \mathbf{q}} \Delta \mathbf{q}
    \label{eq:bidirectional_newton_schur_solutions2}
    \end{aligned}
\end{equation}
where we have defined the following auxiliary matrices
\begin{equation}
    \begin{aligned}
    	\mathbf{V} & = \frac{\partial^2 \mathcal{D}_b}{\partial^2 \Delta \mathbf{p}} - \frac{\partial^2 \mathcal{D}_b}{\partial \Delta \mathbf{p} \Delta \mathbf{c}} \frac{\partial^2 \mathcal{D}_b}{\partial \Delta \mathbf{c} \Delta \mathbf{p}}
    	\\
    	\mathbf{W} & = \frac{\partial^2 \mathcal{D}_b}{\partial \Delta \mathbf{q} \partial \Delta \mathbf{p}} - \frac{\partial^2 \mathcal{D}_b}{\partial \Delta \mathbf{q} \Delta \mathbf{c}} \frac{\partial^2 \mathcal{D}_b}{\partial \Delta \mathbf{c} \Delta \mathbf{p}}
    	\\
    	\mathbf{U} & = \frac{\partial^2 \mathcal{D}_b}{\partial^2 \Delta \mathbf{q}} - \frac{\partial^2 \mathcal{D}_b}{\partial \Delta \mathbf{q} \Delta \mathbf{c}} \frac{\partial^2 \mathcal{D}_b}{\partial \Delta \mathbf{c} \Delta \mathbf{q}}
    \label{eq:auxiliar_matrixes}
    \end{aligned}
\end{equation}
and vectors
\begin{equation}
    \begin{aligned}
    	\mathbf{v} & = \frac{\partial \mathcal{D}_b}{\partial \Delta \mathbf{p}} - \frac{\partial^2 \mathcal{D}_b}{\partial \Delta \mathbf{p} \Delta \mathbf{c}} \frac{\partial \mathcal{D}_b}{\partial \Delta \mathbf{c}}
    	\\
    	\mathbf{u} & = \frac{\partial \mathcal{D}_b}{\partial^2 \Delta \mathbf{q}} - \frac{\partial^2 \mathcal{D}_b}{\partial \Delta \mathbf{q} \Delta \mathbf{c}} \frac{\partial \mathcal{D}_b}{\partial \Delta \mathbf{c}}
    \label{eq:auxiliar_matrixes}
    \end{aligned}
\end{equation}
The complexity of the previous solutions is of:
\begin{equation}
    \begin{aligned}
        O( &
        \underbrace{nmF}_{\mathbf{v}}
        +
        \underbrace{2nmF}_{\mathbf{u}}
        +
        \underbrace{4n^2F + 2n^2m}_{\mathbf{U} \, \& \, \mathbf{V}}
        +
        \\
        &
        \underbrace{2n^2F + n^2m}_{\mathbf{W}}
        +
        \underbrace{(2n)^3}_{
        \begin{pmatrix}
            \mathbf{V} & \mathbf{W}^T
            \\
            \mathbf{W} & \mathbf{U}
            \\
        \end{pmatrix}^{-1}}
        )
    \label{eq:complexity_sim_bidirectional2_newton}
    \end{aligned}
\end{equation}
and
\begin{equation}
    \begin{aligned}
        O( &
        \underbrace{nmF}_{\mathbf{v}}
        +
        \underbrace{2nmF}_{\mathbf{u}}
        +
        \\
        &
        \underbrace{4n^2F + 2n^2m}_{\mathbf{U} \, \& \, \mathbf{V}}
        +
        \underbrace{2n^2F + n^2m}_{\mathbf{W}}
        +
        \\
        &
        \underbrace{4n^3}_{\mathbf{V}^{-1}  \, \& \, \left( \mathbf{U} - \mathbf{W} \mathbf{V}^{-1} \mathbf{W}^T \right)^{-1} }
        )
    \label{eq:complexity_sim_bidirectional2_newton}
    \end{aligned}
\end{equation}
respectively.

The solutions using the Project-Out cost function are:
\begin{itemize}
	\item For \emph{asymmetric} composition:
	\begin{equation}
	    \begin{aligned}
	        \Delta \mathbf{p} & = -\left( \frac{\partial \mathcal{W}}{\Delta \mathbf{p}}^T \nabla^2\mathbf{t} \frac{\partial \mathcal{W}}{\Delta \mathbf{p}}\bar{\mathbf{A}}\mathbf{r} + \mathbf{J}_{\mathbf{t}}^T\bar{\mathbf{A}}\mathbf{J}_{\mathbf{t}} \right)^{-1}
	        \\
	        & \quad \, \mathbf{J}_{\mathbf{t}}^T\bar{\mathbf{A}}\mathbf{r}
	    \label{eq:asymmetric_newton_po_solution}
	    \end{aligned}
	\end{equation}
	with complexity\footnote{\label{foot:ic_newton}In practice, the solutions for the project-out cost function can also be computed slightly faster because they do not need to explicitly solve for $\Delta\mathbf{c}$. However, in this case, using \emph{inverse} composition we can only precompute terms of the form $\mathbf{J}^T\mathbf{U}$ and  $\mathbf{J}^T\mathbf{U}\mathbf{J}$ but not the entire $\mathbf{H}^{-1}\mathbf{J}^T\mathbf{U}$ because of the explicit dependence between $\mathbf{H}$ and the current residual $\mathbf{r}$.} given by Equation \ref{eq:complexity_sim_asymmetric_newton}.

	\item For \emph{bidirectional} composition:
	\begin{equation}
	    \begin{aligned}
	        \Delta \mathbf{q} & = \left( \frac{\partial \mathcal{W}}{\Delta \mathbf{p}}^T \nabla^2\bar{\mathbf{a}} \frac{\partial \mathcal{W}}{\Delta \mathbf{p}}\bar{\mathbf{A}}\mathbf{r} + \mathbf{J}_{\bar{\mathbf{a}}}^T\tilde{\mathbf{P}}\mathbf{J}_{\bar{\mathbf{a}}}\right)^{-1}
	        \\
	        & \quad \, \mathbf{J}_{\bar{\mathbf{a}}}^T\tilde{\mathbf{P}}\mathbf{r}
	        \\
	        \Delta \mathbf{p} & = -\mathbf{H}_\mathbf{i}^{-1} \mathbf{J}_\mathbf{i}^T \bar{\mathbf{A}} \left( \mathbf{r} - \mathbf{J}_{\mathbf{a}} \Delta\mathbf{q} \right)
	    \label{eq:bidirectional_newton_po_solution}
	    \end{aligned}
	\end{equation}
	where the projection operator $\tilde{\mathbf{P}}$ is defined as:
	\begin{equation}
	    \begin{aligned}
			\tilde{\mathbf{P}} &= \bar{\mathbf{A}} - \bar{\mathbf{A}}\mathbf{J}_{\mathbf{i}}^T \mathbf{H}_\mathbf{i}^{-1} \mathbf{J}_{\mathbf{i}}^T\bar{\mathbf{A}}
			\label{eq:bidirectional_projection}
	    \end{aligned}
	\end{equation}
	and where we have defined:
	\begin{equation}
	    \begin{aligned}
			\mathbf{H}_\mathbf{i} = \left( \frac{\partial \mathcal{W}}{\Delta \mathbf{p}}^T \nabla^2\mathbf{i}[\mathbf{p}] \frac{\partial \mathcal{W}}{\Delta \mathbf{p}}\bar{\mathbf{A}}\mathbf{r} + \mathbf{J}_{\mathbf{i}}^T\bar{\mathbf{A}}\mathbf{J}_{\mathbf{i}}^T \right)
	    \end{aligned}
    \end{equation} to unclutter the notation. The complexity per iteration\footnoteref{foot:ic_newton} is given by Equation \ref{eq:complexity_sim_bidirectional2_newton}.
\end{itemize}

Note that, the derivations of the previous solutions, for both types of composition, are analogous to the ones shown in Section \ref{sec:gauss_newton_simultaneous} for the Gauss-Newton method and, consequently, have been omitted here.

\subsubsection*{Alternated}
\label{sec:newton_alternated}

Alternated optimization rules can also be derived for the Newton method following the strategy shown in Section \ref{sec:gauss_newton_alternated} for the Gauss-Newton case. Again, we will simply provide update rules and computational complexity for both types of composition and will omit the details of their full derivation.

For \emph{asymmetric} composition the alternated rules are defined as:
\begin{equation}
    \begin{aligned}
        \Delta \mathbf{c} & = \frac{\partial \mathcal{D}_a}{\partial \Delta \mathbf{c}} - \frac{\partial^2 \mathcal{D}_a}{\partial \Delta \mathbf{c} \partial \Delta \mathbf{p}} \Delta \mathbf{p}
        \\
        \Delta \mathbf{p} & = \frac{\partial^2 \mathcal{D}_a}{\partial^2 \Delta \mathbf{p}}^{-1} \left( \frac{\partial \mathcal{D}_a}{\partial \Delta \mathbf{p}} - \frac{\partial^2 \mathcal{D}_a}{\partial \Delta \mathbf{p} \partial \Delta \mathbf{c}} \Delta \mathbf{c} \right)
        \label{eq:asymmetric_newton_alternated_solution}
    \end{aligned}
\end{equation}
with complexity:
\begin{equation}
    \begin{aligned}
        O(
        \underbrace{nmF}_{\frac{\partial^2 \mathcal{D}_a}{\partial \Delta \mathbf{p} \partial \Delta \mathbf{c}}}
        +
        \underbrace{2n^2F + n^3}_{\frac{\partial^2 \mathcal{D}_a}{\partial^2 \Delta \mathbf{p}}^{-1}}
        )
    \label{eq:complexity_alt_asymmetric_newton}
    \end{aligned}
\end{equation}

The alternated rules for \emph{bidirectional} composition case are given either by:
\begin{equation}
    \begin{aligned}
        \Delta \mathbf{c} & =  \frac{\partial \mathcal{D}_b}{\partial \Delta \mathbf{c}} - \frac{\partial^2 \mathcal{D}_b}{\partial \Delta \mathbf{c} \partial \Delta \mathbf{p}} \Delta \mathbf{p} - \\
        & \quad \, \frac{\partial^2 \mathcal{D}_b}{\partial \Delta \mathbf{c} \partial \Delta \mathbf{q}} \Delta \mathbf{q}
        \\
        \begin{pmatrix}
            \Delta\mathbf{p}
            \\
            \Delta\mathbf{q}
        \end{pmatrix} & =
        \begin{pmatrix}
            \frac{\partial^2 \mathcal{D}_b}{\partial^2 \Delta \mathbf{p}} & \frac{\partial^2 \mathcal{D}_b}{\partial \Delta \mathbf{p} \partial \Delta \mathbf{q}}
            \\
            \frac{\partial^2 \mathcal{D}_b}{\partial \Delta \mathbf{q} \partial \Delta \mathbf{p}} & \frac{\partial^2 \mathcal{D}_b}{\partial^2 \Delta \mathbf{p}}
            \\
        \end{pmatrix}^{-1}
        \\
        & \quad \,
        \begin{pmatrix}
            \frac{\partial \mathcal{D}_b}{\partial \Delta \mathbf{p}} - \frac{\partial^2 \mathcal{D}_b}{\partial \Delta \mathbf{p} \partial \Delta \mathbf{c}} \Delta \mathbf{c}
            \\
            \frac{\partial \mathcal{D}_b}{\partial \Delta \mathbf{q}} - \frac{\partial^2 \mathcal{D}_b}{\partial \Delta \mathbf{q} \partial \Delta \mathbf{c}} \Delta \mathbf{c}
        \end{pmatrix}
        \label{eq:bidirectional_newton_alternated_solution1}
    \end{aligned}
\end{equation}
with complexity:
\begin{equation}
    \begin{aligned}
        O( &
        \underbrace{nmF}_{\frac{\partial^2\mathcal{D}}{\partial\Delta\mathbf{p}\partial\Delta\mathbf{p}}}
        +
        \underbrace{4n^2F}_{\frac{\partial^2\mathcal{D}}{\partial^2\Delta\mathbf{p}} \, \& \, \frac{\partial^2\mathcal{D}}{\partial^2\Delta\mathbf{q}}}
        +
        \\
        &
        \underbrace{(2n)^3}_{
        \begin{pmatrix}
            \frac{\partial^2 \mathcal{D}_b}{\partial^2 \Delta \mathbf{p}} & \frac{\partial^2 \mathcal{D}_b}{\partial \Delta \mathbf{p} \partial \Delta \mathbf{q}}
            \\
            \frac{\partial^2 \mathcal{D}_b}{\partial \Delta \mathbf{q} \partial \Delta \mathbf{p}} & \frac{\partial^2 \mathcal{D}_b}{\partial^2 \Delta \mathbf{p}}
            \\
        \end{pmatrix}^{-1}}
        )
    \label{eq:complexity_alt_bidirectional1_newton}
    \end{aligned}
\end{equation}
or:
\begin{equation}
    \begin{aligned}
        \Delta \mathbf{c} & =  \frac{\partial \mathcal{D}_b}{\partial \Delta \mathbf{c}} - \frac{\partial^2 \mathcal{D}_b}{\partial \Delta \mathbf{c} \partial \Delta \mathbf{p}} \Delta \mathbf{p} - \frac{\partial^2 \mathcal{D}_b}{\partial \Delta \mathbf{c} \partial \Delta \mathbf{q}} \Delta \mathbf{q}
        \\
        \Delta \mathbf{p} & = \frac{\partial^2 \mathcal{D}_b}{\partial^2 \Delta \mathbf{p}}^{-1}
        \\
        & \quad \left( \frac{\partial \mathcal{D}_b}{\partial \Delta \mathbf{p}} - \frac{\partial^2 \mathcal{D}_b}{\partial \Delta \mathbf{p} \partial \Delta \mathbf{c}} \Delta \mathbf{c} - \frac{\partial^2 \mathcal{D}_b}{\partial \Delta \mathbf{p} \partial \Delta \mathbf{q}} \Delta \mathbf{q} \right)
        \\
        \Delta \mathbf{q} & = \frac{\partial^2 \mathcal{D}_b}{\partial^2 \Delta \mathbf{q}}^{-1}
        \\
        & \quad \left( \frac{\partial \mathcal{D}_b}{\partial \Delta \mathbf{q}} - \frac{\partial^2 \mathcal{D}_b}{\partial \Delta \mathbf{q} \partial \Delta \mathbf{c}} \Delta \mathbf{c} - \frac{\partial^2 \mathcal{D}_b}{\partial \Delta \mathbf{q} \partial \Delta \mathbf{p}} \Delta \mathbf{p} \right)
        \label{eq:bidirectional_newton_alternated_solution2}
    \end{aligned}
\end{equation}
with complexity:
\begin{equation}
    \begin{aligned}
        O( &
        \underbrace{nmF}_{\frac{\partial^2\mathcal{D}}{\partial\Delta\mathbf{p}\partial\Delta\mathbf{p}}}
        +
        \underbrace{4n^2F}_{\frac{\partial^2\mathcal{D}}{\partial^2\Delta\mathbf{p}} \, \& \, \frac{\partial^2\mathcal{D}}{\partial^2\Delta\mathbf{q}}}
        +
        \underbrace{2n^3}_{\frac{\partial^2 \mathcal{D}_b}{\partial^2 \Delta \mathbf{p}}^{-1} \, \& \, \frac{\partial^2 \mathcal{D}_b}{\partial^2 \Delta \mathbf{q}}^{-1} }
        )
    \label{eq:complexity_alt_bidirectional2_newton}
    \end{aligned}
\end{equation}

On the other hand, the alternated update rules for the Newton method using the project-out cost function are:
\begin{itemize}
	\item For \emph{asymmetric} composition:
	Again, there is no proper alternated rule because the project-out cost function only depends on one set of parameters, $\Delta \mathbf{p}$.
	\item For \emph{bidirectional} composition:
	\begin{equation}
	    \begin{aligned}
	        \Delta \mathbf{q} & = \mathbf{H}_\mathbf{a}^{-1} \mathbf{J}_{\bar{\mathbf{a}}}^T\bar{\mathbf{A}} \left( \mathbf{r} + \mathbf{J}_{\mathbf{i}} \Delta\mathbf{p} \right)
	        \\
	        \Delta \mathbf{p} & = -\mathbf{H}_\mathbf{i}^{-1} \mathbf{J}_\mathbf{i}^T \bar{\mathbf{A}} \left( \mathbf{r} - \mathbf{J}_{\mathbf{a}} \Delta\mathbf{q} \right)
	    \label{eq:bidirectional_alternated_po_solution}
	    \end{aligned}
	\end{equation}
	where we have defined:
	\begin{equation}
	    \begin{aligned}
	        \mathbf{H}_\mathbf{a} = \left( \frac{\partial \mathcal{W}}{\Delta \mathbf{p}}^T \nabla^2\bar{\mathbf{a}} \frac{\partial \mathcal{W}}{\Delta \mathbf{p}}\bar{\mathbf{A}}\mathbf{r} + \mathbf{J}_{\bar{\mathbf{a}}}^T\bar{\mathbf{A}}\mathbf{J}_{\bar{\mathbf{a}}} \right)
	    \label{eq:po_hessian_q}
	    \end{aligned}
	\end{equation}
	and the complexity at every iteration is given by the following expression complexity:
	\begin{equation}
	    \begin{aligned}
	        O(
	        \underbrace{nmF}_{\mathbf{J}_\mathbf{i}^T \bar{\mathbf{A}}}
	        +
	        \underbrace{3n^2F + 2n^3}_{
                \mathbf{H}_\mathbf{i}^{-1}
                 \, \& \,
                 \mathbf{H}_\mathbf{a}^{-1}
                 }
	        )
	    \label{eq:complexity_po_alt_bidirectional_newton}
	    \end{aligned}
	\end{equation}
\end{itemize}

\subsubsection*{Efficient Second-order Minimization (ESM)}
\label{sec:esm}

Notice that, the \emph{second} order Taylor expanison used by the Newton method means that \emph{Newton} algorithms are second order optimizations algorithms with respect to the incremental warps. However, as shown in the previous section, this property comes at expenses of a significant increase in computational complexity with respect to (\emph{first} order)\emph{Gauss-Newton} algorithms. In this section, we show that the \emph{Asymmetric Gauss-Newton} algorithms derived in Section \ref{sec:gauss_newton} are, in fact, also true \emph{second} order optimization algorithms with respect to the incremental warp $\Delta \mathbf{p}$.

The use of \emph{asymmetric} composition together with the Gauss-Newton method has been proven to naturally lead to Efficient Second order Minimization (ESM) algorithms in the related field of parametric image alignment \cite{Malis2004, Benhimane2004, Megret2008, Megret2010}. Following a similar line of reasoning, we will show that \emph{Asymmetric Gauss-Newton} algorithms for fitting AAMs can also be also interpreted as ESM algorithms. 

In order to show the previous relationship we will make use of the simplified data term\footnote{Notice that similar derivations can also be obtained using the SSD and Project-Out data terms, but we use the simplified one here for clarity.} introduced by Equation \ref{eq:ssd_shape}. Using \emph{forward} composition, the optimization problem defined by:
\begin{equation}
    \begin{aligned}
        \Delta \mathbf{p}^* & = \underset{\Delta \mathbf{p}}{\mathrm{arg\,min\;}} \frac{1}{2} \mathbf{r}_f^T\bar{\mathbf{A}}\mathbf{r}_f
    \label{eq:po_forward}
    \end{aligned}
\end{equation}
where the forward residual $\mathbf{r}_f$ is defined as:
\begin{equation}
    \begin{aligned}
        \mathbf{r}_f & = \mathbf{i}[\mathbf{p} \circ \Delta \mathbf{p}] - \bar{\mathbf{a}}
    \label{eq:po_forward_residual}
    \end{aligned}
\end{equation}
As seen before, Gauss-Newton solves the previous optimization problem by performing a \emph{first} order Taylor expansion of the residual around $\Delta \mathbf{p}$:
\begin{equation}
    \begin{aligned}
        \hat{\mathbf{r}}_f(\Delta\mathbf{p}) & = \mathbf{r}_f + \frac{\partial \mathbf{r}_f}{\partial\Delta \mathbf{p}}\Delta\mathbf{p} + \underbrace{O_{\mathbf{r}_f}(\Delta\mathbf{p}^2)}_{\textrm{remainder}}
        \\
        & = \mathbf{i}[\mathbf{p}] - \bar{\mathbf{a}} + \mathbf{J}_\mathbf{i}\Delta\mathbf{p} + O_{\mathbf{r}_f}(\Delta\mathbf{p}^2)
    \label{eq:po_forward_residual_taylor}
    \end{aligned}
\end{equation}
and solving the following approximation of the original problem:
\begin{equation}
    \begin{aligned}
        \Delta\mathbf{p}^* & = \underset{\Delta\mathbf{p}}{\mathrm{arg\,min\;}} \frac{1}{2} \hat{\mathbf{r}}_f^T\hat{\mathbf{r}}_f
    \label{eq:po_forward_taylor}
    \end{aligned}
\end{equation}

However, note that, instead of performing a first order Taylor expansion, we can also perform a \emph{second} order Taylor expansion of the residual:
\begin{equation}
    \begin{aligned}
        \check{\mathbf{r}}_f(\Delta\mathbf{p}) & = \mathbf{r}_f + \frac{\partial \mathbf{r}_f}{\partial\Delta \mathbf{p}}\Delta\mathbf{p} \,+
        \\
        & \quad \, \frac{1}{2} \Delta\mathbf{p}^T\frac{\partial^2 \mathbf{r}_f}{\partial^2\Delta \mathbf{p}}\Delta\mathbf{p} + O_{\mathbf{r}_f}(\Delta\mathbf{p}^3)
        \\
        & = \mathbf{i}[\mathbf{p}] - \bar{\mathbf{a}} + \mathbf{J}_\mathbf{i}\Delta\mathbf{p} \, +
        \\
        & \quad \, \frac{1}{2} \Delta\mathbf{p}^T\mathbf{H}_\mathbf{i}\Delta\mathbf{p} + O_{\mathbf{r}_f}(\Delta\mathbf{p}^3)
    \label{eq:po_forward_residual_taylor2}
    \end{aligned}
\end{equation}
Then, given the second main assumption behind AAMs (Equation \ref{eq:aam_2}) the following approximation must hold:
\begin{equation}
    \begin{aligned}
        \nabla\mathbf{i}[\mathbf{p}] \frac{\partial\mathcal{W}}{\partial\Delta\mathbf{p}} & \approx \nabla\mathbf{a} \frac{\partial\mathcal{W}}{\partial\Delta\mathbf{p}}
        \\
        \mathbf{J}_\mathbf{i} & \approx \mathbf{J}_\mathbf{a}
        \label{eq:jacobian_aproximation}
    \end{aligned}
\end{equation}
and, because the previous $\mathbf{J}_\mathbf{i}$ and $\mathbf{J}_\mathbf{a}$ are functions of $\Delta\mathbf{p}$, we can perform a \emph{first} order Taylor expansion of $\mathbf{J}_\mathbf{i}$ to obtain:
\begin{equation}
    \begin{aligned}
    \mathbf{J}_\mathbf{i} (\Delta\mathbf{p}) & \approx \mathbf{J}_\mathbf{i} + \Delta\mathbf{p}^T \frac{\partial\mathbf{J}_\mathbf{i}}{\partial\Delta\mathbf{p}} + \underbrace{O_{\mathbf{J}_\mathbf{i}}(\Delta\mathbf{p}^2)}_{\textrm{remainder}}
    \\
    & \approx \mathbf{J}_\mathbf{i} + \Delta\mathbf{p}^T \mathbf{H}_\mathbf{i} + O_{\mathbf{J}_\mathbf{i}}(\Delta\mathbf{p}^2)
    \\
    \mathbf{J}_\mathbf{a} & \approx  \mathbf{J}_\mathbf{i} + \Delta\mathbf{p}^T \mathbf{H}_\mathbf{i} + O_{\mathbf{J}_\mathbf{i}}(\Delta\mathbf{p}^2)
    \\
    \Delta\mathbf{p}^T \mathbf{H}_\mathbf{i} & \approx \mathbf{J}_\mathbf{a} - \mathbf{J}_\mathbf{i} - O_{\mathbf{J}_\mathbf{i}}(\Delta\mathbf{p}^2)
    \label{eq:jacobian_taylor}
    \end{aligned}
\end{equation}

Finally, substituting the previous approximation for $\Delta\mathbf{p}^T \mathbf{H}_\mathbf{i}$ into Equation \ref{eq:po_forward_residual_taylor2} we arrive at:
\begin{equation}
    \begin{aligned}
        \check{\mathbf{r}}_f(\Delta\mathbf{p}) & = \mathbf{i}[\mathbf{p}] - \bar{\mathbf{a}} + \mathbf{J}_\mathbf{i}\Delta\mathbf{p} \, +
        \\
        & \quad \, \frac{1}{2} \Delta\mathbf{p}^T\mathbf{H}_\mathbf{i}\Delta\mathbf{p} + O_{\mathbf{r}_f}(\Delta\mathbf{p}^3)
        \\
        & = \mathbf{i}[\mathbf{p}] - \bar{\mathbf{a}} + \mathbf{J}_\mathbf{i}\Delta\mathbf{p} \, +
        \\
        & \quad \, \frac{1}{2} \left( \mathbf{J}_\mathbf{a} - \mathbf{J}_\mathbf{i} - O_{\mathbf{J}_\mathbf{i}}(\Delta\mathbf{p}^2) \right) \Delta\mathbf{p} \, +
        \\
        & \quad \, O_{\mathbf{r}_f}(\Delta\mathbf{p}^3)
        \\
        & = \mathbf{i}[\mathbf{p}] - \bar{\mathbf{a}} + \frac{1}{2} \left( \mathbf{J}_\mathbf{i} + \mathbf{J}_\mathbf{a} \right) \Delta\mathbf{p} \, +
        \\
        & \quad \, O_{\textrm{total}}(\Delta\mathbf{p}^3)
    \label{eq:po_asymmetric_residual_approximation}
    \end{aligned}
\end{equation}
where the total remainder is cubic with respect to $\Delta \mathbf{p}$:
\begin{equation}
    \begin{aligned}
        O_{\textrm{total}}(\Delta\mathbf{p}^3) = O_{\mathbf{r}_f}(\Delta\mathbf{p}^3) - O_{\mathbf{J}_\mathbf{i}}(\Delta\mathbf{p}^2) \Delta\mathbf{p}
    \label{eq:po_asymmetric_residual_approximation}
    \end{aligned}
\end{equation}
The previous expression constitutes a true \emph{second} order approximation of the forward residual $\mathbf{r}_f$ where the term $\frac{1}{2} \left( \mathbf{J}_\mathbf{i} + \mathbf{J}_\mathbf{a} \right)$ is equivalent to the asymmetric Jacobian in Equation \ref{eq:asymmetric_ssd_taylor} when $\alpha = \beta = 0.5$:
\begin{equation}
    \begin{aligned}
        \frac{1}{2} \left( \mathbf{J}_\mathbf{i} + \mathbf{J}_\mathbf{a} \right) & = \left( \frac{1}{2} \mathbf{J}_\mathbf{i} + \frac{1}{2} \mathbf{J}_\mathbf{a} \right)
        \\
        & = \left( \frac{1}{2} \nabla \mathbf{i}[\mathbf{p}] \frac{\partial\mathcal{W}}{\partial\Delta\mathbf{p}} + \frac{1}{2} \nabla \mathbf{a} \frac{\partial\mathcal{W}}{\partial\Delta\mathbf{p}} \right)
        \\
        & = \left( \frac{1}{2} \nabla \mathbf{i}[\mathbf{p}] + \frac{1}{2} \nabla \mathbf{a} \right) \frac{\partial\mathcal{W}}{\partial\Delta\mathbf{p}}
        \\
        & = \left( \nabla \mathbf{t} \right) \frac{\partial\mathcal{W}}{\partial\Delta\mathbf{p}}
        \\
        & = \mathbf{J}_\mathbf{t}
    \label{eq:equivalent_jacobians}
    \end{aligned}
\end{equation}
and, consequently, \emph{Asymmetric Gauss-Newton} algorithms for fitting AAMs can be viewed as ESM algorithms that only require \emph{first} order partial derivatives of the residual and that have the same computational complexity as \emph{first} order algorithms.

\subsubsection{Wiberg}
\label{sec:wiberg}

The idea behind the Wiberg method is similar to the one used by the alternated Gauss-Newton method in Section \ref{sec:gauss_newton_alternated}, i.e. solving for one set of parameters at a time while keeping the other sets fixed. However, Wiberg does so by rewriting the asymmetric $\mathbf{r}_a(\Delta\mathbf{c}, \Delta\mathbf{p})$ and bidirectional $\mathbf{r}_b(\Delta\mathbf{c}, \Delta\mathbf{p},\Delta\mathbf{q})$ residuals as functions that only depend on $\Delta\mathbf{p}$ and $\Delta\mathbf{q}$ respectively.

For \emph{asymmetric} composition, the residual $\bar{\mathbf{r}}_a (\Delta \mathbf{p})$ is defined as follows:
\begin{equation}
    \begin{aligned}
        \bar{\mathbf{r}}_a (\Delta \mathbf{p}) & = \mathbf{r}_a(\bar{\Delta \mathbf{c}}, \Delta \mathbf{p})
        \\
        & = \mathbf{i}[\mathbf{p} \circ \alpha \Delta \mathbf{p}] - (\mathbf{a} + \mathbf{A}(\mathbf{c} + \bar{\Delta \mathbf{c}}_a)) [\beta \Delta\mathbf{p}]
    \label{eq:asymmetric_wiberg_residual}
    \end{aligned}
\end{equation}
where the function $\bar{\Delta \mathbf{c}}_a(\Delta \mathbf{p})$ is obtained by solving for $\Delta\mathbf{c}$ while keeping $\Delta\mathbf{p}$ fixed:
\begin{equation}
    \begin{aligned}
        \bar{\Delta \mathbf{c}}_a(\Delta \mathbf{p}) & = \mathbf{A}^T \mathbf{r}_a
        \label{eq:asymmetric_wiberg_c_function}
    \end{aligned}
\end{equation}
Given the previous residual, the Wiberg method proceeds to define the following optimization problem with respect to $\Delta\mathbf{p}$:
\begin{equation}
    \begin{aligned}
        \Delta\mathbf{p}^* & = \underset{\Delta\mathbf{p}}{\mathrm{arg\,min\;}} \bar{\mathbf{r}}_a^T\bar{\mathbf{r}}_a
    \label{eq:asymmetric_wiberg_problem1}
    \end{aligned}
\end{equation}
which then solves approximately by performing a first order Taylor of the residual around the incremental warp:
\begin{equation}
    \begin{aligned}
        \Delta\mathbf{p}^* & = \underset{\Delta\mathbf{p}}{\mathrm{arg\,min\;}}  \left\| \bar{\mathbf{r}}_a(\Delta\mathbf{p}) + \frac{\partial \bar{\mathbf{r}}_a}{\partial\Delta\mathbf{p}} \Delta\mathbf{p} \right\|^2
    \label{eq:asymmetric_wiberg_problem2}
    \end{aligned}
\end{equation}
In this case, the Jacobian $\frac{\partial \bar{\mathbf{r}}}{\partial \Delta \mathbf{p}}$ can be obtain by direct application of the \emph{chain rule} and it is defined as follows:
\begin{equation}
    \begin{aligned}
        \frac{d \bar{\mathbf{r}}_a}{d \Delta \mathbf{p}} & = \frac{\partial \bar{\mathbf{r}}_a}{\partial \Delta \mathbf{p}} + \frac{\partial \bar{\mathbf{r}}_a}{\partial \bar{\Delta \mathbf{c}}_a} \frac{\partial \bar{\Delta \mathbf{c}}_a}{\partial \Delta \mathbf{p}}
        \\
        & = \mathbf{J}_{\mathbf{t}} - \mathbf{A}\mathbf{A}^T\mathbf{J}_{\mathbf{t}}
        \\
        & = \bar{\mathbf{A}}\mathbf{J}_{\mathbf{t}}
    \label{eq:asymmetric_wiberg_jacobian}
    \end{aligned}
\end{equation}
The solution for $\Delta\mathbf{p}$ is obtained as usual by equating the derivative of \ref{eq:asymmetric_wiberg_problem1} with respect to $\Delta\mathbf{p}$ to 0:
\begin{equation}
    \begin{aligned}
    	\Delta \mathbf{p}^* & = - \left( \left( \bar{\mathbf{A}}\mathbf{J}_{\mathbf{t}} \right)^T \bar{\mathbf{A}} \mathbf{J}_{\mathbf{t}} \right)^{-1} \left( \bar{\mathbf{A}}\mathbf{J}_{\mathbf{t}} \right)^T \bar{\mathbf{r}}_a
    	\\
    	& = - \left( \mathbf{J}_{\mathbf{t}}^T \bar{\mathbf{A}} \mathbf{J}_{\mathbf{t}} \right)^{-1} \mathbf{J}_{\mathbf{t}}^T \bar{\mathbf{A}} \bar{\mathbf{r}}_a
    \label{eq:asymmetric_wiberg_solution}
    \end{aligned}
\end{equation}
where we have used the fact that the matrix $\bar{\mathbf{A}}$ is idempotent\footnote{$\bar{\mathbf{A}}$ is idempotent:
 \begin{equation*}
    \begin{aligned}
    	\bar{\mathbf{A}}\bar{\mathbf{A}} & = \left( \mathbf{I} - \mathbf{A}\mathbf{A}^T \right) \left( \mathbf{I} - \mathbf{A}\mathbf{A}^T \right)
    	\\
    	& = \mathbf{I}^T\mathbf{I} - 2\mathbf{A}\mathbf{A}^T + \mathbf{A}\underbrace{\mathbf{A}^T\mathbf{A}}_{\mathbf{I}}\mathbf{A}^T
    	\\
    	& = \mathbf{I} - 2\mathbf{A}\mathbf{A}^T + \mathbf{A}\mathbf{A}^T
    	\\
    	& = \mathbf{I} - \mathbf{A}\mathbf{A}^T 
        \\
        & =\bar{\mathbf{A}} 
    \label{eq:idempotent_bar_A}
    \end{aligned}
\end{equation*}
}.

Therefore, the Wiberg method solves explicitly, at each iteration,  for $\Delta\mathbf{p}$ using the previous expression and implicitly for $\Delta\mathbf{c}$ (through $\bar{\Delta \mathbf{c}}_a(\Delta \mathbf{p})$) using Equation \ref{eq:asymmetric_wiberg_c_function}. The complexity per iteration of the Wiberg method is the same as the one of the Gauss-Newton method after applying the Schur complement, Equation \ref{eq:complexity_schur_asymmetric}. In fact, note that the Wiberg solution for $\Delta\mathbf{p}$ (Equation \ref{eq:asymmetric_wiberg_solution}) is the same as the one of the Gauss-Newton method after applying the Schur complement, Equation \ref{eq:asymmetric_schur_solution1}; and also note the similarity between the solutions for $\Delta\mathbf{c}$ of both methods, Equations \ref{eq:asymmetric_wiberg_c_function} and \ref{eq:asymmetric_schur_solution2}. Finally, note that, due to the close relation between the \emph{Wiberg} and \emph{Gauss-Newton} methods, \emph{Asymmetric Wiberg} algorithms are also ESM algorithms for fitting AAMs.

On the other hand, for \emph{bidirectional} composition, the residual $\bar{\mathbf{r}}_b (\Delta \mathbf{p})$ is defined as:
\begin{equation}
    \begin{aligned}
        \bar{\mathbf{r}}_b(\Delta\mathbf{q}) & = \mathbf{r}_b(\bar{\Delta \mathbf{c}}_b, \bar{\Delta \mathbf{p}}_b, \Delta\mathbf{q})
        \\
        & = \mathbf{i}[\mathbf{p} \circ \bar{\Delta \mathbf{p}}_b] - (\mathbf{a} - \mathbf{A}(\mathbf{c} + \bar{\Delta \mathbf{c}}_b))[\Delta \mathbf{q}]
    \label{eq:bidirectional_wiberg_residual}
    \end{aligned}
\end{equation}
where, similarly as before, the function $\bar{\Delta \mathbf{c}}_b(\Delta \mathbf{p}, \Delta \mathbf{q})$ is obtained solving for $\Delta \mathbf{c}$ while keeping both $\Delta \mathbf{p}$ and $\Delta \mathbf{q}$ fixed:
\begin{equation}
    \begin{aligned}
        \bar{\Delta \mathbf{c}}_b(\Delta \mathbf{p}, \Delta \mathbf{q}) & = \mathbf{A}^T \mathbf{r}_b
        \label{eq:bidirectional_wiberg_c_function}
    \end{aligned}
\end{equation}
and the function $\bar{\Delta \mathbf{p}}_b(\bar{\Delta \mathbf{c}}_b, \Delta\mathbf{q})$ is obtained by solving for $\Delta\mathbf{p}$ using the Wiberg method while keeping $\Delta \mathbf{q}$ fixed:
\begin{equation}
    \begin{aligned}
        \bar{\Delta \mathbf{p}}_b(\bar{\Delta \mathbf{c}}_b, \Delta\mathbf{q}) & = -\left( \mathbf{J}_{\mathbf{i}}^T \bar{\mathbf{A}} \mathbf{J}_{\mathbf{i}} \right)^{-1} \mathbf{J}_{\mathbf{i}}^T \bar{\mathbf{A}} \bar{\mathbf{r}}_b
        \label{eq:bidirectional_wiberg_p_function}
    \end{aligned}
\end{equation}
At this point, the Wiberg method proceeds to define the following optimization problem with respect to $\Delta\mathbf{q}$:
\begin{equation}
    \begin{aligned}
        \Delta\mathbf{q}^* & = \underset{\Delta\mathbf{q}}{\mathrm{arg\,min\;}} \bar{\mathbf{r}}_b^T\bar{\mathbf{r}}_b
    \label{eq:bidirectional_wiberg_problem1}
    \end{aligned}
\end{equation}
which, as before, then solves approximately by performing a first order Taylor expansion around $\Delta\mathbf{q}$:
\begin{equation}
    \begin{aligned}
        \Delta\mathbf{q}^* & = \underset{\Delta\mathbf{q}}{\mathrm{arg\,min\;}}  \left\| \bar{\mathbf{r}}_b(\Delta\mathbf{q}) + \frac{\partial \bar{\mathbf{r}}_b}{\partial\Delta\mathbf{q}} \Delta\mathbf{q} \right\|^2
    \label{eq:bidirectional_wiberg_problem2}
    \end{aligned}
\end{equation}
In this case, the Jacobian of the residual can also be obtained by direct application of the chain rule and takes the following form:
\begin{equation}
    \begin{aligned}
        \frac{d \bar{\mathbf{r}}_b}{d \Delta \mathbf{q}} & = \frac{\partial \bar{\mathbf{r}}_b}{\partial \Delta \mathbf{q}} + \frac{\partial \bar{\mathbf{r}}_b}{\partial \bar{\Delta \mathbf{p}}_b} \frac{\partial \bar{\Delta \mathbf{p}}_b}{\partial \Delta \mathbf{q}} +
        \\
        & \quad \left( \frac{\partial\bar{\mathbf{r}}_b}{\partial \bar{\Delta \mathbf{c}}_b} + \frac{\partial \bar{\mathbf{r}}_b}{\partial \bar{\Delta \mathbf{p}}_b} \frac{\partial \bar{\Delta \mathbf{p}}_b}{\partial \bar{\Delta \mathbf{c}}} \right) \frac{\partial \bar{\Delta \mathbf{c}}_b}{\partial \Delta \mathbf{q}}
        \\
        & = - \mathbf{J}_{\mathbf{a}} + \mathbf{J}_{\mathbf{i}} \left( \mathbf{J}_i^T \bar{\mathbf{A}} \mathbf{J}_i \right)^{-1} \mathbf{J}_i^T \bar{\mathbf{A}} \mathbf{J}_{\mathbf{a}} +
        \\
        & \quad \, \left(\mathbf{A} - \mathbf{J}_{\mathbf{i}} \left( \mathbf{J}_i^T \bar{\mathbf{A}} \mathbf{J}_i \right)^{-1} \mathbf{J}_i^T \bar{\mathbf{A}} \mathbf{A} \right) \mathbf{A}^T \mathbf{J}_{\mathbf{a}}
        \\
        & = - \mathbf{J}_{\mathbf{a}} + \mathbf{A}\mathbf{A}^T \mathbf{J}_{\mathbf{a}} +
        \\
        & \quad \, \mathbf{J}_{\mathbf{i}} \left( \mathbf{J}_i^T \bar{\mathbf{A}} \mathbf{J}_i \right)^{-1} \mathbf{J}_i^T \bar{\mathbf{A}} \mathbf{J}_{\mathbf{a}} -
        \\
        & \quad \, \mathbf{J}_{\mathbf{i}} \left( \mathbf{J}_i^T \bar{\mathbf{A}} \mathbf{J}_i \right)^{-1} \mathbf{J}_i^T \bar{\mathbf{A}} \mathbf{A} \mathbf{A}^T \mathbf{J}_{\mathbf{a}}
        \\
        & = - \left( \mathbf{I} - \mathbf{A}\mathbf{A}^T \right) \mathbf{J}_{\mathbf{a}} +
        \\
        & \quad \, \mathbf{J}_{\mathbf{i}} \left( \mathbf{J}_i^T \bar{\mathbf{A}} \mathbf{J}_i \right)^{-1} \mathbf{J}_i^T \bar{\mathbf{A}} \left( \mathbf{I} - \mathbf{A}\mathbf{A}^T \right) \mathbf{J}_{\mathbf{a}}
        \\
        & = - \bar{\mathbf{A}} \mathbf{J}_{\mathbf{a}} + \mathbf{J}_{\mathbf{i}} \left( \mathbf{J}_i^T \bar{\mathbf{A}} \mathbf{J}_i \right)^{-1} \mathbf{J}_i^T \bar{\mathbf{A}} \bar{\mathbf{A}} \mathbf{J}_{\mathbf{a}}
        \\
        & = \left( -\mathbf{I} + \mathbf{J}_{\mathbf{i}} \left( \mathbf{J}_i^T \bar{\mathbf{A}} \mathbf{J}_i \right)^{-1} \mathbf{J}_i^T \bar{\mathbf{A}} \right) \bar{\mathbf{A}} \mathbf{J}_{\mathbf{a}}
        \\
        & = - \mathbf{P} \mathbf{J}_{\mathbf{a}}
    \label{eq:bidirectional_wiberg_jacobian}
    \end{aligned}
\end{equation}
 And, again, the solution for $\Delta\mathbf{q}$ is obtained as usual by equating the derivative of \ref{eq:bidirectional_wiberg_problem2} with respect to $\Delta\mathbf{q}$ to 0:
 \begin{equation}
    \begin{aligned}
    	\Delta \mathbf{q}^* & = \left( \left( \mathbf{P} \mathbf{J}_{\mathbf{t}} \right)^T \mathbf{P} \mathbf{J}_{\mathbf{t}} \right)^{-1} \left( \mathbf{P} \mathbf{J}_{\mathbf{t}} \right)^T \bar{\mathbf{r}}_a
    \label{eq:bidirectional_wiberg_solution}
    \end{aligned}
\end{equation}
In this case, the Wiberg method solves explicitly, at each iteration, for $\Delta\mathbf{p}$ using the previous expression and implicitly for $\Delta\mathbf{p}$ and $\Delta\mathbf{c}$ (through $\bar{\Delta \mathbf{p}}_b(\bar{\Delta \mathbf{c}}_b, \Delta\mathbf{q})$ and $\bar{\Delta \mathbf{c}}_b(\Delta \mathbf{p}, \Delta \mathbf{q})$) using Equations \ref{eq:bidirectional_wiberg_p_function} and \ref{eq:bidirectional_wiberg_c_function} respectively. Again, the complexity per iteration is the same as the one of the Gauss-Newton method after applying the Schur complement, Equation \ref{eq:complexity_schur_bidirectional2}; and the solutions for both methods are almost identical, Equations \ref{eq:bidirectional_wiberg_solution}, \ref{eq:bidirectional_wiberg_p_function} and \ref{eq:bidirectional_wiberg_c_function} and Equations \ref{eq:bidirectional_schur_solution1}, \ref{eq:bidirectional_schur_solution2} and \ref{eq:bidirectional_schur_solution3}.

On the other hand, the Wiberg solutions for the project-out cost function are:
\begin{itemize}
	\item For \emph{asymmetric} composition:
	Because the project-out cost function only depends on one set of parameters, $\Delta \mathbf{p}$, in this case Wiberg reduces to Gauss-Newton.

	\item For \emph{bidirectional} composition:
	\begin{equation}
	    \begin{aligned}
	    	\bar{\Delta \mathbf{p}} & = - \left( \mathbf{J}_\mathbf{i}^T \bar{\mathbf{A}} \mathbf{J}_\mathbf{i} \right)^{-1} \mathbf{J}_\mathbf{i}^T \bar{\mathbf{A}} \mathbf{r}
	    	\\
	        \Delta \mathbf{q} & = \left( \mathbf{J}_{\bar{\mathbf{a}}}^T \mathbf{P} \mathbf{J}_{\bar{\mathbf{a}}} \right)^{-1} \mathbf{J}_{\bar{\mathbf{a}}}^T \mathbf{P} \mathbf{r}
	    \label{eq:bidirectional_wiberg_po_solution}
	    \end{aligned}
	\end{equation}
    Again, in this case, the solutions obtained with the Wiberg method are almost identical to the ones obtained using Gauss-Newton after applying the Schur complement, Equation \ref{eq:bidirectional_schur_po_solution}.
\end{itemize}
	

\section{Relation to Prior Work}
\label{sec:relaltion}

In this section we relate relevant prior work on CGD algorithms for fitting AAMs \cite{Matthews2004, Gross2005, Papandreou2008, Amberg2009, Martins2010, Tzimiropoulos2013, Kossaifi2014} to the unified and complete view introduced in the previous Section.

\subsection{Project-Out algorithms}

In their seminal work \cite{Matthews2004}, Matthews and Baker proposed the first CGD algorithm for fitting AAMs, the so-called Project-out Inverse Compositional (PIC) algorithm. This algorithm uses Gauss-Newton to solve the optimization problem posed by the project-out cost function using inverse composition. The use of the project-out norm removes the need to solve for the appearance parameters and the use of inverse composition allows for the precomputation of the pseudo-inverse of the Jacobian with respect to $\Delta\mathbf{p}$, i.e. $\left( \mathbf{J}_{\bar{\mathbf{a}}}^T\bar{\mathbf{A}}\mathbf{J}_{\bar{\mathbf{a}}} \right)^{-1}\mathbf{J}_{\bar{\mathbf{a}}}\bar{\mathbf{A}}$. The PIC algorithm is very efficient ($O(nF)$) but it has been shown to perform poorly in generic and unconstrained scenarios \cite{Gross2005, Papandreou2008}. In this paper, we refer to this algorithm as the \emph{Project-Out Inverse Gauss-Newton} algorithm.

The forward version of the previous algorithm, i.e. the \emph{Project-Out Forward Gauss-Newton} algorithm, was proposed by Amberg et al. in \cite{Amberg2009}. In this case, the use of forward composition prevents the precomputation of the Jacobian pseudo-inverse and its complexity increases to $O(nmF + n^2F + n^3)$. However, this algorithm has been shown to largely outperform its inverse counterpart, and obtains good performance under generic and unconstrained conditions \cite{Amberg2009, Tzimiropoulos2013}\footnote{Notice that, in \cite{Amberg2009}, Amberg et al. also introduced a hybrid forward/inverse algorithm, coined CoLiNe. This algorithm is a compromise between the previous two algorithms in terms of both complexity and accuracy. Due to its rather ad-hoc derivation, this algorithm was not considered in this paper.}

To the best of our knowledge, the rest of Project-Out algorithms derived in Section \ref{sec:fitting}, i.e.:
\begin{itemize}
\item \emph{Project-Out Forward Newton}
\item \emph{Project-Out Inverse Newton}
\item \emph{Project-Out Asymmetric Gauss-Newton}
\item \emph{Project-Out Asymmetric Newton}
\item \emph{Project-Out Bidirectional Gauss-Newton Schur}
\item \emph{Project-Out Bidirectional Gauss-Newton Alternated}
\item \emph{Project-Out Bidirectional Newton Schur}
\item \emph{Project-Out Bidirectional Newton Alternated}
\item \emph{Project-Out Bidirectional Wiberg}
\end{itemize}
have never been published before and are a significant contribution of this work.

\subsection{SSD algorithms}

In \cite{Gross2005} Gross et al. presented the Simultaneous Inverse Compositional (SIC) algorithm and show that it largely outperforms the \emph{Project-Out Inverse Gauss-Newton} algorithm in terms of fitting accuracy. This algorithm uses Gauss-Newton to solve the optimization problem posed by the SSD cost function using inverse composition. In this case, the Jacobian with respect to $\Delta\mathbf{p}$, depends on the current value of the appearance parameters and needs to be recomputed at every iteration. Moreover, the inclusion of the Jacobian with respect to the appearance increments $\delta\mathbf{c}$, increases the size of the simultaneous Jacobian to $\frac{\partial\mathbf{r}}{\partial\Delta\boldsymbol{\ell}} = \left( -\mathbf{A}, -\mathbf{J}_\mathbf{a} \right) \in \mathbb{R}^{F \times (m + n)}$ and, consequently, the computational cost per iteration of the algorithm is $O((m + n)^2F + (m + n)^3)$.

As we shown in Sections \ref{sec:gauss_newton_simultaneous}, \ref{sec:gauss_newton_alternated} and \ref{sec:wiberg} the previous complexity can be dramatically reduced by taking advantage of the problem structure in order to derive more efficient and exact algorithm by:
\begin{inparaenum}[\itshape a\upshape)]
\item applying the Schur complement;
\item adopting an alternated optimization approach; or
\item or using the Wiberg method.
\end{inparaenum}
Papandreou and Maragos \cite{Papandreou2008} proposed an algorithm that is equivalent to the solution obtained by applying the Schur complement to the problem, as described in Section \ref{sec:gauss_newton_simultaneous}. The same algorithm was reintroduced in \cite{Tzimiropoulos2013} using a somehow ad-hoc derivation (reminiscent of the Wiberg method) under the name Fast-SIC. This algorithm has a computational cost per iteration of $O(nmF + n^2F + n^3)$. In this paper, following our unified view on CGD algorithm, we refer to the previous algorithm as the \emph{SSD Inverse Gauss-Newton Schur} algorithm. The alternated optimization approach was used in \cite{Tzimiropoulos2012} and \cite{Antonakos2014} with complexity $O(n^2F + n^3)$ per iteration. We refer to it as the \emph{SSD Inverse Gauss-Newton Alternated} algorithm.

On the other hand, the forward version of the previous algorithm was first proposed by Martins et al. in \cite{Martins2010}\footnote{Note that Martins et al.  used an additive update rule for the shape parameters, $\mathbf{p}^* =  \mathbf{p} + \Delta\mathbf{p}$, so strictly speaking they derived an additive version of the algorithm i.e the \emph{Simultaneous Forward Additive} (SFA) algorithm.}. In this case, the Jacobian with respect to $\Delta\mathbf{p}$ depends on the current value of the shape parameters $\mathbf{p}$ through the warped image $\mathbf{i}[\mathbf{p}]$ and also needs to be recomputed at every iteration. Consequently, the complexity if the algorithm is the same as in the naive inverse approach of Gross et al. In this paper, we refer to this algorithm as the \emph{SSD Forward Gauss-Newton} algorithm. It is important to notice that Tzimiropoulos and Pantic \cite{Tzimiropoulos2013} derived a more efficient version of this algorithm ($O(nmF + n^2F + n^3)$), coined Fast-Forward, by applying the same derivation used to obtain their Fast-SIC algorithm. They showed that in the forward case their derivation removed the need to explicitly solve for the appearance parameters. Their algorithm is equivalent to the previous \emph{Project-Out Forward Gauss-Newton}.

\begin{figure*}[t!]
	\centering
	\begin{subfigure}{0.16\textwidth}
		\includegraphics[width=\textwidth]{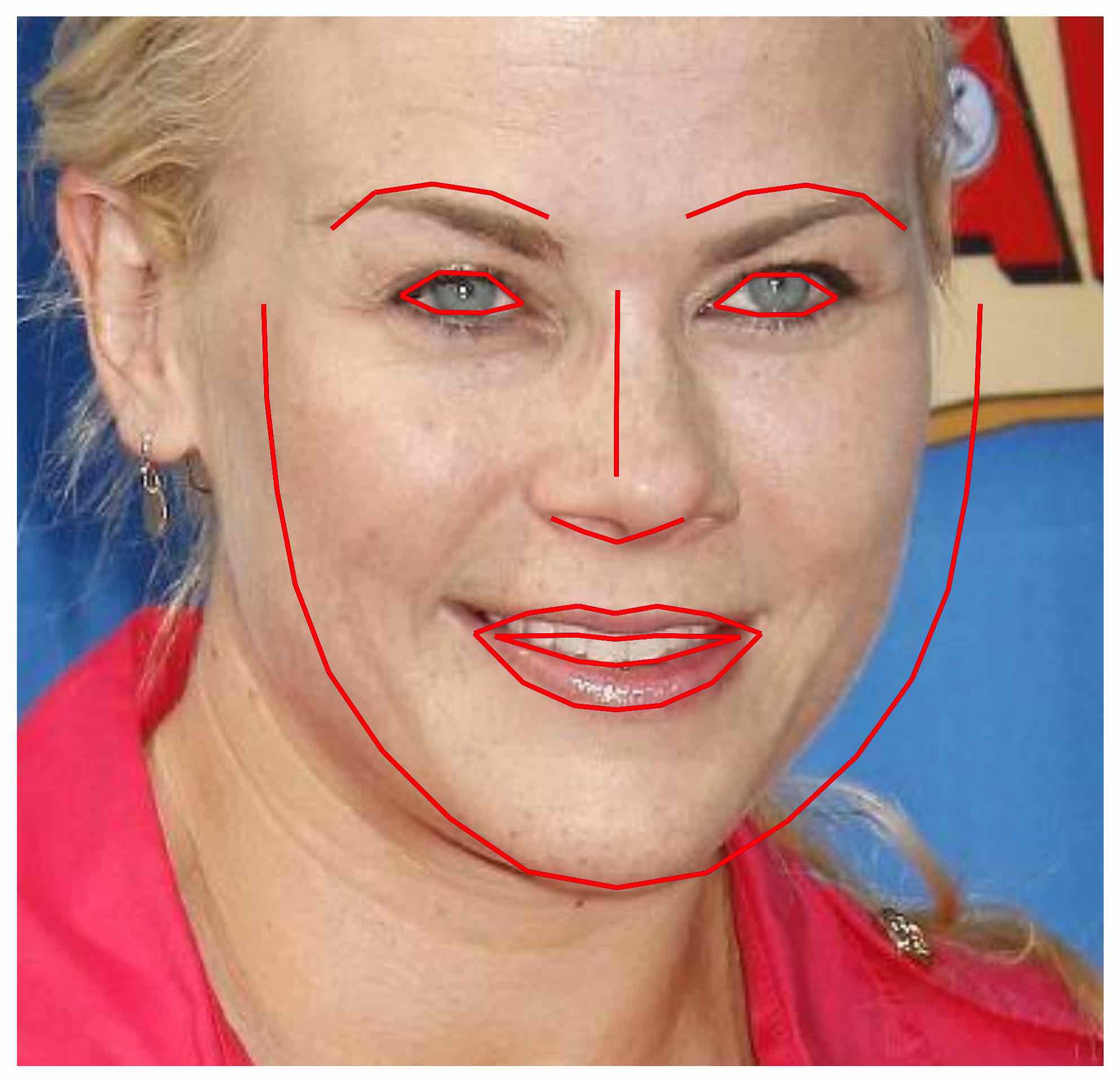}
		\caption{$0\%$}
		\label{fig:ini_0}
	\end{subfigure}
	\begin{subfigure}{0.16\textwidth}
		\includegraphics[width=\textwidth]{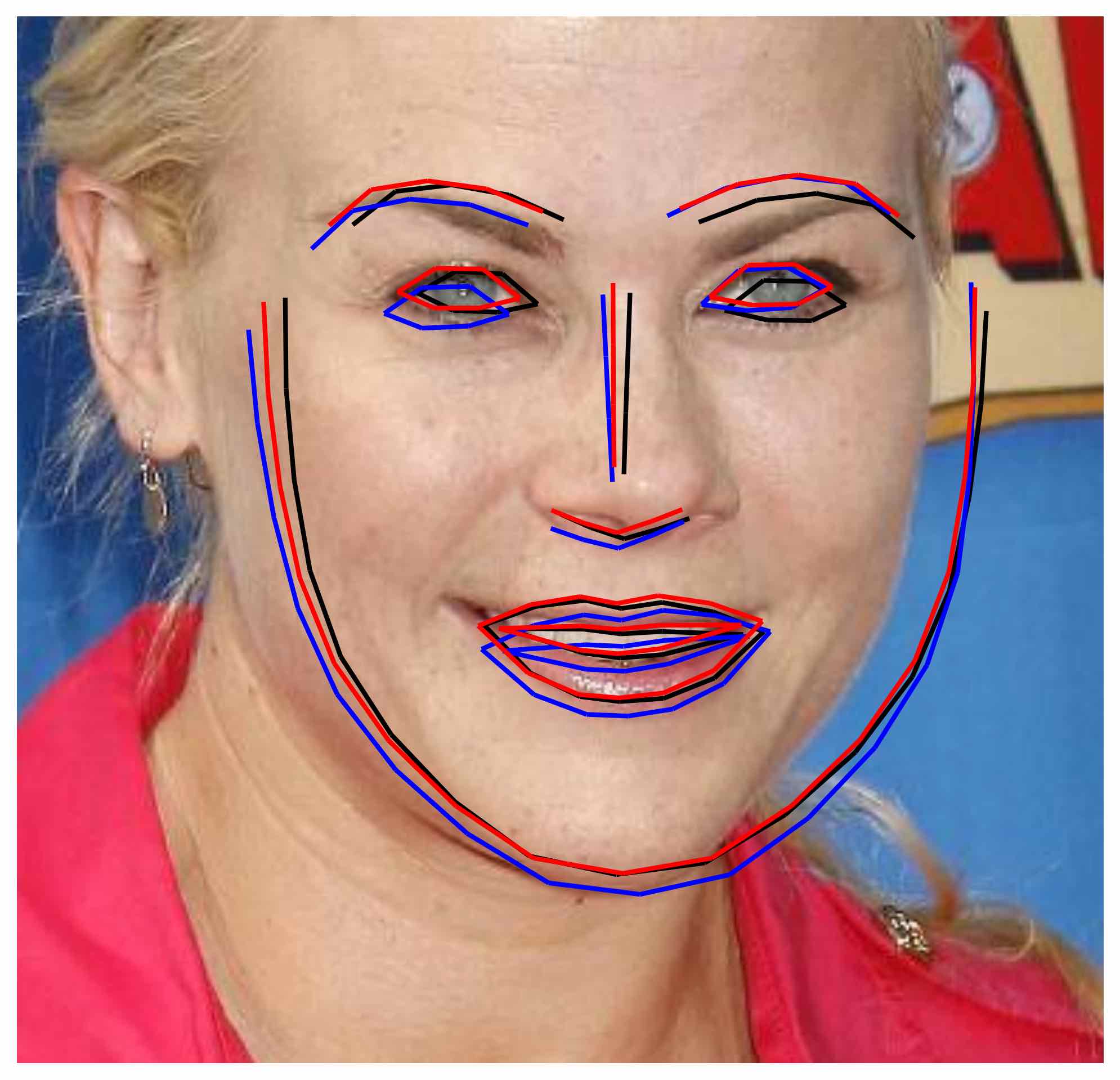}
		\caption{$2.5\%$}
		\label{fig:ini_1}
	\end{subfigure}
	\begin{subfigure}{0.16\textwidth}
		\includegraphics[width=\textwidth]{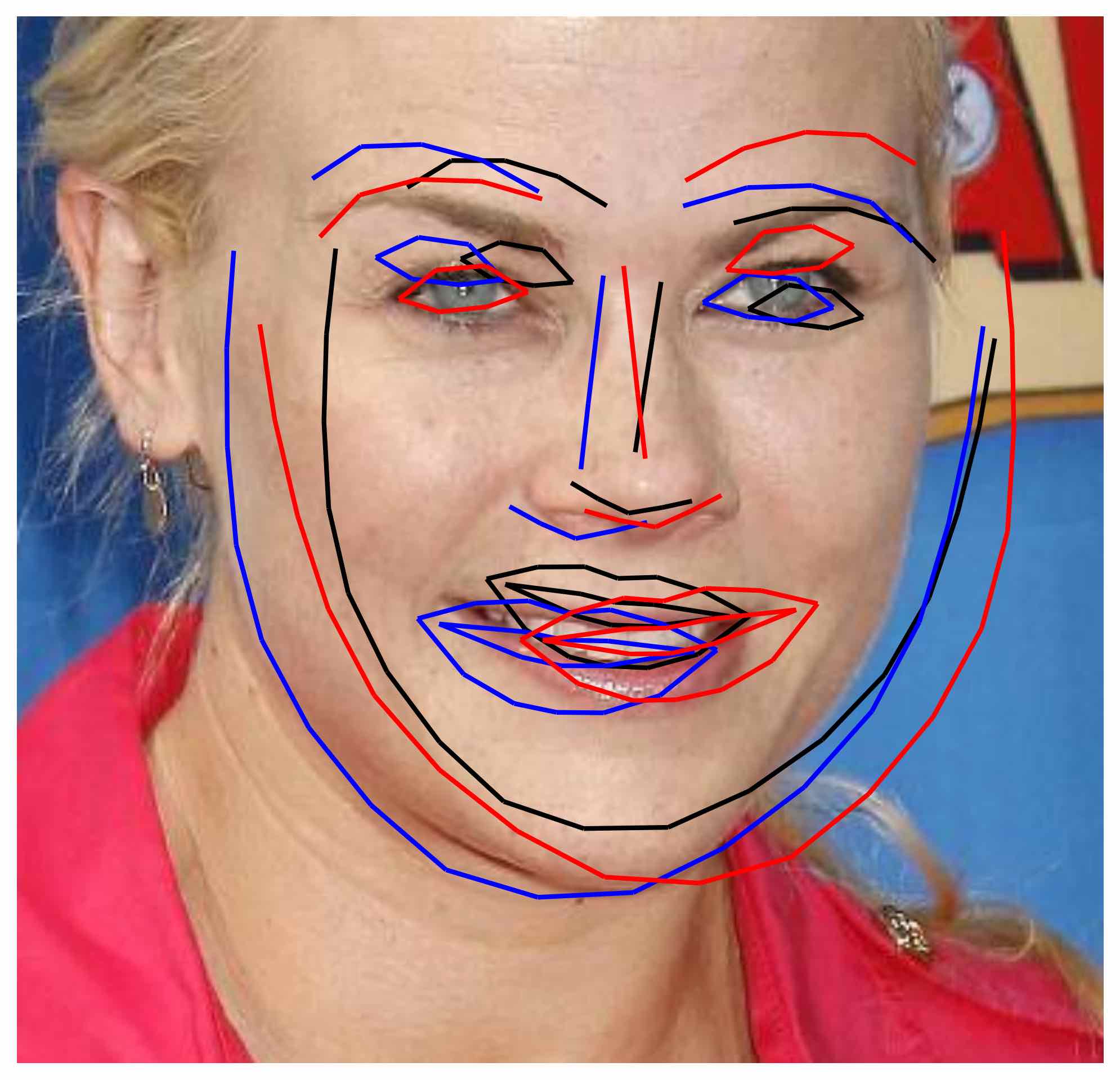}
		\caption{$5\%$}
		\label{fig:ini_2}
	\end{subfigure}
	\begin{subfigure}{0.16\textwidth}
		\includegraphics[width=\textwidth]{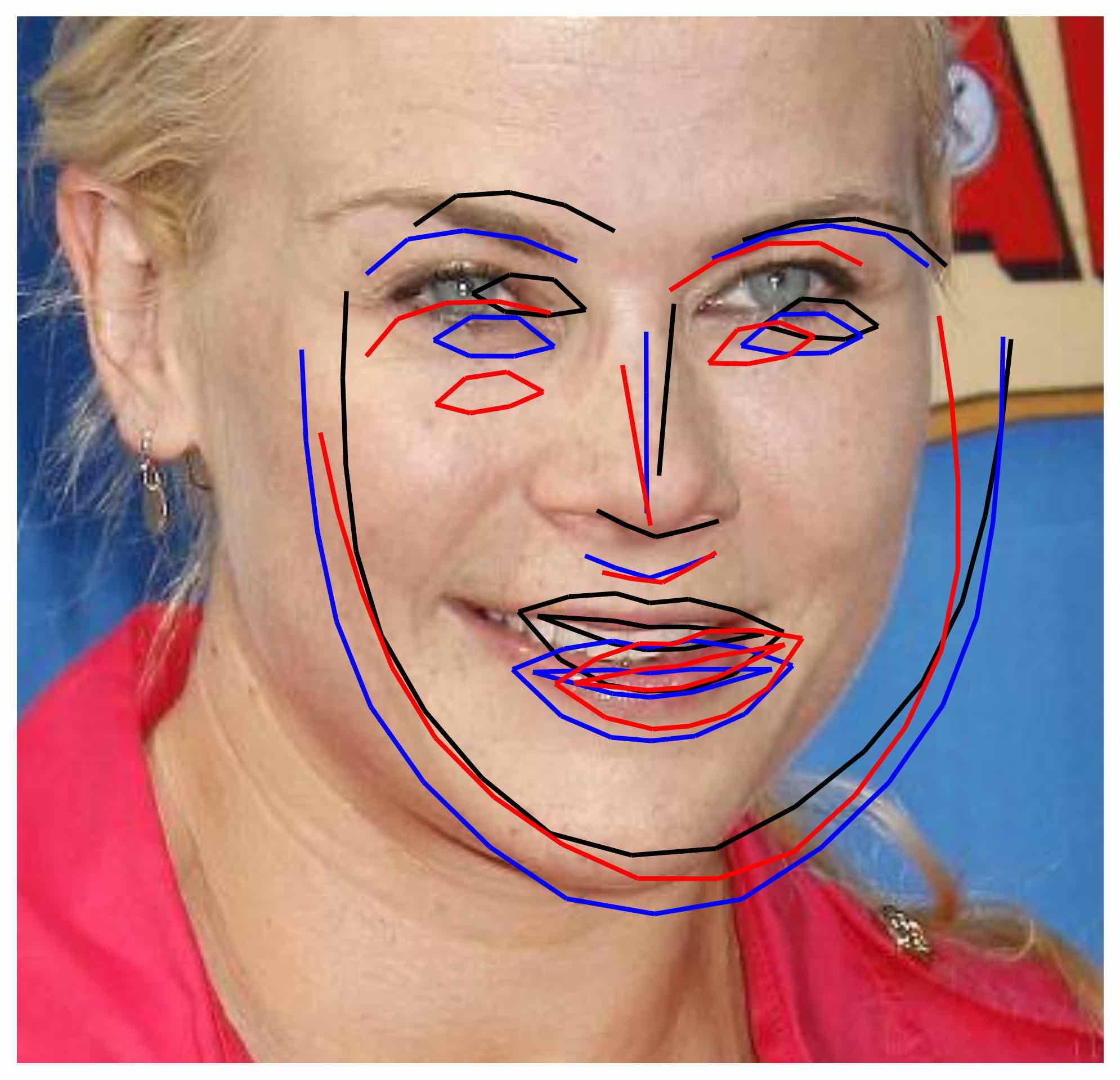}
		\caption{$7.5\%$}
		\label{fig:ini_3}
	\end{subfigure}
	\begin{subfigure}{0.16\textwidth}
		\includegraphics[width=\textwidth]{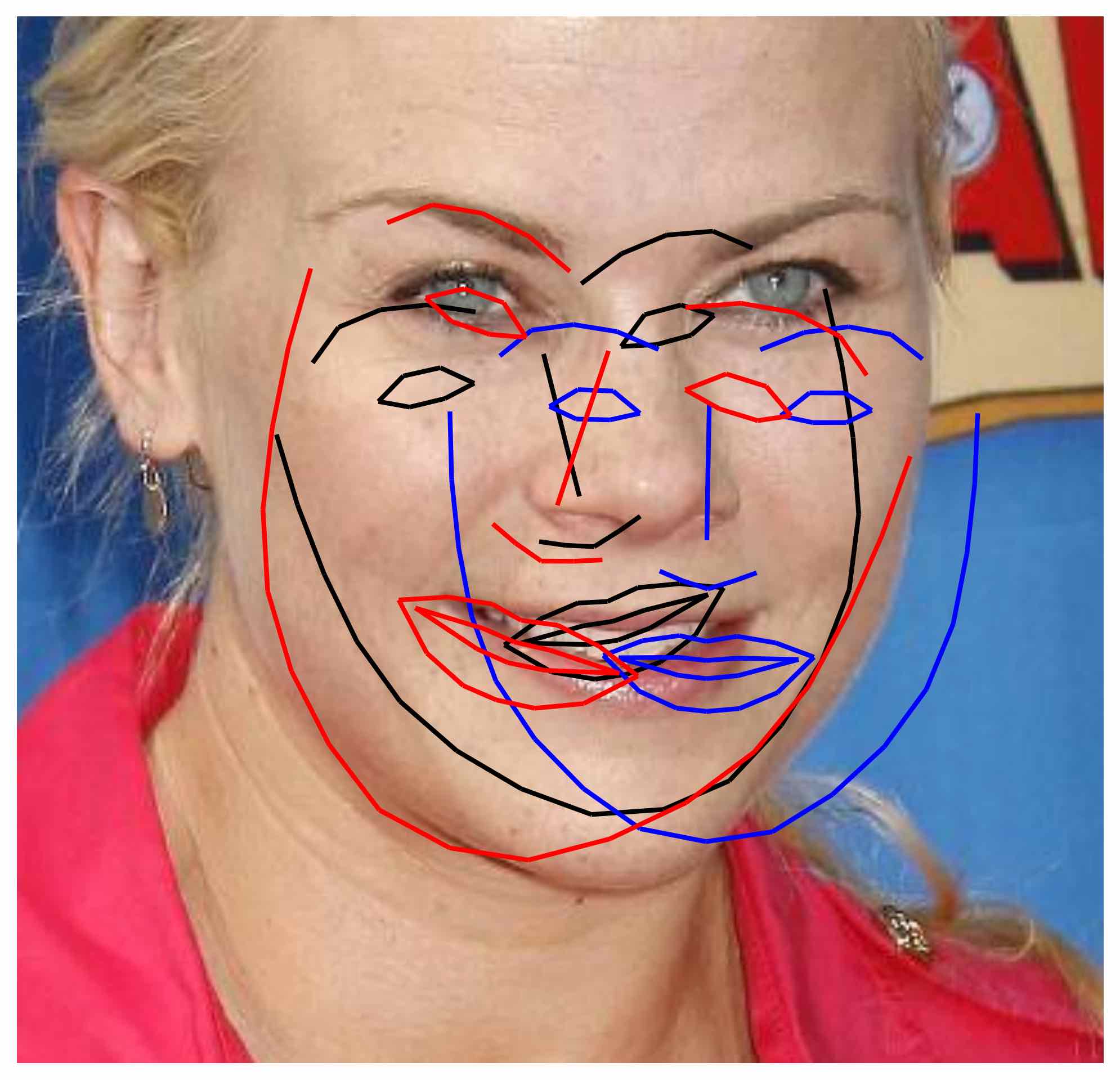}
		\caption{$10\%$}
		\label{fig:ini_4}
	\end{subfigure}
    \caption{Exemplar initializations obtained by varying the percentage of uniform noise added to the similarity parameters. Note that, increasing the percentage of noise produces more challenging initialization.}
    \label{fig:ini}
\end{figure*}

Finally, Kossaifi et al. derived the \emph{SSD Inverse Newton Schur} algorithm in \cite{Kossaifi2014}. This algorithm has a total complexity per iteration of $O(nmF + n^2m + 2n^2F + n^3)$ and was shown to slightly underperform its equivalent Gauss-Newton counterpart.

The remaining SSD algorithms derived in Section \ref{sec:fitting}, i.e.:
\begin{itemize}
\item \emph{SSD Inverse Wiberg}
\item \emph{SSD Forward Gauss-Newton Alternated}
\item \emph{SSD Forward Newton Schur}
\item \emph{SSD Forward Newton Alternated}
\item \emph{SSD Forward Wiberg}
\item \emph{SSD Asymmetric Gauss-Newton Schur}
\item \emph{SSD Asymmetric Gauss-Newton Alternated}
\item \emph{SSD Asymmetric Newton Schur}
\item \emph{SSD Asymmetric Newton Alternated}
\item \emph{SSD Asymmetric Wiberg}
\item \emph{SSD Bidirectional Gauss-Newton Schur}
\item \emph{SSD Bidirectional Gauss-Newton Alternated}
\item \emph{SSD Bidirectional Newton Schur}
\item \emph{SSD Bidirectional Newton Alternated}
\item \emph{SSD Bidirectional Wiberg}
\end{itemize}
have never been published before and are also a key contribution of the presented work.

Notice that, the iterative solutions of all CGD algorithms studied in this paper are given in Appendix \ref{sec:app2}.

\section{Experiments}
\label{sec:experiment}

In this section, we analyze the performance of the CGD algorithms derived in Section \ref{sec:fitting} on the specific problems of non-rigid face alignment in-the-wild. Results for five experiments are reported. The first experiment compares the fitting accuracy and convergence properties of all algorithms on the test set of the popular Labeled Faces in-the-Wild (LFPW) \cite{Belhumeur2011} database. The second experiment quantifies the importance of the two terms in the Bayesian project-out cost function in relation to the fitting accuracy obtained by \emph{Project-Out} algorithms. In the third experiment, we study the effect that varying the value of the parameters $\alpha$ and $\beta$ has on the performance of \emph{Asymmetric} algorithms. The fourth experiment explores the effect of optimizing the cost functions using reduced subsets of the total number of pixels and quantifies the impact that this has on the accuracy and computational efficiency of CGD algorithms. Finally, in the fifth experiment, we report the performance of the most accurate CGD algorithms on the test set of the Helen \cite{Le2012} database and on the entire Annotated Faces in-the-Wild (AFW) \cite{Zhu2012} database. 

Throughout this section, we abbreviate CGD algorithms using the following convention: \emph{CF\_TC\_OM(\_OS)} where: 
\begin{inparaenum}[\itshape a\upshape)]
\item \emph{CF} stands for \emph{Cost Function} and can be either \emph{SSD} or \emph{PO} depending on whether the algorithm uses the \emph{Sum of Squared Differences} or the \emph{Project Out} cost function;
\item \emph{TC} stands for \emph{Type of Composition} and can be \emph{For}, \emph{Inv}, \emph{Asy} or \emph{Bid} depending on whether the algorithm uses \emph{Forward}, \emph{Inverse}, \emph{Asymmetric} or \emph{Bidirectional} compositions;
\item \emph{OM} stands for \emph{Optimization Method} and can be \emph{GN}, \emph{N} or \emph{W} depending on whether the algorithm uses the \emph{Gauss-Newton}, \emph{Newton} or \emph{Wiberg} optimization methods; and, finally,
\item if \emph{Gauss-Newton} or \emph{Newton} methods are used, the optional field \emph{OS}, which stands for \emph{Optimization Strategy}, can be \emph{Sch} or \emph{Alt} depending on whether the algorithm solves for the parameters simultaneously using the \emph{Schur complement} or using \emph{Alternated optimization}.
\end{inparaenum}
For example, following the previous convention the \emph{Project Out Bidirectional Gauss-Newton Schur} algorithm is denoted by \emph{PO\_Bid\_GN\_Sch}.

Landmark annotations for all databases are provided by the iBUG group\footnote{\url{http://ibug.doc.ic.ac.uk/resources/300-W/}} \cite{Sagonas2013,Sagonas2013b} and fitting accuracy is reported using the point-to-point error measure normalized by the \emph{face size}\footnote{\label{facesize}The face size is computed as the mean of the height and width of the bounding box containing a face.}  proposed in \cite{Zhu2012} over the 49 interior points of the iBug annotation scheme.

In all face alignment experiments, we use a single AAM, trained using the $\sim800$ and $\sim2000$ training images of the LFPW and Helen databases. Similar to \cite{Tzimiropoulos2014}, we use a modified version of the \emph{Dense} Scale Invariant Feature Transform (DSIFT) \cite{Lowe1999, Dalal2005} to define the appearance representation of the previous AAM. In particular, we describe each pixel with a reduced SIFT descriptor of length $8$ using the public implementation provided by the authors of \cite{Vedaldi2008vlfeat}. All algorithms are implemented in a coarse to fine manner using a Gaussian pyramid with $2$ levels (face images are normalized to a \emph{face size}\footnoteref{facesize} of roughly $150$ pixels at the top level). In all experiments, we optimize over $7$ shape parameters ($4$ similarity transform and $3$ non-rigid shape parameters) at the first pyramid level and over $16$ shape parameters ($4$ similarity transform and $12$ non-rigid shape parameters) at the second one. The dimensionality of the appearance models is kept to represent $75\%$ of the total variance in both levels. This results in $225$ and $280$ appearance parameters at the first and second pyramid levels respectively. The previous choices were determined by testing on a small hold out set of the training data. 

In all experiments, algorithms are initialized by perturbing the similarity transform that perfectly aligns the model's mean shape (a frontal pose and neutral expression looking shape) with the ground truth shape of each image. These transforms are perturbed by adding uniformly distributed random noise to their scale, rotation and translation parameters. Exemplar initializations obtained by this procedure for different amounts of noise are shown in Figure \ref{fig:ini}. Notice that, we found that initializing using $5\%$ uniform noise is (statistically) equivalent to initializing with the popular OpenCV \cite{opencv_library} implementation of the well-known Viola and Jones face detector \cite{Viola2001} on the test images of the LFPW database.

Unless stated otherwise:
\begin{inparaenum}[\itshape i\upshape)] 
\item algorithms are initialized with $5\%$ uniform noise
\item test images are fitted three times using different random initializations (the same exact random initializations are used for all algorithms);
\item algorithms are left to run for 40 iterations (24 iterations at the first pyramid level and 16 at the second);  
\item results for \emph{Project-Out} algorithms are obtained using the Bayesian project-out cost function defined by Equation \ref{eq:prob_po}; and
\item results for \emph{Asymmetric} algorithms are reported for the special case of symmetric composition i.e. $\alpha=\beta=0.5$ in Equation \ref{eq:ssd_ac}.
\end{inparaenum}

In order to showcase the broader applicability of AAMs, we complete the previous performance analysis by performing a sixth and last experiment on the problem of non-rigid car alignment in-the-wild. To this end, we report the fitting accuracy of the best performing CGD algorithms on the MIT StreetScene\footnote{\label{car_db_url}\url{http://cbcl.mit.edu/software-datasets/streetscenes}} database.

Finally, in order to encourage open research and facilitate future comparisons with the results presented in this section, we make the implementation of all algorithms publicly available as part of the Menpo Project\footnoteref{menpo_url} \cite{Menpo2014}.

\subsection{Comparison on LFPW}
\label{exp:1}

In this experiment, we report the fitting accuracy and convergence properties of all CGD algorithms studied in this paper. Results are reported on the $\sim220$ test images of the LFPW database. In order to keep the information easily readable and interpretable, we group algorithms by cost function (i.e. \emph{SSD} or \emph{Project-Out}), and optimization method (i.e. \emph{Gauss-Newton}, \emph{Newton} or \emph{Wiberg}).

\begin{figure*}[p]
	\centering
	\begin{subfigure}{0.48\textwidth}
	    \includegraphics[width=\textwidth]{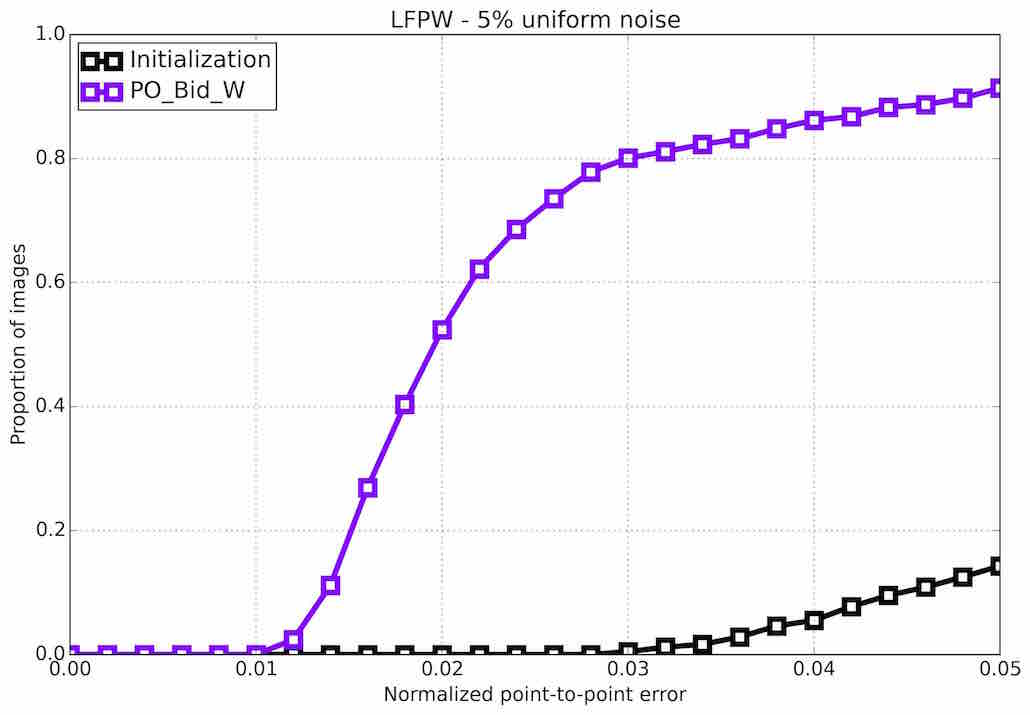}
	    \caption{Cumulative Error Distribution graph on the LFPW test dataset for all Project-Out Wiberg algorithms initialized with $5\%$ uniform noise.}
	    \label{fig:ced_bpo_w_5}
	\end{subfigure}
	\hfill
	\begin{subfigure}{0.48\textwidth}
	    \includegraphics[width=\textwidth]{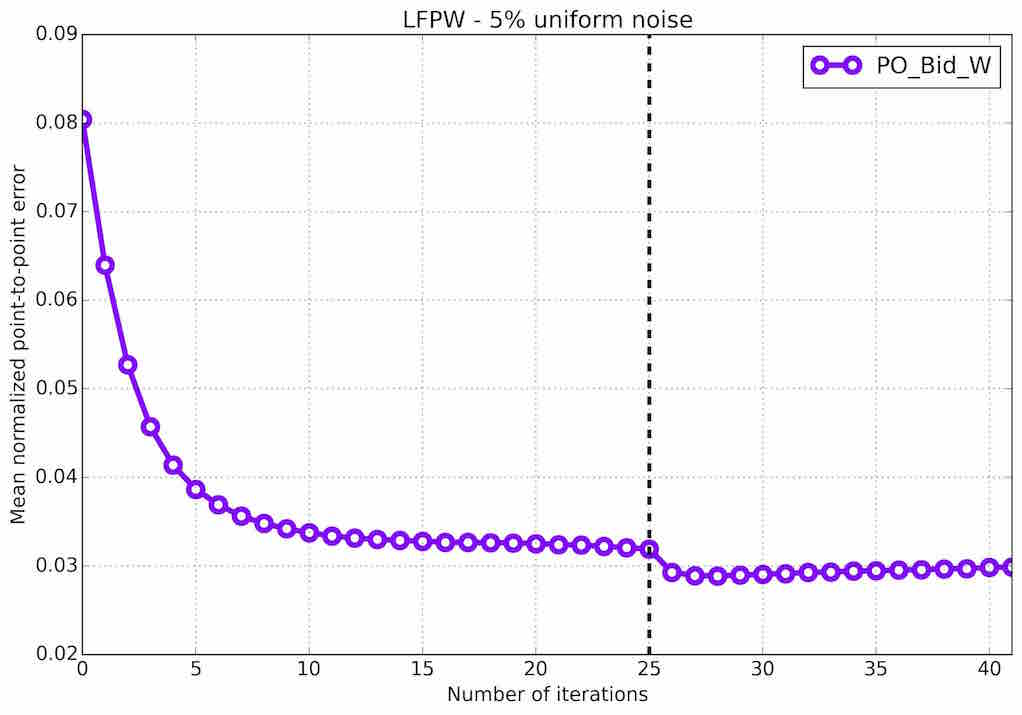}
	    \caption{Mean normalized point-to-point error vs number of iterations graph on the LFPW test dataset for all Project-Out Wiberg algorithms initialized with $5\%$ uniform noise.}
	    \label{fig:mean_error_vs_iters_bpo_w_5}
	\end{subfigure}
	\par\bigskip\bigskip
	\begin{subfigure}{0.48\textwidth}
	    \includegraphics[width=\textwidth]{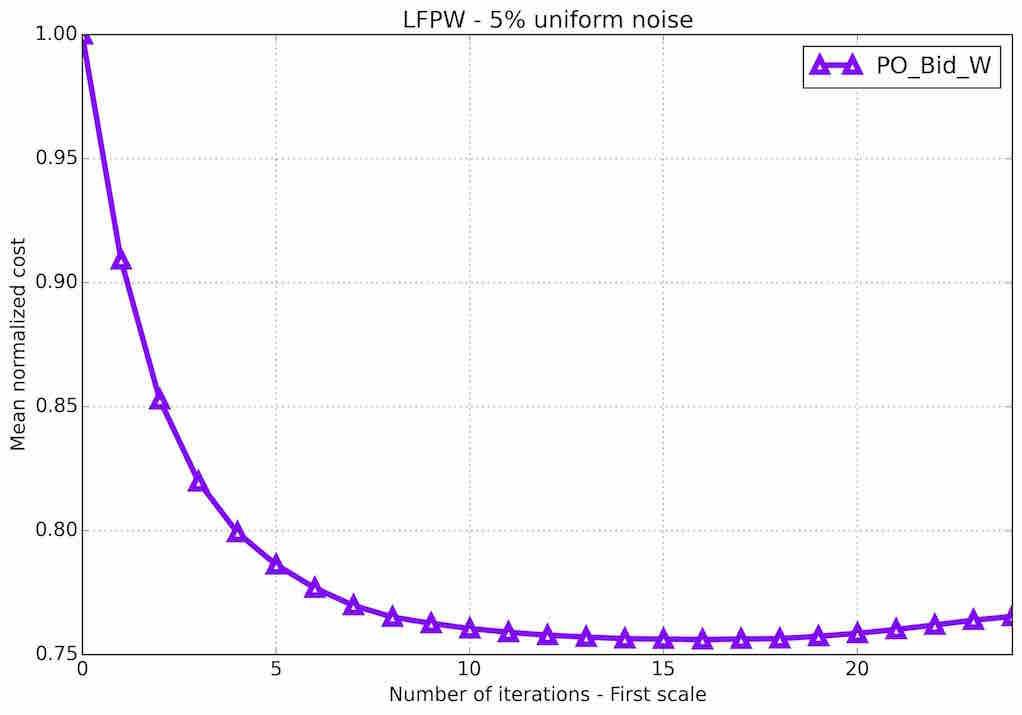}
	    \caption{Mean normalized cost vs number of first scale iterations graph on the LFPW test dataset for all Project-Out Wiberg algorithms initialized with $5\%$ uniform noise.}
	    \label{fig:mean_cost_vs_iters1_bpo_w_5}
	\end{subfigure}
	\hfill
	\begin{subfigure}{0.48\textwidth}
	    \includegraphics[width=\textwidth]{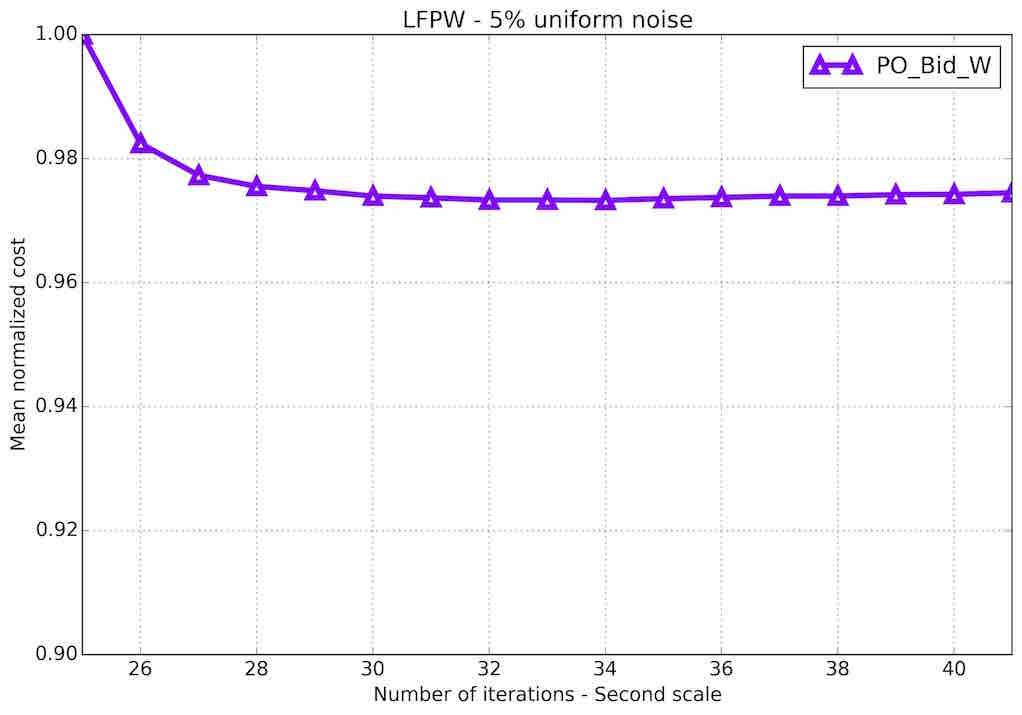}
	    \caption{Mean normalized cost vs number of second scale iterations graph on the LFPW test dataset for all Project-Out Wiberg algorithms initialized with $5\%$ uniform noise.}
	    \label{fig:mean_cost_vs_iters2_bpo_w_5}
	\end{subfigure}
	\par\bigskip\bigskip
	\begin{subfigure}{\textwidth}
		\center
		\begin{tabular}{lcccccc}
		    \toprule
		    Algorithm & $<0.02$ & $<0.03$ & $<0.04$ & Mean & Std & Median 
		    \\
		    \midrule
		    Initialization & 0.000 & 0.004 & 0.055 & 0.080 & 0.028 & 0.078
		    \\ 
		    PO\_Bid\_W\_Sch & 0.524 & 0.801 & 0.862 & 0.030 & 0.039 & 0.020
		    \\
		    \bottomrule
	  	\end{tabular}
	  	\caption{Table showing the proportion of images fitted with a normalized point-to-point error below $0.02$, $0.03$ and $0.04$ together with the normalized point-to-point error mean, std and median for all Project-Out Wiberg algorithms initialized with $5\%$ uniform noise.}
	    \label{tab:stats_bpo_w_5}
	\end{subfigure}
	\caption{Results showing the fitting accuracy and convergence properties of the Project-Out Wiberg algorithms on the LFPW test dataset.}
	\label{fig:bpo_w_5}
\end{figure*}

Results for this experiment are reported in Figures \ref{fig:ssd_gn_5}, \ref{fig:ssd_n_5}, \ref{fig:ssd_w_5}, \ref{fig:bpo_gn_5}, \ref{fig:bpo_n_5} and \ref{fig:bpo_w_5}. These figures have all the same structure and are composed of four figures and a table. Figures \ref{fig:ced_ssd_gn_5}, \ref{fig:ced_ssd_n_5}, \ref{fig:ced_ssd_w_5}, \ref{fig:ced_bpo_gn_5}, \ref{fig:ced_bpo_n_5} and \ref{fig:ced_bpo_w_5} report the Cumulative Error Distribution (CED), i.e the proportion of images vs normalized point-to-point error for each of the algorithms' groups. Tables \ref{tab:stats_ssd_gn_5}, \ref{tab:stats_ssd_n_5}, \ref{tab:stats_ssd_w_5}, \ref{tab:stats_bpo_gn_5}, \ref{tab:stats_bpo_n_5}, and \ref{tab:stats_bpo_w_5} summarize and complete the information on the previous CEDs by stating the proportion of images fitted with a normalized point-to-point error smaller than $0.02$, $0.03$ and $0.04$; and by stating the mean, std and median of the final normalized point-to-point error. The aim of the previous figures and tables is to help us compare the final fitting accuracy obtained by each algorithm. On the other hand, Figures \ref{fig:mean_error_vs_iters_ssd_gn_5}, \ref{fig:mean_error_vs_iters_ssd_n_5}, \ref{fig:mean_error_vs_iters_ssd_w_5}, \ref{fig:mean_error_vs_iters_bpo_gn_5}, \ref{fig:mean_error_vs_iters_bpo_n_5} and \ref{fig:mean_error_vs_iters_bpo_w_5} report the mean normalized point-to-point error at each iteration while Figures \ref{fig:mean_cost_vs_iters1_ssd_gn_5}, \ref{fig:mean_cost_vs_iters2_ssd_gn_5}, \ref{fig:mean_cost_vs_iters1_ssd_n_5}, \ref{fig:mean_cost_vs_iters2_ssd_n_5}, \ref{fig:mean_cost_vs_iters1_ssd_w_5}, \ref{fig:mean_cost_vs_iters2_ssd_w_5}, \ref{fig:mean_cost_vs_iters1_bpo_gn_5}, \ref{fig:mean_cost_vs_iters2_bpo_gn_5}, \ref{fig:mean_cost_vs_iters1_bpo_n_5}, \ref{fig:mean_cost_vs_iters2_bpo_n_5} and \ref{fig:mean_cost_vs_iters1_bpo_w_5}, \ref{fig:mean_cost_vs_iters2_bpo_w_5} report the mean normalized cost at each iteration\footnote{These figures are produced by dividing the value of the cost function at each iteration by its initial value and averaging for all images.}. The aim of these figures is to help us compare the convergence properties of every algorithm.

\subsubsection{SSD Gauss-Newton algorithms}

\begin{figure*}[p]
	\centering
	\begin{subfigure}{0.48\textwidth}
	    \includegraphics[width=\textwidth]{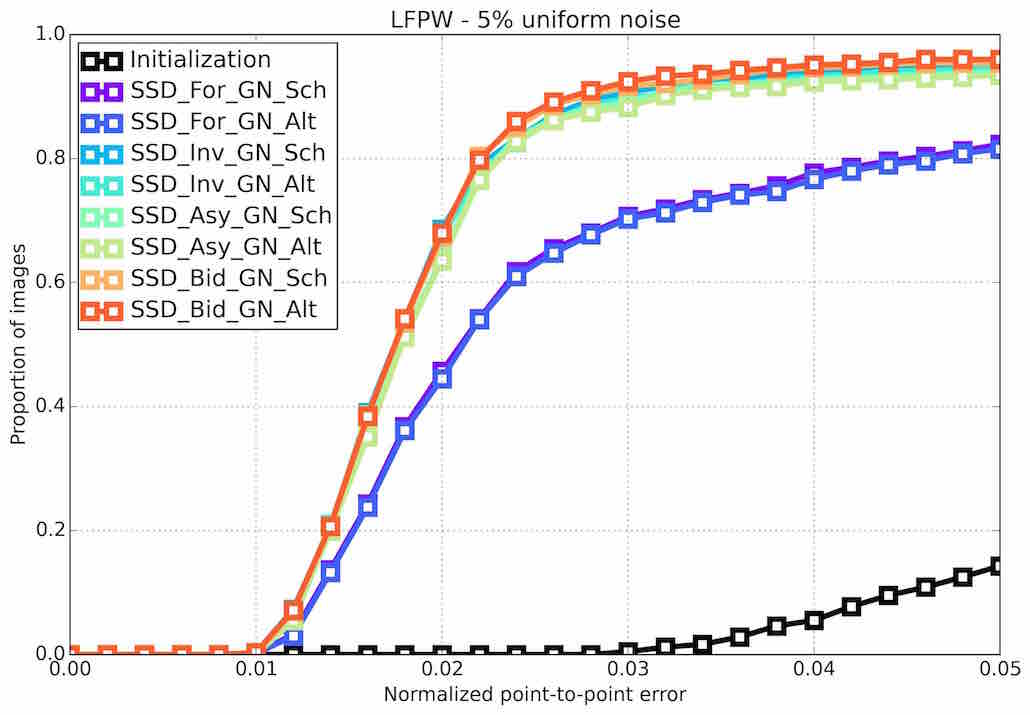}
	    \caption{CED on the LFPW test dataset for all SSD Gauss-Newton algorithms initialized with $5\%$ uniform noise.}
	    \label{fig:ced_ssd_gn_5}
	\end{subfigure}
	\hfill
	\begin{subfigure}{0.48\textwidth}
	    \includegraphics[width=\textwidth]{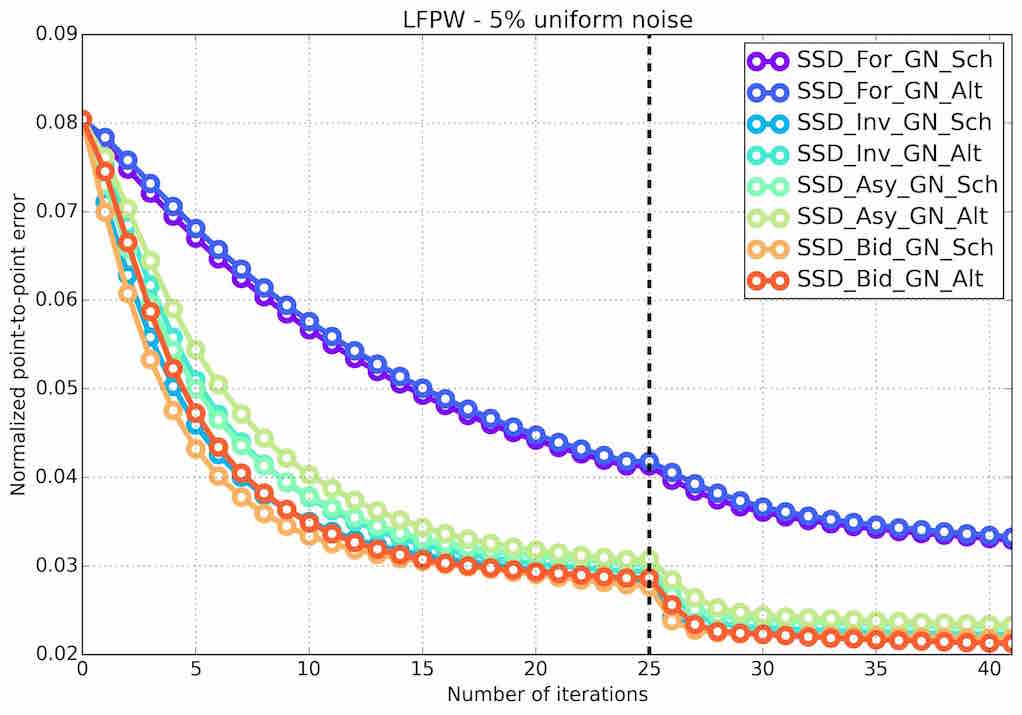}
	    \caption{Mean normalized point-to-point error vs number of iterations on the LFPW test dataset for all SSD Gauss-Newton algorithms initialized with $5\%$ uniform noise.}
	    \label{fig:mean_error_vs_iters_ssd_gn_5}
	\end{subfigure}
	\par\bigskip\bigskip
	\begin{subfigure}{0.48\textwidth}
	    \includegraphics[width=\textwidth]{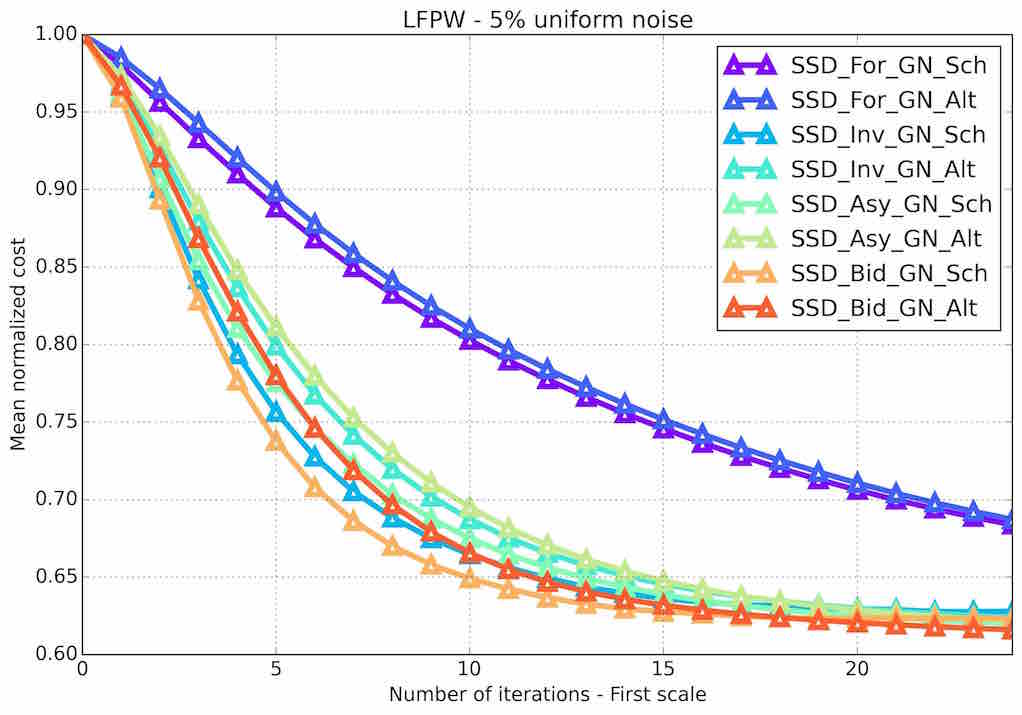}
	    \caption{Mean normalized cost vs number of first scale iterations on the LFPW test dataset for all SSD Gauss-Newton algorithms initialized with $5\%$ uniform noise.}
	    \label{fig:mean_cost_vs_iters1_ssd_gn_5}
	\end{subfigure}
	\hfill
	\begin{subfigure}{0.48\textwidth}
	    \includegraphics[width=\textwidth]{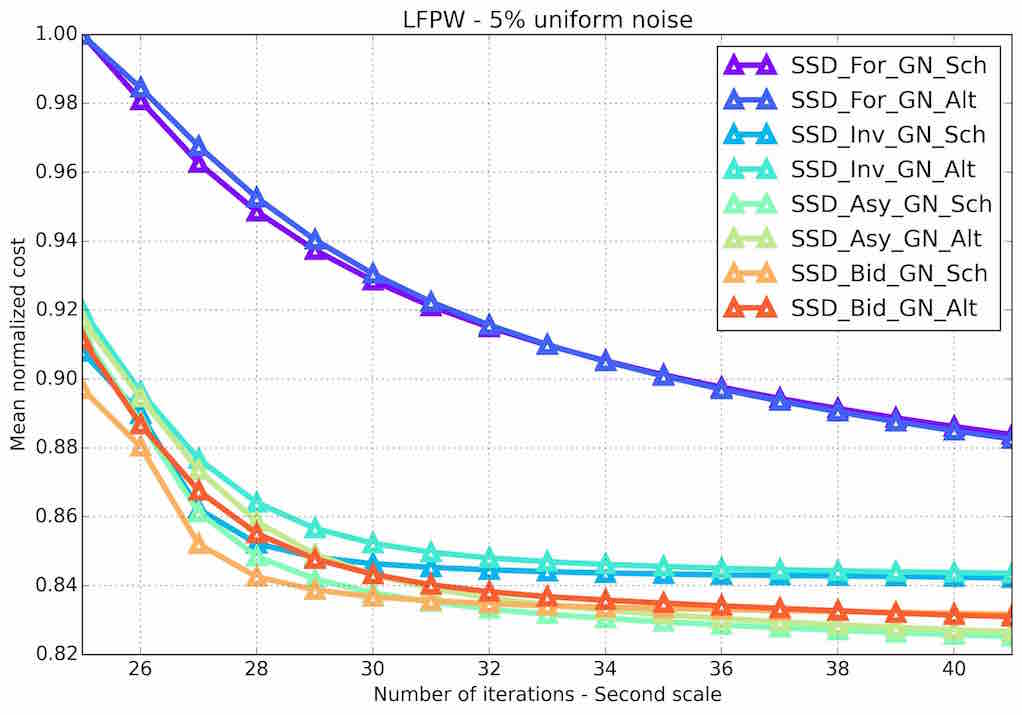}
	    \caption{Mean normalized cost vs number of second scale iterations on the LFPW test dataset for all SSD Gauss-Newton algorithms initialized with $5\%$ uniform noise.}
	    \label{fig:mean_cost_vs_iters2_ssd_gn_5}
	\end{subfigure}
	\par\bigskip\bigskip
	\begin{subfigure}{\textwidth}
		\center
		\begin{tabular}{lcccccc}
			\toprule
		    Algorithm & $<0.02$ & $<0.03$ & $<0.04$ & Mean & Std & Median 
		    \\
		    \midrule
		    Initialization & 0.000 & 0.004 & 0.055 & 0.080 & 0.028 & 0.078
		    \\
		    SSD\_For\_GN\_Sch & 0.456 & 0.707 & 0.777 & 0.033 & 0.030 & 0.021 
		    \\
		    SSD\_For\_GN\_Alt & 0.445 & 0.702 & 0.766 & 0.033 & 0.030 & 0.021
		    \\
		    SSD\_Inv\_GN\_Sch & 0.686 & 0.906 & 0.939 & 0.022 & \textbf{0.019} & \textbf{0.017}
		    \\
		    SSD\_Inv\_GN\_Alt & 0.673 & 0.897 & 0.933 & 0.022 & 0.020 & \textbf{0.017}
		    \\
		    SSD\_Asy\_GN\_Sch & 0.640 & 0.891 & 0.929 & 0.023 & 0.021 & 0.018
		    \\
		    SSD\_Asy\_GN\_Alt & 0.635 & 0.882 & 0.924 & 0.023 & 0.021 & 0.018
		    \\
		    SSD\_Bid\_GN\_Sch & 0.674 & 0.917 & 0.946 & 0.022 & \textbf{0.019} & \textbf{0.017}
		    \\
		    SSD\_Bid\_GN\_Alt & \textbf{0.680} & \textbf{0.924} & \textbf{0.951} & \textbf{0.021} & \textbf{0.019} & \textbf{0.017} 
		    \\
		    \bottomrule
	  	\end{tabular}
	  	\caption{Table showing the proportion of images fitted with a normalized point-to-point error below $0.02$, $0.03$ and $0.04$ together with the normalized point-to-point error mean, std and median for all SSD Gauss-Newton algorithms initialized with $5\%$ uniform noise.}
	    \label{tab:stats_ssd_gn_5}
	\end{subfigure}
	\caption{Results showing the fitting accuracy and convergence properties of the SSD Gauss-Newton algorithms on the LFPW test dataset initialized with $5\%$ uniform noise.}
	\label{fig:ssd_gn_5}
\end{figure*}

Results for \emph{SSD Gauss-Newton} algorithms are reported in Figure \ref{fig:ssd_gn_5}. We can observe that \emph{Inverse}, \emph{Asymmetric} and \emph{Bidirectional} algorithms obtain a similar performance and significantly outperform \emph{Forward} algorithms in terms of fitting accuracy, Figure \ref{fig:ced_ssd_gn_5} and Table \ref{tab:stats_ssd_gn_5}. In absolute terms, \emph{Bidirectional} algorithms slightly outperform \emph{Inverse} and \emph{Asymmetric} algorithms. On the other hand, the difference in performance between the \emph{Simultaneous Schur} and \emph{Alternated} optimizations strategies are minimal for all algorithms and they were found to have no statistical significance.

Looking at Figures \ref{fig:mean_error_vs_iters_ssd_gn_5}, \ref{fig:mean_cost_vs_iters1_ssd_gn_5} and \ref{fig:mean_cost_vs_iters2_ssd_gn_5} there seems to be a clear (and obviously expected) correlation between the normalized point-to-point error and the normalized value of the cost function at each iteration. In terms of convergence, it can be seen that \emph{Forward} algorithms converge slower than \emph{Inverse}, \emph{Asymmetric} and \emph{Bidirectional}. \emph{Bidirectional} algorithms converge slightly faster than \emph{Inverse} algorithms and these slightly faster than \emph{Asymmetric} algorithms. In this case, the \emph{Simultaneous Schur} optimization strategy seems to converge slightly faster than the \emph{Alternated} one for all \emph{SSD Gauss-Newton} algorithms.

\subsubsection{SSD Newton algorithms}

\begin{figure*}[p]
	\centering
	\begin{subfigure}{0.48\textwidth}
	    \includegraphics[width=\textwidth]{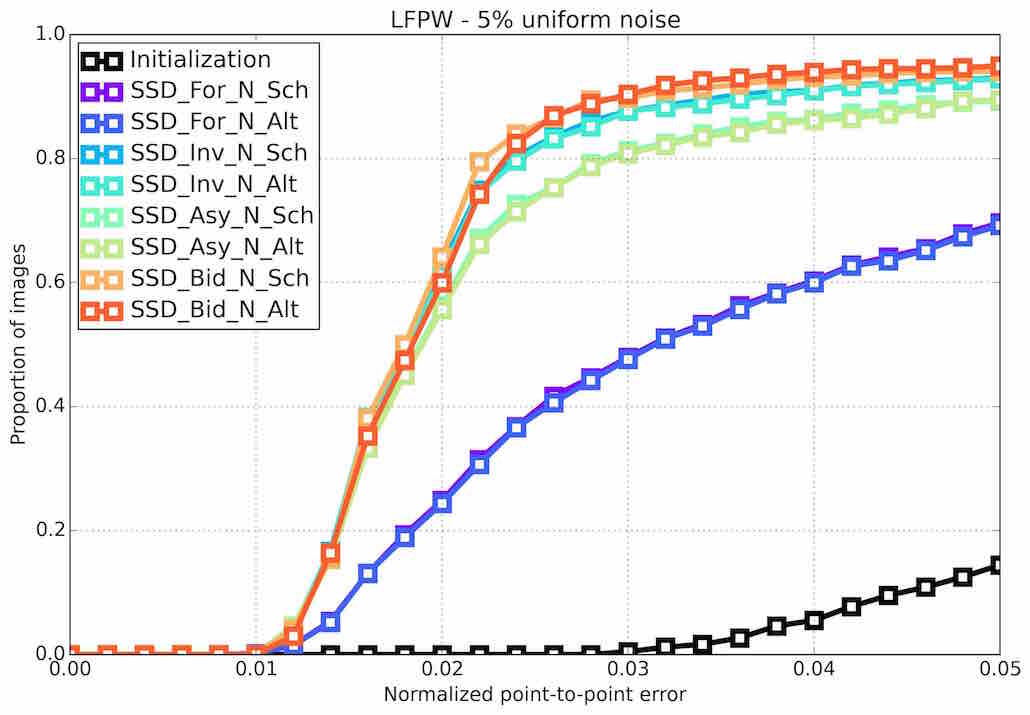}
	    \caption{Cumulative error distribution on the LFPW test dataset for all SSD Newton algorithms initialized with $5\%$ uniform noise.}
	    \label{fig:ced_ssd_n_5}
	\end{subfigure}
	\hfill
	\begin{subfigure}{0.48\textwidth}
	    \includegraphics[width=\textwidth]{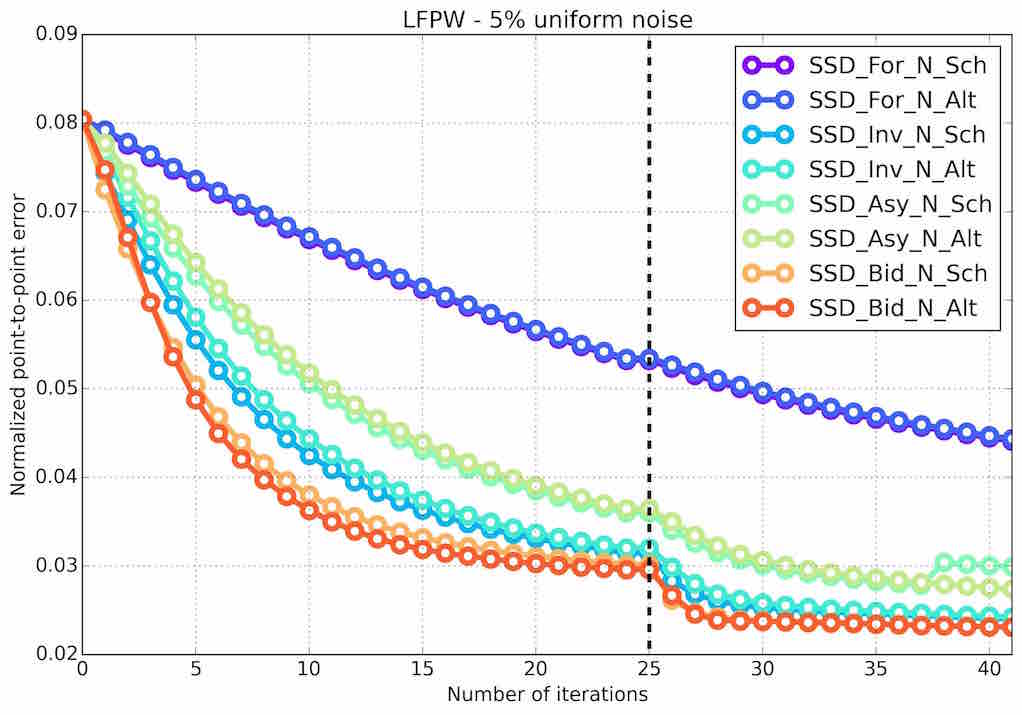}
	    \caption{Mean normalized point-to-point error vs number of iterations on the LFPW test dataset for all SSD Newton algorithms initialized with $5\%$ uniform noise.}
	    \label{fig:mean_error_vs_iters_ssd_n_5}
	\end{subfigure}
	\par\bigskip\bigskip
	\begin{subfigure}{0.48\textwidth}
	    \includegraphics[width=\textwidth]{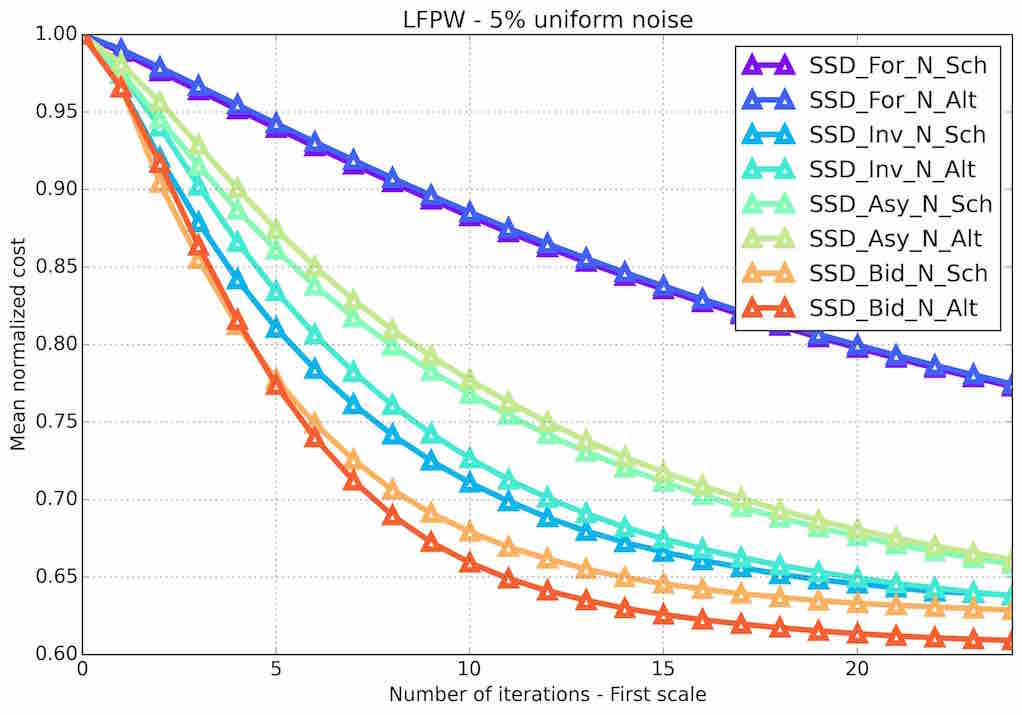}
	    \caption{Mean normalized cost vs number of first scale iterations on the LFPW test dataset for all SSD Newton algorithms initialized with $5\%$ uniform noise.}
	    \label{fig:mean_cost_vs_iters1_ssd_n_5}
	\end{subfigure}
	\hfill
	\begin{subfigure}{0.48\textwidth}
	    \includegraphics[width=\textwidth]{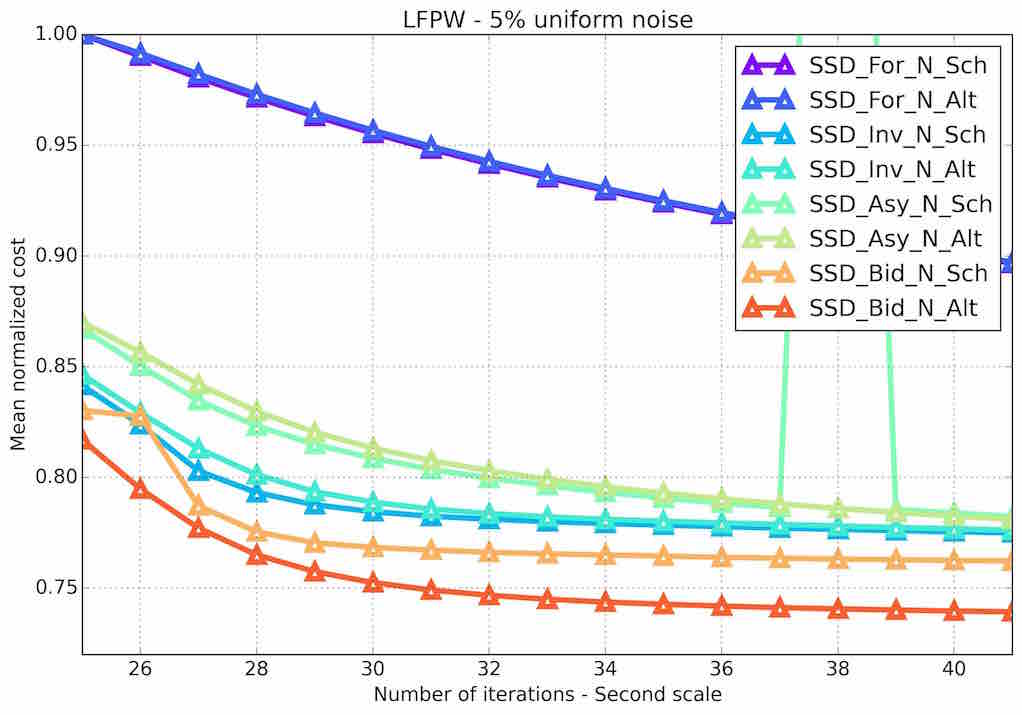}
	    \caption{Mean normalized cost vs number of second scale iterations on the LFPW test dataset for all SSD Newton algorithms initialized with $5\%$ uniform noise.}
	    \label{fig:mean_cost_vs_iters2_ssd_n_5}
	\end{subfigure}
	\par\bigskip\bigskip
	\begin{subfigure}{\textwidth}
		\center
		\begin{tabular}{lcccccc}
			\toprule
		    Algorithm & $<0.02$ & $<0.03$ & $<0.04$ & Mean & Std & Median 
		    \\
		    \midrule
		    Initialization & 0.000 & 0.004 & 0.055 & 0.080 & 0.028 & 0.078
		    \\
		    SSD\_For\_N\_Sch & 0.249 & 0.479 & 0.603 & 0.044 & 0.033 & 0.031
		    \\
		    SSD\_For\_N\_Alt & 0.244 & 0.476 & 0.600 & 0.044 & 0.033 & 0.032
		    \\
		    SSD\_Inv\_N\_Sch & 0.626 & 0.876 & 0.909 & 0.024 & 0.022 & \textbf{0.018}
		    \\
		    SSD\_Inv\_N\_Alt & 0.613 & 0.876 & 0.909 & 0.024 & 0.022 & \textbf{0.018}
		    \\
		    SSD\_Asy\_N\_Sch & 0.562 & 0.812 & 0.863 & 0.030 & 0.076 & 0.019
		    \\
		    SSD\_Asy\_N\_Alt & 0.557 & 0.808 & 0.862 & 0.027 & 0.025 & 0.019
		    \\
		    SSD\_Bid\_N\_Sch & \textbf{0.641} & 0.897 & 0.932 & \textbf{0.023} & 0.022 & \textbf{0.018}
		    \\
		    SSD\_Bid\_N\_Alt & 0.600 & \textbf{0.903} & \textbf{0.939} & \textbf{0.023} & \textbf{0.021} & \textbf{0.018}
		    \\
		    \bottomrule
	  	\end{tabular}
	  	\caption{Table showing the proportion of images fitted with a normalized point-to-point error below $0.02$, $0.03$ and $0.04$ together with the normalized point-to-point error Mean, Std and Median for all SSD Newton algorithms initialized with $5\%$ uniform noise.}
	    \label{tab:stats_ssd_n_5}
	\end{subfigure}
	\caption{Results showing the fitting accuracy and convergence properties of the SSD Newton algorithms on the LFPW test dataset initialized with $5\%$ uniform noise.}
	\label{fig:ssd_n_5}
\end{figure*}

Results for \emph{SSD Newton} algorithms are reported on Figure \ref{fig:ssd_n_5}. In this case, we can observe that the fitting performance of all algorithms decreases with respect to their \emph{Gauss-Newton} counterparts Figure \ref{fig:ced_ssd_n_5} and Table \ref{tab:stats_ssd_n_5}. This is most noticeable in the case of \emph{Forward} algorithms for which there is $\sim20\%$ drop in the proportion of images fitted below $0.02$, $0.03$ and $0.04$ with respect to its \emph{Gauss-Newton} equivalents. For these algorithms there is also a significant increase in the mean and median of the normalized point-to-point error. \emph{Asymmetric Newton} algorithms also perform considerably worse, between $5\%$ and $10\%$, than their \emph{Gauss-Newton} versions. The drop in performance is reduced for \emph{Inverse} and \emph{Bidirectional Newton} algorithms for which accuracy is only reduced by around $3\%$ with respect their \emph{Gauss-Newton} equivalent. 

Within \emph{Newton} algorithms, there are clear differences in terms of speed of convergence \ref{fig:mean_error_vs_iters_ssd_n_5}, \ref{fig:mean_cost_vs_iters1_ssd_n_5} and \ref{fig:mean_cost_vs_iters2_ssd_n_5}. \emph{Bidirectional} algorithms are the fastest to converge followed by \emph{Inverse} and \emph{Asymmetric} algorithms, in this order, and lastly \emph{Forward} algorithms. In this case, the \emph{Simultaneous Schur} optimization strategy seems to converge again slightly faster than the \emph{Alternated} one for all algorithms but \emph{Bidirectional} algorithms, for which the \emph{Alternated} strategy converges slightly faster. Overall, \emph{SSD Newton} algorithms converge slower than \emph{SSD Gauss-Newton} algorithms.

\subsubsection{SSD Wiberg algorithms}

\begin{figure*}[p]
	\centering
	\begin{subfigure}{0.48\textwidth}
	    \includegraphics[width=\textwidth]{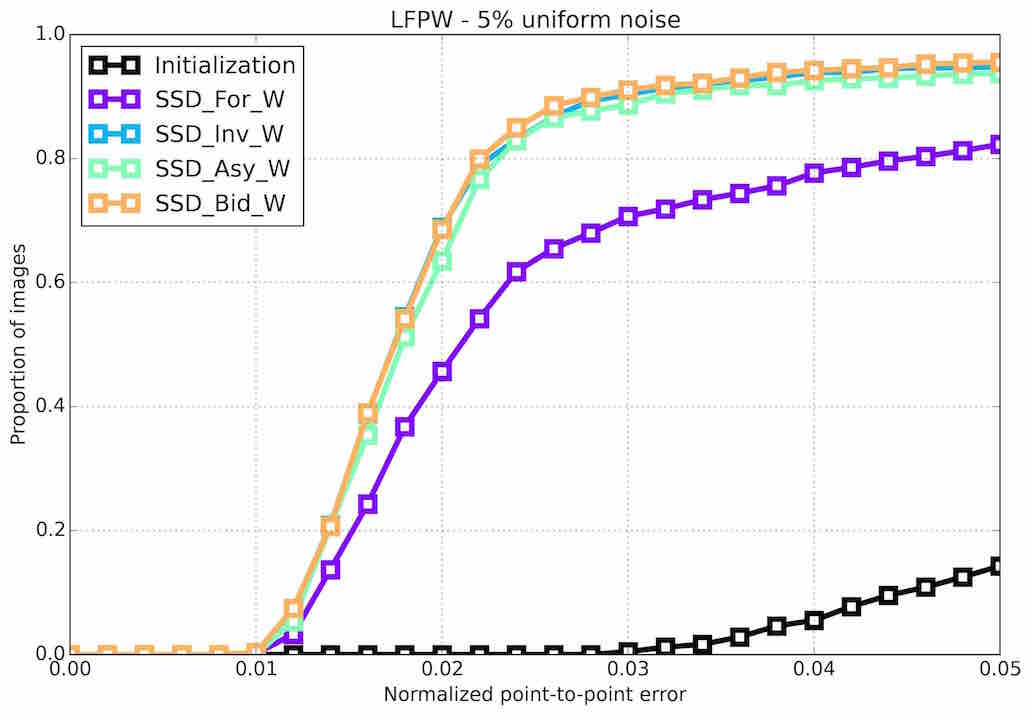}
	    \caption{CED on the LFPW test dataset for all SSD Wiberg algorithms initialized with $5\%$ uniform noise.}
	    \label{fig:ced_ssd_w_5}
	\end{subfigure}
	\hfill
	\begin{subfigure}{0.48\textwidth}
	    \includegraphics[width=\textwidth]{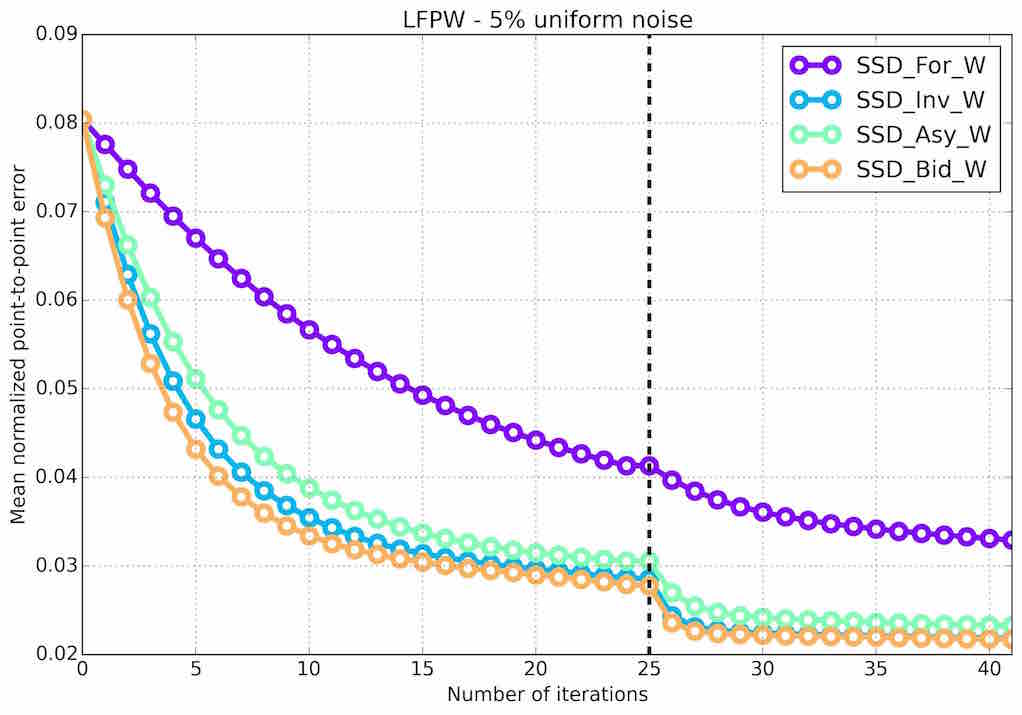}
	    \caption{Mean normalized point-to-point error vs number of iterations on the LFPW test dataset for all SSD Wiberg algorithms initialized with $5\%$ uniform noise.}
	    \label{fig:mean_error_vs_iters_ssd_w_5}
	\end{subfigure}
	\par\bigskip\bigskip
	\begin{subfigure}{0.48\textwidth}
	    \includegraphics[width=\textwidth]{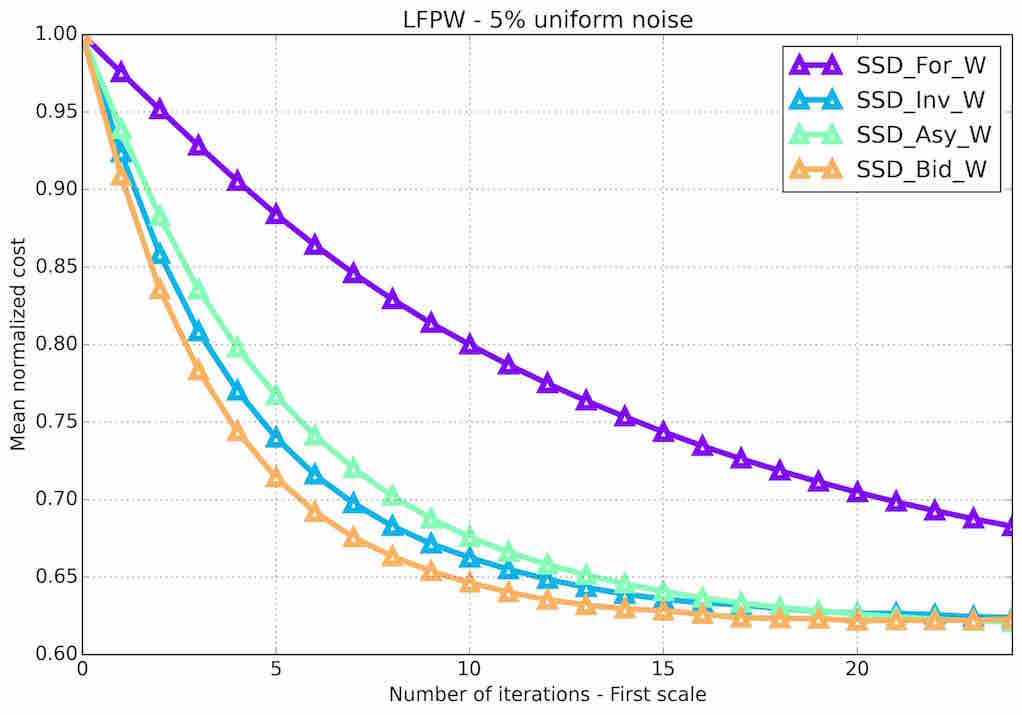}
	    \caption{Mean normalized cost vs number of first scale iterations on the LFPW test dataset for all SSD Wiberg algorithms initialized with $5\%$ uniform noise.}
	    \label{fig:mean_cost_vs_iters1_ssd_w_5}
	\end{subfigure}
	\hfill
	\begin{subfigure}{0.48\textwidth}
	    \includegraphics[width=\textwidth]{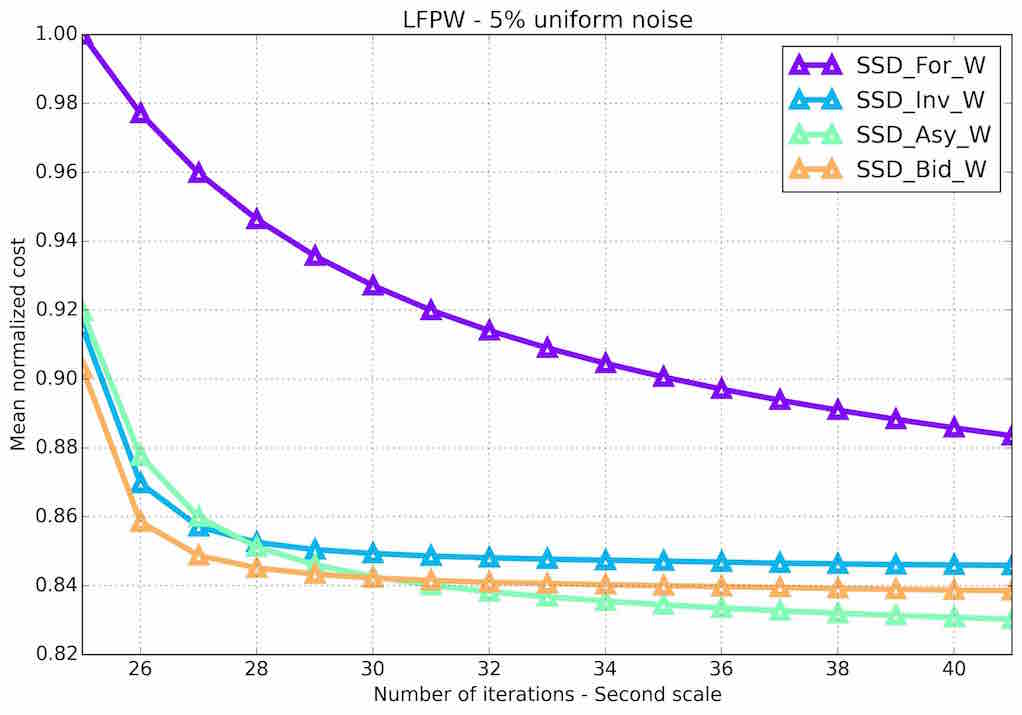}
	    \caption{Mean normalized cost vs number of second scale iterations on the LFPW test dataset for all SSD Wiberg algorithms initialized with $5\%$ uniform noise.}
	    \label{fig:mean_cost_vs_iters2_ssd_w_5}
	\end{subfigure}
	\par\bigskip\bigskip
	\begin{subfigure}{\textwidth}
		\center
		\begin{tabular}{lcccccc}
		    \toprule
		    Algorithm & $<0.02$ & $<0.03$ & $<0.04$ & Mean & Std & Median 
		    \\
		    \midrule
		    Initialization & 0.000 & 0.004 & 0.055 & 0.080 & 0.028 & 0.078
		    \\ 
		    SSD\_For\_W & 0.457 & 0.707 & 0.777 & 0.33 & 0.030 & 0.021
		    \\
		    SSD\_Inv\_W & \textbf{0.689} & 0.903 & 0.939 & \textbf{0.22} & \textbf{0.019} & \textbf{0.017}
		    \\
		    SSD\_Asy\_W & 0.635 & 0.887 & 0.926 & 0.23 & 0.021 & 0.018
		    \\
		    SSD\_Bid\_W & 0.686 & \textbf{0.911} & \textbf{0.942} & \textbf{0.22} & \textbf{0.019} & \textbf{0.017}
		    \\
		    \bottomrule
	  	\end{tabular}
	  	\caption{Table showing the proportion of images fitted with a normalized point-to-point error below $0.02$, $0.03$ and $0.04$ together with the normalized point-to-point error mean, std and median for all SSD Wiberg algorithms initialized with $5\%$ uniform noise.}
	    \label{tab:stats_ssd_w_5}
	\end{subfigure}
	\caption{Results showing the fitting accuracy and convergence properties of the SSD Wiberg algorithms on the LFPW test dataset.}
	\label{fig:ssd_w_5}
\end{figure*}

Results for \emph{SSD Wiberg} algorithms are reported on Figure \ref{fig:ssd_w_5}. Figure \ref{fig:ced_ssd_w_5} and Table \ref{tab:stats_ssd_w_5} and Figures \ref{fig:mean_error_vs_iters_ssd_w_5}, \ref{fig:mean_cost_vs_iters1_ssd_w_5} and \ref{fig:mean_cost_vs_iters2_ssd_w_5} show that these results are (as one would expect) virtually equivalent to those obtained by their \emph{Gauss-Newton} counterparts.

\subsubsection{Project-Out Gauss-Newton algorithms}

\begin{figure*}[p]
	\centering
	\begin{subfigure}{0.48\textwidth}
	    \includegraphics[width=\textwidth]{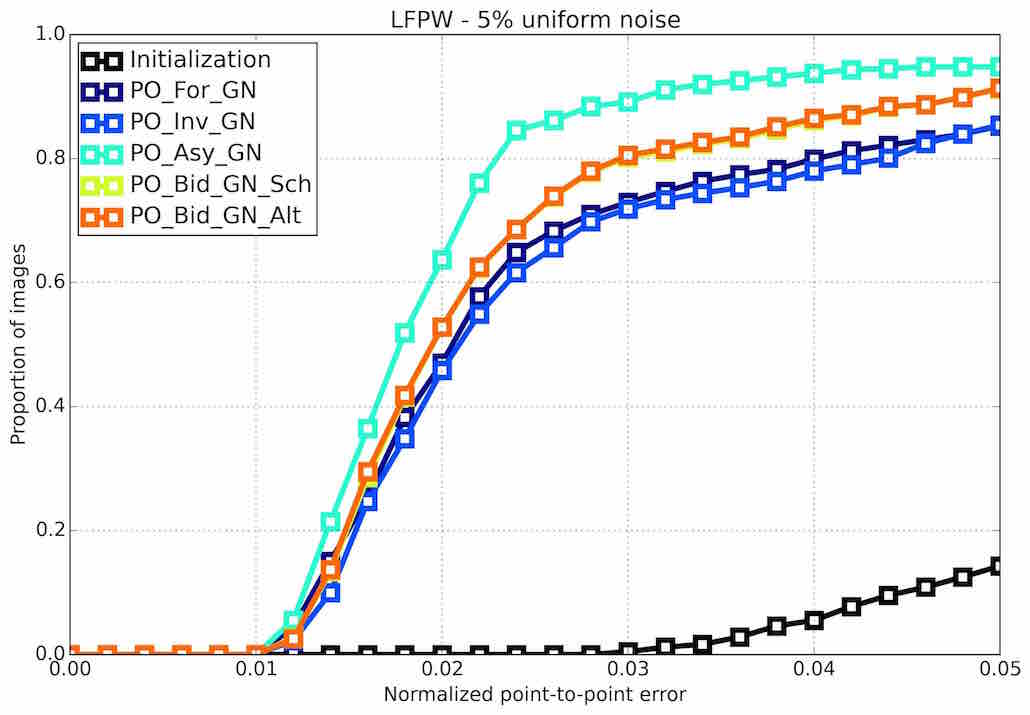}
	    \caption{CED graph on the LFPW test dataset for all Project-Out Gauss-Newton algorithms initialized with $5\%$ uniform noise.}
	    \label{fig:ced_bpo_gn_5}
	\end{subfigure}
	\hfill
	\begin{subfigure}{0.48\textwidth}
	    \includegraphics[width=\textwidth]{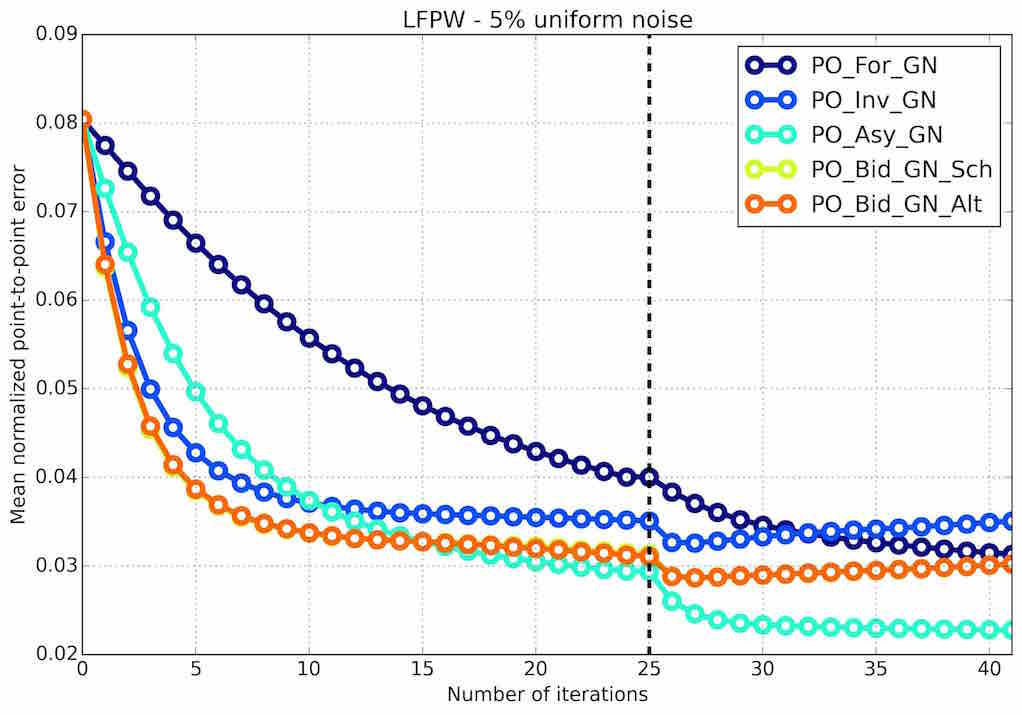}
	    \caption{Mean normalized point-to-point error vs number of iterations on the LFPW test dataset for all Project-Out Gauss-Newton algorithms initialized with $5\%$ uniform noise.}
	    \label{fig:mean_error_vs_iters_bpo_gn_5}
	\end{subfigure}
	\par\bigskip\bigskip
	\begin{subfigure}{0.48\textwidth}
	    \includegraphics[width=\textwidth]{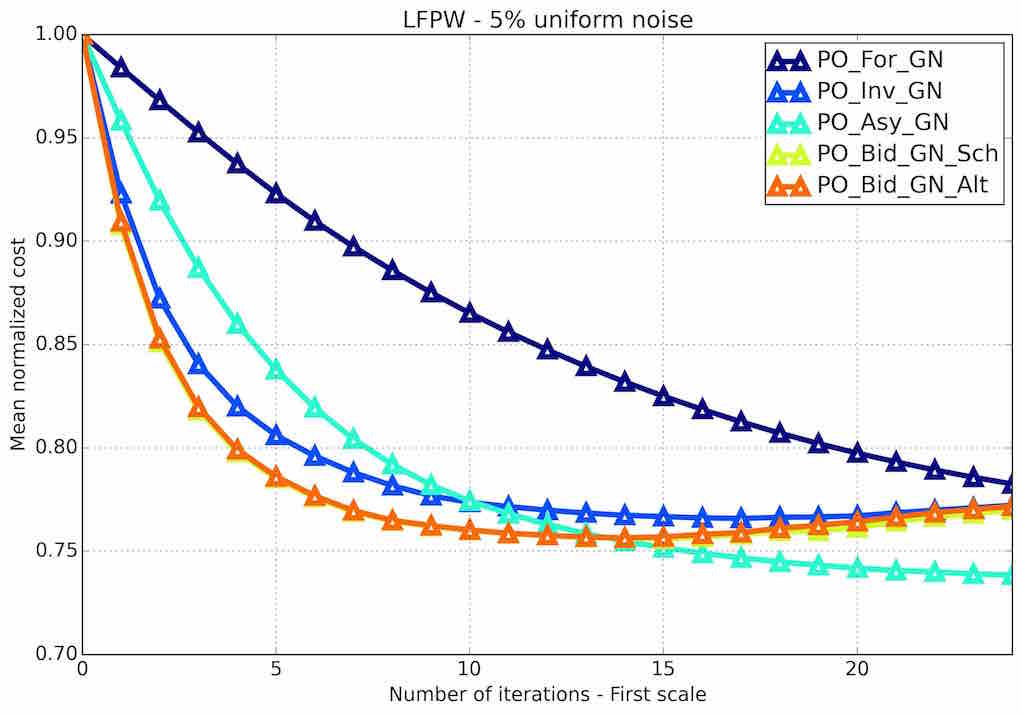}
	    \caption{Mean normalized cost vs number of first scale iterations on the LFPW test dataset for all Project-Out Gauss-Newton algorithms initialized with $5\%$ uniform noise.}
	    \label{fig:mean_cost_vs_iters1_bpo_gn_5}
	\end{subfigure}
	\hfill
	\begin{subfigure}{0.48\textwidth}
	    \includegraphics[width=\textwidth]{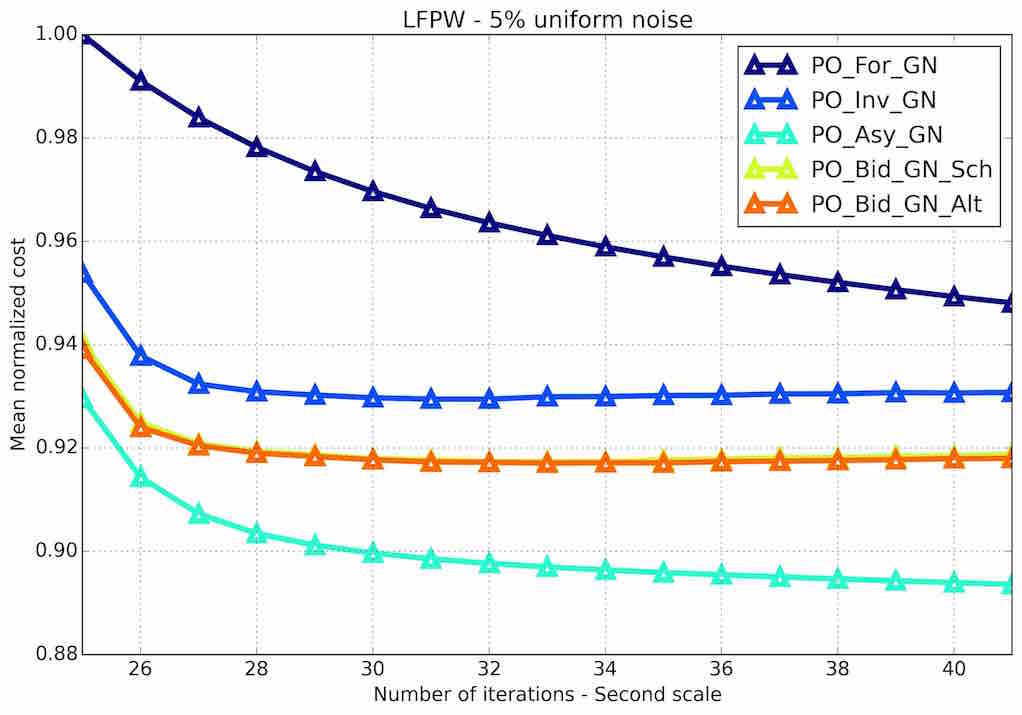}
	    \caption{Mean normalized cost vs number of second scale iterations on the LFPW test dataset for all Project-Out Gauss-Newton algorithms initialized with $5\%$ uniform noise.}
	    \label{fig:mean_cost_vs_iters2_bpo_gn_5}
	\end{subfigure}
	\par\bigskip\bigskip
	\begin{subfigure}{\textwidth}
		\center
		\begin{tabular}{lcccccc}
		    \toprule
		    Algorithm & $<0.02$ & $<0.03$ & $<0.04$ & Mean & Std & Median 
		    \\
		    \midrule
		    Initialization & 0.000 & 0.004 & 0.055 & 0.080 & 0.028 & 0.078
		    \\ 
		    PO\_For\_GN\_Sch & 0.470 & 0.729 & 0.799 & 0.031 & 0.029 & 0.021
		    \\
		    PO\_For\_GN\_Alt & 0.458 & 0.719 & 0.780 & 0.035 & 0.044 & 0.021
		    \\
		    PO\_Inv\_GN\_Sch & \textbf{0.637} & \textbf{0.891} & \textbf{0.938} & \textbf{0.023} & \textbf{0.021} & \textbf{0.018}
		    \\
		    PO\_Bid\_GN\_Sch & 0.528 & 0.802 & 0.862 & 0.030 & 0.039 & 0.020
		    \\
		    PO\_Bid\_GN\_Alt & 0.528 & 0.805 & 0.865 & 0.030 & 0.040 & 0.019
		    \\
		    \bottomrule
	  	\end{tabular}
	  	\caption{Table showing the proportion of images fitted with a normalized point-to-point error below $0.02$, $0.03$ and $0.04$ together with the normalized point-to-point error mean, std and median for all Project-Out Gauss-Newton algorithms initialized with $5\%$ uniform noise.}
	    \label{tab:stats_bpo_gn_5}
	\end{subfigure}
	\caption{Results showing the fitting accuracy and convergence properties of the Project-Out Gauss-Newton algorithms on the LFPW test dataset.}
	\label{fig:bpo_gn_5}
\end{figure*}

Results for \emph{Project-Out Gauss-Newton} algorithms are reported on Figure \ref{fig:bpo_gn_5}. We can observe that, there is significant drop in terms of fitting accuracy for \emph{Inverse} and \emph{Bidirectional} algorithms with respect to their \emph{SSD} versions, \ref{fig:ced_bpo_gn_5} and Table \ref{tab:stats_bpo_gn_5}. As expected, the \emph{Forward} algorithm achieves virtually the same results as its \emph{SSD} counterpart. The \emph{Asymmetric} algorithm obtains similar accuracy to that of the best performing \emph{SSD} algorithms.

Looking at Figures \ref{fig:mean_error_vs_iters_bpo_gn_5}, \ref{fig:mean_cost_vs_iters1_bpo_gn_5} and \ref{fig:mean_cost_vs_iters2_bpo_gn_5} we can see that \emph{Inverse} and \emph{Bidirectional} algorithms converge slightly faster than the \emph{Asymmetric} algorithm. However, the \emph{Asymmetric} algorithm ends up descending to a significant lower value of the mean normalized cost which also translates to a lower value for the final mean normalized point-to-point error. Similar to \emph{SSD} algorithms, the \emph{Forward} algorithm  is the worst convergent algorithm.

Finally notice that, in this case, there is virtually no difference, in terms of both final fitting accuracy and speed of convergence, between the \emph{Simultaneous Schur} and \emph{Alternated} optimizations strategies used by the \emph{Bidirectional} algorithm.

\subsubsection{Project-Out Newton algorithms}

\begin{figure*}[p]
	\centering
	\begin{subfigure}{0.48\textwidth}
	    \includegraphics[width=\textwidth]{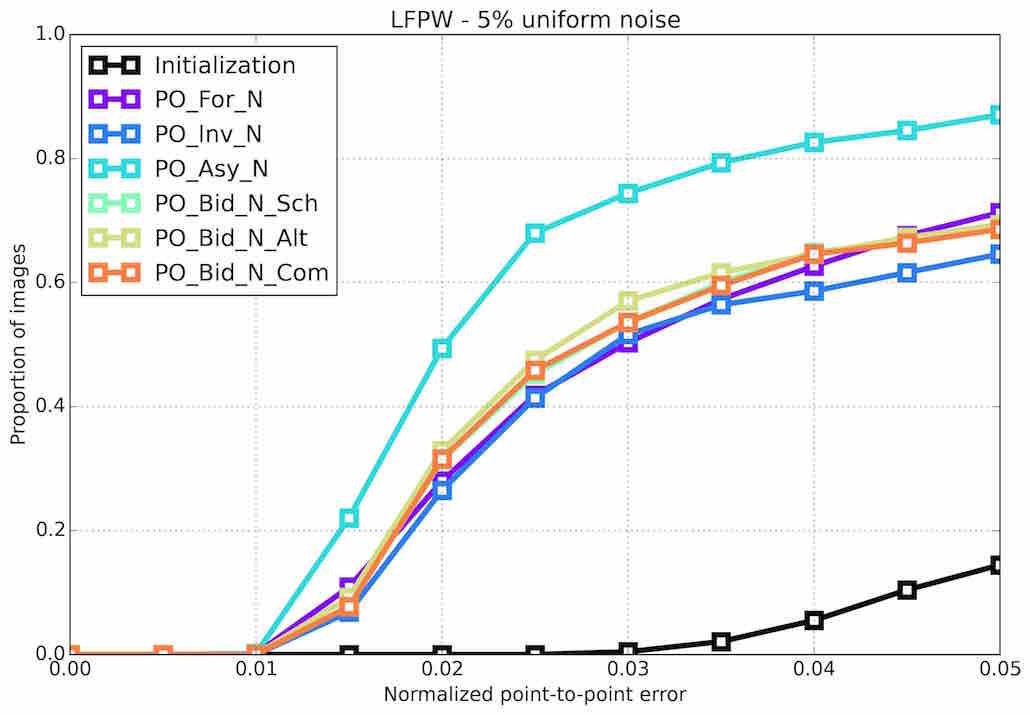}
	    \caption{CED graph on the LFPW test dataset for all Project-Out Newton algorithms initialized with $5\%$ uniform noise.}
	    \label{fig:ced_bpo_n_5}
	\end{subfigure}
	\hfill
	\begin{subfigure}{0.48\textwidth}
	    \includegraphics[width=\textwidth]{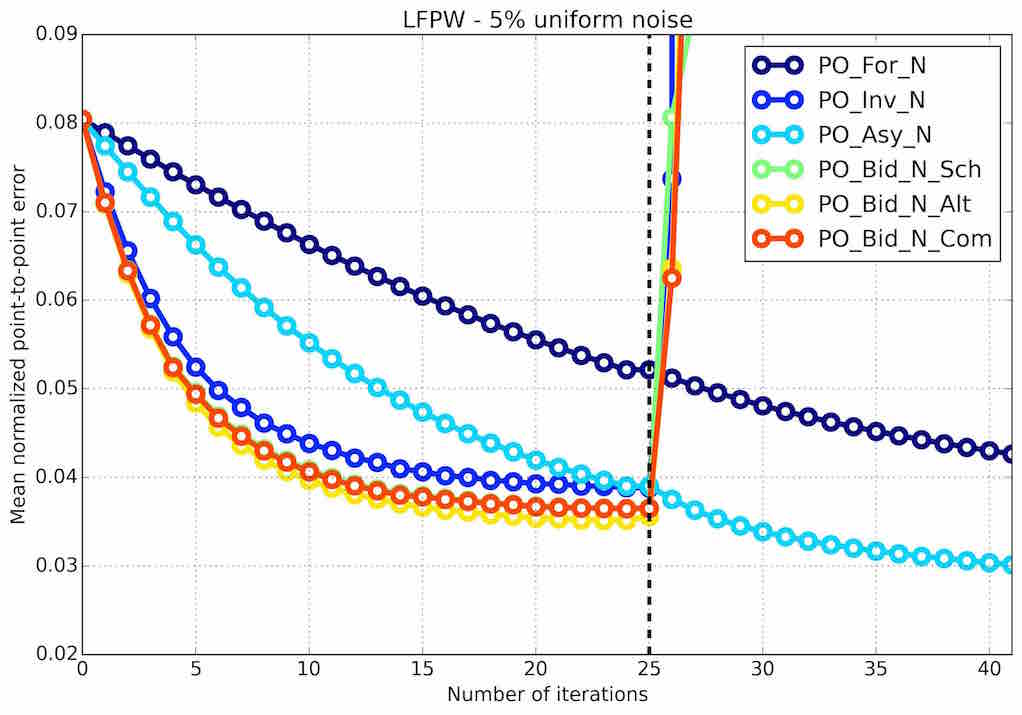}
	    \caption{Mean normalized point-to-point error vs number of iterations on the LFPW test dataset for all Project-Out Newton algorithms initialized with $5\%$ uniform noise.}
	    \label{fig:mean_error_vs_iters_bpo_n_5}
	\end{subfigure}
	\par\bigskip\bigskip
	\begin{subfigure}{0.48\textwidth}
	    \includegraphics[width=\textwidth]{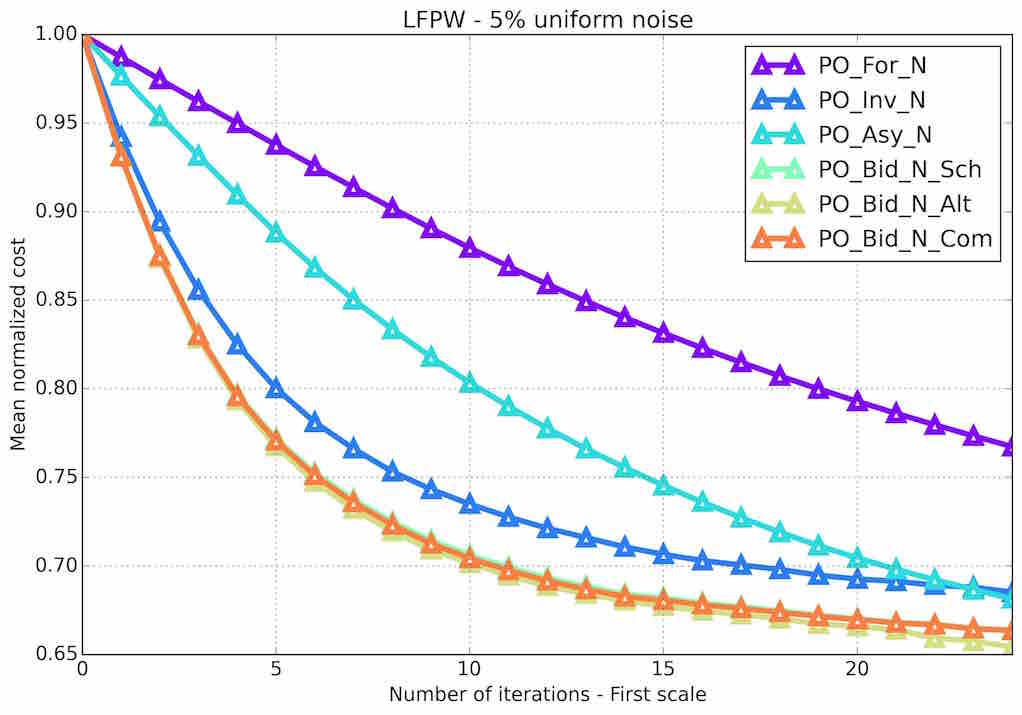}
	    \caption{Mean normalized cost vs number of first scale iterations on the LFPW test dataset for all Project-Out Newton algorithms initialized with $5\%$ uniform noise.}
	    \label{fig:mean_cost_vs_iters1_bpo_n_5}
	\end{subfigure}
	\hfill
	\begin{subfigure}{0.48\textwidth}
	    \includegraphics[width=\textwidth]{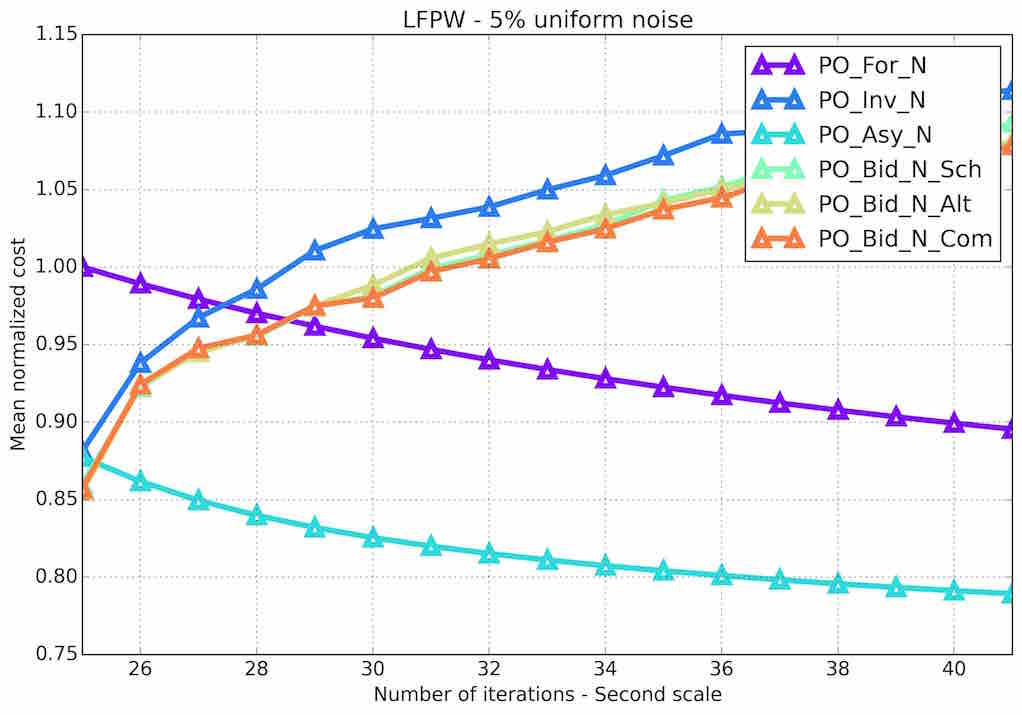}
	    \caption{Mean normalized cost vs number of second scale iterations on the LFPW test dataset for all Project-Out Newton algorithms initialized with $5\%$ uniform noise.}
	    \label{fig:mean_cost_vs_iters2_bpo_n_5}
	\end{subfigure}
	\par\bigskip\bigskip
	\begin{subfigure}{\textwidth}
		\center
		\begin{tabular}{lcccccc}
		    \toprule
		    Algorithm & $<0.02$ & $<0.03$ & $<0.04$ & Mean & Std & Median 
		    \\
		    \midrule
		    Initialization & 0.000 & 0.004 & 0.055 & 0.080 & 0.028 & 0.078
		    \\ 
		    PO\_For\_N\_Sch & 0.280 & 0.503 & 0.626 & 0.043 & 0.033 & 0.030
		    \\
		    PO\_Inv\_N\_Alt & 0.265 & 0.516 & 0.586 & 11.929 & 179.525 & 0.029
		    \\
		    PO\_Asy\_N\_Sch & \textbf{0.494} & \textbf{0.744} & \textbf{0.826} & \textbf{0.030} & \textbf{0.028} & \textbf{0.020}
		    \\
		    PO\_Bid\_N\_Sch & 0.314 & 0.536 & 0.649 & 0.287 & 1.347 & 0.027
		    \\
		    PO\_Bid\_N\_Alt & 0.329 & 0.570 & 0.649 & 0.280 & 1.465 & 0.026
		    \\
		    \bottomrule
	  	\end{tabular}
	  	\caption{Table showing the proportion of images fitted with a normalized point-to-point error below $0.02$, $0.03$ and $0.04$ together with the normalized point-to-point error mean, std and median for all Project-Out Newton algorithms initialized with $5\%$ uniform noise.}
	    \label{tab:stats_bpo_n_5}
	\end{subfigure}
	\caption{Results showing the fitting accuracy and convergence properties of the Project-Out Newton algorithms on the LFPW test dataset.}
	\label{fig:bpo_n_5}
\end{figure*}

Results for \emph{Project-Out Newton} algorithms are reported on Figure \ref{fig:bpo_n_5}. It can be clearly seen that \emph{Project-Out Newton} algorithms perform much worse than their \emph{Gauss-Newton} and \emph{SSD} counterparts. The final fitting accuracy obtained by these algorithms is very poor compared to the one obtained by the best \emph{SSD} and \emph{Project-Out Gauss-Newton} algorithms, Figures \ref{fig:ced_bpo_n_5} and Table \ref{tab:stats_bpo_n_5}. In fact, by looking at Figures \ref{fig:mean_error_vs_iters_bpo_n_5}, \ref{fig:mean_cost_vs_iters1_bpo_n_5} and \ref{fig:mean_cost_vs_iters2_bpo_n_5} only the \emph{Forward} and \emph{Asymmetric} algorithms seem to be stable at the second level of the Gaussian pyramid with \emph{Inverse} and \emph{Bidirectional} algorithms completely diverging for some of the images as shown by the large mean and std of their final normalized point-to-point errors.

\subsubsection{Project-Out Wiberg algorithms}

Results for the \emph{Project-Out Bidirectional Wiberg} algorithm are reported on Figure \ref{fig:bpo_n_5}. As expected, the results are virtually identical to those of the obtained by \emph{Project-Out Bidirectional Gauss-Newton} algorithms.  



\begin{figure*}[t!]
	\centering
	\begin{subfigure}{0.16\textwidth}
		\includegraphics[width=\textwidth]{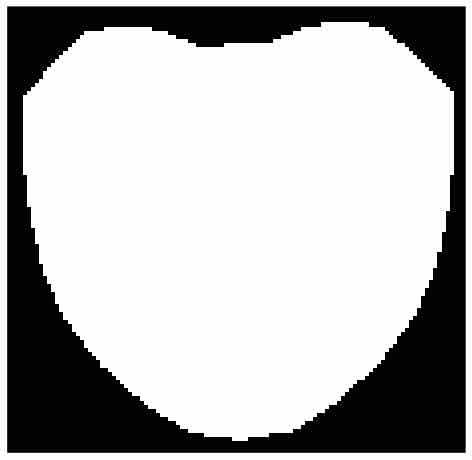}
		\caption{$100\%$}
		\label{fig:sampling_100}
	\end{subfigure}
	\begin{subfigure}{0.16\textwidth}
		\includegraphics[width=\textwidth]{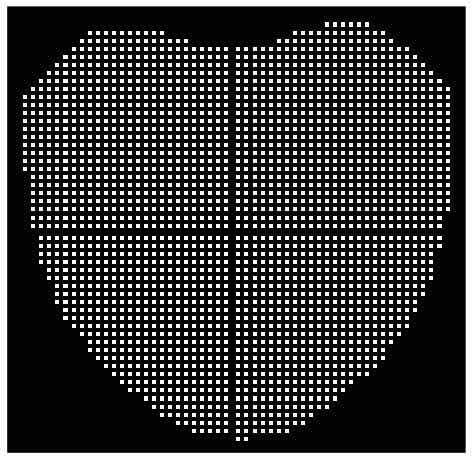}
		\caption{$50\%$}
		\label{fig:sampling_50}
	\end{subfigure}
	\begin{subfigure}{0.16\textwidth}
		\includegraphics[width=\textwidth]{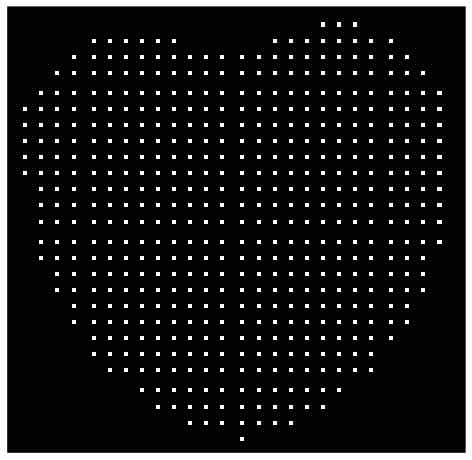}
		\caption{$25\%$}
		\label{fig:sampling_25}
	\end{subfigure}
	\begin{subfigure}{0.16\textwidth}
		\includegraphics[width=\textwidth]{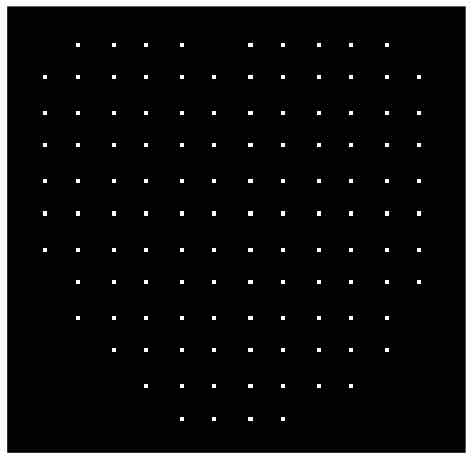}
		\caption{$12\%$}
		\label{fig:ini_12}
	\end{subfigure}
	\caption{Subset of pixels on the reference frame used to optimize the SSD and Project-Out cost functions for different sampling rates.}
    \label{fig:sampling_masks}
\end{figure*}

\subsection{Weighted Bayesian project-out}
\label{exp:2}

In this experiment, we quantify the importance of each of the two terms in our Bayesian project-out cost function, Equation \ref{eq:prob_po}. To this end, we introduce the parameters, $\rho \in [0, 1]$ and $\gamma = 1 - \rho$, to weight up the relative contribution of both terms:
\begin{equation}
    \begin{aligned}
        \rho|| \mathbf{i}[\mathbf{p}] - \mathbf{\bar{a}} ||^2_{\mathbf{A}\mathbf{D}^{-1}\mathbf{A}^T} 
        + 
        \frac{\gamma}{\sigma^2}|| \mathbf{i}[\mathbf{p}] - \mathbf{\bar{a}} ||^2_{\bar{\mathbf{A}}}
    \end{aligned}
    \label{eq:weighted_po}
\end{equation}
Setting $\rho=0$, $\gamma=1$ reduces the previous cost function to the original project-out loss proposed in \cite{Matthews2004}; completely disregarding the contribution of the prior distribution over the appearance parameters i.e the Mahalanobis distance \emph{within} the appearance subspace. On the contrary, setting $\rho=1$, $\gamma=0$ reduces the cost function to the first term; completely disregarding the contribution of the project-out term i.e. the distance \emph{to} the appearance subspace. Finally setting $\rho=\gamma=0.5$ leads to the standard Bayesian project-out cost function proposed in Section \ref{sec:po_pi}.
 
In order to assess the impact that each term has on the fitting accuracy obtained by the previous \emph{Project-Out} algorithm we repeat the experimental set up of the first experiment and test all \emph{Project-Out Gauss-Newton} algorithms for different values of the parameters $\rho = 1 -\gamma$. Notice that, in this case, we only report the performance of \emph{Gauss-Newton} algorithms because they were shown to vastly outperform \emph{Newton} algorithms and to be virtually equivalent to \emph{Wiberg} algorithms in the first experiment.

\begin{figure*}[p]
	\centering
	\begin{subfigure}{\textwidth}
	    \includegraphics[width=\textwidth]{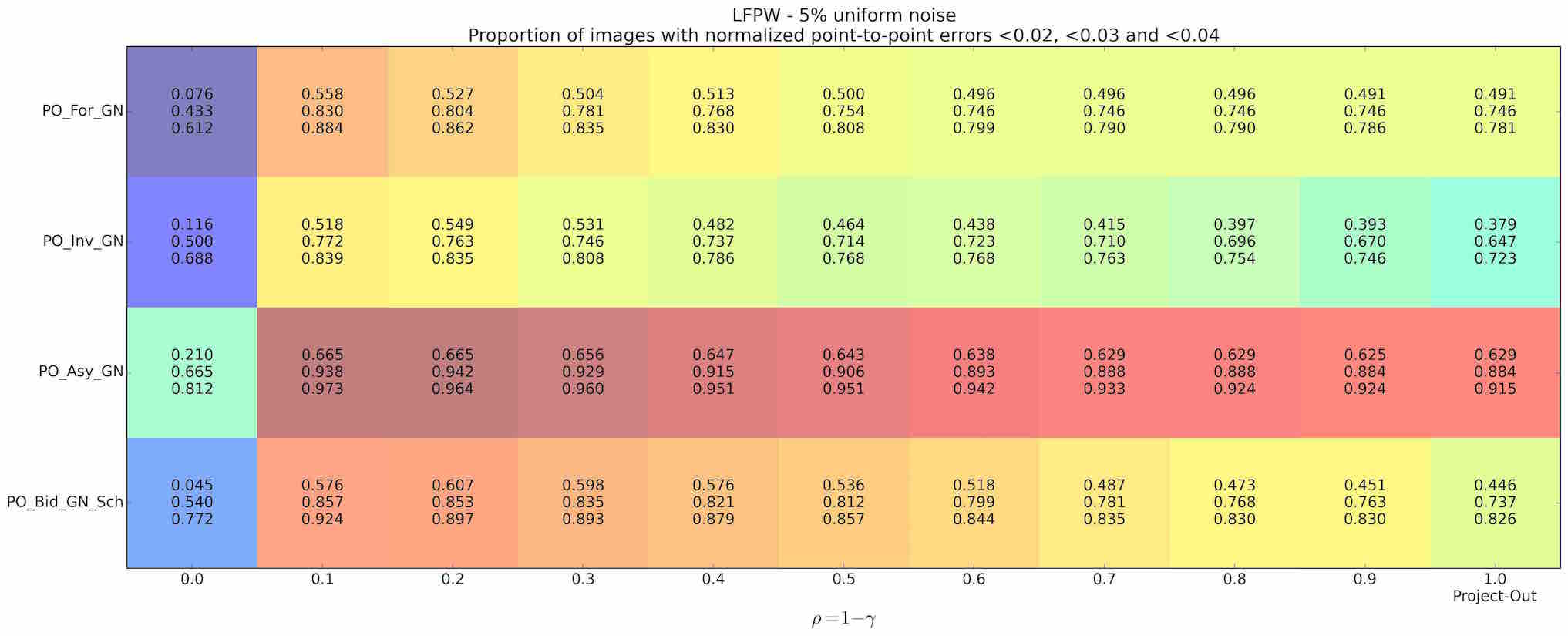}
	    \caption{Proportion of images with normalized point-to-point errors smaller than $0.02$, $0.03$ and $0.04$ for the Project-Out and SSD Asymmetric Gauss-Newton algorithms for different values of $\rho = 1 -\gamma$  and initialized with $5\%$ noise. Colors encode overall fitting accuracy, from highest to lowest: red, orange, yellow, green, blue and purple.}
	    \label{fig:convergence_vs_rho_po_gn}
	\end{subfigure}
	\par\bigskip
	\begin{subfigure}{0.48\textwidth}
	    \includegraphics[width=\textwidth]{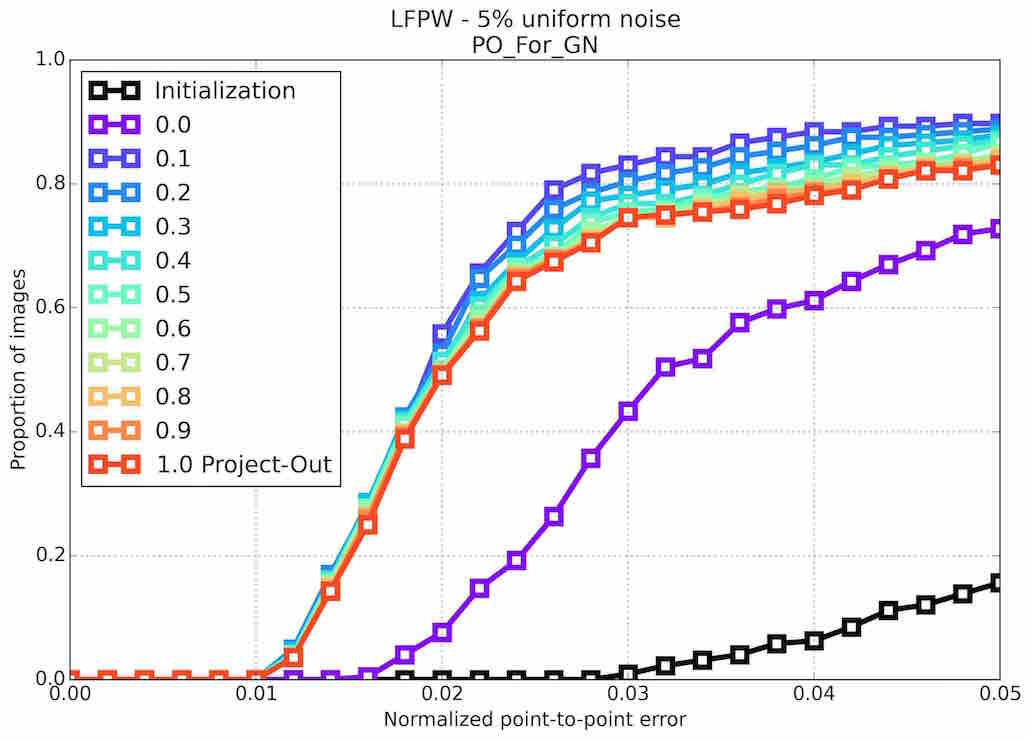}
	    \caption{CED on the LFPW test dataset for Project-Out Forward Gauss-Newton algorithms for different values of $\rho = 1 -\gamma$  and initialized with $5\%$ noise.}
	    \label{fig:ced_po_for_gn}
	\end{subfigure}
	\hfill
	\begin{subfigure}{0.48\textwidth}
	    \includegraphics[width=\textwidth]{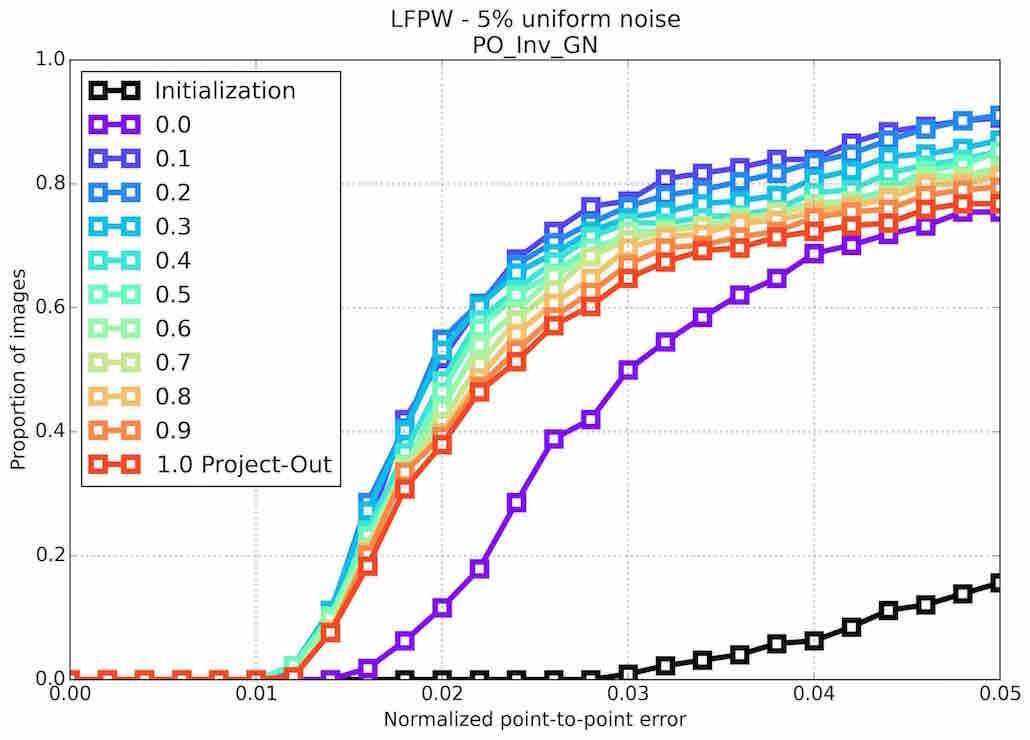}
	    \caption{CED on the LFPW test dataset for Project-Out Inverse Gauss-Newton algorithms for different values of $\rho = 1 -\gamma$ and initialized with $5\%$ noise.}
	    \label{fig:ced_po_inv_gn}
	\end{subfigure}
	\par\bigskip
	\begin{subfigure}{0.48\textwidth}
	    \includegraphics[width=\textwidth]{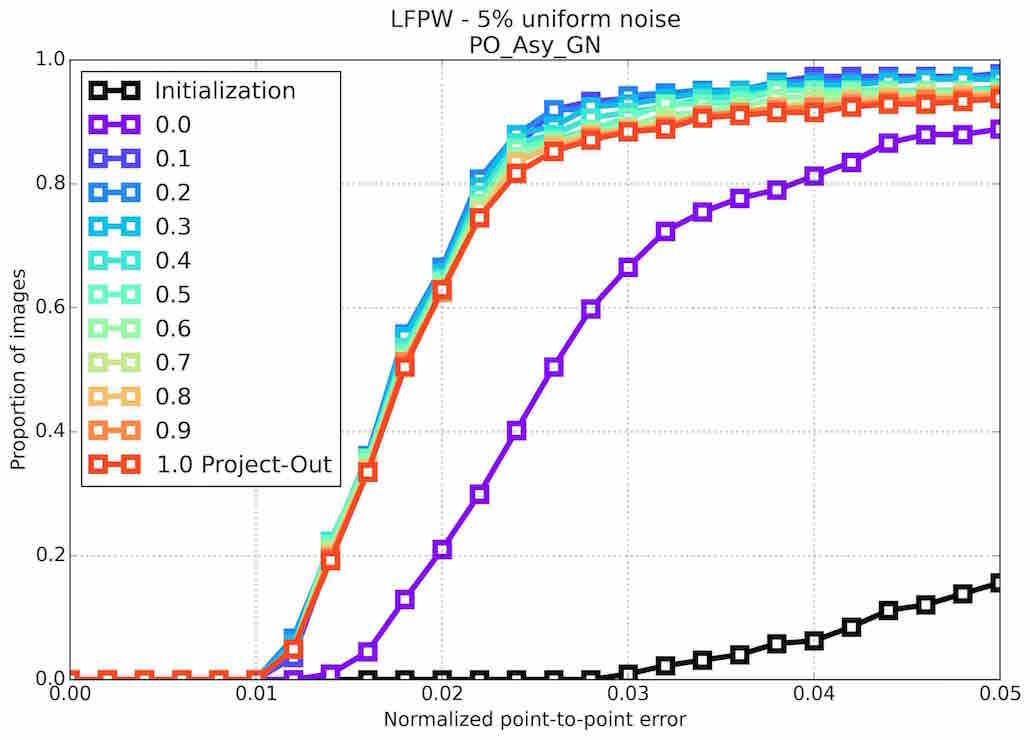}
	    \caption{CED on the LFPW test dataset for Project-Out Asymmetric Gauss-Newton algorithms for different values of $\rho = 1 -\gamma$  and initialized with $5\%$ noise.}
	    \label{fig:ced_po_asy_gn}
	\end{subfigure}
	\hfill
	\begin{subfigure}{0.48\textwidth}
	    \includegraphics[width=\textwidth]{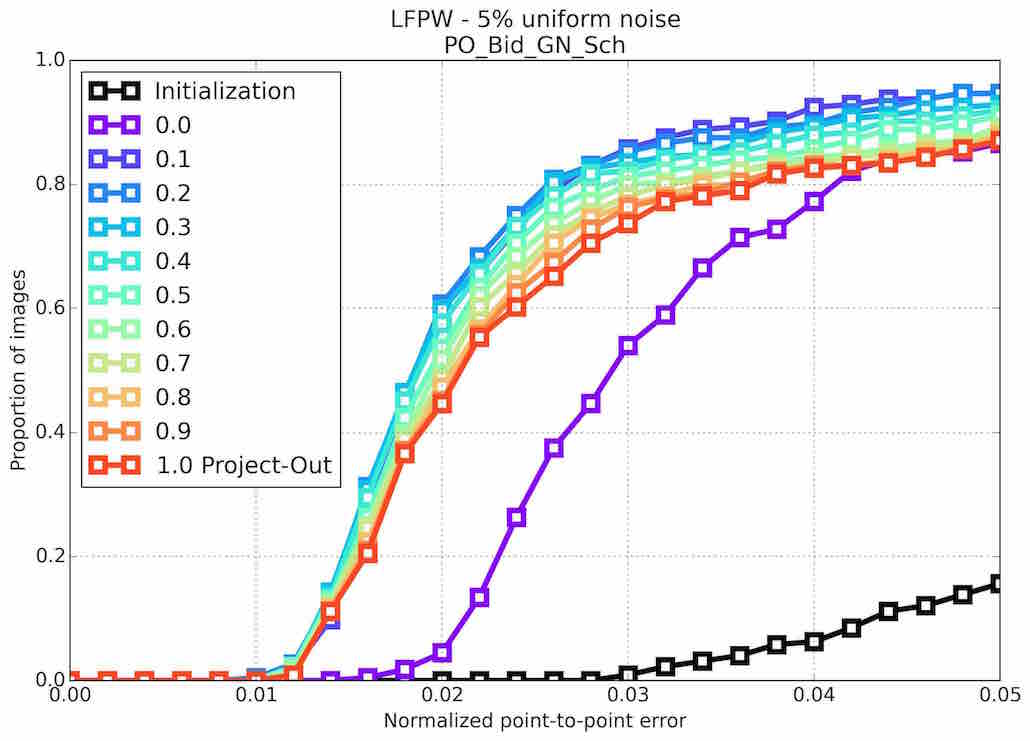}
	    \caption{CED on the LFPW test dataset for Project-Out Bidirectional Gauss-Newton algorithms for different values of $\rho = 1 -\gamma$  and initialized with $5\%$ noise.}
	    \label{fig:ced_po_bid_gn}
	\end{subfigure}
	\caption{Results quantifying the effect of varying the value of the parameters $\rho = 1 -\gamma$ in Project-Out Gauss-Newton algorithms.}
	\label{fig:rho}
\end{figure*}

Results for this experiment are reported by Figure \ref{fig:rho}. We can see that, regardless of the type of composition, a weighted combination of the two previous terms always leads to a smaller mean normalized point-to-point error compared to either term on its own. Note that the final fitting accuracy obtained with the standard Bayesian project-out cost function is substantially better than the one obtained by the original project-out loss (this is specially noticeable for the \emph{Inverse} and \emph{Bidirectional} algorithms); fully justifying the inclusion of the first term, i.e the Mahalanobis distance \emph{within} the appearance subspace, into the cost function. Finally, in this particular experiment, the final fitting accuracy of all algorithms is maximized by setting $\rho=0.1$, $\gamma=0.9$, further highlighting the importance of the first term in the Bayesian formulation.


\subsection{Optimal asymmetric composition}
\label{exp:3}

This experiment quantifies the effect that varying the value of the parameters $\alpha \in [0, 1]$ and $\beta = 1 -\alpha$ in Equation \ref{eq:ssd_ac} has in the fitting accuracy obtained by the \emph{Asymmetric} algorithms. Note that for $\alpha=1$, $\beta=0$ and $\alpha=0$, $\beta=1$ these algorithms reduce to their \emph{Forward} and \emph{Inverse} versions respectively. Recall that, in previous experiments, we used the \emph{Symmetric} case $\alpha=\beta=0.5$ to generate the results reported for \emph{Asymmetric} algorithms. Again, we only report performance for \emph{Gauss-Newton} algorithms.

\begin{figure*}[p]
	\centering
	\begin{subfigure}{\textwidth}
	    \includegraphics[width=\textwidth]{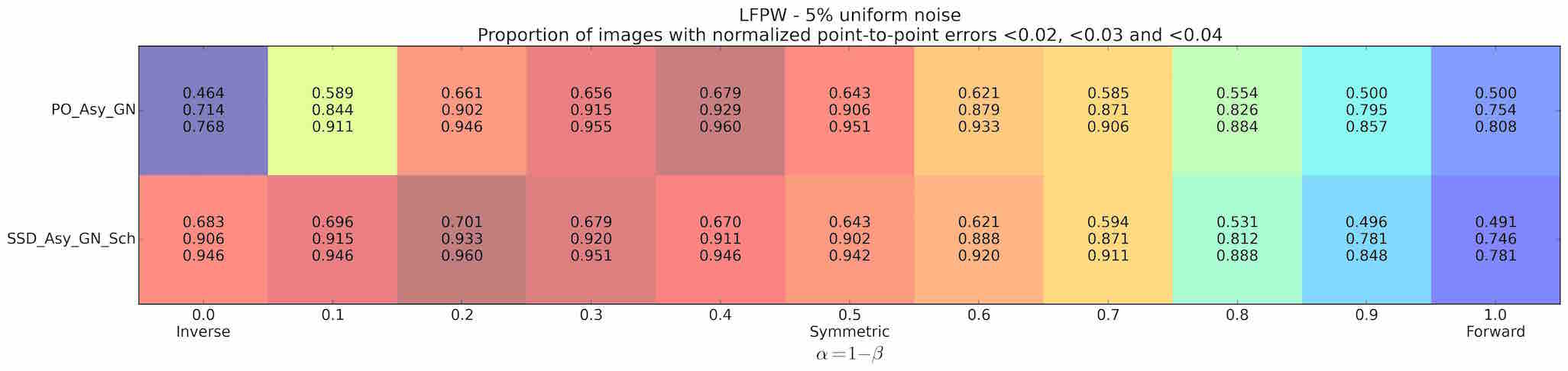}
	    \caption{Proportion of images with normalized point-to-point errors smaller than $0.02$, $0.03$ and $0.04$ for the Project-Out and SSD Asymmetric Gauss-Newton algorithms for different values of $\alpha = 1 - \beta$ and initialized with $5\%$ noise. Colors encode overall fitting accuracy, from highest to lowest: red, orange, yellow, green, blue and purple.}
	    \label{fig:asy_gn_vs_alpha_5}
	\end{subfigure}
	\par\bigskip\bigskip
	\begin{subfigure}{0.48\textwidth}
	    \includegraphics[width=\textwidth]{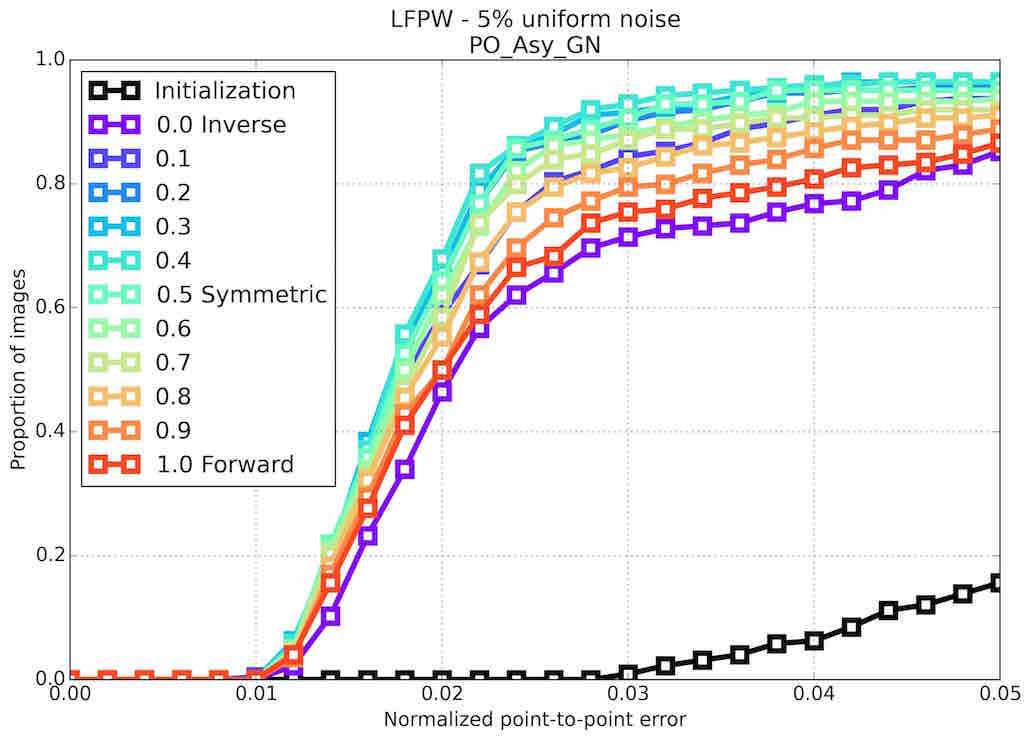}
	    \caption{CED on the LFPW test dataset for Project-Out Asymmetric Gauss-Newton algorithm for different values of $\alpha = 1 - \beta$ and initialized with $5\%$ noise.}
	    \label{fig:ced_po_asy_gn_5}
	\end{subfigure}
	\hfill
	\begin{subfigure}{0.48\textwidth}
	    \includegraphics[width=\textwidth]{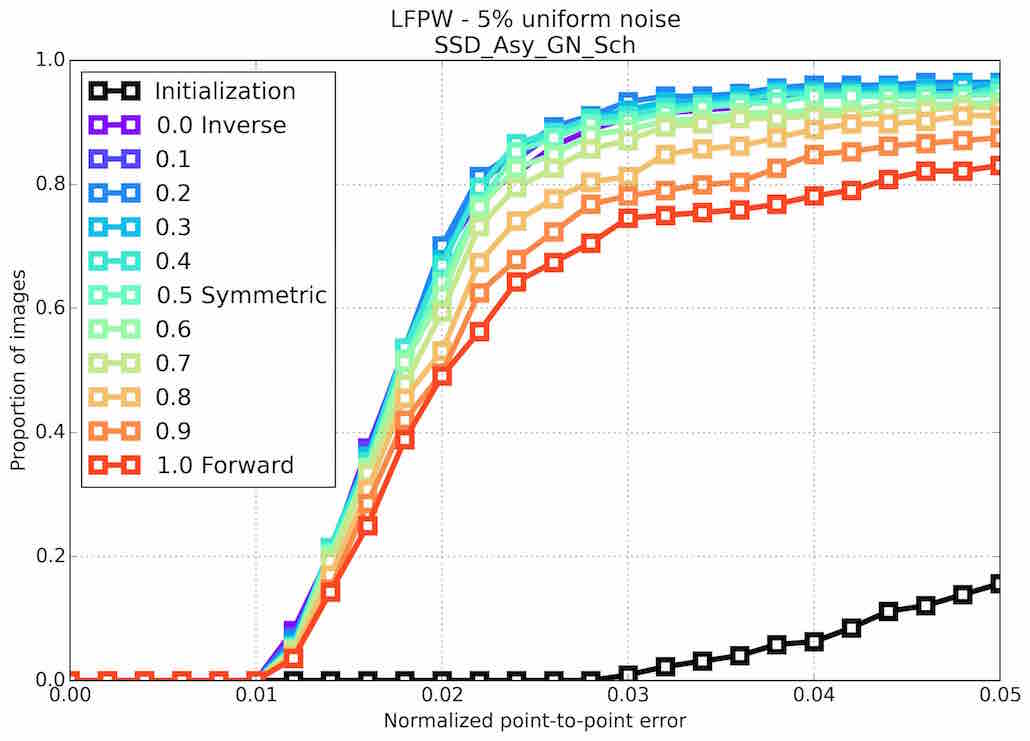}
	    \caption{CED on the LFPW test dataset for the the SSD Asymmetric Gauss-Newton algorithm for different values of $\alpha = 1 - \beta$ and initialized with $5\%$ noise.}
	    \label{fig:ced_ssd_asy_gn_5}
	\end{subfigure}
	\caption{Results quantifying the effect of varying the value of the parameters $\alpha = 1 - \beta$ in Asymmetric algorithms.}
	\label{fig:alpha}
\end{figure*}

We again repeat the experimental set up described in the first experiments and report the fitting accuracy obtained by the \emph{Project Out} and \emph{SSD Asymmetric Gauss-Newton} algorithms for different values of the parameters $\alpha = 1 - \beta$. Results are shown in Figure \ref{fig:alpha}. For the \emph{BPO Asymmetric} algorithm, the best results are obtain by setting $\alpha=0.4$, $\beta=0.6$, Figures \ref{fig:asy_gn_vs_alpha_5} (top) and \ref{fig:ced_po_asy_gn_5}. These results slightly outperform those obtain by the default \emph{Symmetric} algorithm and this particular configuration of the \emph{BPO Asymmetric} algorithm is the best performing one on the LFPW test dataset. For the \emph{SSD Asymmetric Gauss-Newton} algorithm the best results are obtained by setting $\alpha=0.2$, $\beta=0.8$, Figures \ref{fig:asy_gn_vs_alpha_5} (bottom) and \ref{fig:ced_ssd_asy_gn_5}. In this case, the boost in performance with respect to the default \emph{Symmetric} algorithm is significant and, with this particular configuration, the \emph{SSD Asymmetric Gauss-Newton} algorithm is the best performing \emph{SSD} algorithm on the LFPW test dataset, outperforming \emph{Inverse} and \emph{Bidirectional} algorithms.


\subsection{Sampling and Number of Iterations}
\label{exp:4}

In this experiment, we explore two different strategies to reduce the running time of the previous CGD algorithms. 

The first one consists of optimizing the SSD and Project-Out cost functions using only a subset of all pixels in the reference frame. In AAMs the total number of pixels on the reference frame, $F$, is typically several orders of magnitude bigger than the number of shape, $n$, and appearance, $m$, components i.e. $F>>m>>n$. Therefore, a significant reduction in the complexity (and running time) of CGD algorithms can be obtained by decreasing the number of pixels that are used to optimize the previous cost functions. To this end, we compare the accuracy obtained by using $100\%$, $50\%$, $25\%$ and $12\%$ of the total number of pixels on the reference frame. Note that, pixels are (approximately) evenly sampled across the reference frame in all cases, Figure \ref{fig:sampling_masks}.

The second strategy consists of simply reducing the number of iterations that each algorithm is run. Based on the figures used to assess the convergence properties of CGD algorithms in previous experiments, we compare the accuracy obtained by running the algorithms for $40$ $(24 + 16)$ and $20$ $(12 + 8)$ iterations.

Note that, in order to further highlight the advantages and disadvantages of using the previous strategies we report the fitting accuracy obtained by initializing the algorithms using different amounts of uniform noise.

\begin{figure*}[p]
	\centering
	\begin{subfigure}{0.48\textwidth}
	    \includegraphics[width=\textwidth]{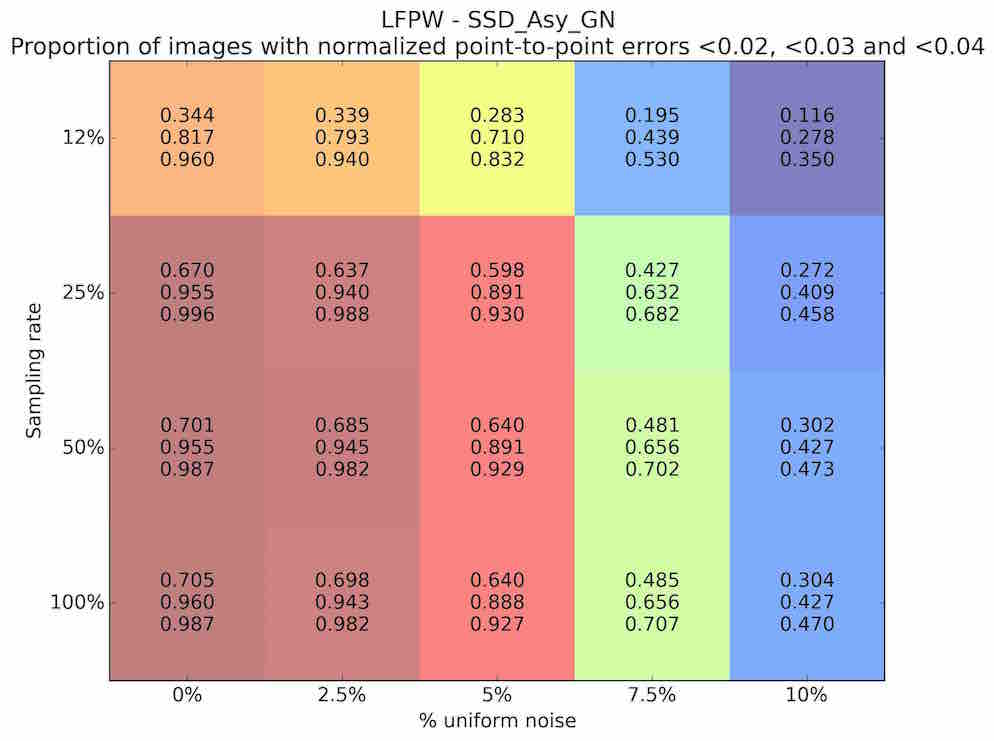}
	    \caption{Proportion of images with normalized point-to-point errors smaller than $0.02$, $0.03$ and $0.04$ for the SSD Asymmetric Gauss-Newton algorithm using different sampling rates, $40$ $(24 + 16)$ iterations, and initialized with different amounts of noise. Colors encode overall fitting accuracy, from highest to lowest: red, orange, yellow, green, blue and purple.}
	    \label{fig:sampling_vs_noise_ssd_asy_gn}
	\end{subfigure}
	\hfill
	\begin{subfigure}{0.48\textwidth}
	    \includegraphics[width=\textwidth]{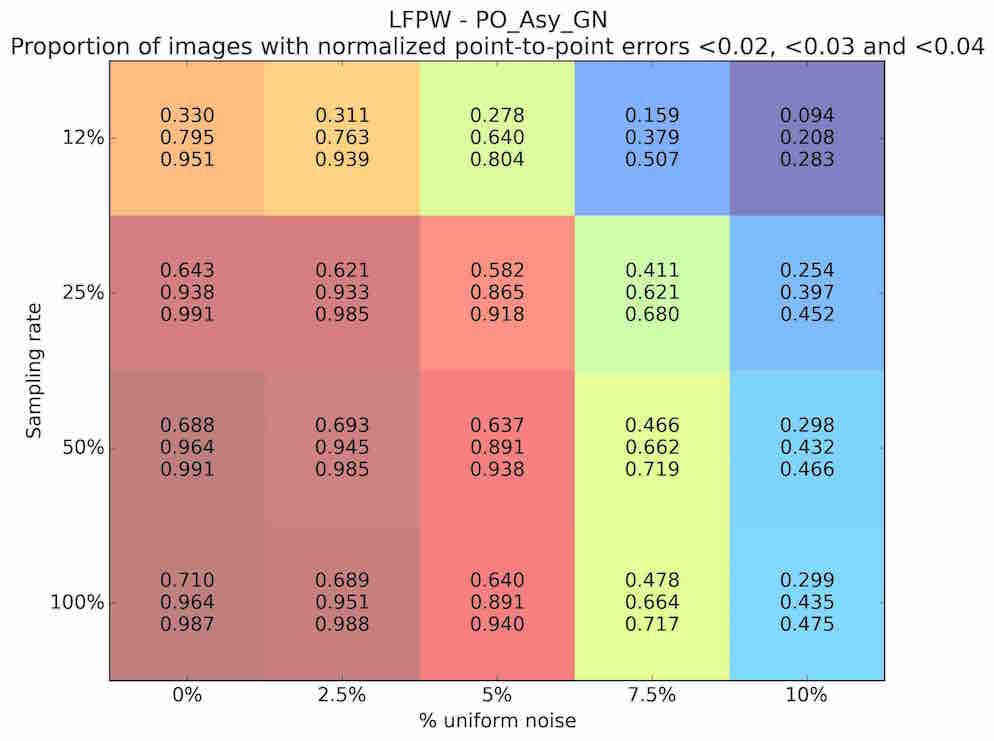}
	    \caption{Proportion of images with normalized point-to-point errors smaller than $0.02$, $0.03$ and $0.04$ for the Project-Out Asymmetric Gauss-Newton algorithm using different sampling rates, $40$ $(24 + 16)$ iterations, and initialized with different amounts of noise. Colors encode overall fitting accuracy, from highest to lowest: red, orange, yellow, green, blue and purple.}
	    \label{fig:sampling_vs_noise_po_asy_gn}
	\end{subfigure}
	\par\bigskip
	\begin{subfigure}{\textwidth}
		\center
		\begin{tabular}{lcccccc}
		    \toprule
		    & $100\%$ & $<50\%$ & $<25\%$ & $<12\%$ 
		    \\
		    \midrule
		    SSD\_Asy\_GN\_Sch & $\sim1680$ ms & $\sim930$ ms & $\sim650$ ms & $\sim590$ ms
		    \\ 
		    PO\_Asy\_GN & $\sim1400$ ms & $\sim680$ ms & $\sim480$ ms & $\sim380$ ms
		    \\
		    \bottomrule
	  	\end{tabular}
	  	\caption{Table showing run time of each algorithm for different amounts of sampling and $40$ $(24 + 16)$ iterations.}
	    \label{tab:runtime_40}
	\end{subfigure}
	\par\bigskip
	\begin{subfigure}{0.48\textwidth}
	    \includegraphics[width=\textwidth]{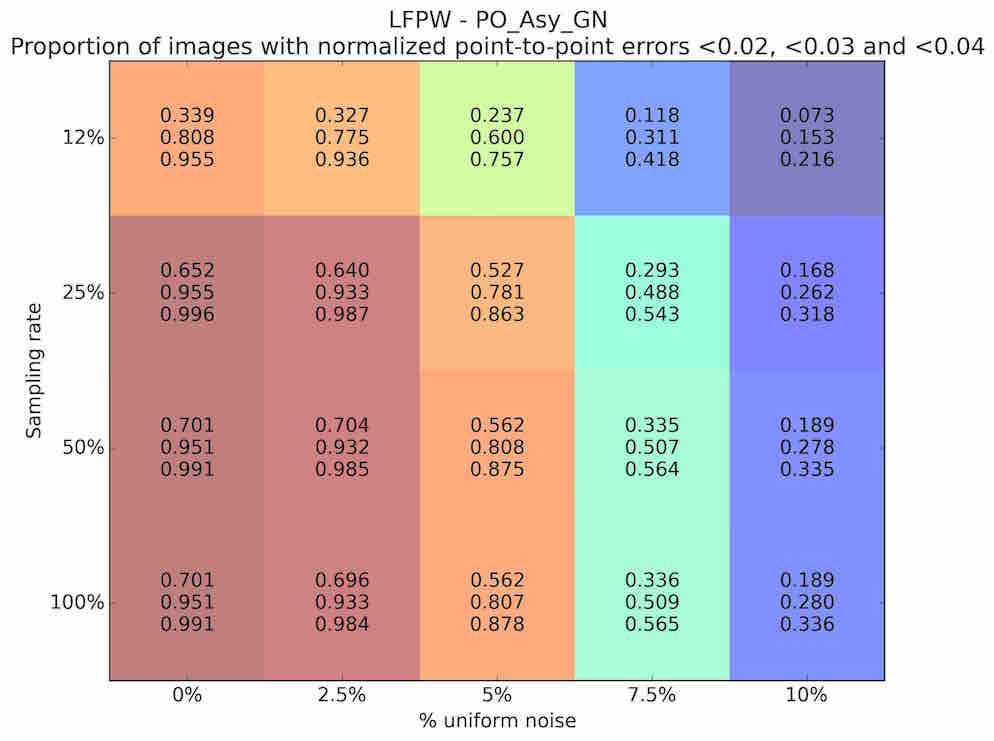}
	    \caption{Proportion of images with normalized point-to-point errors smaller than $0.02$, $0.03$ and $0.04$ for the Project-Out Asymmetric Gauss-Newton algorithm using different sampling rates, $20$ $(12 + 8)$ iterations, and initialized with different amounts of noise. Colors encode overall fitting accuracy, from highest to lowest: red, orange, yellow, green, blue and purple.}
	    \label{fig:sampling_vs_noise_ssd_asy_gn_20}
	\end{subfigure}
	\hfill
	\begin{subfigure}{0.48\textwidth}
	    \includegraphics[width=\textwidth]{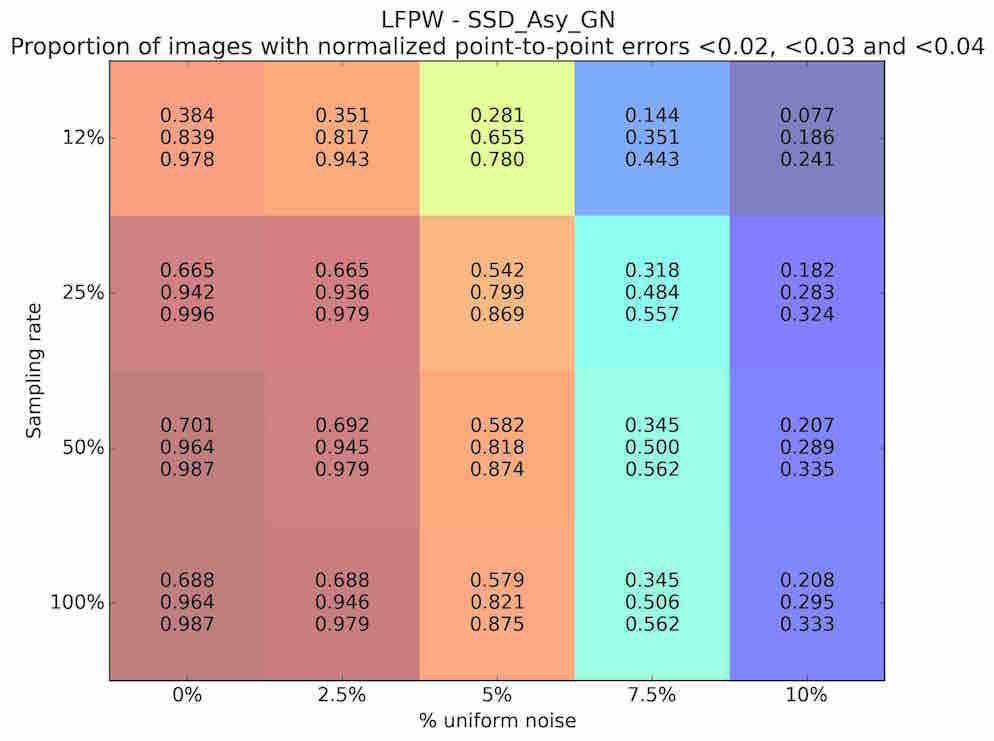}
	    \caption{Proportion of images with normalized point-to-point errors smaller than $0.02$, $0.03$ and $0.04$ for the SSD Asymmetric Gauss-Newton algorithm using different sampling rates, $20$ $(12 + 8)$ iterations, and initialized with different amounts of noise. Colors encode overall fitting accuracy, from highest to lowest: red, orange, yellow, green, blue and purple.}
	    \label{fig:sampling_vs_noise_po_asy_gn_20}
	\end{subfigure}
	\par\bigskip
	\begin{subfigure}{\textwidth}
		\center
		\begin{tabular}{lcccccc}
		    \toprule
		    & $100\%$ & $<50\%$ & $<25\%$ & $<12\%$ 
		    \\
		    \midrule
		    SSD\_Asy\_GN\_Sch & $\sim892$ ms & $\sim519$ ms & $\sim369$ ms & $\sim331$ ms
		    \\
		    PO\_Asy\_GN & $\sim707$ ms & $\sim365$ ms & $\sim235$ ms & $\sim211$ ms
		    \\ 
		    \bottomrule
	  	\end{tabular}
	  	\caption{Table showing run time of each algorithm for different amounts of sampling and $20$ $(12 + 8)$ iterations.}
	    \label{tab:runtime_20}
	\end{subfigure}
	\caption{Results assessing the effectiveness of sampling for the best performing Project-Out and SSD algorithms on the LFPW database.}
	\label{fig:sampling}
\end{figure*}

Once more we repeat the experimental set up of the first experiment and report the fitting accuracy obtained by the Project Out and SSD Asymmetric Gauss-Newton algorithms. Results for this experiment are shown in Figure \ref{fig:sampling}. It can be seen that reducing the number of pixels up to $25\%$ while maintaining the original number of iterations to $40$ $(24 + 16)$ has little impact on the fitting accuracy achieved by both algorithms while reducing them to $12\%$ has a clear negative impact, Figures \ref{fig:sampling_vs_noise_ssd_asy_gn} and \ref{fig:sampling_vs_noise_po_asy_gn}. Also, performance seems to be consistent along the amount of noise. In terms of run time, Table \ref{tab:runtime_40}, reducing the number of pixels to $50\%$, $25\%$ and $12\%$ offers speed ups of $\sim2.0$x, $\sim2.9$x and $\sim3.7$x for the \emph{BPO} algorithm and of $\sim1.8$x, $\sim2.6$x and $\sim2.8$x for the \emph{SSD} algorithm respectively.  

On the other hand, reducing the number of iterations from $40$ $(24 + 16)$ to $20$ $(12 + 8)$ has no negative impact in performance for levels of noise smaller than $2\%$ but has a noticeable negative impact for levels of noise bigger than $5\%$. Notice that remarkable speed ups, Table \ref{tab:runtime_20}, can be obtain for both algorithms by combining the previous two strategies at the expenses of small but noticeable decreases in fitting accuracy.

\subsection{Comparison on Helen and AFW}
\label{exp:5}

In order to facilitate comparisons with recent prior work on AAMs \cite{Tzimiropoulos2013, Antonakos2014, Kossaifi2014} and with other state-of-the-art approaches in face alignment \cite{Xiong2013, Asthana2013}, in this experiment, we report the fitting accuracy of the \emph{SSD} and \emph{Project-Out Asymmetric Gauss-Newton} algorithms on the widely used test set of the Helen database and on the entire AFW database. Furthermore we compare the performance of the previous two algorithms with the one obtained by the recently proposed Gauss-Newton Deformable Part Models (GN-DPMs) proposed by Tzimiropoulos and Pantic in \cite{Tzimiropoulos2014}; which was shown to achieve state-of-the-art results in the problem of face alignment in-the-wild.

For both our algorithms, we report two different types of results:
\begin{inparaenum}[\itshape i\upshape)]
\item sampling rate of $25\%$ and $20$ $(12 + 8)$ iterations; and
\item sampling rate of $50\%$ and $40$ $(24 + 16)$ iterations,
\end{inparaenum}. For GN-DPMs we use the authors public implementation to generate the results. In this case, we report, again, two different types of results by letting the algorithm run for $20$ and $40$ iterations.

\begin{figure*}[p]
	\centering
	\begin{subfigure}{0.48\textwidth}
	    \includegraphics[width=\textwidth]{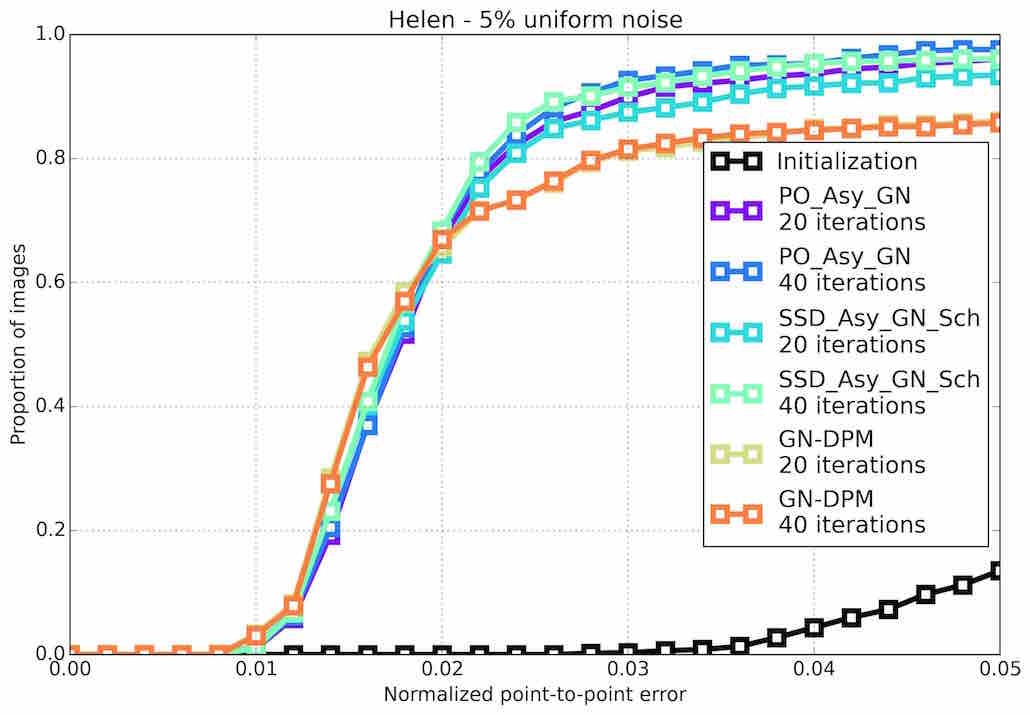}
	    \caption{CED on the Helen test dataset for the Project-Out and SSD Asymmetric Gauss-Newton algorithms initialized with $5\%$ noise.}
	    \label{fig:ced_helen}
	\end{subfigure}
	\hfill
	\begin{subfigure}{0.48\textwidth}
	    \includegraphics[width=\textwidth]{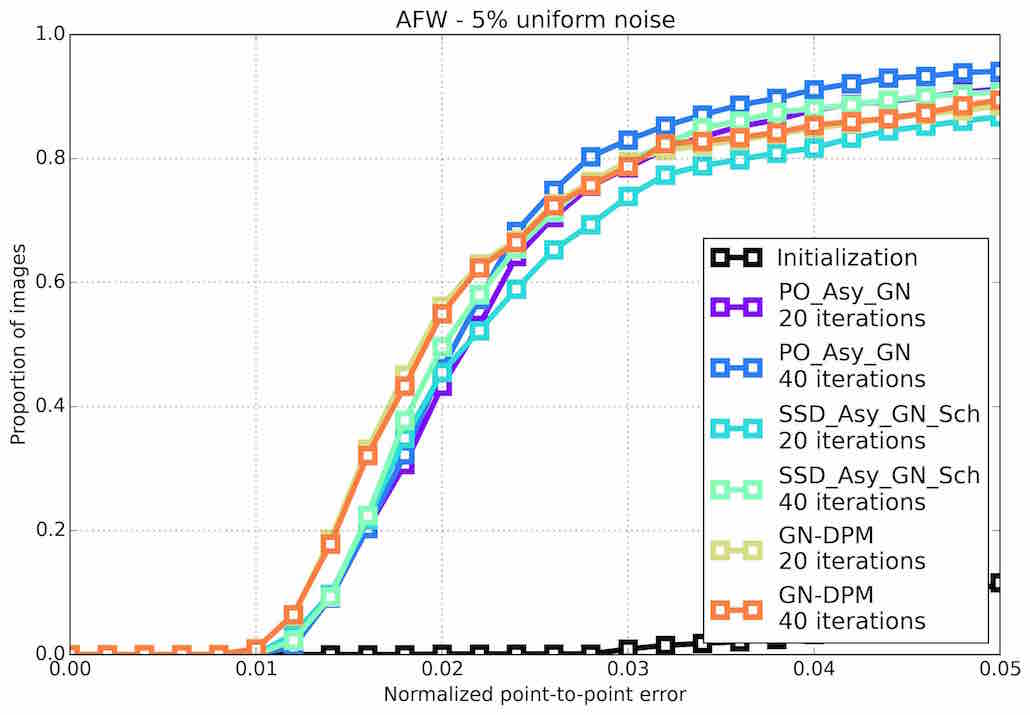}
	    \caption{CED on the AFW database for the Project-Out and SSD Asymmetric Gauss-Newton algorithm initialized with $5\%$ noise.}
	    \label{fig:ced_afw}
	\end{subfigure}
	\caption{Results showing the fitting accuracy of the SSD and Project-Out Asymmetric Gauss-Newton algorithms on the Helen and AFW databases.}
	\label{fig:helen_afw}
\end{figure*}

Result for this experiment are shown in Figure \ref{fig:helen_afw}. Looking at Figure \ref{fig:ced_helen} we can see that both, \emph{SSD} and \emph{Project-Out Asymmetric Gauss-Newton} algorithms, obtain similar fitting accuracy on the Helen test dataset. Note that, in all cases, their accuracy is comparable to the one achieved by GN-DPMs for normalized point-to-point errors $<0.2$ and significantly better for $<0.3$, $<0.4$. As expected, the best results for both our algorithms are obtained using $50\%$ of the total amount of pixels and $40$ $(24 + 16)$ iterations. However, the results obtained by using only $25\%$ of the total amount of pixels and $20$ $(12 + 8)$ iterations are comparable to the previous ones; specially for the \emph{Project-Out Asymmetric Gauss-Newton}. In general, these results are consistent with the ones obtained on the LFPW test dataset, Experiments \ref{exp:1} and \ref{exp:3}. 

On the other hand, the performance of both algorithms drops significantly on the AFW database, Figure \ref{fig:ced_afw} . In this case, GN-DPMs achieves slightly better results than the \emph{SSD} and \emph{Project-Out Asymmetric Gauss-Newton} algorithms for normalized point-to-point errors $<0.2$ and slightly worst for $<0.3$, $<0.4$. Again, both our algorithms obtain better results by using $50\%$ sampling rate and $40$ $(24 + 16)$ iterations and the difference in accuracy with respect to the versions using $25\%$ sampling rate and $20$ $(12 + 8)$ iterations slightly widens when compared to the results obtained on the Helen test dataset. This drop in performance is consistent with other recent works on AAMs \cite{Tzimiropoulos2014, Alabort2014, Antonakos2014, Alabort2015} and it is attributed to large difference in terms of shape and appearance statistics between the images of the AFW dataset and the ones of the training sets of the LFPW and Helen datasets where the AAM model was trained on. 

\begin{figure*}
	\begin{subfigure}{\textwidth}
	    \includegraphics[width=0.16\textwidth, height=0.16\textwidth]{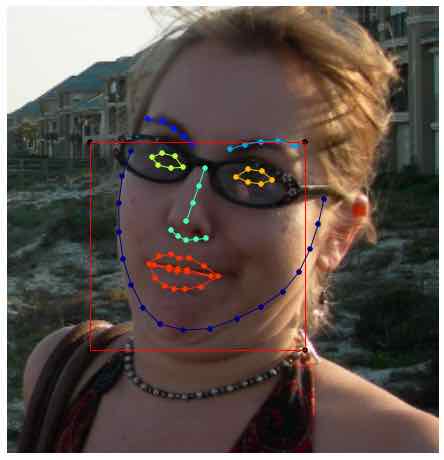}
		\includegraphics[width=0.16\textwidth, height=0.16\textwidth]{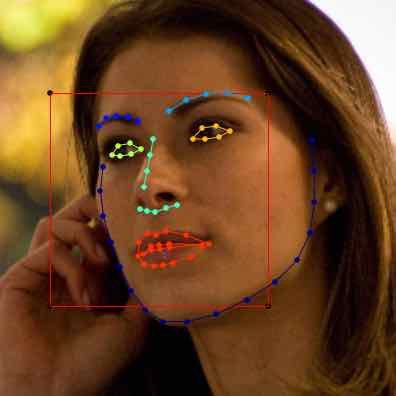}
		\includegraphics[width=0.16\textwidth, height=0.16\textwidth]{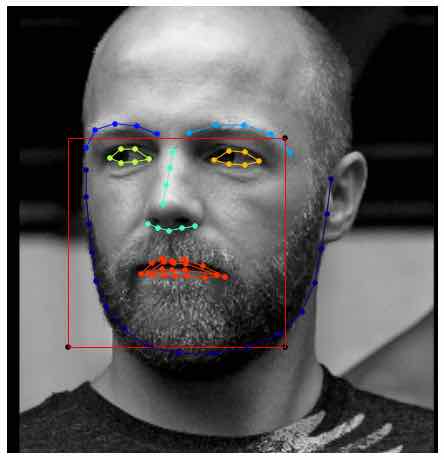}
		\includegraphics[width=0.16\textwidth, height=0.16\textwidth]{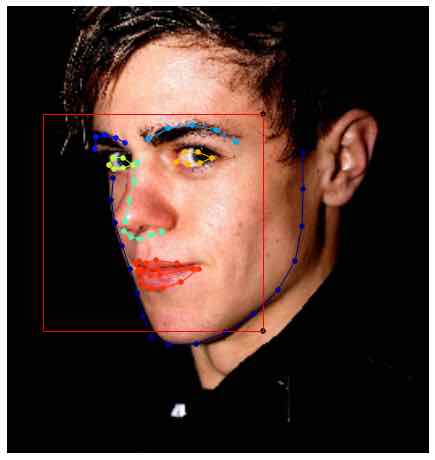}
		\includegraphics[width=0.16\textwidth, height=0.16\textwidth]{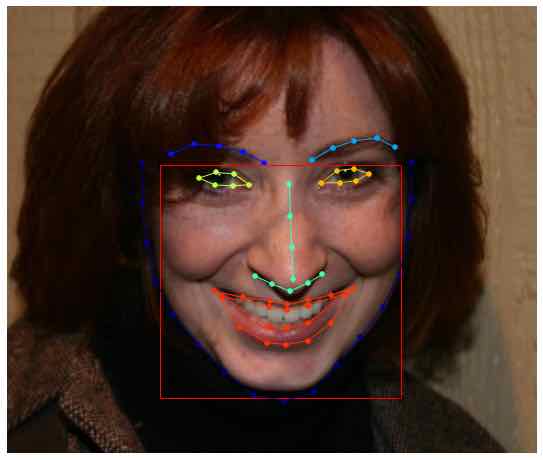}
		\includegraphics[width=0.16\textwidth, height=0.16\textwidth]{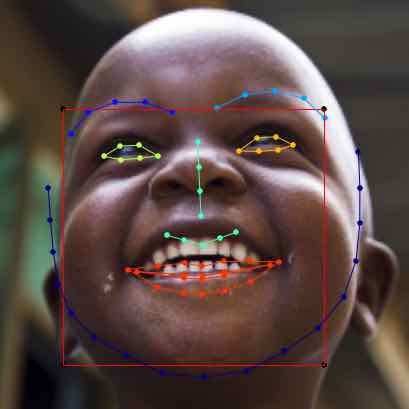}
		\caption{Exemplar results from the Helen test dataset obtained by the Project-Out Asymmetric Gauss-Newton Schur algorithm.}
		\label{fig:helen_po}
	\end{subfigure}
	\begin{subfigure}{\textwidth}
	    \includegraphics[width=0.16\textwidth, height=0.16\textwidth]{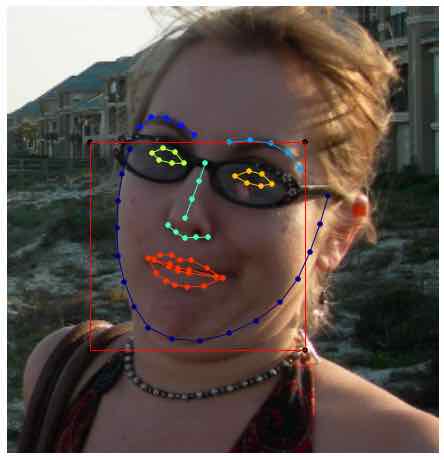}
		\includegraphics[width=0.16\textwidth, height=0.16\textwidth]{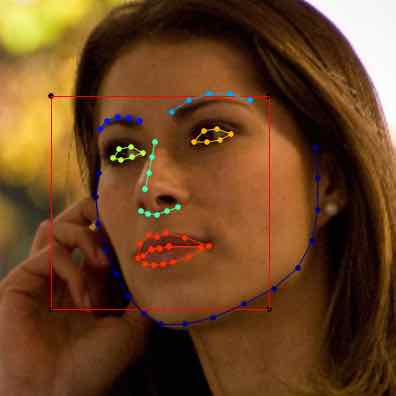}
		\includegraphics[width=0.16\textwidth, height=0.16\textwidth]{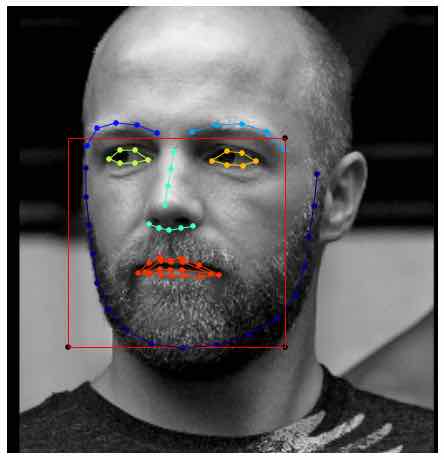}
		\includegraphics[width=0.16\textwidth, height=0.16\textwidth]{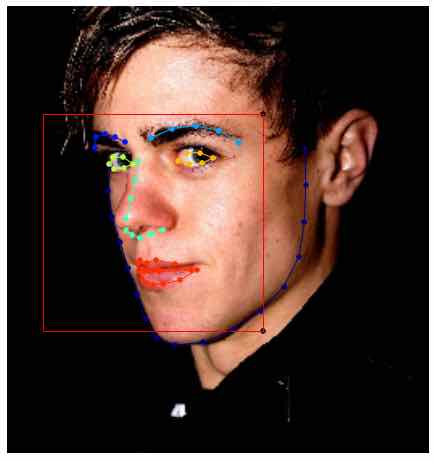}
		\includegraphics[width=0.16\textwidth, height=0.16\textwidth]{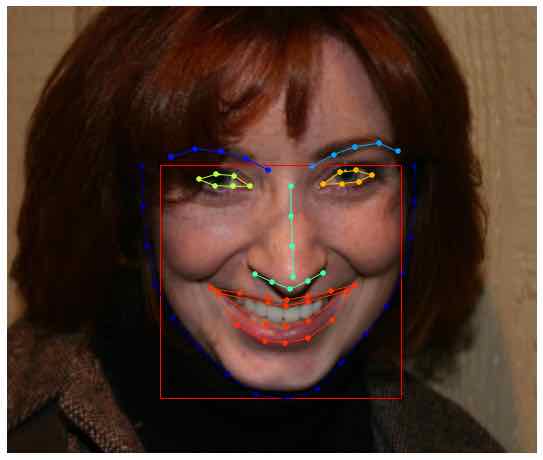}
		\includegraphics[width=0.16\textwidth, height=0.16\textwidth]{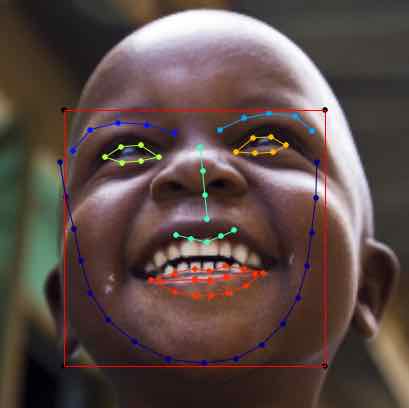}
		\caption{Exemplar results from the Helen test dataset obtained by the SSD Asymmetric Gauss-Newton Schur algorithm.}
		\label{fig:helen_ssd}
	\end{subfigure}
	\caption{Exemplar results from the Helen test dataset.}
	\label{fig:helen}
\end{figure*}

\begin{figure*}
	\begin{subfigure}{\textwidth}
	    \includegraphics[width=0.16\textwidth, height=0.16\textwidth]{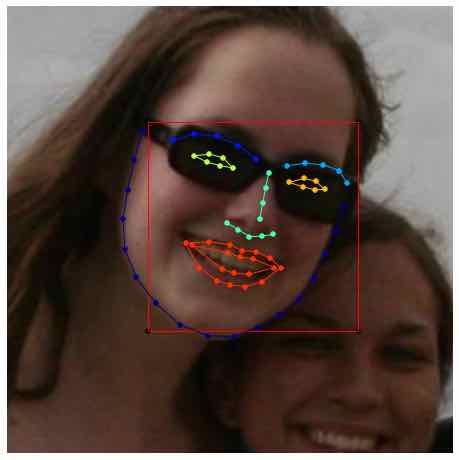}
		\includegraphics[width=0.16\textwidth, height=0.16\textwidth]{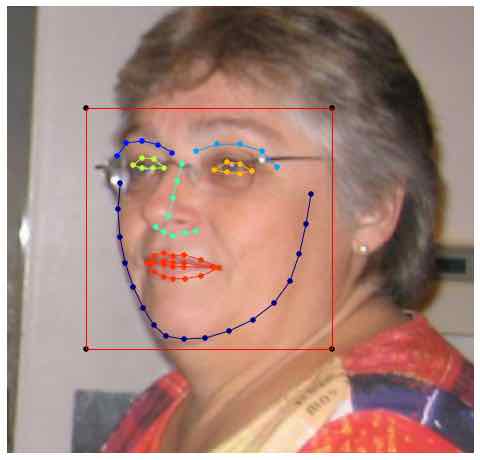}
		\includegraphics[width=0.16\textwidth, height=0.16\textwidth]{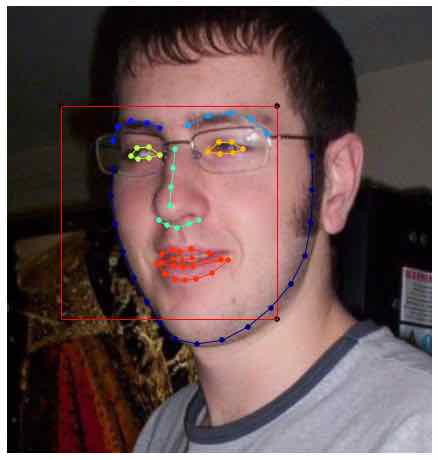}
		\includegraphics[width=0.16\textwidth, height=0.16\textwidth]{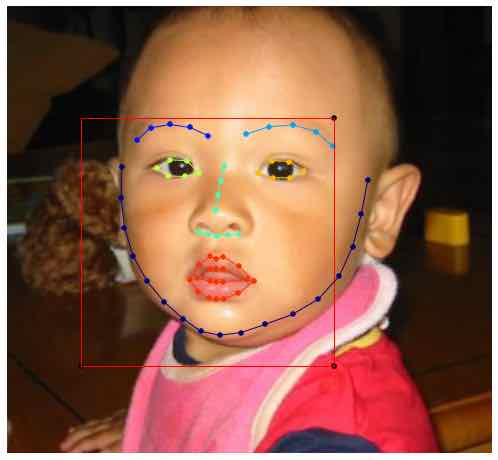}
		\includegraphics[width=0.16\textwidth, height=0.16\textwidth]{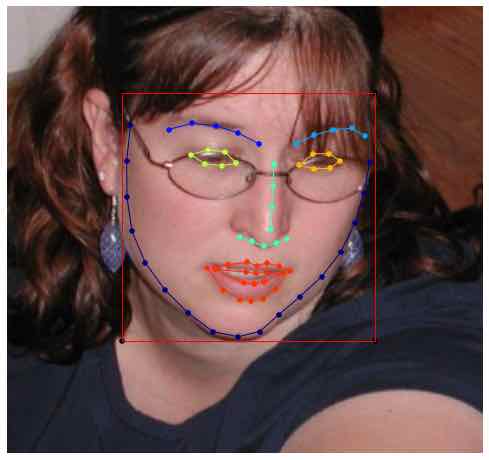}
		\includegraphics[width=0.16\textwidth, height=0.16\textwidth]{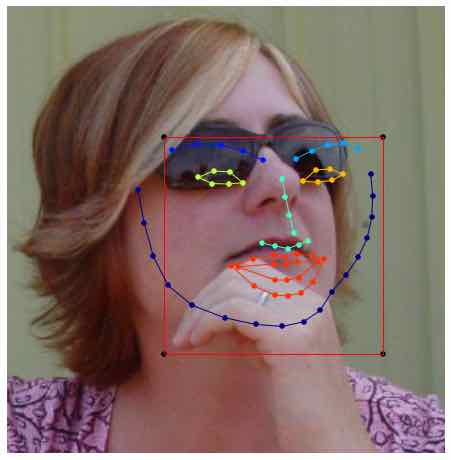}
		\caption{Exemplar results from the Helen test dataset obtained by the Project-Out Asymmetric Gauss-Newton Schur algorithm.}
		\label{fig:afw_po}
	\end{subfigure}
	\begin{subfigure}{\textwidth}
	    \includegraphics[width=0.16\textwidth, height=0.16\textwidth]{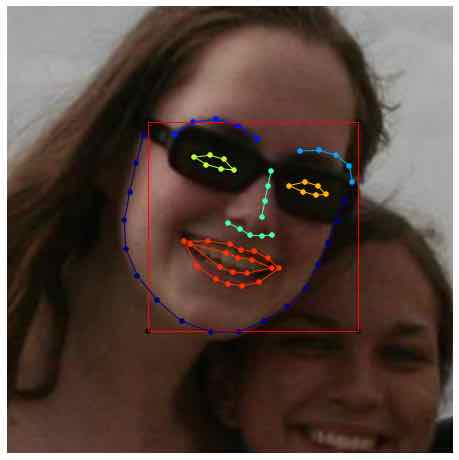}
		\includegraphics[width=0.16\textwidth, height=0.16\textwidth]{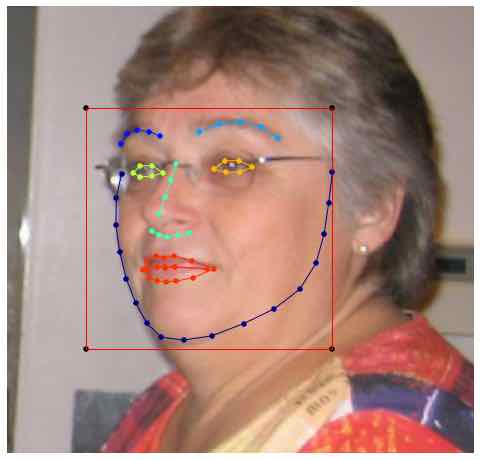}
		\includegraphics[width=0.16\textwidth, height=0.16\textwidth]{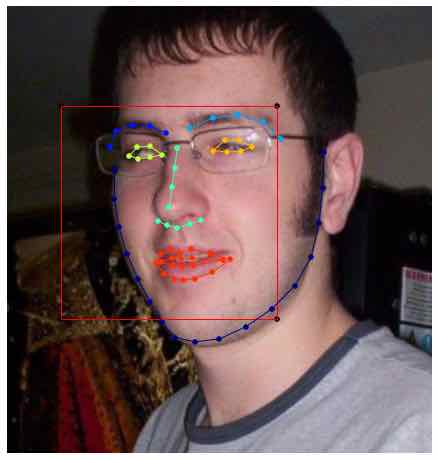}
		\includegraphics[width=0.16\textwidth, height=0.16\textwidth]{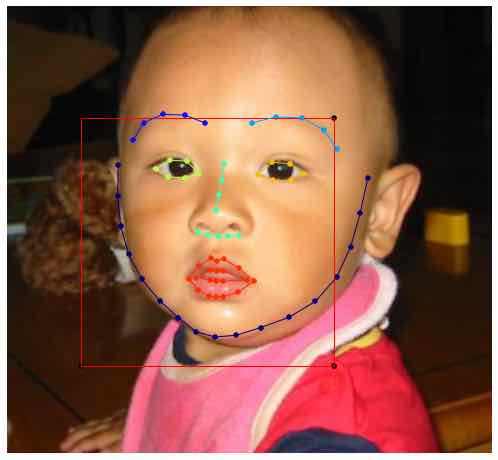}
		\includegraphics[width=0.16\textwidth, height=0.16\textwidth]{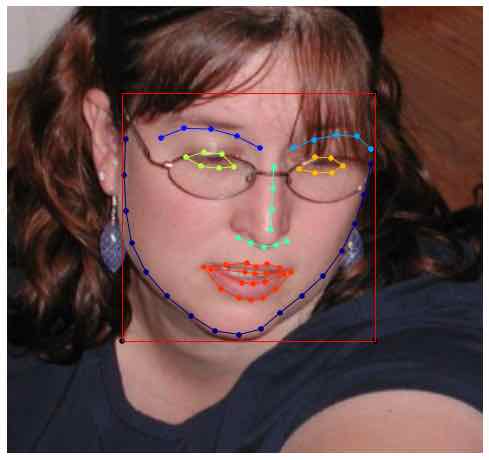}
		\includegraphics[width=0.16\textwidth, height=0.16\textwidth]{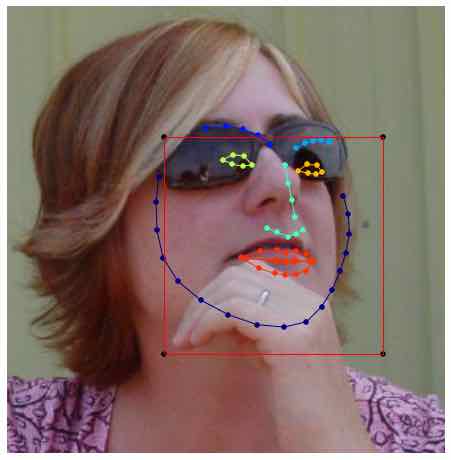}
		\caption{Exemplar results from the AFW dataset obtained by the SSD Asymmetric Gauss-Newton Schur algorithm.}
		\label{fig:afw_ssd}
	\end{subfigure}
	\caption{Exemplar results from the AFW dataset.}
	\label{fig:afw}
\end{figure*}

Exemplar results for this experiment are shown in Figures \ref{fig:helen} and \ref{fig:afw}.

\subsection{Comparison on MIT StreetScene}
\label{exp:6}

In this final experiment, we present results for a different type of object: cars. To this end, we use the first view of the MIT StreetScene\footnoteref{car_db_url} dataset containing a wide variety of frontal car images obtained in the wild. We use 10-fold cross-validation on the $\sim 500$ images of the previous dataset to train and test our algorithms. We report results for the two versions of the \emph{SSD Asymmetric Gauss-Newton} and the \emph{Project-Out Asymmetric Gauss-Newton} algorithms used in Experiment \ref{exp:5}.

\begin{figure*}[p]
	\centering
	\begin{subfigure}{0.48\textwidth}
    	\includegraphics[width=\textwidth]{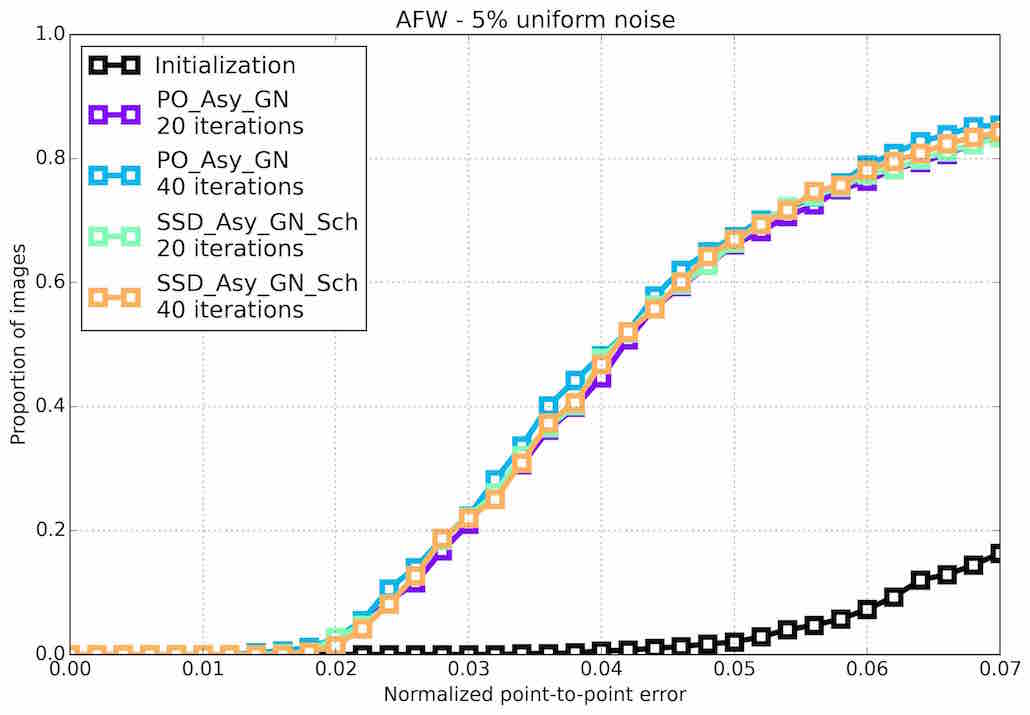}
    \end{subfigure}
    \caption{CED on the first view of the MIT StreetScene test dataset for the Project-Out and SSD Asymmetric Gauss-Newton algorithms initialized with $5\%$ noise.}
    \label{fig:ced_cars}
\end{figure*}

Result for this experiment are shown in Figure \ref{fig:ced_cars}. We can observe that all algorithms obtain similar performance and that they vastly improve upon the original initialization. 

\begin{figure*}
	\begin{subfigure}{\textwidth}
	    \includegraphics[width=0.16\textwidth]{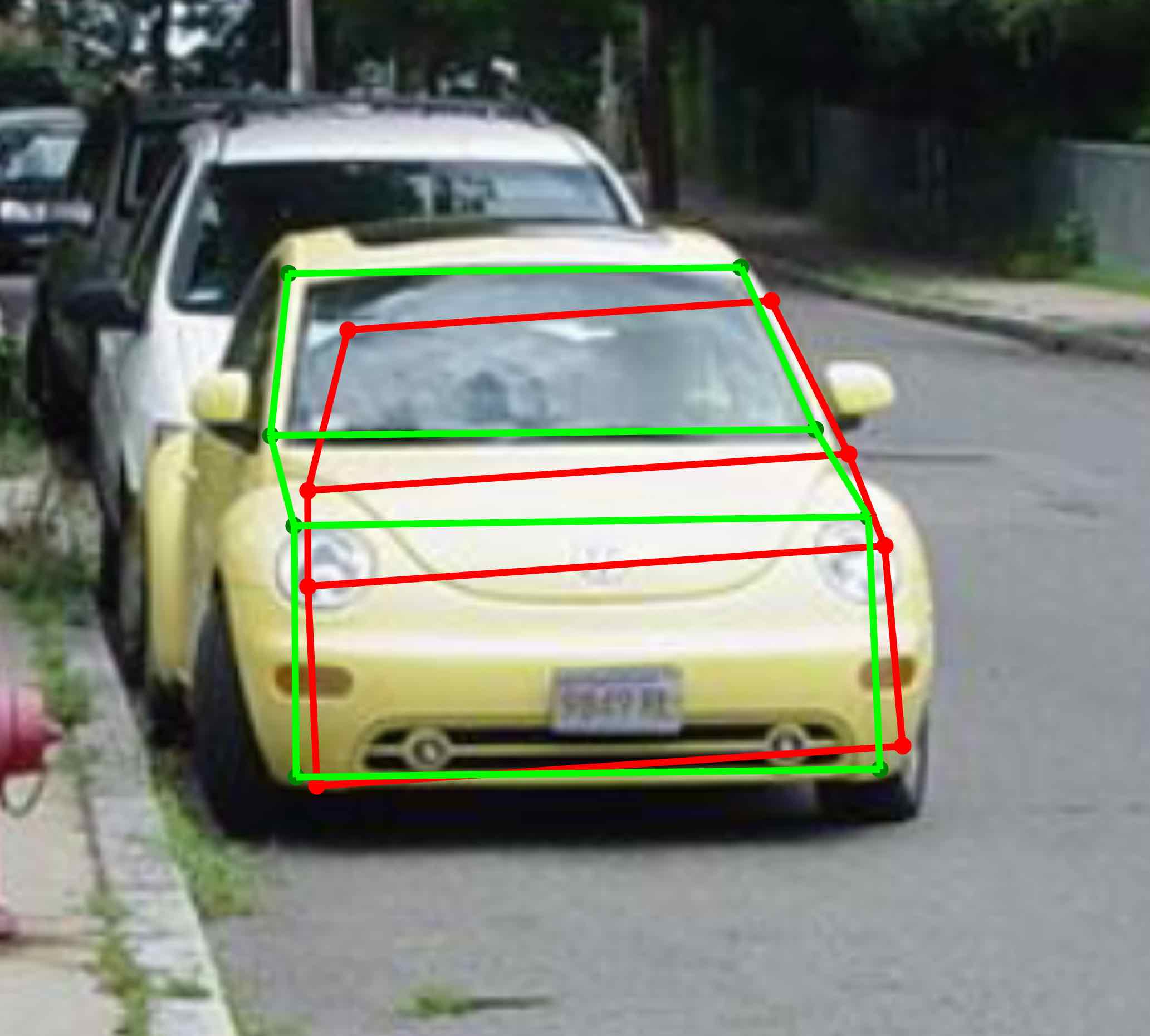}
		\includegraphics[width=0.16\textwidth]{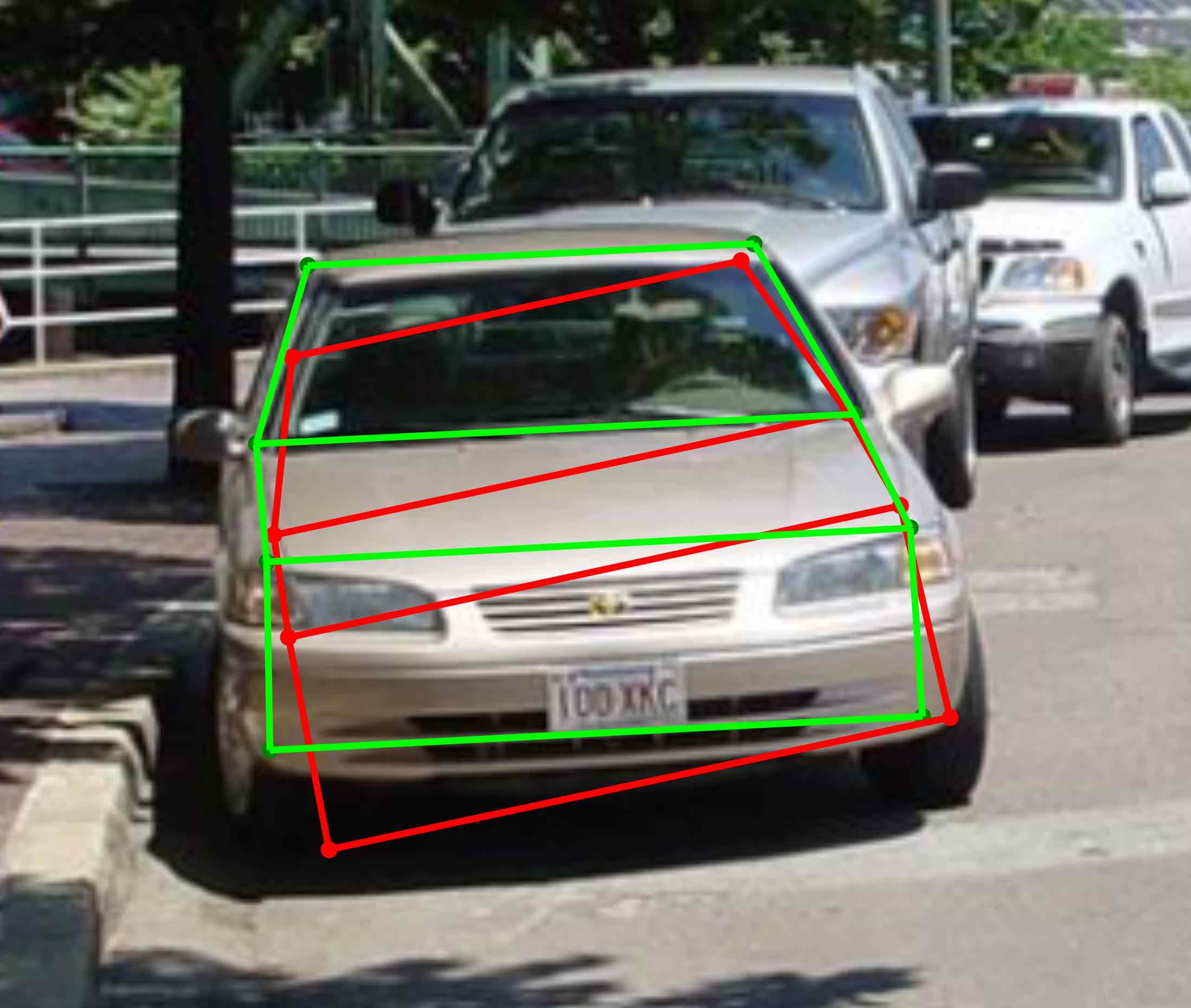}
		\includegraphics[width=0.16\textwidth]{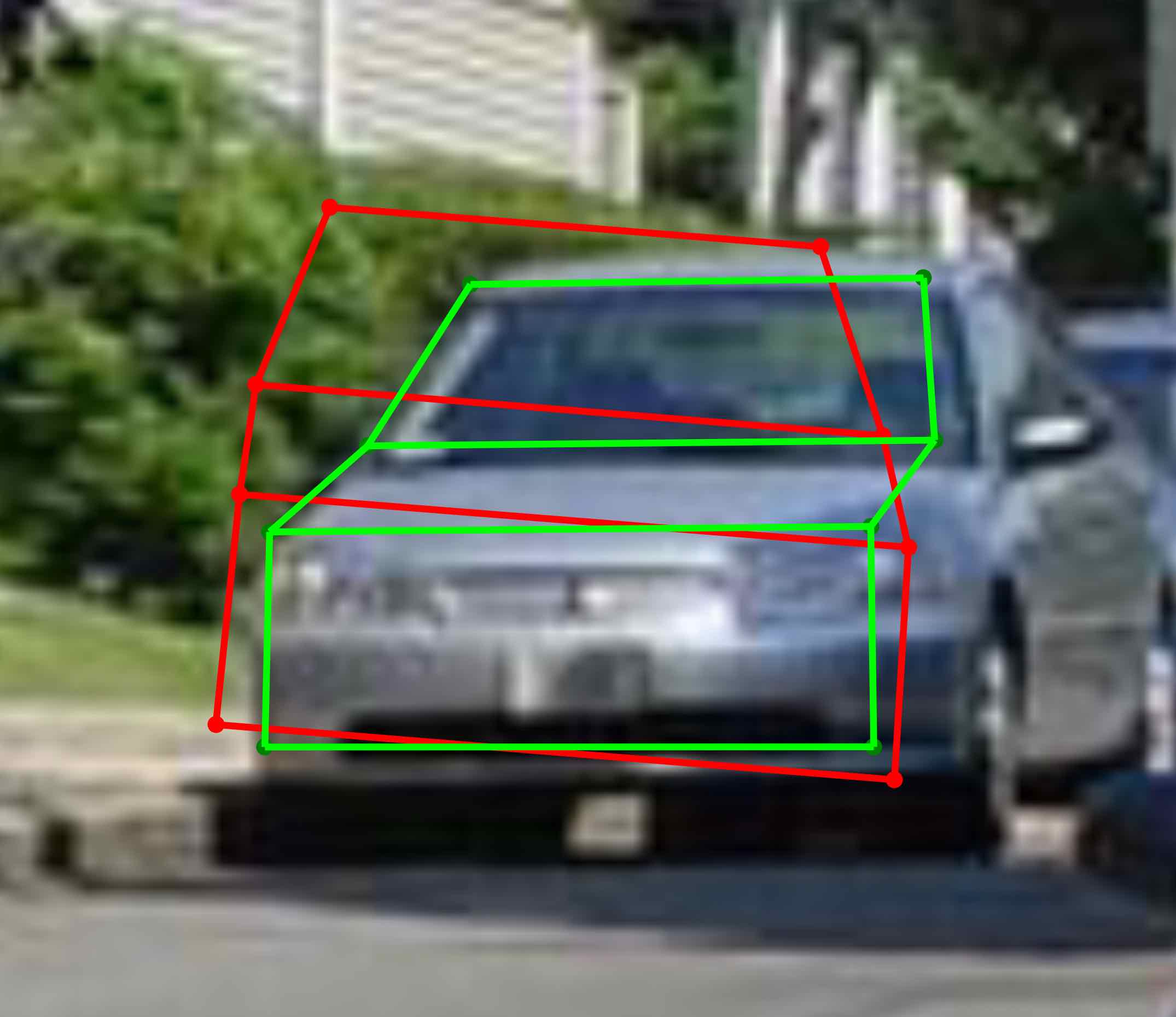}
		\includegraphics[width=0.16\textwidth]{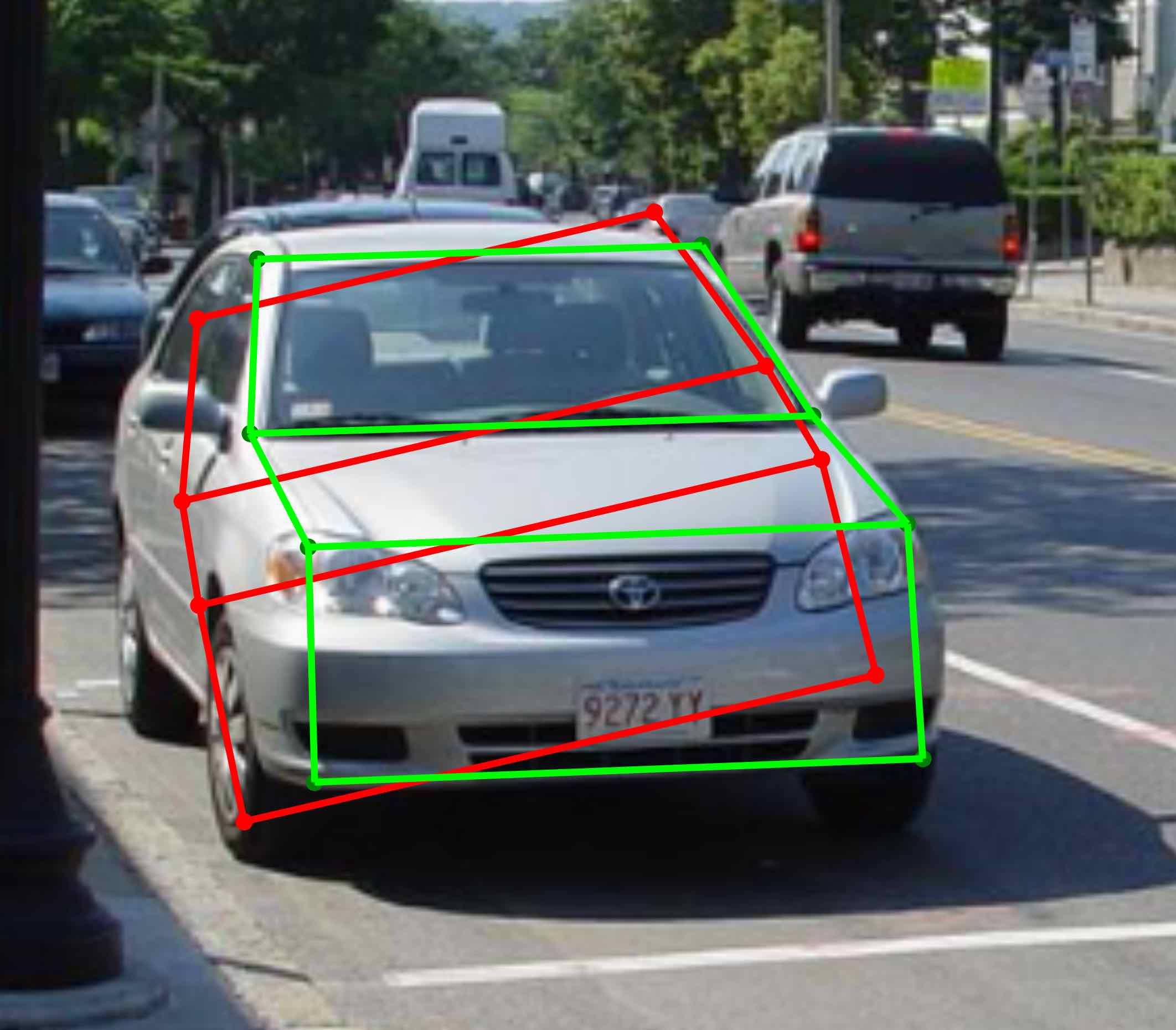}
		\includegraphics[width=0.16\textwidth]{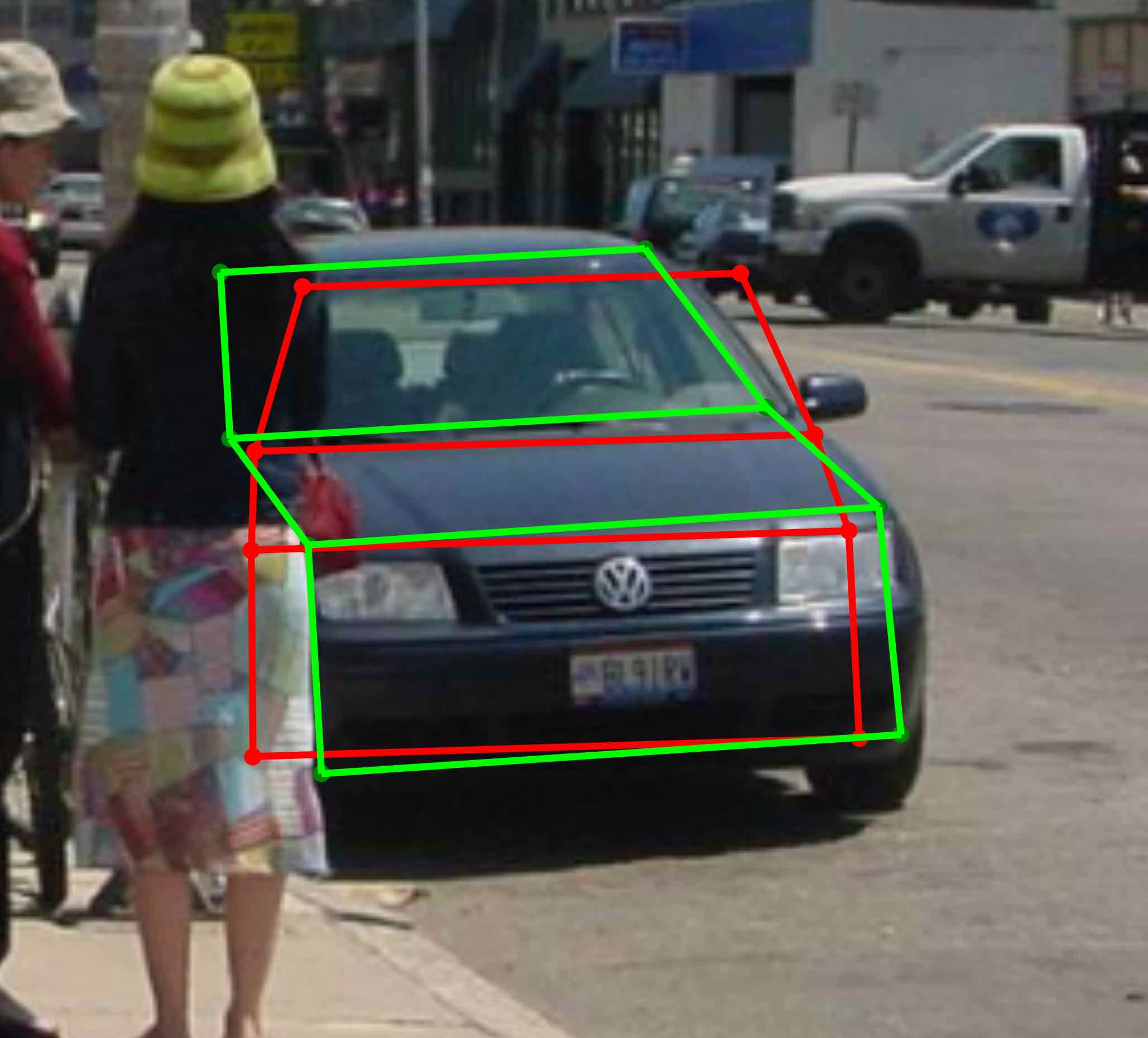}
		\includegraphics[width=0.16\textwidth]{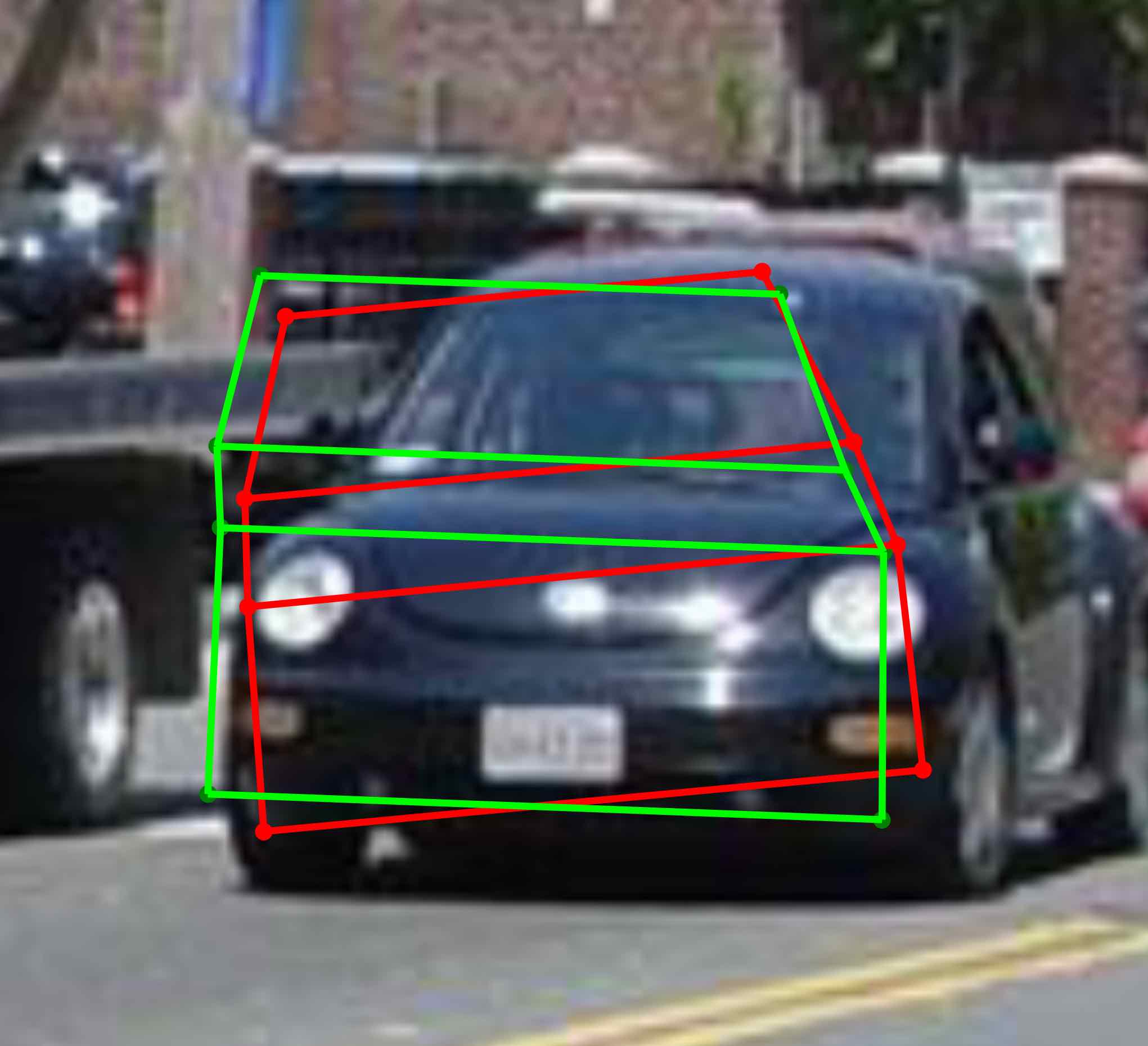}
		\caption{Exemplar results from the MIT StreetScene test dataset obtained by the Project-Out Asymmetric Gauss-Newton Schur algorithm.}
		\label{fig:cars_po}
	\end{subfigure}
	\begin{subfigure}{\textwidth}
	    \includegraphics[width=0.16\textwidth]{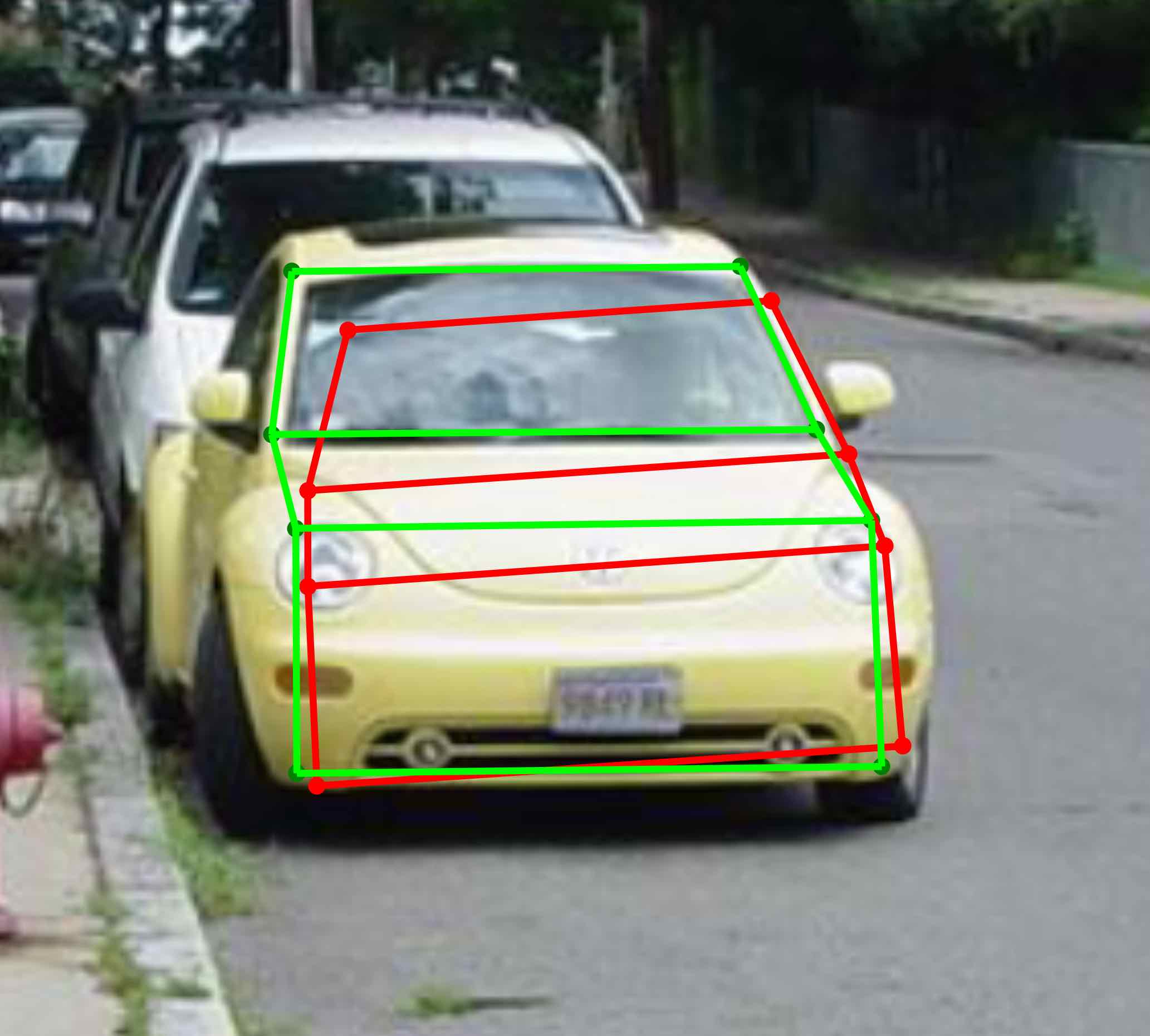}
		\includegraphics[width=0.16\textwidth]{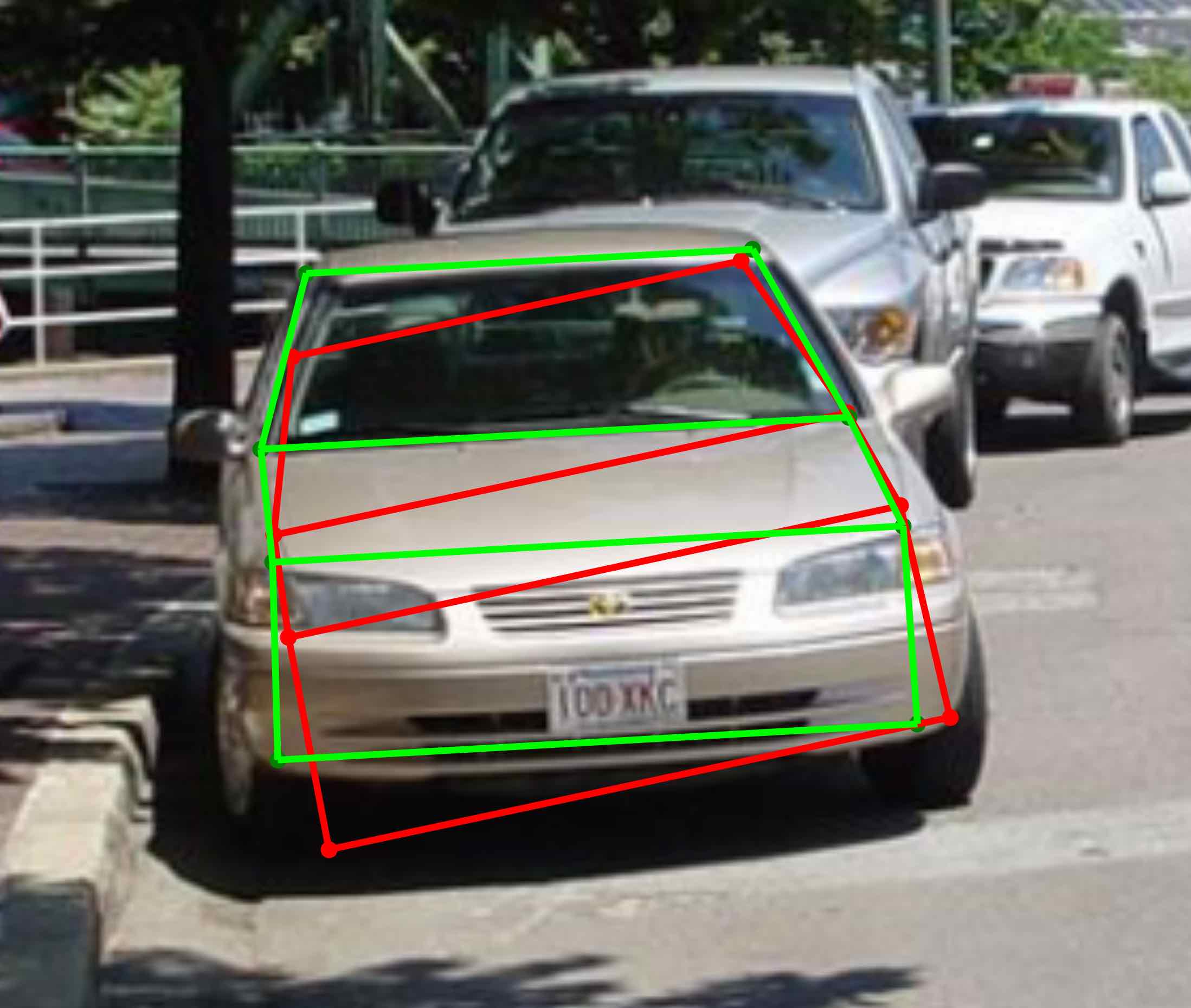}
		\includegraphics[width=0.16\textwidth]{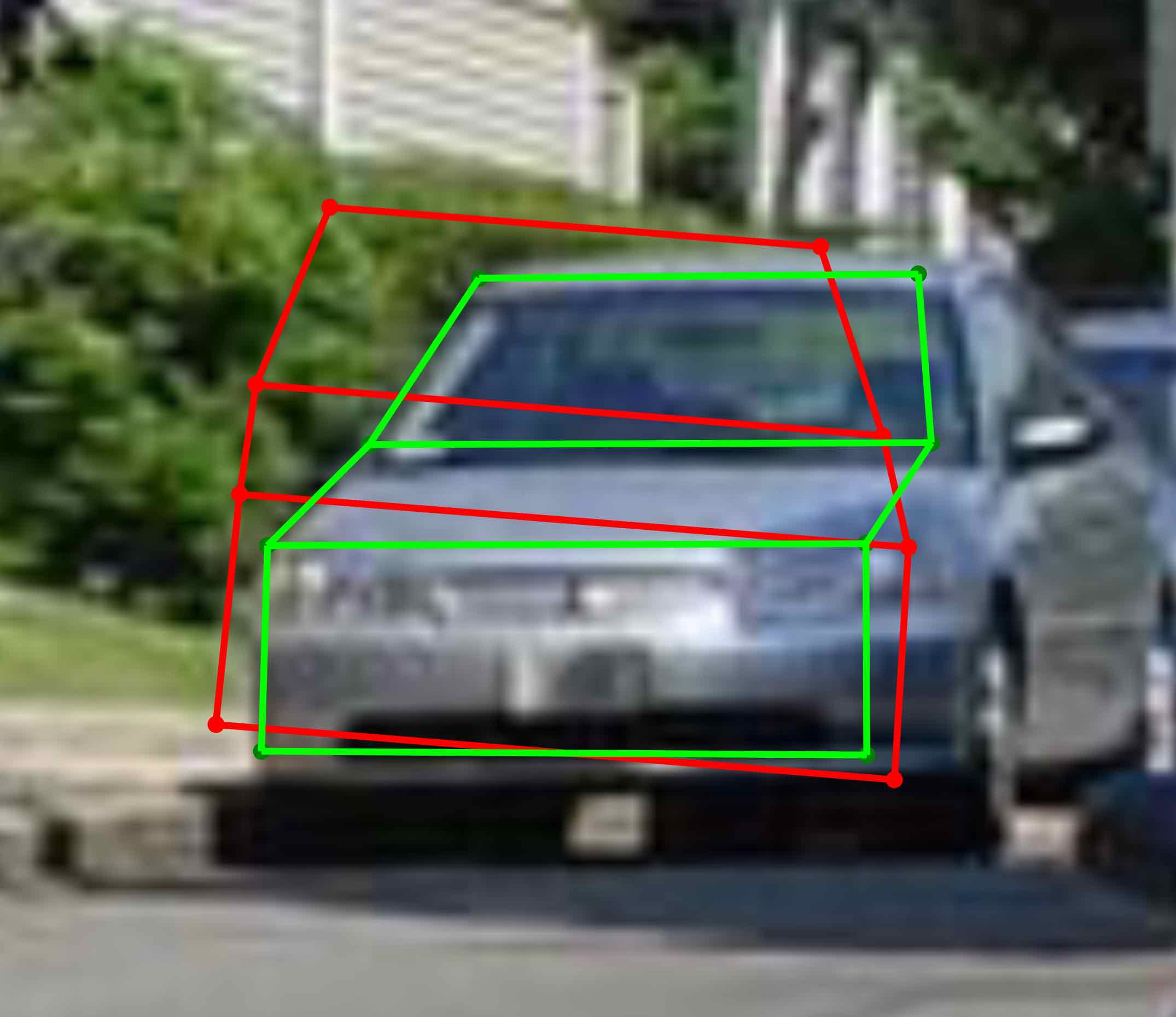}
		\includegraphics[width=0.16\textwidth]{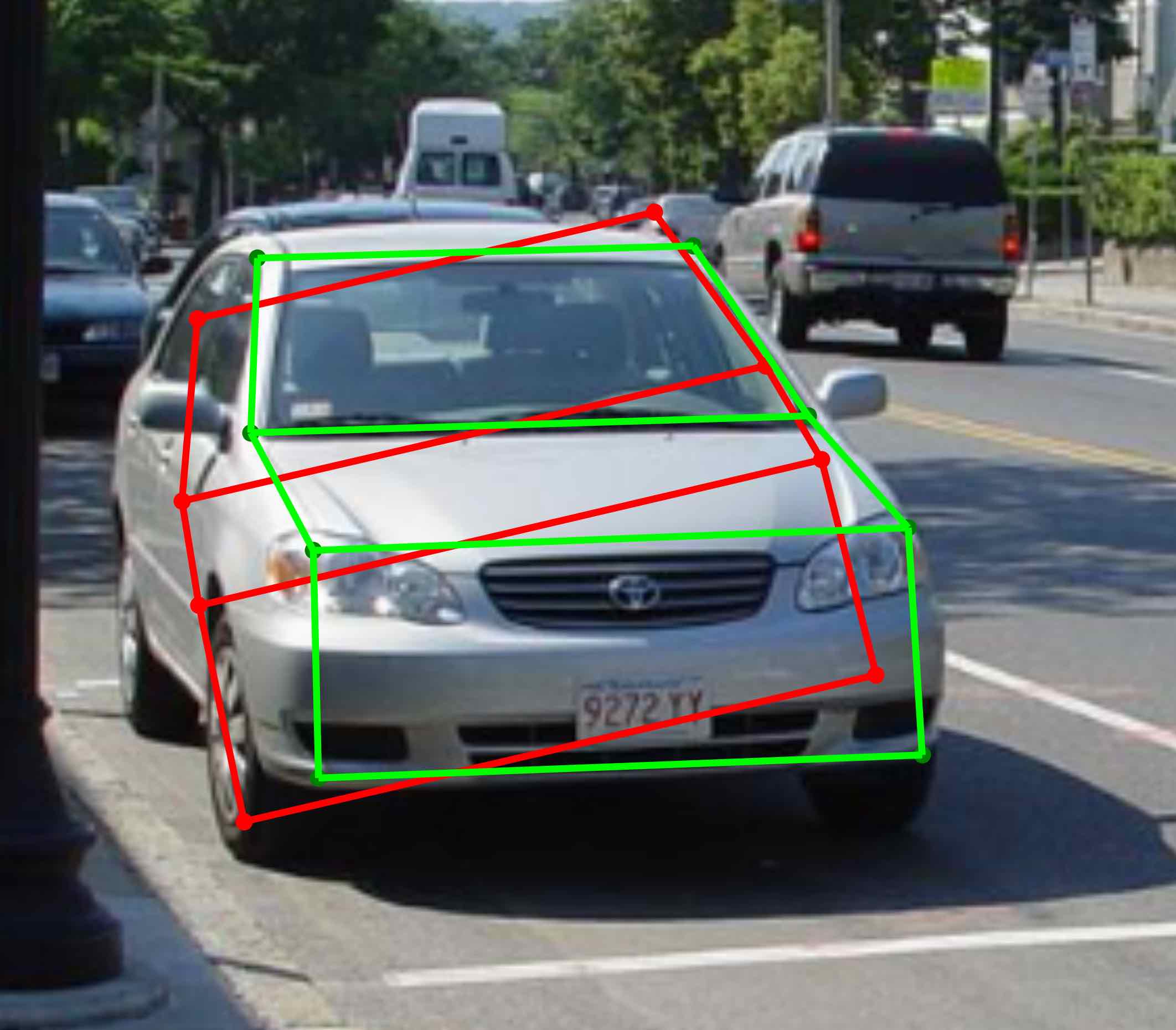}
		\includegraphics[width=0.16\textwidth]{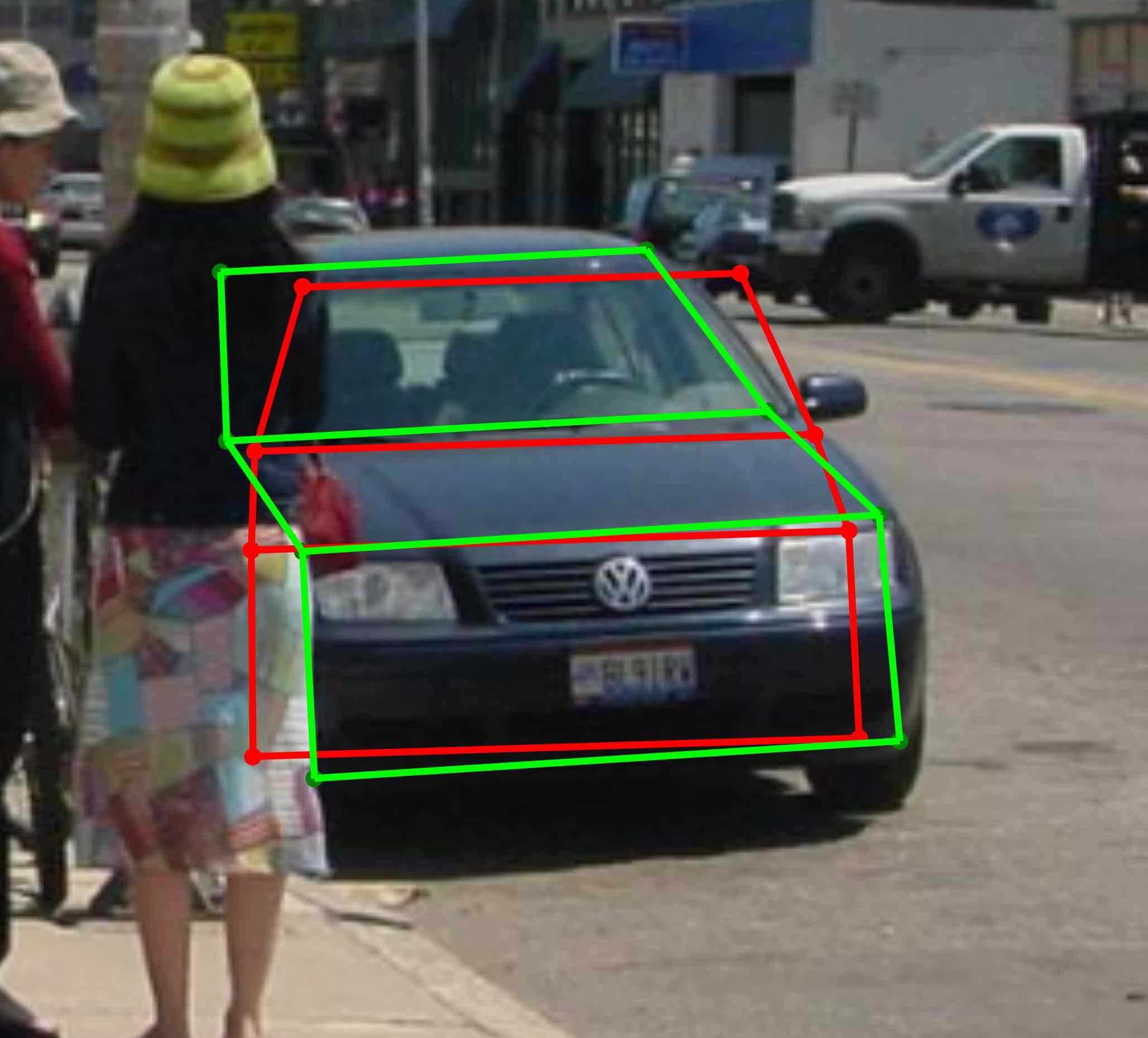}
		\includegraphics[width=0.16\textwidth]{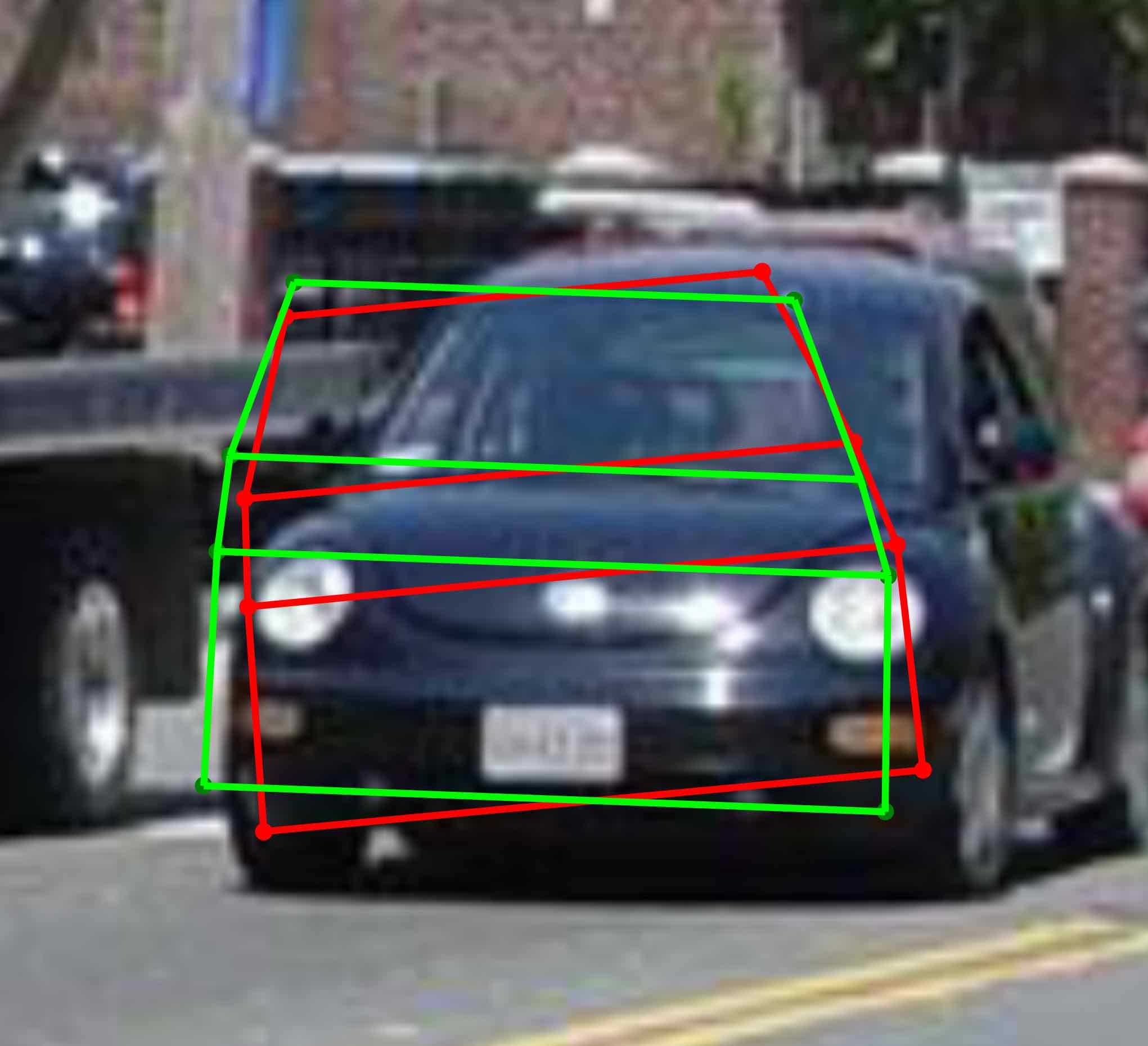}
		\caption{Exemplar results from the MIT StreetScene test dataset obtained by the SSD Asymmetric Gauss-Newton Schur algorithm.}
		\label{fig:cars_ssd}
	\end{subfigure}
	\caption{Exemplar results from the MIT StreetScene test dataset.}
	\label{fig:cars}
\end{figure*}

Exemplar results for this experiment are shown in Figure \ref{fig:cars}.

\subsection{Analysis}
\label{exp:analysis}

Given the results reported by the previous six experiments we conclude that:

\begin{enumerate}
\item Overall, \emph{Gauss-Newton} and \emph{Wiberg} algorithms vastly outperform \emph{Newton} algorithms for fitting AAMs. Experiment \ref{exp:1} clearly shows that the former algorithms provide significantly higher levels of fitting accuracy at considerably lower computational complexities and run times. These findings are consistent with existent literature in the related field of parametric image alignment \cite{Matthews2004} and also, to certain extend, with prior work on \emph{Newton} algorithms for AAM fitting \cite{Kossaifi2014}. We attribute the bad performance of \emph{Newton} algorithms to the difficulty of accurately computing a (noiseless) estimate of the full Hessian matrix using finite differences.

\item \emph{Gauss-Newton} and \emph{Wiberg} algorithms are virtually equivalent in performance. The results in Experiment \ref{exp:1} show that the difference in accuracy between both types of algorithms is minimal and the small differences in their respective solutions are, in practice, insignificant.

\item Our \emph{Bayesian} project-out formulation leads to significant improvements in fitting accuracy without adding extra computational cost. Experiment \ref{exp:2} shows that a weighted combination of the two terms forming \emph{Bayesian} project-out loss always outperforms the \emph{classic} project out formulation.

\item The \emph{Asymmetric} composition proposed in this work leads to CGD algorithms that are more accurate and that converge faster. In particular, the \emph{SSD} and \emph{Project-Out Asymmetric Gauss-Newton} algorithms are shown to achieve significantly better performance than their \emph{Forward} and \emph{Inverse} counterparts in Experiments \ref{exp:1} and \ref{exp:3}.

\item Finally, a significant reduction in the computational complexity and runtime of CDG algorithms can be obtained by limiting the number of pixels considered during optimization of the loss function and by adjusting the number of iterations that the algorithms are run for, Experiment \ref{exp:4}.
\end{enumerate}

\section{Conclusion}
\label{sec:conclusion}

In this paper we have thoroughly studied the problem of fitting AAMs using CGD algorithms. We have presented a unified and complete framework for these algorithms and classified them with respect to three of their main characteristics:
\begin{inparaenum}[\itshape i\upshape)]
	\item \emph{cost function};
	\item type of \emph{composition}; and
	\item \emph{optimization method}.
\end{inparaenum}

Furthermore, we have extended the previous framework by:
\begin{itemize}
	\item Proposing a novel \emph{Bayesian cost function} for fitting AAMs that can be interpreted as a more general formulation of the well-known project-out loss. We have assumed a probabilistic model for appearance generation with both Gaussian noise and a Gaussian prior over a latent appearance space. Marginalizing out the latent appearance space, we have derived a novel cost function that only depends on shape parameters and that can be interpreted as a valid and more general probabilistic formulation of the well-known project-out cost function \cite{Matthews2004}. In the experiments, we have showed that our Bayesian formulation considerably outperforms the original project-out cost function.
	
	\item Proposing \emph{asymmetric} and \emph{bidirectional} compositions for CGD algorithms. We have shown the connection between Gauss-Newton Asymmetric algorithms and ESM algorithms and experimentally proved that these two novel types of composition lead to better convergent and more robust CGD algorithm for fitting AAMs.
	
	\item Providing new valuable insights into existent CGD algorithms by reinterpreting them as direct applications of the \emph{Schur complement} and the \emph{Wiberg method}.
\end{itemize} 

Finally, in terms of future work, we plan to:
\begin{itemize}
	\item Adapt existent Supervised Descent (SD) algorithms for face alignment \cite{Xiong2013, Tzimiropoulos2015} to AAMs and investigate their relationship with the CGD algorithms studied in this paper. 
	
	\item Investigate if our Bayesian cost function and the proposed asymmetric and bidirectional compositions can also be successfully applied to similar generative parametric models, such as the Gauss-Newton Parts-Based Deformable Model (GN-DPM) proposed in \cite{Tzimiropoulos2014}.   
\end{itemize} 

\begin{acknowledgements}
The work of Joan Alabort-i-Medina
is funded by a DTA studentship from Imperial College London
and by the Qualcomm Innovation Fellowship. The work of S. Zafeiriou has been partly funded by the EPSRC project Adaptive Facial Deformable Models for Tracking (ADAManT), EP/L026813/1.
\end{acknowledgements}

\bibliographystyle{spbasic}      
\bibliography{main.bib}   

\appendix 

\pagebreak

\section{Terms in SSD Newton Hessians}
\label{sec:app1}

In this section we define the individual terms of the Hessian matrices used by the \emph{SSD Asymmetric} and \emph{Bidirectional Newton} optimization algorithms derived in Section \ref{sec:newton}.

\subsection{Asymmetric}
\label{sec:app11}

The individual terms forming the Hessian matrix of the \emph{SSD Asymmetric Newton} algorithm defined by Equation \ref{eq:asymmetric_newton_hessian} are defined as follows:
\begin{equation}
    \begin{aligned}
		\frac{\partial^2 \mathcal{D}_a}{\partial^2 \Delta \mathbf{c}} & = \frac{\partial -\mathbf{A}^T \mathbf{r}_a}{\partial \Delta \mathbf{c}}
		\\
		& = -\mathbf{A}^T \frac{\partial \mathbf{r}_a}{\partial \Delta \mathbf{c}}
		\\
		& = \underbrace{\mathbf{A}^T \mathbf{A}}_{\mathbf{I}}
    \label{eq:asymmetric_hessian_term1}
    \end{aligned}
\end{equation}
\begin{equation}
    \begin{aligned}
		\frac{\partial^2 \mathcal{D}_a}{\partial \Delta \mathbf{c} \partial \Delta \mathbf{p}} & = \frac{\partial -\mathbf{A}^T \mathbf{r}_a}{\partial \Delta \mathbf{p}}
		\\
		& = \frac{\partial -\mathbf{A}^T}{\partial \Delta \mathbf{p}} \mathbf{r}_a - \mathbf{A}^T \frac{\partial\mathbf{r}_a}{\partial \Delta \mathbf{p}}
		\\
		& = -\beta \mathbf{J}_\mathbf{A}^T \mathbf{r}_a - \mathbf{A}^T \mathbf{J}_{\mathbf{t}}
    \label{eq:asymmetric_hessian_term2}
    \end{aligned}
\end{equation}
where we have defined $\mathbf{J}_\mathbf{A} = [\nabla \mathbf{a}_1, \cdots, \nabla \mathbf{a}_m]^T \frac{\partial\mathcal{W}}{\partial \Delta \mathbf{p}}$.
\begin{equation}
    \begin{aligned}
		\frac{\partial^2 \mathcal{D}_a}{\partial^2 \Delta \mathbf{p}} & =  \frac{\partial \mathbf{J}_{\mathbf{t}}^T \mathbf{r}_a}{\partial \Delta \mathbf{p}}
		\\
		& = \frac{\partial \mathbf{J}_{\mathbf{t}}^T}{\partial \Delta \mathbf{p}} \mathbf{r}_a + \mathbf{J}_{\mathbf{t}}^T \frac{\partial \mathbf{r}_a}{\partial \Delta \mathbf{p}}
		\\
		& = \left( \frac{\partial\mathcal{W}}{\partial \Delta \mathbf{p}}^T \nabla^2 \mathbf{t} \frac{\partial\mathcal{W}}{\partial \Delta \mathbf{p}} + \underbrace{\nabla \mathbf{t} \overbrace{\frac{\partial^2 \mathcal{W}}{\partial^2 \mathbf{p}}}^{\mathbf{0}}}_{\mathbf{0}} \right) \mathbf{r}_a +
		\\
		& \qquad \, \mathbf{J}_{\mathbf{t}}^T \mathbf{J}_{\mathbf{t}}
		\\
		& = \left( \frac{\partial\mathcal{W}}{\partial \Delta \mathbf{p}}^T \nabla^2 \mathbf{t} \frac{\partial\mathcal{W}}{\partial \Delta \mathbf{p}} \right) \mathbf{r}_a + \mathbf{J}_{\mathbf{t}}^T \mathbf{J}_{\mathbf{t}}
    \label{eq:asymmetric_hessian_term3}
    \end{aligned}
\end{equation}

\subsection{Bidirectional}
\label{sec:app12}

The individual terms forming the Hessian matrix of the \emph{SSD Bidirectional Newton} algorithm defined by Equation \ref{eq:bidirectional_newton_hessian} are defined as follows:
\begin{equation}
    \begin{aligned}
		\frac{\partial^2 \mathcal{D}_b}{\partial^2 \Delta \mathbf{c}} & = \frac{\partial -\mathbf{A}^T \mathbf{r}_b}{\partial \Delta \mathbf{c}}
		\\
		& = -\mathbf{A}^T \frac{\partial \mathbf{r}_b}{\partial \Delta \mathbf{c}}
		\\
		& = \underbrace{\mathbf{A}^T \mathbf{A}}_{\mathbf{I}}
    \label{eq:bidirectional_hessian_term1}
    \end{aligned}
\end{equation}
\begin{equation}
    \begin{aligned}
		\frac{\partial^2 \mathcal{D}_b}{\partial \Delta \mathbf{c} \partial \Delta \mathbf{p}} & = \frac{\partial -\mathbf{A}^T \mathbf{r}_b}{\partial \Delta \mathbf{p}}
		\\
		& = -\mathbf{A}^T \frac{\partial \mathbf{r}_b}{\partial \Delta \mathbf{p}}
		\\
		& = -\mathbf{A}^T \mathbf{J}_{\mathbf{i}}
    \label{eq:bidirectional_hessian_term2}
    \end{aligned}
\end{equation}
\begin{equation}
    \begin{aligned}
		\frac{\partial^2 \mathcal{D}_b}{\partial \Delta \mathbf{c} \partial \Delta \mathbf{q}} & =  \frac{\partial -\mathbf{A}^T \mathbf{r}_b}{\partial \Delta \mathbf{q}}
		\\
		&= \frac{\partial -\mathbf{A}^T}{\partial \Delta \mathbf{q}} \mathbf{r}_b - \mathbf{A}^T \frac{\partial \mathbf{r}_b}{\partial \Delta \mathbf{q}}
		\\
		& = -\mathbf{J}_{\mathbf{A}}^T \mathbf{r}_b + \mathbf{A}^T \mathbf{J}_{\mathbf{a}}
    \label{eq:bidirectional_hessian_term3}
    \end{aligned}
\end{equation}
\begin{equation}
    \begin{aligned}
		\frac{\partial^2 \mathcal{D}_b}{\partial^2 \Delta \mathbf{p}} & =  \frac{\partial \mathbf{J}_{\mathbf{i}}^T \mathbf{r}_b}{\partial \Delta \mathbf{p}}
		\\
		& = \frac{\partial \mathbf{J}_{\mathbf{i}}^T}{\partial \Delta \mathbf{p}} \mathbf{r}_b + \mathbf{J}_{\mathbf{i}}^T \frac{\partial \mathbf{r}_b}{\partial \Delta \mathbf{p}}
		\\
		& = \left( \frac{\partial\mathcal{W}}{\partial \Delta \mathbf{p}}^T \nabla^2 \mathbf{i}[\mathbf{p}] \frac{\partial\mathcal{W}}{\partial \Delta \mathbf{p}} \right) \mathbf{r}_b + \mathbf{J}_{\mathbf{i}}^T \mathbf{J}_{\mathbf{i}}
    \label{q:bidirectional_hessian_term4}
    \end{aligned}
\end{equation}
\begin{equation}
    \begin{aligned}
		\frac{\partial^2 \mathcal{D}_b}{\partial \Delta \mathbf{p} \partial \Delta \mathbf{q}} & =  \frac{\partial \mathbf{J}_{\mathbf{i}}^T \mathbf{r}_b}{\partial \Delta \mathbf{q}}
		\\
		& = -\mathbf{J}_{\mathbf{i}}^T \mathbf{J}_{\mathbf{a}}
    \label{eq:bidirectional_hessian_term5}
    \end{aligned}
\end{equation}
\begin{equation}
    \begin{aligned}
		\frac{\partial^2 \mathcal{D}_b}{\partial^2 \Delta \mathbf{q}} & =  \frac{\partial -\mathbf{J}_{\mathbf{a}}^T \mathbf{r}_b}{\partial \Delta \mathbf{q}}
		\\
		& = \frac{\partial -\mathbf{J}_{\mathbf{a}}^T}{\partial \Delta \mathbf{q}} \mathbf{r}_b - \mathbf{J}_{\mathbf{a}}^T \frac{\partial \mathbf{r}_b}{\partial \Delta \mathbf{q}}
		\\
		& = -\left( \frac{\partial\mathcal{W}}{\partial \Delta \mathbf{q}}^T \nabla^2 (\mathbf{a} + \mathbf{A}\mathbf{c}) \frac{\partial\mathcal{W}}{\partial \Delta \mathbf{q}} \right) \mathbf{r}_b + \mathbf{J}_{\mathbf{a}}^T \mathbf{J}_{\mathbf{a}}
    \label{q:bidirectional_hessian_term6}
    \end{aligned}
\end{equation}

\section{Iterative solutions of all algorithms}
\label{sec:app2}

In this section we report the iterative solutions of all CGD algorithms studied in this paper. In order to keep the information structured algorithms are grouped by their cost function. Consequently, iterative solutions for all SSD and Project-Out algorithms are stated in Tables \ref{tab:ssd_solution} and \ref{tab:po_solution}.

\begin{table*}
\centering
\begin{tabular}{l|l|l|l}
	\toprule
	\multirow{2}{*}{SSD algorithms}  & \multicolumn{3}{c}{Iterative solutions}
	\\
	& \multicolumn{1}{c|}{$\Delta\mathbf{p}$} & \multicolumn{1}{c|}{$\Delta\mathbf{q}$} & \multicolumn{1}{c}{$\Delta\mathbf{c}$}
	\\
	\midrule
	\multirow{2}{*}{SSD\_For\_GN\_Sch \cite{Amberg2009, Tzimiropoulos2013}} 
	& 
	$
	\Delta\mathbf{p} = -\hat{\mathbf{H}}_{\mathbf{i}}^{-1} \mathbf{J}_{\mathbf{i}}^T\bar{\mathbf{A}}\mathbf{r}
	$
	&
	&
	\multirow{2}{*}{
	$
	\Delta\mathbf{c} = \mathbf{A} \left (\mathbf{r} + \mathbf{J}_{\mathbf{i}}\Delta\mathbf{p} \right)
	$
	}
	\\
	&
	$
	\hat{\mathbf{H}}_{\mathbf{i}} = \mathbf{J}_{\mathbf{i}}^T\bar{\mathbf{A}}\mathbf{J}_{\mathbf{i}}
	$
	&
	&
	\\
	\midrule
	\multirow{2}{*}{SSD\_For\_GN\_Alt} 
	& 
	$
	\Delta\mathbf{p} = -\mathbf{H}_{\mathbf{i}}^{-1} \mathbf{J}_{\mathbf{i}}^T \left( \mathbf{r} - \mathbf{A}\Delta\mathbf{c} \right)
	$
	&
	&
	\multirow{2}{*}{
	$
	\Delta\mathbf{c} = \mathbf{A} \left (\mathbf{r} + \mathbf{J}_{\mathbf{i}}\Delta\mathbf{p} \right)
	$
	}
	\\
	&
	$
	\mathbf{H}_{\mathbf{i}} = \mathbf{J}_{\mathbf{i}}^T\mathbf{J}_{\mathbf{i}}
	$
	&
	&
	\\
	\midrule
	\multirow{2}{*}{SSD\_For\_N\_Sch} 
	& 
	$
	\Delta\mathbf{p} = -\left(\hat{\mathbf{H}}_{\mathbf{i}}^{\textrm{N}}\right)^{-1} \mathbf{J}_{\mathbf{i}}^T\bar{\mathbf{A}}\mathbf{r}
	$
	&
	&
	\multirow{2}{*}{
	$
	\Delta\mathbf{c} = \mathbf{A} \left (\mathbf{r} + \mathbf{J}_{\mathbf{i}}\Delta\mathbf{p} \right)
	$
	}
	\\
	&
	$
	\hat{\mathbf{H}}_{\mathbf{i}}^{\textrm{N}} = \frac{\partial \mathcal{W}}{\Delta \mathbf{p}}^T \nabla^2\mathbf{i} \frac{\partial \mathcal{W}}{\Delta \mathbf{p}}\mathbf{r} + \hat{\mathbf{H}}_{\mathbf{i}}
	$
	&
	&
	\\
	\midrule
	\multirow{2}{*}{SSD\_For\_N\_Alt} 
	& 
	$
	\Delta\mathbf{p} = -\left(\mathbf{H}_{\mathbf{i}}^{\textrm{N}}\right)^{-1} \mathbf{J}_{\mathbf{i}}^T\bar{\mathbf{A}}\left( \mathbf{r} - \mathbf{A}\Delta\mathbf{c} \right)
	$
	&
	&
	\multirow{2}{*}{
	$
	\Delta\mathbf{c} = \mathbf{A} \left (\mathbf{r} + \mathbf{J}_{\mathbf{i}}\Delta\mathbf{p} \right)
	$
	}
	\\
	&
	$
	\mathbf{H}_{\mathbf{i}}^{\textrm{N}} = \frac{\partial \mathcal{W}}{\Delta \mathbf{p}}^T \nabla^2\mathbf{i} \frac{\partial \mathcal{W}}{\Delta \mathbf{p}}\mathbf{r} + \mathbf{H}_{\mathbf{i}}
	$
	&
	&
	\\
	\midrule
	SSD\_For\_W
	& 
	$
	\Delta\mathbf{p} = -\hat{\mathbf{H}}_{\mathbf{i}}^{-1} \mathbf{J}_{\mathbf{i}}^T\bar{\mathbf{A}}\mathbf{r}
	$
	&
	&
	$
	\Delta\mathbf{c} = \mathbf{A} \mathbf{r}
	$
	\\
	\midrule
	\multirow{2}{*}{SSD\_Inv\_GN\_Sch \cite{Papandreou2008, Tzimiropoulos2013}} 
	& 
	$
	\Delta\mathbf{p} = \hat{\mathbf{H}}_{\mathbf{a}}^{-1} \mathbf{J}_{\mathbf{a}}^T\bar{\mathbf{A}}\mathbf{r}
	$
	&
	&
	\multirow{2}{*}{
	$
	\Delta\mathbf{c} = \mathbf{A} \left (\mathbf{r} - \mathbf{J}_{\mathbf{a}}\Delta\mathbf{p} \right)
	$
	}
	\\
	&
	$
	\hat{\mathbf{H}}_{\mathbf{a}} = \mathbf{J}_{\mathbf{a}}^T\bar{\mathbf{A}}\mathbf{J}_{\mathbf{a}}
	$
	&
	&
	\\
	\midrule
	\multirow{2}{*}{SSD\_Inv\_GN\_Alt \cite{Tzimiropoulos2012, Antonakos2014}} 
	& 
	$
	\Delta\mathbf{p} = \mathbf{H}_{\mathbf{a}}^{-1} \mathbf{J}_{\mathbf{a}}^T \left( \mathbf{r} - \mathbf{A}\Delta\mathbf{c} \right)
	$
	&
	&
	\multirow{2}{*}{
	$
	\Delta\mathbf{c} = \mathbf{A} \left (\mathbf{r} - \mathbf{J}_{\mathbf{a}}\Delta\mathbf{p} \right)
	$
	}
	\\
	&
	$
	\mathbf{H}_{\mathbf{a}} = \mathbf{J}_{\mathbf{a}}^T\mathbf{J}_{\mathbf{a}}
	$
	&
	&
	\\
	\midrule
	\multirow{2}{*}{SSD\_Inv\_N\_Sch} 
	& 
	$
	\Delta\mathbf{p} = \left(\hat{\mathbf{H}}_{\mathbf{a}}^{\textrm{N}}\right)^{-1} \mathbf{J}_{\mathbf{a}}^T\bar{\mathbf{A}}\mathbf{r}
	$
	&
	&
	\multirow{2}{*}{
	$
	\Delta\mathbf{c} = \mathbf{A} \left (\mathbf{r} - \mathbf{J}_{\mathbf{a}}\Delta\mathbf{p} \right)
	$
	}
	\\
	&
	$
	\hat{\mathbf{H}}_{\mathbf{a}}^{\textrm{N}} = \frac{\partial \mathcal{W}}{\Delta \mathbf{p}}^T \nabla^2\mathbf{a} \frac{\partial \mathcal{W}}{\Delta \mathbf{p}}\mathbf{r} + \hat{\mathbf{H}}_{\mathbf{a}}
	$
	&
	&
	\\
	\midrule
	\multirow{2}{*}{SSD\_Inv\_N\_Alt} 
	& 
	$
	\Delta\mathbf{p} = \left(\mathbf{H}_{\mathbf{a}}^{\textrm{N}}\right)^{-1} \mathbf{J}_{\mathbf{a}}^T\bar{\mathbf{A}}\left( \mathbf{r} - \mathbf{A}\Delta\mathbf{c} \right)
	$
	&
	&
	\multirow{2}{*}{
	$
	\Delta\mathbf{c} = \mathbf{A} \left (\mathbf{r} - \mathbf{J}_{\mathbf{a}}\Delta\mathbf{p} \right)
	$
	}
	\\
	&
	$
	\mathbf{H}_{\mathbf{a}}^{\textrm{N}} = \frac{\partial \mathcal{W}}{\Delta \mathbf{p}}^T \nabla^2\mathbf{i} \frac{\partial \mathcal{W}}{\Delta \mathbf{p}}\mathbf{r} + \mathbf{H}_{\mathbf{a}}
	$
	&
	&
	\\
	\midrule
	SSD\_Inv\_W
	& 
	$
	\Delta\mathbf{p} = \hat{\mathbf{H}}_{\mathbf{a}}^{-1} \mathbf{J}_{\mathbf{a}}^T\bar{\mathbf{A}}\mathbf{r}
	$
	&
	&
	$
	\Delta\mathbf{c} = \mathbf{A} \mathbf{r}
	$
	\\
	\midrule
	\multirow{2}{*}{SSD\_Asy\_GN\_Sch} 
	& 
	$
	\Delta\mathbf{p} = -\hat{\mathbf{H}}_{\mathbf{t}}^{-1} \mathbf{J}_{\mathbf{t}}^T\bar{\mathbf{A}}\mathbf{r}
	$
	&
	&
	\multirow{2}{*}{
	$
	\Delta\mathbf{c} = \mathbf{A} \left (\mathbf{r} + \mathbf{J}_{\mathbf{t}}\Delta\mathbf{p} \right)
	$
	}
	\\
	&
	$
	\hat{\mathbf{H}}_{\mathbf{t}} = \mathbf{J}_{\mathbf{t}}^T\bar{\mathbf{A}}\mathbf{J}_{\mathbf{t}}
	$
	&
	&
	\\
	\midrule
	\multirow{2}{*}{SSD\_Asy\_GN\_Alt} 
	& 
	$
	\Delta\mathbf{p} = -\mathbf{H}_{\mathbf{t}}^{-1} \mathbf{J}_{\mathbf{t}}^T \left( \mathbf{r} - \mathbf{A}\Delta\mathbf{c} \right)
	$
	&
	&
	\multirow{2}{*}{
	$
	\Delta\mathbf{c} = \mathbf{A} \left (\mathbf{r} + \mathbf{J}_{\mathbf{t}}\Delta\mathbf{p} \right)
	$
	}
	\\
	&
	$
	\mathbf{H}_{\mathbf{t}} = \mathbf{J}_{\mathbf{t}}^T\mathbf{J}_{\mathbf{t}}
	$
	&
	&
	\\
	\midrule
	\multirow{2}{*}{SSD\_Asy\_N\_Sch} 
	& 
	$
	\Delta\mathbf{p} = -\left(\hat{\mathbf{H}}_{\mathbf{t}}^{\textrm{N}}\right)^{-1} \mathbf{J}_{\mathbf{t}}^T\bar{\mathbf{A}}\mathbf{r}
	$
	&
	&
	\multirow{2}{*}{
	$
	\Delta\mathbf{c} = \mathbf{A} \left (\mathbf{r} + \mathbf{J}_{\mathbf{t}}\Delta\mathbf{p} \right)
	$
	}
	\\
	&
	$
	\hat{\mathbf{H}}_{\mathbf{t}}^{\textrm{N}} = \frac{\partial \mathcal{W}}{\Delta \mathbf{p}}^T \nabla^2\mathbf{t} \frac{\partial \mathcal{W}}{\Delta \mathbf{p}}\mathbf{r} + \hat{\mathbf{H}}_{\mathbf{t}}
	$
	&
	&
	\\
	\midrule
	\multirow{2}{*}{SSD\_Asy\_N\_Alt} 
	& 
	$
	\Delta\mathbf{p} = -\left(\mathbf{H}_{\mathbf{t}}^{\textrm{N}}\right)^{-1} \mathbf{J}_{\mathbf{t}}^T\bar{\mathbf{A}}\left( \mathbf{r} - \mathbf{A}\Delta\mathbf{c} \right)
	$
	&
	&
	\multirow{2}{*}{
	$
	\Delta\mathbf{c} = \mathbf{A} \left (\mathbf{r} + \mathbf{J}_{\mathbf{t}}\Delta\mathbf{p} \right)
	$
	}
	\\
	&
	$
	\mathbf{H}_{\mathbf{t}}^{\textrm{N}} = \frac{\partial \mathcal{W}}{\Delta \mathbf{p}}^T \nabla^2\mathbf{t} \frac{\partial \mathcal{W}}{\Delta \mathbf{p}}\mathbf{r} + \mathbf{H}_{\mathbf{t}}
	$
	&
	&
	\\
	\midrule
	SSD\_Asy\_W
	& 
	$
	\Delta\mathbf{p} = -\hat{\mathbf{H}}_{\mathbf{t}}^{-1} \mathbf{J}_{\mathbf{t}}^T\bar{\mathbf{A}}\mathbf{r}
	$
	&
	&
	$
	\Delta\mathbf{c} = \mathbf{A} \mathbf{r}
	$
	\\
	\midrule
	\multirow{3}{*}{SSD\_Bid\_GN\_Sch} 
	& 
	\multirow{2}{*}{
	$
	\Delta\mathbf{p} = -\hat{\mathbf{H}}_{\mathbf{i}}^{-1} \mathbf{J}_{\mathbf{i}}^T\bar{\mathbf{A}} \mathbf{r}_1
	$
	}
	&
	$
	\Delta\mathbf{q} = \check{\mathbf{H}}_{\mathbf{a}}^{-1} \mathbf{J}_{\mathbf{a}}^T\mathbf{P}\mathbf{r}
	$
	&
	\multirow{2}{*}{
	$
	\Delta\mathbf{c} = \mathbf{A} \mathbf{r}_2
	$
	}
	\\
	&
	\multirow{2}{*}{
	$
	\mathbf{r}_1 = \left(\mathbf{r} -\mathbf{J}_{\mathbf{a}}\Delta\mathbf{q}\right)
	$
	}
	&
	$
	\check{\mathbf{H}}_{\mathbf{a}} = \mathbf{J}_{\mathbf{a}}^T\mathbf{P}\mathbf{J}_{\mathbf{a}}
	$
	&
	\multirow{2}{*}{
	$
	\mathbf{r}_2 = \left (\mathbf{r} + \mathbf{J}_{\mathbf{i}}\Delta\mathbf{p} - \mathbf{J}_{\mathbf{a}}\Delta\mathbf{q}\right)
	$
	}
	\\
	&
	&
	$
	\mathbf{P} = \bar{\mathbf{A}} - \bar{\mathbf{A}}\mathbf{J}_{\mathbf{i}}\hat{\mathbf{H}}_\mathbf{i}^{-1}\mathbf{J}^T_{\mathbf{i}}\bar{\mathbf{A}} 
	$
	&
	\\
	\midrule
	\multirow{2}{*}{SSD\_Bid\_GN\_Alt}
	& 
	$
	\Delta\mathbf{p} = -\mathbf{H}_{\mathbf{i}}^{-1} \mathbf{J}_{\mathbf{i}}^T \mathbf{r}_3
	$
	&
	$
	\Delta\mathbf{q} = \mathbf{H}_{\mathbf{a}}^{-1} \mathbf{J}_{\mathbf{a}}^T \mathbf{r}_4
	$
	&
	$
	\Delta\mathbf{c} = \mathbf{A} \mathbf{r}_2
	$
	\\
	&
	$
	\mathbf{r}_3 = \left(\mathbf{r} - \mathbf{A}\Delta\mathbf{c} - \mathbf{J}_{\mathbf{a}}\Delta\mathbf{q} \right)
	$
	&
	$
	\mathbf{r}_4 = \left(\mathbf{r} - \mathbf{A}\Delta\mathbf{c} + \mathbf{J}_{\mathbf{i}}\Delta\mathbf{p} \right)
	$
	&
	\\
	\midrule
	\multirow{3}{*}{SSD\_Bid\_N\_Sch} 
	& 
	\multirow{3}{*}{
	$
	\Delta\mathbf{p} = -\left( \hat{\mathbf{H}}_{\mathbf{i}}^{\textrm{N}} \right)^{-1} \mathbf{J}_{\mathbf{i}}^T\bar{\mathbf{A}} \mathbf{r}_1
	$
	}
	&
	$
	\Delta\mathbf{q} = \left( \check{\mathbf{H}}_{\mathbf{a}}^{\textrm{N}} \right)^{-1} \mathbf{J}_{\mathbf{a}}^T\mathbf{P}^{\textrm{N}}\mathbf{r}
	$
	&
	\multirow{3}{*}{
	$
	\Delta\mathbf{c} = \mathbf{A} \mathbf{r}_2
	$
	}
	\\
	&
	&
	$
	\check{\mathbf{H}}_{\mathbf{a}}^{\textrm{N}} = \frac{\partial \mathcal{W}}{\Delta \mathbf{p}}^T \nabla^2\mathbf{t} \frac{\partial \mathcal{W}}{\Delta \mathbf{p}}\mathbf{r} + \check{\mathbf{H}}_{\mathbf{a}}
	$
	&
	\\
	&
	&
	$
	\mathbf{P}^{\textrm{N}} = \bar{\mathbf{A}} - \bar{\mathbf{A}}\mathbf{J}_{\mathbf{i}}\left(\hat{\mathbf{H}}_\mathbf{i}^{\textrm{N}}\right)^{-1}\mathbf{J}^T_{\mathbf{i}}\bar{\mathbf{A}} 
	$
	&
	\\
	\midrule
	SSD\_Bid\_N\_Alt
	& 
	$
	\Delta\mathbf{p} = -\left(\mathbf{H}_{\mathbf{i}}^{\textrm{N}}\right)^{-1} \mathbf{J}_{\mathbf{i}}^T \mathbf{r}_3
	$
	&
	$
	\Delta\mathbf{q} = \left(\mathbf{H}_{\mathbf{a}}^{\textrm{N}}\right)^{-1} \mathbf{J}_{\mathbf{a}}^T \mathbf{r}_4
	$
	&
	$
	\Delta\mathbf{c} = \mathbf{A} \mathbf{r}_2
	$
	\\
	\midrule
	SSD\_Bid\_W
	& 
	$
	\Delta\mathbf{p} = -\hat{\mathbf{H}}_{\mathbf{i}}^{-1} \mathbf{J}_{\mathbf{i}}^T\bar{\mathbf{A}} \mathbf{r}
	$
	&
	$
	\Delta\mathbf{q} = \check{\mathbf{H}}_{\mathbf{a}}^{-1} \mathbf{J}_{\mathbf{a}}^T\mathbf{P}\mathbf{r}
	$
	&
	$
	\Delta\mathbf{c} = \mathbf{A} \mathbf{r}
	$
	\\
	\bottomrule
\end{tabular}
\caption{Iterative solutions of all SSD algorithms studied in this paper.}
\label{tab:ssd_solution}
\end{table*}

\begin{table*}
\centering
\begin{tabular}{l|l|l}
	\toprule
	\multirow{2}{*}{Project-Out algorithms} & \multicolumn{2}{c}{Iterative solutions}
	\\
	& \multicolumn{1}{c|}{$\Delta\mathbf{p}$} & \multicolumn{1}{c}{$\Delta\mathbf{q}$}
	\\
	\midrule
	\multirow{2}{*}{PO\_For\_GN \cite{Amberg2009, Tzimiropoulos2013}} 
	& 
	$
	\Delta\mathbf{p} = -\hat{\mathbf{H}}_{\mathbf{i}}^{-1} \mathbf{J}_{\mathbf{i}}^T\bar{\mathbf{A}}\mathbf{r}
	$
	&
	\\
	& 
	$
	\hat{\mathbf{H}}_{\mathbf{i}} = \mathbf{J}_{\mathbf{i}}^T\bar{\mathbf{A}}\mathbf{J}_{\mathbf{i}} 
	$
	&
	\\
	\midrule
	\multirow{2}{*}{PO\_For\_N}
	& 
	$
	\Delta\mathbf{p} = -\left( \hat{\mathbf{H}}_{\mathbf{i}}^\textrm{N} \right)^{-1} \mathbf{J}_{\mathbf{i}}^T\bar{\mathbf{A}}\mathbf{r}
	$
	&
	\\
	& 
	$
	\hat{\mathbf{H}}_{\mathbf{i}}^{\textrm{N}} = \frac{\partial \mathcal{W}}{\partial\Delta \mathbf{p}}^T \nabla^2\mathbf{i} \frac{\partial \mathcal{W}}{\partial\Delta \mathbf{p}}\bar{\mathbf{A}}\mathbf{r} + \hat{\mathbf{H}}_{\mathbf{i}}
	$
	&
	\\
	\midrule
	\multirow{2}{*}{PO\_Inv\_GN \cite{Matthews2004}}
	& 
	$
	\Delta\mathbf{p} = \hat{\mathbf{H}}_{\mathbf{a}}^{-1} \mathbf{J}_{\bar{\mathbf{a}}}^T\bar{\mathbf{A}}\mathbf{r}
	$
	&
	\\
	& 
	$
	\hat{\mathbf{H}}_{\bar{\mathbf{a}}} = \mathbf{J}_{\bar{\mathbf{a}}}^T\bar{\mathbf{A}}\mathbf{J}_{\bar{\mathbf{a}}}
	$
	&
	\\
	\midrule
	\multirow{2}{*}{PO\_Inv\_N}
	&
	$
	\Delta\mathbf{p} = \left( \hat{\mathbf{H}}_{\bar{\mathbf{a}}}^{\textrm{N}} \right)^{-1} \mathbf{J}_{\bar{\mathbf{a}}}^T\bar{\mathbf{A}}\mathbf{r}
	$
	&
	\\
	& 
	$
	\hat{\mathbf{H}}_{\bar{\mathbf{a}}}^{\mathrm{N}} = \frac{\partial \mathcal{W}}{\Delta \mathbf{p}}^T \nabla^2\bar{\mathbf{a}} \frac{\partial \mathcal{W}}{\Delta \mathbf{p}}\bar{\mathbf{A}}\mathbf{r} + \hat{\mathbf{H}}_{\bar{\mathbf{a}}}
	$
	&
	\\
	\midrule
	\multirow{2}{*}{PO\_Asy\_GN} 
	& 
	$
	\Delta\mathbf{p} = -\hat{\mathbf{H}}_{\mathbf{t}}^{-1} \mathbf{J}_{\mathbf{t}}^T\bar{\mathbf{A}}\mathbf{r}
	$
	&
	\\
	& 
	$
	\hat{\mathbf{H}}_{\mathbf{t}} = \mathbf{J}_{\mathbf{t}}^T\bar{\mathbf{A}}\mathbf{J}_{\mathbf{t}} 
	$
	&
	\\
	\midrule
	\multirow{2}{*}{PO\_Asy\_N}
	& 
	$
	\Delta\mathbf{p} = -\left( \hat{\mathbf{H}}_{\mathbf{t}}^\textrm{N} \right)^{-1} \mathbf{J}_{\mathbf{t}}^T\bar{\mathbf{A}}\mathbf{r}
	$
	&
	\\
	& 
	$
	\hat{\mathbf{H}}_{\mathbf{t}}^{\textrm{N}} = \frac{\partial \mathcal{W}}{\partial\Delta \mathbf{p}}^T \nabla^2\mathbf{t} \frac{\partial \mathcal{W}}{\partial\Delta \mathbf{p}}\bar{\mathbf{A}}\mathbf{r} + \hat{\mathbf{H}}_{\mathbf{t}}
	$
	&
	\\
	\midrule
	\multirow{3}{*}{PO\_Bid\_GN\_Sch} 
	& 
	\multirow{3}{*}{
	$
	\Delta\mathbf{p} = -\hat{\mathbf{H}}_{\mathbf{i}}^{-1} \mathbf{J}_{\mathbf{i}}^T\bar{\mathbf{A}} \left( \mathbf{r} - \mathbf{J}_{\bar{\mathbf{a}}}\Delta\mathbf{q} \right)
	$
	}
	&
	$
	\Delta\mathbf{q} = \check{\mathbf{H}}_{\bar{\mathbf{a}}}^{-1} \mathbf{J}_{\mathbf{i}}^T\mathbf{P}\mathbf{r}
	$
	\\
	& 
	&
	$
	\check{\mathbf{H}}_{\bar{\mathbf{a}}} = \mathbf{J}_{\bar{\mathbf{a}}}^T\mathbf{P}\mathbf{J}_{\bar{\mathbf{a}}}
	$
	\\
	& 
	&
	$
	\mathbf{P} = \bar{\mathbf{A}} - \bar{\mathbf{A}}\mathbf{J}_{\mathbf{i}}\hat{\mathbf{H}}_\mathbf{i}^{-1}\mathbf{J}^T_{\mathbf{i}}\bar{\mathbf{A}} 
	$
	\\
	\midrule
	PO\_Bid\_GN\_Alt 
	& 
	$
	\Delta\mathbf{p} = -\hat{\mathbf{H}}_{\mathbf{i}}^{-1} \mathbf{J}_{\mathbf{i}}^T\bar{\mathbf{A}} \left( \mathbf{r} - \mathbf{J}_{\bar{\mathbf{a}}}\Delta\mathbf{q} \right)
	$
	&
	$
	\Delta\mathbf{q} = \hat{\mathbf{H}}_{\bar{\mathbf{a}}}^{-1} \mathbf{J}_{\bar{\mathbf{a}}}^T\bar{\mathbf{A}} \left( \mathbf{r} + \mathbf{J}_{\mathbf{i}}\Delta\mathbf{p} \right)
	$
	\\
	\midrule
	\multirow{3}{*}{PO\_Bid\_N\_Sch} 
	& 
	\multirow{3}{*}{
	$
	\Delta\mathbf{p} = -\left( \hat{\mathbf{H}}_\mathbf{i}^{\textrm{N}}\right)^{-1} \mathbf{J}_{\mathbf{i}}^T\bar{\mathbf{A}}\left( \mathbf{r} - \mathbf{J}_{\bar{\mathbf{a}}} \Delta\mathbf{q} \right) 
	$
	}
	&
	$
	\Delta\mathbf{q} = \left(\check{\mathbf{H}}_{\bar{\mathbf{a}}}^{\textrm{N}}\right)^{-1} \mathbf{J}_{\bar{\mathbf{a}}}^T{\mathbf{P}}^{\mathrm{N}}\mathbf{r}
	$
	\\
	& 
	&
	$
	\check{\mathbf{H}}_{\bar{\mathbf{a}}}^\textrm{N} = \frac{\partial \mathcal{W}}{\Delta \mathbf{p}}^T \nabla^2\bar{\mathbf{a}} \frac{\partial \mathcal{W}}{\Delta \mathbf{p}}\bar{\mathbf{A}}\mathbf{r} + \check{\mathbf{H}}_{\bar{\mathbf{a}}}
	$
	\\
	& 
	&
	$
	\mathbf{P}^{\mathrm{N}} = \bar{\mathbf{A}} - \bar{\mathbf{A}}\mathbf{J}_{\mathbf{i}}\left(\hat{\mathbf{H}}_\mathbf{i}^{\textrm{N}}\right)^{-1}\mathbf{J}^T_{\mathbf{i}}\bar{\mathbf{A}} 
	$
	\\
	\midrule
	PO\_Bid\_N\_Alt
	& 
	$
	\Delta\mathbf{p} = -\left(\hat{\mathbf{H}}_{\mathbf{i}}^{\mathrm{N}}\right)^{-1} \mathbf{J}_{\mathbf{i}}^T\bar{\mathbf{A}} \left( \mathbf{r} - \mathbf{J}_{\bar{\mathbf{a}}} \Delta\mathbf{q} \right) 
	$
	&
	$
	\Delta\mathbf{q} = \left(\hat{\mathbf{H}}_{\bar{\mathbf{a}}}^{\mathrm{N}}\right)^{-1} \mathbf{J}_{\bar{\mathbf{a}}}^T\bar{\mathbf{A}} \left( \mathbf{r} + \mathbf{J}_{\mathbf{i}} \Delta\mathbf{p} \right) 
	$
	\\
	\midrule
	PO\_Bid\_W
	& 
	$
	\Delta\mathbf{p} = -\hat{\mathbf{H}}_{\mathbf{i}}^{-1} \mathbf{J}_{\mathbf{i}}^T\bar{\mathbf{A}}\mathbf{r}
	$
	&
	$
	\Delta\mathbf{q} = \check{\mathbf{H}}_{\bar{\mathbf{a}}}^{-1} \mathbf{J}_{\bar{\mathbf{a}}}^T\mathbf{P}\mathbf{r}
	$
	\\
	\bottomrule
\end{tabular}
\caption{Iterative solutions of all Project-Out algorithms studied in this paper.}
\label{tab:po_solution}
\end{table*}

\end{document}